\PassOptionsToPackage{table}{xcolor}
\documentclass[acmtog,screen]{acmart}

\AtBeginDocument{%
  }

\copyrightyear{2026}
\acmYear{2026}
\setcopyright{cc}
\setcctype{by-nc-nd}
\acmConference[SA Conference Papers '26]{SIGGRAPH Asia 2026 Conference Papers}{December 01--04, 2026}{Kuala Lumpur, Malaysia}
\acmBooktitle{SIGGRAPH Asia 2026 Conference Papers (SA Conference Papers '26), December 01--04, 2026, Kuala Lumpur, Malaysia}
\acmDOI{10.1145/3829340.3842331}
\acmISBN{979-8-4007-2842-6/2026/12}

\usepackage{enumitem}
\usepackage{multirow}
\usepackage{algorithm}
\usepackage{algorithmic}

\definecolor{section}{HTML}{00960a}
\definecolor{highlight}{HTML}{ff4821}
\definecolor{green}{HTML}{6b9a6e}
\definecolor{lightgreen}{HTML}{a4e0a8}
\definecolor{yellow}{HTML}{ffeebd}
\definecolor{orange}{HTML}{ff9e7b}
\definecolor{red}{HTML}{eb6548}
\definecolor{orange2}{HTML}{fbb69d}
\definecolor{darkred}{HTML}{c3375a}
\definecolor{purple}{HTML}{a779e9}
\definecolor{best}{HTML}{ffeebd}
\definecolor{secondbest}{HTML}{a4e0a8}
\definecolor{catblackbox}{HTML}{D6E8F5}
\definecolor{catmulti}{HTML}{E8DDF5}
\definecolor{catfine}{HTML}{F5E0DD}
\colorlet{catmodular}{catfine}

\newif\ifarxiv
\arxivtrue

\newcommand{\myref}[2]{%
\ifarxiv
    \ref{#1}%
\else
    #2%
\fi
}

\begin{document}

\title{Modular Neural Image Signal Processing}

\author{Mahmoud Afifi}
\email{m.3afifi@gmail.com}
\affiliation{%
  \institution{AI Center-Toronto, Samsung Electronics}
  \city{Toronto}
  \country{Canada}
}

\author{Zhongling Wang}
\email{z.wang2@samsung.com}
\affiliation{%
  \institution{AI Center-Toronto, Samsung Electronics}
  \city{Toronto}
  \country{Canada}
}

\author{Ran Zhang}
\email{ran.zhang@samsung.com}
\affiliation{%
  \institution{AI Center-Toronto, Samsung Electronics}
  \city{Toronto}
  \country{Canada}
}

\author{Michael S. Brown}
\email{michael.b1@samsung.com}
\affiliation{%
  \institution{AI Center-Toronto, Samsung Electronics}
  \city{Toronto}
  \country{Canada}
}

\begin{abstract}

  Neural image signal processing (ISP) pipelines have largely abandoned the modular design of classical ISPs in favor of black-box networks, gaining expressiveness but losing interpretability, controllability, and the ability to swap or extend stages. Reintroducing modularity exposes a tradeoff: pipelines built from fixed operators or fixed neural proxies preserve interpretability but limit expressiveness, while jointly optimizing the proxies recovers expressiveness but often entangles functional roles across stages. We present a practical role-preserving and yet expressive neural ISP built on a two-level training strategy. At the \emph{module} level, components with available ground truth are optimized separately to preserve functional decomposition. At the \emph{operator} level, where intermediate ground truth is impractical to obtain, operators are trained jointly under final-image supervision but kept aligned with their roles via architectural design and role-specific losses. The resulting framework is simultaneously \textbf{interpretable}, \textbf{controllable}, and \textbf{expressive}, enabling operator-level adjustment, targeted debugging, multi-style extension without full-pipeline retraining, and rendering of raw images from unseen cameras by swapping the camera-specific modules. Lightweight at $\sim$0.5--$3.9$M parameters, our ISP achieves state-of-the-art results on most S24 picture styles and competitive performance on MIT-Adobe FiveK at a fraction of the parameter count of the closest neural ISPs. Code and the photo-editing tool are available at \url{https://github.com/mahmoudnafifi/modular_neural_isp}.

\end{abstract} 



\begin{teaserfigure}
  \centering
  \captionsetup{type=figure} %
  \includegraphics[clip,trim={1.7cm 6.5cm 5.3cm 0cm},width=\textwidth]{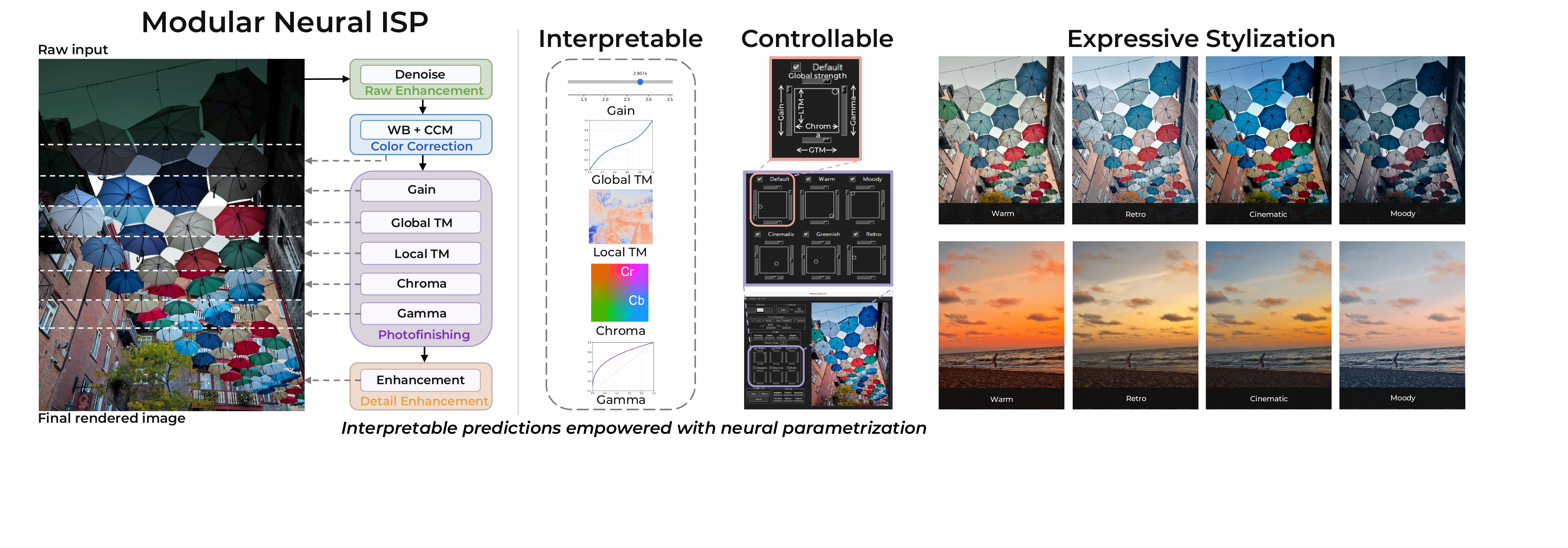}
  \captionof{figure}{
    We present a modular neural ISP that decomposes raw-to-sRGB rendering into a sequence of role-aligned operators. From a single raw input (iPhone~13), our pipeline produces the final image while remaining \textbf{interpretable}---each operator exposes compact parameters such as tone curves and a 2D chroma LuT; \textbf{controllable}---these parameters can be edited interactively to adjust the look; and \textbf{expressive}---supporting diverse picture styles and user-personalized renderings from the same raw. The framework was not trained on any iPhone captures, and this rendering uses our generic denoiser and cross-camera white balance. See Sec.~\ref{sec:photo-editing-tool} for details on the UI. Abbreviations: white-balance (WB), color correction matrix (CCM), tone mapping (TM).
  \label{fig:teaser}}
\end{teaserfigure} 
\maketitle
\section{Introduction and Related Work}
\label{sec:intro}

Classical image signal processing (ISP) pipelines are modular: raw sensor data is transformed into display-referred images through a sequence of well-defined stages~\cite{delbracio2021mobile, karaimer2016software}---raw image enhancement~\cite{brooks2019unprocessing, liba2019handheld, liang2020raw, wang2020practical, conde2024bsraw}, white balancing and color-space conversion~\cite{barron2015convolutional, hu2017fc4, bianco2013color, brown2023color, afifi2024optimizing}, and quality enhancement~\cite{hdrnet, vinker2021unpaired, wang2021dual, zhang2019zoom, abuolaim2020defocus, wu2021train, kee2025removing}. Traditional decomposition makes pipelines tunable and transparent, though engineering-heavy. Modern neural ISPs avoid this ``calibration tax'' through black-box networks, yet they lack the transparency that makes classical designs easy to scale and support.

\paragraph{Single-stage Neural ISP}
Recent learning-based approaches replace the entire ISP with a single black-box neural network trained end-to-end on a specific camera (e.g.,~\cite{cycleisp, zurich, invertible, lan, microisp, lite-isp, fourier, ispdiffuser}). Such single-stage black-box designs face three limitations. First, they generalize poorly to unseen cameras since their mappings are tightly coupled to the training device~\cite{afifi2021semi, perevozchikov2024rawformer}. Second, several models (e.g.,~\cite{zurich, fourier, ispdiffuser}) demand substantial memory and compute, impractical for on-device or interactive rendering. Third, they are difficult to interpret, debug, or extend---a barrier to supporting new picture styles or fixing user-specific corner cases~\cite{petapixel2020wildfirewb, petapixel2023iphone15, adobe2024hdrdenoise, adobe2024expertraw, gray2024photostyles}.

\paragraph{Coarse-grained Multi-stage Neural ISP}
\emph{Coarse-grained} multi-stage ISPs (e.g.,~\cite{cameranet, afifi2021cie, flexisp, paramisp, pixtalk}) group many ISP operations into one stage, leaving control at the stage boundary, and only partially address the three limitations above. For example, CameraNet~\cite{cameranet} has two stages---restoration then enhancement---each supervised by its own intermediate ground truth from third-party software. Its enhancement stage bundles exposure, tone, and color into one black-box network. Some require end-to-end fine-tuning that compromises interpretability~\cite{flexisp}, while others operate strictly in the nonlinear domain without a clear strategy for linear raw input~\cite{pixtalk}.

\paragraph{Fine-grained Multi-stage Neural ISP}
\emph{Fine-grained} designs assign a single operator to each stage, making every operation individually inspectable. They also make the stage order a design choice, as coarse stages follow the sequence dictated by image formation.
Exposure~\cite{hu2018exposure}, ReconfigISP~\cite{yu2021reconfigisp}, and AdaptiveISP~\cite{adaptiveisp} adopt this level of granularity and share our high-level idea of predicting parameters for differentiable operators, but determine the pipeline order dynamically rather than fixing it. Exposure uses reinforcement learning to select and order handcrafted global filters and predict their parameters at each step, limiting expressiveness for complex styles. AdaptiveISP shares this design but optimizes object-detection accuracy rather than rendering quality. ReconfigISP performs neural-proxy topology search, training each proxy to imitate a handcrafted algorithm, and replaces them with the handcrafted functions at deployment. Because their pipeline orders are image- or dataset-adaptive, the resulting pipelines lack a stable stage-to-role mapping. These methods also omit local tone mapping (LTM) and chroma mapping---essential for high dynamic range and realistic photofinishing in commercial ISPs---and require full retraining to add or modify operators. Fig.~\ref{fig:photofinishing_pipeline} visually compares our photofinishing module against Exposure.

Neural Photo Finishing (NPF)~\cite{neural_photo_finishing} is positioned by its authors as a differentiable proxy of Adobe Camera Raw (ACR), inheriting ACR's fixed order, with one proxy per stage. We differ in three ways: (a)~NPF inherits ACR's stages (no LTM, vendor-fixed color tables) and freezes proxies trained to imitate ACR; we design our pipeline and train learnable LTM and chroma-LuT operators end-to-end with role-specific architectural and loss constraints. (b)~NPF's parameter space is confined to ACR's sliders, which cannot represent spatially-varying tone or complex chroma transforms; we predict image-adaptive LTM grids and a chroma LuT. (c)~NPF's modularity is inherited from ACR---training requires proprietary tap-outs from each stage in a closed-source ISP---while ours is built into the framework and needs no such access.

\paragraph{Role-Preserving Modular Design}
Our design adds two properties to the fine-grained setting: (i) a fixed, conventional pipeline order, and (ii) each stage constrained to perform its intended functional role after joint training. A fixed order is important for role preservation: an operator's functional role is only well-defined relative to its upstream input distribution. Under dynamic ordering, this distribution varies per image, forcing operators to absorb different upstream contexts and entangling roles across stages---which in turn breaks~(ii) and prevents stable module replacement, debugging, or extension. Different datasets may favor different stage orders (Table~\ref{tab:order-ablation}), and we adopt one that generalizes well. Prior fine-grained designs do not learn their operators end-to-end, using either hand-designed operators whose roles are fixed by construction~\cite{hu2018exposure, adaptiveisp} or neural operators supervised against intermediate targets~\cite{yu2021reconfigisp, neural_photo_finishing}. Our operators are learned jointly from final-image supervision, with role preservation built into the architecture and the losses.

\begin{figure*}[!tp]
  \centering
  \includegraphics[width=\linewidth]{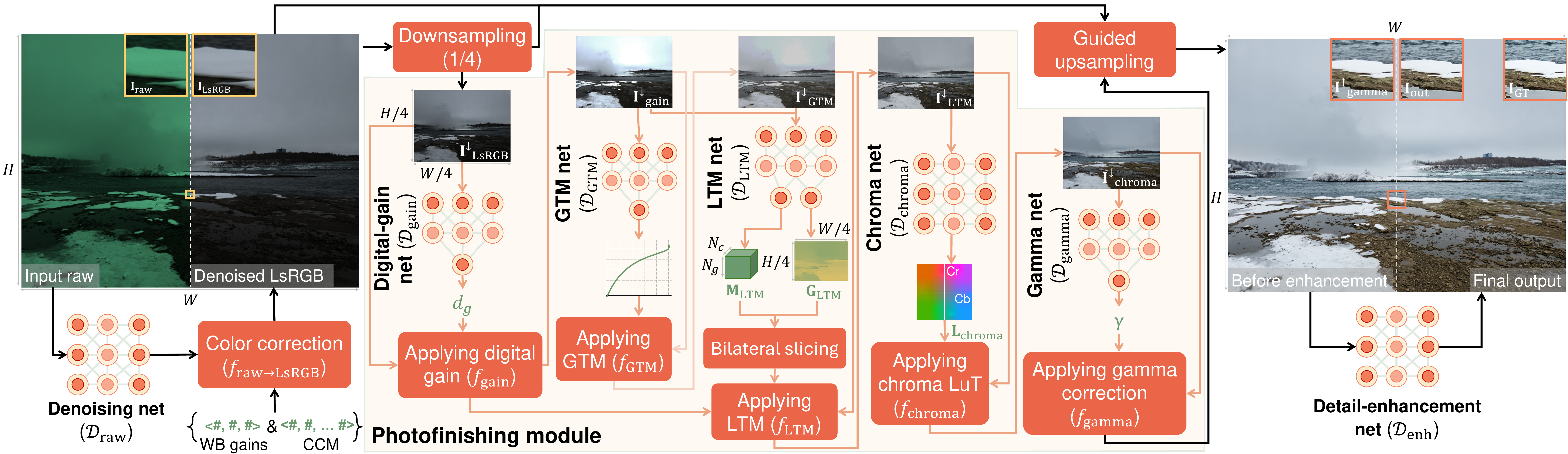}
  \caption{Overview of our modular framework. The pipeline begins with two camera-dependent modules: raw denoising and color correction. Then, the photofinishing module processes a downsampled version of the linear sRGB image through five operators. A guided upsampling step reconstructs the full-resolution output, which is then refined by a detail-enhancement module to produce the final image.
  }
  \label{fig:main}
\end{figure*}

\paragraph{Our Method}
Our modular ISP (Fig.~\ref{fig:teaser} and ~\ref{fig:main}) consists of five modules---raw enhancement, color correction, photofinishing, guided upsampling, and detail-enhancement---with photofinishing further decomposed into five learnable operators: digital gain, global tone mapping (GTM), local tone mapping (LTM), chroma mapping, and gamma correction. To preserve functional decomposition while still approximating complex photographic styles, we train at two levels. Where ground-truth supervision is available (module level), we use direct losses and module-wise optimization. Where intermediate ground truth is impractical to obtain (operator level), the operators within a module are trained end-to-end from final-image supervision but constrained to their intended roles through architectural design and role-specific losses. Our framework combines a fixed-order, role-aligned design with end-to-end training to support practical control and replacement at both module and operator levels while remaining competitive in expressiveness.

This design enables fine-grained control over each module, isolation of artifact causes, replacement of camera-specific modules with generic ones for unseen cameras, and efficient multi-style extension without retraining the full pipeline. To showcase this, we build a lightweight photo-editing tool atop our modular ISP. Details are in Sec.~\ref{sec:photo-editing-tool} and
\ifarxiv
the accompanying \href{https://youtu.be/ByhQjQSjxVM}{video (click to view)}.
\else
the accompanying supplementary video.
\fi

\paragraph{Contributions}
(1)~A practical, role-preserving modular neural ISP that follows a meaningful raw-to-sRGB processing order, supports independent module and operator control, and enables efficient multi-style rendering without full-pipeline retraining. A lightweight photo-editing tool showcases this flexibility.
(2)~A method for training role-constrained operators end-to-end from final-image supervision, with no proprietary intermediate tap-outs, instantiated as a photofinishing module of five operators (digital gain, GTM, LTM, chroma mapping, and gamma correction) that remain interpretable and independently controllable after joint training.
(3)~Extensive evaluation across module/operator/loss ablations, stage-order ablations, multi-style and cross-camera renderings, and a user study against proprietary ISPs.
(4)~Additional implementation contributions detailed in the supplementary materials: a GPU-accelerated Bilateral Guided Upsampling~\cite{bgu} with per-channel gated regularization (Sec.~\myref{sec:supp-bgu}{\textcolor{section}{E}}), a GPU-accelerated Fast Bilateral Solver~\cite{barron2016fast} (Sec.~\myref{sec:accelerated_bilateral_solver}{\textcolor{section}{A}}), multi-scale LTM-coefficient refinement for halo suppression (Sec.~\myref{sec:artifacts}{\textcolor{section}{B.1}}), and calibrated synthetic-noise augmentation for cross-camera robustness (Sec.~\myref{sec:cross-camera}{\textcolor{section}{H}}).
\section{Method}
\label{sec:method}

Given a demosaiced raw image $\mathbf{I}_{\texttt{raw}} \in \mathbb{R}^{H\times W\times 3}$,
we render a display-referred output $\mathbf{I}_{\texttt{out}} \in \mathbb{R}^{H\times W\times 3}$ in sRGB that matches a ground-truth reference $\mathbf{I}_{\texttt{GT}}$ rendered under a specific picture style, while maintaining modularity and fine-grained control across the pipeline. We consider demosaicing a swappable upstream module (an active research area in its own right), outside the scope of this paper. We use Menon~\cite{menon2007demosaicing} for mosaiced inputs.

Rather than searching for stage orderings~\cite{hu2018exposure, yu2021reconfigisp}, we adopt a fixed order aligned with the conventional raw-to-sRGB image formation process. Our framework (Fig.~\ref{fig:main}) comprises five major modules. The raw image first passes through raw enhancement (Sec.~\ref{sec:method-raw-enhancement}) and color correction (Sec.~\ref{sec:method-color-correction}), producing the linear sRGB image $\mathbf{I}_{\texttt{LsRGB}}$, which is downsampled to $\mathbf{I}^{\downarrow}_{\texttt{LsRGB}} \in \mathbb{R}^{\frac{H}{4}\times\frac{W}{4}\times 3}$. A photofinishing module (Sec.~\ref{sec:method-photofinishing}) then applies five parametric operators---digital gain, global tone mapping (GTM), local tone mapping (LTM), chroma mapping, and gamma correction---whose image-specific parameters are predicted by lightweight networks. Guided upsampling (Sec.~\ref{sec:method-upsampling}) restores full resolution, and a detail-enhancement step (Sec.~\ref{sec:method-detail-enhancement}) produces $\mathbf{I}_{\texttt{out}}$. Example intermediate outputs are shown in Fig.~\ref{fig:intermediate-example-style0}.

The major modules are trained separately to preserve module-level decomposition. Within photofinishing, where per-operator ground truth is unavailable, the operators are trained jointly from final-image supervision but kept aligned with their intended roles through operator parameterization and role-specific losses. At inference the pipeline takes only the raw image, and the style is selected by loading the corresponding photofinishing and detail-enhancement weights rather than by conditioning on a target image.
\subsection{Raw Enhancement}
\label{sec:method-raw-enhancement}

Raw images suffer from noise and detail loss due to low photon counts and sensor imperfections~\cite{chen2018learning, abdelhamed2018high}, making denoising an important first module of any ISP. We adopt single-image raw denoising for broad applicability:
\begin{equation}
  \label{eq:denoising}
  \mathbf{I}_{\texttt{enh-raw}} = f_{\texttt{enh-raw}}\!\left(\mathbf{I}_{\texttt{raw}}\right),
\end{equation}
where $f_{\texttt{enh-raw}}(\cdot)$ is implemented as a fully convolutional network $\mathcal{D}_{\texttt{raw}}$, trained with a pixel-wise $\ell_1$ loss against ground-truth denoised raws (see Sec.~\myref{sec:denoising-training}{\textcolor{section}{G.1}}). We provide two versions: a \emph{camera-specific} variant trained per device for highest fidelity, and a \emph{generic} variant trained on synthetic noise from multiple sensors for cross-camera robustness (Sec.~\myref{sec:cross-camera}{\textcolor{section}{H}}).
Ground-truth denoised raws (referred to as pseudo GT) are obtained from Adobe Lightroom AI denoiser rather than physically captured ones~\cite{abdelhamed2018high}, which are time-consuming to acquire and lack scene diversity. When no such denoiser is available, we provide a workflow that synthesizes pseudo ground truths by inverting the (already denoised by an in-camera ISP) target sRGB images back to the raw domain via inverse gamma and a fitted polynomial color mapping (Sec.~\myref{sec:misalignment}{\textcolor{section}{B.2}}).

\subsection{Color Correction}
\label{sec:method-color-correction}

After raw denoising, we apply a color-correction function $f_{\texttt{raw}\rightarrow\texttt{LsRGB}}(\cdot)$ that maps the denoised raw image into the linear sRGB space---a camera-agnostic space that decouples downstream operators from camera-specific characteristics:
\begin{equation}
  \label{eq:color-correction}
  \mathbf{I}_{\texttt{LsRGB}(x,y)} =
  \mathbf{M}_{\texttt{CCM}}\!\left( \mathbf{D}_{\texttt{WB}}\,\mathbf{I}_{\texttt{enh-raw}(x,y)} \right),
\end{equation}
where $\mathbf{D}_{\texttt{WB}}$ is a diagonal matrix of red and blue white-balance (WB) gains, and $\mathbf{M}_{\texttt{CCM}}$ is the color correction matrix (CCM) interpolated from pre-calibrated camera-specific matrices using these gains~\cite{kim2025ccmnet}. WB gains are typically read from DNG metadata (estimated by the camera's auto white balance, AWB), or can be predicted by learning-based AWB methods~(e.g., \cite{barron2017fast, lo2021clcc, kim2025ccmnet}). See Sec.~\myref{sec:awb}{\textcolor{section}{I.3}} for integration details.

\subsection{Photofinishing}
\label{sec:method-photofinishing}

\begin{figure*}[!t]
  \centering
  \includegraphics[width=\linewidth]{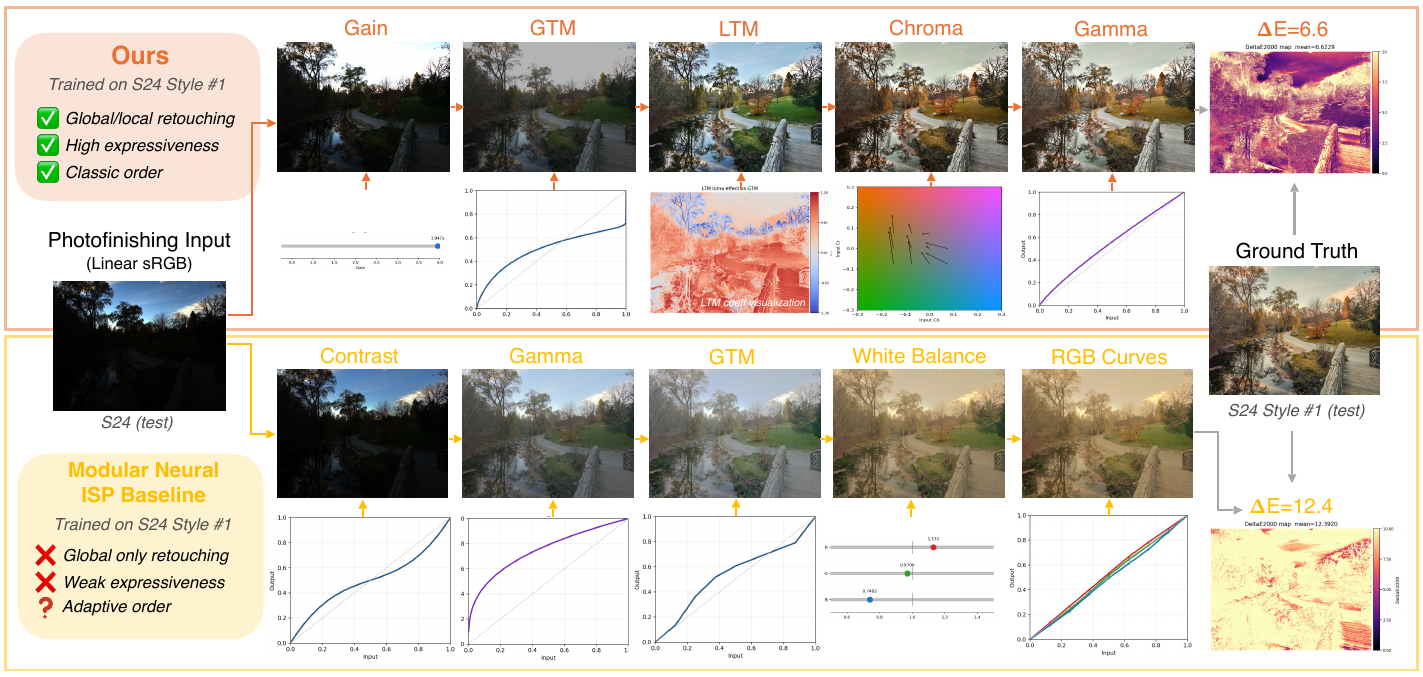}
  \caption{
    Photofinishing comparison on S24 Style~\#1. Both models are trained on Style~\#1 and process the same linear sRGB input. Top: our pipeline---gain, GTM, LTM, chroma LuT, gamma---in a fixed conventional order. Bottom: Exposure~\cite{hu2018exposure} (and similarly ReconfigISP~\cite{yu2021reconfigisp}) predicts an image-adaptive sequence of global operators. The LTM map visualizes the high-dimensional LTM coefficients as the log-ratio between LTM output and input. Detail-enhancement is disabled in our output for a fair $\Delta E$ comparison, since Exposure has no equivalent operator.
  }
  \label{fig:photofinishing_pipeline}
\end{figure*}

The photofinishing module is a core contribution of our work, finalizing the image's `look and feel' through both perceptual enhancement and artistic stylization. Its fixed, well-established operator ordering provides stability when learning photographic styles, unlike Exposure~\cite{hu2018exposure} and ReconfigISP~\cite{yu2021reconfigisp}, which determine the operator order dynamically and force operators to absorb different upstream contexts, thus entangling functional roles across operators. Crucially, our pipeline explicitly includes learnable local tone mapping and a chroma LuT, both absent from prior multi-stage neural ISPs~\cite{hu2018exposure,yu2021reconfigisp,neural_photo_finishing}, substantially enhancing expressiveness as illustrated in Fig.~\ref{fig:photofinishing_pipeline}.

The module operates on the downsampled image $\mathbf{I}^{\downarrow}_{\texttt{LsRGB}}$ for efficiency and comprises five operators: 1) digital gain ($f_{\texttt{gain}}$) for brightness, 2) global tone mapping ($f_{\texttt{GTM}}$) for global contrast and highlight preservation, 3) local tone mapping ($f_{\texttt{LTM}}$) for spatially adaptive contrast and detail, 4) chroma mapping ($f_{\texttt{chroma}}$) for chromaticity, and 5) gamma correction ($f_{\texttt{gamma}}$) for display-referred output. Each operator is parameterized by image-specific coefficient(s) predicted by a lightweight network ($\mathcal{D}_{\texttt{gain}}$, $\mathcal{D}_{\texttt{GTM}}$, $\mathcal{D}_{\texttt{LTM}}$, $\mathcal{D}_{\texttt{chroma}}$, $\mathcal{D}_{\texttt{gamma}}$, totaling $\sim$200K parameters). Since per-operator ground truth is unavailable, we train the photofinishing module end-to-end while keeping each operator aligned with its intended role (e.g., GTM adjusts global tone, not brightness) through architectural design and role-specific losses, as detailed below.

\paragraph{Digital Gain}
A global gain factor $d_g$ predicted by $\mathcal{D}_{\texttt{gain}}$ scales the input brightness:
\begin{equation}
\label{eq:gain}
\mathbf{I}^{\downarrow}_{\texttt{gain}} =
f_{\texttt{gain}}\left(\mathbf{I}^{\downarrow}_{\texttt{LsRGB}}; \,d_g\right)
= d_g \, \mathbf{I}^{\downarrow}_{\texttt{LsRGB}}.
\end{equation}

\paragraph{Global Tone Mapping (GTM)}
Tone mapping refines image contrast while maintaining perceptual brightness. GTM and LTM share a tone-mapping function $f_{\texttt{TM}}$ but differ in parameterization. GTM applies $f_{\texttt{TM}}$ pixel-wise (with $(x,y)$ indexing pixel locations) with image-wise parameters:
\begin{equation}
\label{eq:apply-gtm}
\mathbf{I}^{\downarrow(\rho)}_{\texttt{GTM}(x, y)} =
f_{\texttt{TM}}\big(
\mathbf{I}^{\downarrow(\rho)}_{\texttt{gain}(x, y)};\,
a_{\texttt{GTM}}, b_{\texttt{GTM}}, c_{\texttt{GTM}}
\big),
\end{equation}
where $\rho \in \{$R, G, B$\}$ indexes the color channel and $f_{\texttt{TM}}$ is defined as:
\begin{equation}
\label{eq:tm-function}
f_{\texttt{TM}}(x;\,a_{\texttt{TM}},b_{\texttt{TM}},c_{\texttt{TM}})
=\frac{x^{\,a_{\texttt{TM}}}}{x^{\,a_{\texttt{TM}}}+\big(c_{\texttt{TM}}\,(1{-}x)\big)^{\,b_{\texttt{TM}}}}.
\end{equation}
$x\!\in\![0,1]$ is the normalized RGB intensity. The image-wise parameters $a_{\texttt{GTM}}, b_{\texttt{GTM}}, c_{\texttt{GTM}}$, predicted by $\mathcal{D}_{\texttt{GTM}}$, control 1) midtone contrast (exponent $a_{\texttt{GTM}}$), 2) shadow compression (slope $b_{\texttt{GTM}}$), and 3) highlight roll-off (scaling $c_{\texttt{GTM}}$, modulating the curve's upper-end curvature). Tone mapping is applied directly to the linear RGB channels of $\mathbf{I}^{\downarrow}_{\texttt{gain}}$ rather than to luminance only (which would leave chroma to $f_{\texttt{chroma}}$). The luminance-only variant yielded suboptimal results in our ablations with details in Sec.~\myref{supp-sec:ablation-studies}{\textcolor{section}{K.1}}.

\paragraph{Local Tone Mapping (LTM)}
The LTM operator complements the GTM by providing spatially adaptive tone control.
To achieve this, $\mathcal{D}_{\texttt{LTM}}$ comprises two subnetworks: 1) a multi-scale guidance subnetwork that outputs a guidance map
$\mathbf{G}_{\texttt{guide}} \in \mathbb{R}^{\frac{H}{4}\times\frac{W}{4}\times 1}$, and 2) a grid-prediction subnetwork that processes both
$\mathbf{I}^{\downarrow}_{\texttt{gain}}$ and
$\mathbf{I}^{\downarrow}_{\texttt{GTM}}$ (concatenated along channels) to predict a coarse grid of tone-mapping parameters
$\mathbf{M}_{\texttt{LTM}} \in \mathbb{R}^{N_g \times N_g \times N_c \times 5}$,
where $N_g, N_c \ll H/4, W/4$.
Including both inputs provides the grid-prediction subnetwork with cues about the global tone-mapping behavior (from $\mathbf{I}^{\downarrow}_{\texttt{GTM}}$) and the pre-tonemapped image content (from $\mathbf{I}^{\downarrow}_{\texttt{gain}}$), helping it predict the coefficients applied to both inputs by the LTM function $f_{\texttt{LTM}}$, defined as:
\begin{equation}
\label{eq:apply-ltm}
\small{
\begin{alignedat}{2}
\mathbf{I}^{\downarrow(\rho)}_{\texttt{LTM}(x,y)} &=
(1-\mathbf{W}_{\texttt{LTM}(x,y)})
\,\mathbf{I}^{\downarrow(\rho)}_{\texttt{GTM}(x,y)} \\ %
&\quad+\mathbf{W}_{\texttt{LTM}(x,y)}
\,f_{\texttt{TM}}\,\big(
\mathbf{X}_{\texttt{LTM}(x,y)}^{(\rho)};
\mathbf{A}_{\texttt{LTM}(x,y)},
\mathbf{B}_{\texttt{LTM}(x,y)},
\mathbf{C}_{\texttt{LTM}(x,y)}
\big),
\end{alignedat}
}
\end{equation}
where $\mathbf{X}_{\texttt{LTM}(x,y)}$ is the locally scaled version of the gain-adjusted linear sRGB image:
\begin{equation}
\label{eq:ltm-input}
\mathbf{X}_{\texttt{LTM}(x, y)}^{(\rho)} =
\mathbf{I}^{\downarrow(\rho)}_{\texttt{gain}(x, y)}
\,\mathbf{G}_{\texttt{LTM}(x, y)},
\end{equation}
and the spatial coefficient maps:
\[
\mathbf{A}_{\texttt{LTM}},
\mathbf{B}_{\texttt{LTM}},
\mathbf{C}_{\texttt{LTM}},
\mathbf{G}_{\texttt{LTM}},
\mathbf{W}'_{\texttt{LTM}}
\in
\mathbb{R}^{\frac{H}{4}\times\frac{W}{4}}
\]
are generated by sampling the grid $\mathbf{M}_{\texttt{LTM}}$ via trilinear interpolation using the guidance map $\mathbf{G}_{\texttt{guide}}$.
The preliminary map $\mathbf{W}'_{\texttt{LTM}}$ is then passed through a sigmoid activation to obtain $\mathbf{W}_{\texttt{LTM}}$. This formulation enables the LTM to build upon the globally tone-mapped image while flexibly re-tone-mapping the gain-adjusted image as needed.
Unlike HDRNet~\cite{hdrnet}, which predicts arbitrary local affine color transforms via a bilateral grid, our LTM is constrained to parameterize the same tone-mapping curve $f_{\texttt{TM}}$ used by GTM. This functional constraint preserves the operator's tone-mapping role and achieves higher PSNR/SSIM with fewer parameters than HDRNet-based replacements (Table~\ref{tab:ablation-hdrnet}). See Sec.~\myref{supp-sec:ablation-studies}{\textcolor{section}{K.1}} for ablations.

\paragraph{Chroma Mapping}
After tone mapping, chroma mapping refines image colors for enhancement or artistic stylization. We implement $f_{\texttt{chroma}}$ as an image-specific 2D LuT in CbCr, chosen over a 3D RGB LuT~\cite{zeng2022lut} to reduce memory and ensure this operator affects only chromaticity. The tone-mapped image $\mathbf{I}^{\downarrow}_{\texttt{LTM}}$ is converted to YCbCr\footnote{We use ITU-R~BT.709 YCbCr~$\leftrightarrow$~RGB matrices, approximating our input as 2.2~gamma-encoded sRGB for simplicity and differentiability.}. The chroma network $\mathcal{D}_{\texttt{chroma}}$, an encoder-decoder, extracts features from a differentiable 2D CbCr histogram (Sec.~\myref{sec:hist}{\textcolor{section}{D}}) modulated by a sigmoid-gated latent vector predicted from the Y channel for brightness-dependent modulation, then outputs a residual 2D LuT. Adding this residual to a learnable global base LuT yields $\mathbf{L}_{\texttt{chroma}}$, which is applied to CbCr via bilinear interpolation. The chroma-mapped CbCr is recombined with Y and converted back to a pre-gamma, quasi-linear sRGB image $\mathbf{I}^{\downarrow}_{\texttt{chroma}}$.

For more stylized color manipulations in artistic picture styles, we optionally augment $f_{\texttt{chroma}}$ with an image-independent learnable 3D LuT $\mathbf{L}_{\texttt{RGB}}$ ($11{\times}11{\times}11$), applied via trilinear lookup to the tone-mapped RGB values before $\mathcal{D}_{\texttt{chroma}}$. This 3D LuT captures rich color transformations that complement $\mathbf{L}_{\texttt{chroma}}$, but can interfere with the roles of other tone-mapping operators, so we keep it optional.

\paragraph{Gamma Correction}
As the final operator of photofinishing, we apply gamma correction using $f_{\texttt{gamma}}$, defined as:
\begin{equation}
\label{eq:gamma}
\mathbf{I}^{\downarrow}_{\texttt{gamma}} =
f_{\texttt{gamma}}\,\left(\mathbf{I}^{\downarrow}_{\texttt{chroma}};\,\gamma\right)
= \left(\mathbf{I}^{\downarrow}_{\texttt{chroma}}\right)^{(1/\gamma)},
\end{equation}
where $\gamma$ is an image-specific factor predicted by $\mathcal{D}_{\texttt{gamma}}$.

\paragraph{Training Loss}
We train the five networks of our photofinishing module ($\mathcal{D}_{\texttt{gain}}$, $\mathcal{D}_{\texttt{GTM}}$, $\mathcal{D}_{\texttt{LTM}}$, $\mathcal{D}_{\texttt{chroma}}$, and $\mathcal{D}_{\texttt{gamma}}$), and optionally $\mathbf{L}_{\texttt{RGB}}$, jointly in an end-to-end manner by minimizing:

\begin{equation}
\label{eq:ps-loss-func}
\mathcal{L}_{\texttt{total}} = \mathcal{L}_{\texttt{fid}} + \mathcal{L}_{\texttt{reg}},
\end{equation}
with a fidelity term aggregating pixel, structural, and perceptual losses,
\begin{equation}
\label{eq:ps-loss-fid}
\small{\mathcal{L}_{\texttt{fid}} =
\lambda_1 \, \ell_1
+ \lambda_{\texttt{SSIM}} \, \ell_{\texttt{SSIM}}
+ \lambda_{\Delta E} \, \ell_{\Delta E}
+ \lambda_{\texttt{perc}} \, \ell_{\texttt{perc}}
+ \lambda_{\texttt{CbCr}} \, \ell_{\texttt{CbCr}},}
\end{equation}
and a role-specific constraint term that keeps each operator aligned with its intended function,
\begin{equation}
\label{eq:ps-loss-reg}
\small{\mathcal{L}_{\texttt{reg}} =
\lambda_{\texttt{LuT-s}} \, \ell_{\texttt{LuT-s}}
+ \lambda_{\texttt{LTM-s}} \, \ell_{\texttt{LTM-s}}
+ \lambda_{\texttt{TM}} \, \ell_{\texttt{TM}}
+ \lambda_{\texttt{luma}} \, \ell_{\texttt{luma}}.}
\end{equation}

The fidelity term combines pixel ($\ell_1$), structural ($\ell_{\texttt{SSIM}}$), color ($\ell_{\Delta E}$), and perceptual ($\ell_{\texttt{perc}}$) matching against the target sRGB image. $\ell_{\Delta E}$ uses $\Delta E_{76}$ for faster training, while our evaluation uses the more accurate $\Delta E_{2000}$. $\ell_{\texttt{CbCr}}$ compares the post-LuT CbCr to the de-gammaed ground-truth CbCr (using the predicted $\gamma$). This target shifts as $\gamma$ is learned, but $\gamma$ is bounded and training is stable in practice. The constraint term enforces functional decomposition. $\ell_{\texttt{TM}}$ separates GTM and LTM by anchoring each at its native scale: an $\ell_1$ between an $8{\times}$-downsampled GTM-output luma and the corresponding GT luma confines GTM to global tone, while a full-resolution $\ell_1$ between LTM-output luma and GT luma drives LTM to refine local tone. $\ell_{\texttt{luma}}$ pins the mean luma of the GTM output to that of the gain output, leaving brightness to $f_{\texttt{gain}}$ so GTM only adjusts contrast. $\ell_{\texttt{LTM-s}}$ and $\ell_{\texttt{LuT-s}}$ are total-variation smoothness penalties on the LTM coefficient maps and on $\mathbf{L}_{\texttt{chroma}}$. By construction, $f_{\texttt{gain}}$ (scalar multiply), $f_{\texttt{chroma}}$ (CbCr-only LuT), and $f_{\texttt{gamma}}$ (bounded monotone power) already have disjoint roles. $f_{\texttt{GTM}}$ and $f_{\texttt{LTM}}$ share $f_{\texttt{TM}}$ (smooth curve), so the constraint losses above are what keep their roles distinct. The coefficients~$\lambda_j$ weight each term. Full loss definitions, weighting factors, training setup, and ablations are in Secs.~\myref{sec:loss_funcs}{\textcolor{section}{F}},~\myref{sec:supp-photofinishing-training}{\textcolor{section}{G.2}},~and~\myref{supp-sec:ablation-studies}{\textcolor{section}{K.1}}.

\subsection{Upsampling}
\label{sec:method-upsampling}

For efficiency, photofinishing is performed on a downscaled image, $\mathbf{I}^{\downarrow}_{\texttt{LsRGB}}$, and produces $\mathbf{I}^{\downarrow}_{\texttt{gamma}}$, which is then upsampled using the high-resolution linear sRGB image $\mathbf{I}_{\texttt{LsRGB}}$ as guidance. We implemented a differentiable version of bilateral grid upsampling (BGU)~\cite{bgu}, which computes an affine transform per grid cell by solving a regularized least-squares system. The original Halide BGU constrains regularization to be achromatic (a single scalar gain across channels) and uses grid blurring to handle empty cells---both of which introduce limitations: 1) enforced achromaticity causes color crosstalk, and 2) grid blurring trades off detail for smoothness. We address these issues with per-channel gated regularization that removes the need for grid blurring. Each cell is regularized independently by channel, while empty cells fall back to global per-channel gains derived from pooled grid statistics.
This design yields sharper, more faithful reconstructions (see Sec.~\myref{sec:supp-bgu}{\textcolor{section}{E}} of the supplementary material for details).
The guided upsampling produces the photofinished image $\mathbf{I}^{\uparrow}_{\texttt{gamma}} \in \mathbb{R}^{H\times W\times3}$, which is then refined by the detail-enhancement module.

\subsection{Detail-Enhancement}
\label{sec:method-detail-enhancement}

The final module of our pipeline applies a detail-enhancement step to compensate for residual artifacts from denoising and guided upsampling.
The output image $\mathbf{I}_{\texttt{out}}$ is obtained as:
\begin{equation}
\label{eq:method-enh}
\mathbf{I}_{\texttt{out}} = f_{\texttt{enh}}\left(\mathbf{I}^{\uparrow}_{\texttt{gamma}}\right),
\end{equation}
where $f_{\texttt{enh}}(\cdot)$ is implemented as a compact fully convolutional network $\mathcal{D}_{\texttt{enh}}$.
To train $\mathcal{D}_{\texttt{enh}}$, we first generate $\mathbf{I}^{\uparrow}_{\texttt{gamma}}$ for each training image using all preceding modules (with pretrained models), and use them as inputs.
The network is optimized with a pixel-wise $\ell_1$ loss between the predicted and ground-truth sRGB images.
Additional details are provided in Sec.~\myref{sec:supp-detail-enhancement-training}{\textcolor{section}{G.3}} of the supplementary material.

We explored merging the raw enhancement and detail-enhancement modules by fine-tuning
$\mathcal{D}_{\texttt{raw}}$ after training the photofinishing module, aiming to
pre-enhance image details before resampling and thereby mitigate detail loss. However, this approach was unstable and often failed to converge. Using $\mathcal{D}_{\texttt{enh}}$ as a separate final module (with only $\sim$50K parameters) proved more robust and easier to train.

\subsection{Photo-Editing Tool}
\label{sec:photo-editing-tool}
To demonstrate the advantages of our modular design, we developed a user-interactive photo-editing tool built on top of our neural ISP.
The tool provides full control over the \textit{entire} rendering process, supporting multiple picture styles and additional editing adjustments (e.g., highlights, shadows, contrast, and exposure) directly within the ISP pipeline.
Our tool extends naturally beyond the training camera. By isolating camera-specific processing from photofinishing and enhancement, our modular design renders images from unseen cameras in our styles once they are mapped into the canonical linear sRGB space. To support this, we provide a generic denoiser and cross-camera white-balance options for unseen devices. We also support post-editable JPEG re-rendering
by embedding compressed raw data, and can process external sRGB
images through a lightweight linearization network. The full tool delivers these capabilities with only $\sim$~3.9M parameters.
The UI is shown in Fig.~\ref{fig:teaser} and Fig.~\myref{fig:gui}{\textcolor{section}{25}}. Comprehensive details and additional demonstrations are provided in the supplementary material (Sec.~\myref{sec:gui}{\textcolor{section}{I}}) and
\ifarxiv
in the accompanying \href{https://youtu.be/ByhQjQSjxVM}{video (click to view)}.
\else
the accompanying video.
\fi

Further details on network architectures (Sec.~\myref{sec:network_design}{\textcolor{section}{C}}), loss definitions (Sec.~\myref{sec:loss_funcs}{\textcolor{section}{F}}), weighting factors and training setup (Sec.~\myref{sec:training_details}{\textcolor{section}{G}}), and ablation studies (Sec.~\myref{sec:ablation_studies}{\textcolor{section}{K}}) are included in the supplementary material.

\section{Experimental Results}
\label{sec:experiments}

We compare our modular ISP framework against representative neural ISP methods across three categories: single-stage black-box models (PyNet~\cite{zurich}, LAN~\cite{lan}, LiteISP~\cite{lite-isp}, Invertible-ISP~\cite{invertible}, FourierISP~\cite{fourier}, MicroISP~\cite{microisp}, and ISPDiffuser~\cite{ispdiffuser}), coarse multi-stage architectures (CIE-XYZ Net~\cite{afifi2021cie}, ParamISP~\cite{paramisp}, and Deep-FlexISP~\cite{flexisp}), and fine multi-stage methods, which either order operators dynamically (Exposure~\cite{hu2018exposure} and ReconfigISP~\cite{yu2021reconfigisp}) or fix the order (Neural Photo-Finishing (NPF)~\cite{neural_photo_finishing}), as our design does.

Because a key objective of our framework is the ability to learn diverse picture styles, multi-style evaluation is essential. We therefore use the S24 dataset~\cite{s24}, which provides six Adobe Lightroom sRGB renderings per raw input---one default (Style~\#0) and five artistic styles (Styles~\#1--5). It comprises 2,619 training, 205 validation, and 400 testing raw images (example in Fig.~\ref{fig:qualtative-results}). We also train/evaluate on MIT-Adobe FiveK~\cite{Adobe5K}, which contains images captured with different cameras and has one target rendering style.

We trained three denoising variants ($\mathcal{D}_{\texttt{raw}}$)---lite (0.25~M params), base (0.93~M), and large (3.6~M) using the pseudo ground-truth denoised raws in S24, which were generated by Adobe Lightroom's AI denoiser. Details are in Sec.~\myref{sec:supp-denoising-nets}{\textcolor{section}{C.1}}. Each module (denoising, photofinishing, detail-enhancement) is trained separately. Only photofinishing and detail enhancement are style-specific. The denoiser is shared across styles, substantially reducing per-style training and memory cost relative to methods that retrain the full pipeline. Baselines were retrained for each style using their official implementations, except ReconfigISP and NPF, which are partly reimplemented (their code/assets are incomplete), so their numbers are best-effort adaptations rather than exact reproductions (see Sec.~\myref{sec:evaluation_details}{\textcolor{section}{L}}). Nevertheless, an exact reproduction would still be bounded by their design limitations, as discussed in Sec.~\ref{sec:intro}. For fair comparison, baselines with an explicit denoising stage were retrained on the same noisy/denoised raw pairs we used. Those without one received the noisy raws directly as input, matching their original training protocols.

\subsection{Quantitative Results}
\label{sec:quantitative-results}

We report PSNR, SSIM~\cite{ssim}, LPIPS~\cite{LPIPS}, and $\Delta E_{2000}$~\cite{delta-e}. Table~\ref{tab:s24} summarizes results on the default S24 style with parameter counts for our three denoising variants (lite, base, large), with/without the detail-enhancement network. Our large variant achieves the best PSNR, SSIM, LPIPS, and $\Delta E_{2000}$. The closest competitor, LiteISP~\cite{lite-isp}, uses 9~M parameters and yields $\sim$2~dB lower PSNR. Further ablation experiments on denoising, guided upsampling, and photofinishing loss design are detailed in Sec.~\myref{supp-sec:ablation-studies}{\textcolor{section}{K.1}}.

Table~\ref{tab:s24-psnr-summary} reports PSNR across the five artistic styles (Styles~\#1--5, full metrics in Sec.~\myref{sec:supp-additional-results}{\textcolor{section}{K.2}}) along with the total parameter count required to support all five styles. Because our denoiser is shared, the per-style cost is minimal---a key benefit of our modular design---and our method achieves state-of-the-art PSNR across styles at moderate total parameter count. The optional 3D LuT $\mathbf{L}_{\texttt{RGB}}$ consistently improves artistic styles but can slightly degrade some metrics on the default style (whose target is natural color rendering rather than aggressive stylization). We also include cmKAN~\cite{cmkan} as a post-hoc rendered Style~\#0$\to$Style~\#1$\sim$\#5 color transfer baseline (`PP (cmKAN)'), contrasting direct per-style rendering against post-hoc style transfer.

On MIT-Adobe FiveK~\cite{Adobe5K}, our method remains competitive while preserving modular control, ranking first in SSIM and second in PSNR, trailing LiteISP by only 0.16~dB, with substantially fewer parameters (Table~\ref{tab:adobe5k-main}). We attribute this gap to our 50K enhancement module, compared with black-box baselines of 7--21M parameters: it alone must cover the texture and detail of FiveK's 30+ cameras, and on Zurich the smartphone-to-DSLR detail gap is larger still and the training pairs are not perfectly aligned. We trail by 0.9~dB there (Table~\myref{tab:zurich}{\textcolor{section}{8}} in supp) but lead Exposure by 1.9 and 1.4~dB. See Sec.~\myref{sec:supp-additional-results}{\textcolor{section}{K.2}} for details.

\begin{table}[t]
  \caption{Results on the S24 test set style~\#0. Our method is evaluated with 3 denoising model capacities (lite, base, and large), w/ and w/o the enhancement network. The best results are highlighted in \textbf{\colorbox{best}{yellow}}. Methods are color-coded by category: \colorbox{catblackbox}{single-stage black-box}, \colorbox{catmulti}{coarse multi-stage}, and \colorbox{catfine}{fine multi-stage}. Our methods achieve state-of-the-art (SOTA) performance and are efficient ($\sim$1.4~sec for a 12MP image with the large model on an NVIDIA RTX 4080 SUPER). \label{tab:s24}}
  \centering
  \resizebox{\linewidth}{!}{
    \begin{tabular}{lccccr}
      \toprule
      \textbf{Method} & \textbf{PSNR}\,$\uparrow$ & \textbf{SSIM}\,$\uparrow$ & \textbf{LPIPS}\,$\downarrow$ & $\Delta$\textbf{E 2000}\,$\downarrow$ & \textbf{\# params (K)} \\
      \midrule
      \cellcolor{catblackbox}ISPDiffuser~\cite{ispdiffuser}       & 24.03 & 0.881 & 0.159 & 5.918  & 20,939 \\
      \cellcolor{catblackbox}PyNet~\cite{zurich}                  & 23.80 & 0.869 & 0.100 & 6.277  & 47,548 \\
      \cellcolor{catblackbox}Invertible-ISP~\cite{invertible}     & 22.87 & 0.820 & 0.147 & 7.374  & 1,414  \\
      \cellcolor{catblackbox}LAN~\cite{lan}                       & 23.11 & 0.812 & 0.116 & 6.765  & 47     \\
      \cellcolor{catblackbox}MicroISP~\cite{microisp}             & 20.55 & 0.775 & 0.180 & 11.012 & 14     \\
      \cellcolor{catblackbox}LiteISP~\cite{lite-isp}              & 25.49 & 0.897 & 0.074 & 5.521  & 9,094  \\
      \cellcolor{catblackbox}FourierISP~\cite{fourier}            & 24.50 & 0.913 & 0.096 & 5.928  & 7,590  \\
      \cellcolor{catmulti}CIE-XYZ Net~\cite{afifi2021cie}         & 23.32 & 0.860 & 0.124 & 7.024  & 1,349  \\
      \cellcolor{catmulti}Deep-FlexISP~\cite{flexisp}                  & 24.01 & 0.886 & 0.110 & 6.938  & 25,705 \\
      \cellcolor{catmulti}ParamISP~\cite{paramisp}                & 24.32 & 0.841 & 0.115 & 6.135  & 1,420  \\
      \cellcolor{catfine}Exposure~\cite{hu2018exposure}                                  & 20.03 & 0.791 & 0.153 & 9.305  & 6,124  \\
      \cellcolor{catfine}ReconfigISP~\cite{yu2021reconfigisp}\textsuperscript{\dag}      & 20.07 & 0.808 & 0.182 & 10.027 & -      \\
      \cellcolor{catmodular}NPF~\cite{neural_photo_finishing}\textsuperscript{\dag}      & 21.27 & 0.796 & 0.164 & 8.629  & -      \\
      \midrule
      \cellcolor{catmodular}Ours (lite,$\texttt{ }$ w/o enhancement) & 26.36 & 0.878 & 0.071 & 4.413 & 452   \\
      \cellcolor{catmodular}Ours (base,\hspace{1.5mm}w/o enhancement)& 26.48 & 0.883 & 0.065 & 4.282 & 1,140 \\
      \cellcolor{catmodular}Ours (large, w/o enhancement)           & 26.51 & 0.884 & 0.064 & 4.253 & 3,842 \\
      \addlinespace
      \cellcolor{catmodular}Ours (lite,$\texttt{ }$ w/$\texttt{ }$ enhancement)  & 27.37 & 0.916 & 0.060 & 4.059 & 503   \\
      \cellcolor{catmodular}Ours (base,\hspace{1.5mm}w/$\texttt{ }$ enhancement) & 27.52 & 0.922 & 0.055 & 3.938 & 1,191 \\
      \cellcolor{catmodular}Ours (large, w/$\texttt{ }$ enhancement)            & \textbf{\cellcolor{best}27.57} & \textbf{\cellcolor{best}0.923} & \textbf{\cellcolor{best}0.054} & \textbf{\cellcolor{best}3.913} & 3,892 \\
      \bottomrule
  \end{tabular}}
\end{table}

\begin{table}[t]
  \centering
  \caption{Results across the S24 dataset artistic styles (Styles~\#1--5).
  We report the avg PSNR, along with the total number of parameters (K) required to support all five styles. The best results are highlighted in \textbf{\colorbox{best}{yellow}}. Methods are color-coded by category: \colorbox{catblackbox}{single-stage black-box}, \colorbox{catmulti}{coarse multi-stage}, and \colorbox{catfine}{fine multi-stage}.}
  \label{tab:s24-psnr-summary}
  \resizebox{\linewidth}{!}{
    \begin{tabular}{lcccccr}
      \toprule
      \multirow{2}{*}{\textbf{Method}} &
      \multirow{2}{*}{\textbf{S \#1}} &
      \multirow{2}{*}{\textbf{S \#2}} &
      \multirow{2}{*}{\textbf{S \#3}} &
      \multirow{2}{*}{\textbf{S \#4}} &
      \multirow{2}{*}{\textbf{S \#5}} &
      \textbf{\# params} \\
      & & & & & & \textbf{(K, 5 styles)} \\
      \midrule
      \cellcolor{catblackbox}ISPDiffuser~\cite{ispdiffuser}                              & 25.60 & 27.30 & 25.02 & 25.93 & 26.83 & 104,694 \\
      \cellcolor{catblackbox}PyNet~\cite{zurich}                                         & 24.36 & 25.94 & 24.70 & 24.34 & 26.32 & 237,741 \\
      \cellcolor{catblackbox}Invertible-ISP~\cite{invertible}                            & 23.48 & 26.35 & 23.84 & 23.33 & 24.90 & 7,069   \\
      \cellcolor{catblackbox}LAN~\cite{lan}                                              & 22.98 & 23.74 & 23.47 & 22.80 & 25.38 & 234     \\
      \cellcolor{catblackbox}MicroISP~\cite{microisp}                                    & 20.30 & 23.66 & 21.45 & 20.34 & 22.68 & 68      \\
      \cellcolor{catblackbox}LiteISP~\cite{lite-isp}                                     & 26.66 & 28.33 & 26.31 & 25.04 & 28.07 & 45,470  \\
      \cellcolor{catblackbox}FourierISP~\cite{fourier}                                   & 25.19 & 28.03 & 25.38 & 24.74 & 27.41 & 37,949  \\
      \cellcolor{catmulti}CIE-XYZ Net~\cite{afifi2021cie}                                & 22.40 & 24.05 & 22.00 & 22.26 & 24.67 & 6,744   \\
      \cellcolor{catmulti}Deep-FlexISP~\cite{flexisp}                                         & 24.86 & 27.47 & 25.23 & 23.96 & 24.65 & 128,525 \\
      \cellcolor{catmulti}ParamISP~\cite{paramisp}                                       & 24.97 & 27.11 & 24.77 & 24.18 & 25.43 & 7,100   \\
      \cellcolor{catfine}Exposure~\cite{hu2018exposure}                                  & 19.77 & 22.01 & 19.69 & 17.34 & 22.52 & 30,618  \\
      \cellcolor{catfine}ReconfigISP~\cite{yu2021reconfigisp}\textsuperscript{\dag}      & 19.56 & 23.46 & 19.69 & 19.04 & 22.60 & -       \\
      \cellcolor{catmodular}NPF~\cite{neural_photo_finishing}\textsuperscript{\dag}      & 20.43 & 24.54 & 21.18 & 20.59 & 22.14 & -       \\
      PP (cmKAN)~\cite{cmkan}                                                            & 20.93 & 23.04 & 21.85 & 20.91 & 21.30 & 385     \\
      \midrule
      \cellcolor{catmodular}Ours (lite,$\texttt{ }$ w/o enh.)                            & 25.16 & 28.09 & 25.66 & 25.47 & 27.08 & 1,281 \\
      \cellcolor{catmodular}Ours (base,\hspace{1.5mm}w/o enh.)                           & 25.28 & 28.19 & 25.73 & 25.55 & 27.19 & 1,969 \\
      \cellcolor{catmodular}Ours (large, w/o enh.)                                       & 25.31 & 28.22 & 25.75 & 25.58 & 27.23 & 4,670 \\
      \addlinespace
      \cellcolor{catmodular}Ours (lite,$\texttt{ }$ w/o enh., w/ 3D LuT)                 & 26.39 & 29.21 & 26.69 & 26.35 & 27.93 & 1,348 \\
      \cellcolor{catmodular}Ours (base,\hspace{1.5mm}w/o enh., w/ 3D LuT)                & 26.52 & 29.29 & 26.79 & 26.47 & 28.13 & 2,035 \\
      \cellcolor{catmodular}Ours (large, w/o enh., w/ 3D LuT)                            & 26.56 & \textbf{\cellcolor{best}{29.31}} & \textbf{\cellcolor{best}{26.83}} & 26.51 & 28.19 & 4,737 \\
      \addlinespace
      \cellcolor{catmodular}Ours (lite,$\texttt{ }$ w/\hspace{2mm} enh., w/ 3D LuT)      & 26.56 & 28.92 & 26.78 & 26.66 & 28.73 & 1,550 \\
      \cellcolor{catmodular}Ours (base,\hspace{1.5mm}w/\hspace{2mm} enh., w/ 3D LuT)     & 26.71 & 28.99 & 26.78 & 26.79 & 28.95 & 2,238 \\
      \cellcolor{catmodular}Ours (large, w/\hspace{2mm} enh., w/ 3D LuT)                 & \textbf{\cellcolor{best}{26.75}} & 29.01 & \textbf{\cellcolor{best}{26.83}} & \textbf{\cellcolor{best}{26.84}} & \textbf{\cellcolor{best}{29.03}} & 4,940 \\
      \bottomrule
  \end{tabular}}
\end{table}

\begin{table}[t]
  \centering
  \caption{Results on MIT-Adobe FiveK~\cite{Adobe5K}, with the strongest baselines per category. Best in \textbf{\colorbox{best}{yellow}}. For the full table, see Table~\mbox{\myref{tab:adobe5k}{\textcolor{section}{24}}} in the supplement.}
  \label{tab:adobe5k-main}
  \resizebox{\linewidth}{!}{
    \begin{tabular}{lccccr}
      \toprule
      \textbf{Method} & \textbf{PSNR}\,$\uparrow$ & \textbf{SSIM}\,$\uparrow$ & \textbf{LPIPS}\,$\downarrow$ & $\Delta$\textbf{E 2000}\,$\downarrow$ & \textbf{\# params (K)} \\
      \midrule
      \cellcolor{catblackbox}LiteISP~\cite{lite-isp}       & \textbf{\cellcolor{best}{21.45}} & 0.740 & \textbf{\cellcolor{best}{0.169}} & \textbf{\cellcolor{best}{9.687}} & 9,094 \\
      \cellcolor{catblackbox}FourierISP~\cite{fourier}     & 20.29 & 0.737 & 0.208 & 10.582 & 7,590 \\
      \cellcolor{catmulti}ParamISP~\cite{paramisp}         & 21.17 & 0.735 & 0.184 & 9.729 & 1,420 \\
      \cellcolor{catmulti}CIE-XYZ Net~\cite{afifi2021cie}  & 20.25 & 0.715 & 0.189 & 10.352 & 1,349 \\
      \cellcolor{catfine}Exposure~\cite{hu2018exposure}    & 19.42 & 0.687 & 0.251 & 11.660 & 6,124 \\
      \midrule
      \cellcolor{catmodular}Ours (lite,$\texttt{ }$ w/ enhancement)  & 21.27 & 0.739 & 0.219 & 10.765 & 503 \\
      \cellcolor{catmodular}Ours (base, w/ enhancement)              & 21.28 & 0.740 & 0.221 & 10.762 & 1,191 \\
      \cellcolor{catmodular}Ours (large, w/ enhancement)             & 21.29 & \textbf{\cellcolor{best}{0.742}} & 0.220 & 10.743 & 3,892 \\
      \bottomrule
    \end{tabular}}
\end{table}

\subsection{Modularity Analysis}
\label{sec:module-swapping-main}

Do operators remain functionally aligned with their intended roles under joint training, or entangle? We address this through visual interventions and quantitative confirmations.
$f_{\texttt{gain}}$ (scalar multiply), $f_{\texttt{chroma}}$ (CbCr-only LuT), and $f_{\texttt{gamma}}$ (bounded monotone power) have disjoint interfaces by construction. $f_{\texttt{GTM}}$ and $f_{\texttt{LTM}}$ share the same monotonic curve $f_{\texttt{TM}}$ applied identically across R, G, B (Eqs.~\ref{eq:apply-gtm},~\ref{eq:apply-ltm}), so neither has a channel-specific knob for chroma manipulation. These interfaces enable role-consistent operator substitution \emph{without retraining}: substituting Style~\#1's chroma into the Style~\#0 pipeline (Fig.~\ref{fig:mix-styles_new}) transfers Style~\#1's palette while preserving Style~\#0's tone (e.g., the tree stays as dark as in Style~\#0 rather than brightening as in Style~\#1's full pipeline)---also demonstrating cross-camera generalization on an iPhone~13 input unseen during training.
The role-specific losses $\ell_{\texttt{TM}}$ and $\ell_{\texttt{luma}}$ keep GTM and LTM at distinct scales. This is visible both during 1) training: intermediate GTM/LTM outputs with and without these losses (Fig.~\myref{fig:ablation-loss-terms-1}{\textcolor{section}{43}} in supp.) 2) and at inference in the disable-LTM and disable-GTM panels of Fig.~\ref{fig:mix-styles_new}: dropping GTM removes global tone while local detail remains (e.g., the trees keep their texture); dropping LTM preserves global tone while local lifting is gone (e.g., the trees stay dark). These observations are supported quantitatively: each role-specific loss contributes measurably to final quality (Table ~\myref{tab:ablation_losses-1}{\textcolor{section}{14}}, Table ~\myref{tab:ablation_losses-2}{\textcolor{section}{15}}). Each operator's removal causes a PSNR/SSIM drop consistent with its role (Table~\myref{tab:ablation-photofinishing-design}{\textcolor{section}{17}}), and replacing our role-constrained LTM with an unconstrained HDRNet degrades PSNR by $\sim$2~dB at higher parameter count (Table~\ref{tab:ablation-hdrnet}).

\begingroup
\renewcommand{\thefootnote}{\dag}
\footnotetext[0]{ReconfigISP and NPF are partly reimplemented. See Sec.~\myref{sec:evaluation_details}{\textcolor{section}{L}} in the supplementary material for details.}
\endgroup

\begin{table}[t]
\centering
\caption{Alternative designs using HDRNet~\cite{hdrnet} as a replacement for our LTM operator or the entire photofinishing module, evaluated on S24 Style~\#0 (LsRGB to sRGB). Best results are highlighted in \textbf{\colorbox{best}{yellow}}.}
\label{tab:ablation-hdrnet}
\resizebox{\linewidth}{!}{
\begin{tabular}{lccr}
  \toprule
  \textbf{Method} & \textbf{PSNR}\,$\uparrow$ & \textbf{SSIM}\,$\uparrow$ & \textbf{\# params} \\
  \midrule
  LTM replaced with HDRNet            & 25.35 & 0.920 & 570,462 \\
  Photofinishing replaced with HDRNet & 24.87 & 0.911 & 483,453 \\
  \midrule
  Ours                                & \textbf{\cellcolor{best}{27.49}} & \textbf{\cellcolor{best}{0.939}} & 207,224 \\
  \bottomrule
\end{tabular}}
\end{table}

Although our LTM network and HDRNet~\cite{hdrnet} both predict locally adaptive tone-mapping operators, their formulations and architectures differ substantially (Sec.~\myref{sec:supp-ps-ltm-net}{\textcolor{section}{C.5}} in supp). We test two replacements, both jointly trained under the same configuration: substituting only our LTM network/operator with HDRNet, and replacing the entire photofinishing module with a single HDRNet. As shown in Table~\ref{tab:ablation-hdrnet}, our design achieves higher PSNR/SSIM with fewer parameters. See Fig.~\ref{fig:ours-vs-hdr-net} for qualitative results.

Retraining LiteISP and ISPDiffuser on the same denoised pseudo-ground-truth raw isolates the denoiser (Table~\ref{tab:matched-denoiser}). Our photofinishing core alone beats LiteISP on Style~\#0 with 43$\times$ fewer parameters, and the 50K enhancement module leads on all but one style.

\begin{table}[t]
\centering
\caption{Matched-denoiser comparison on S24 (PSNR\,$\uparrow$). All rows use the same denoised pseudo-ground-truth raw for training and evaluation, isolating photofinishing from denoising. Best results are highlighted in \textbf{\colorbox{best}{yellow}}.}
\label{tab:matched-denoiser}
\resizebox{\linewidth}{!}{
\begin{tabular}{lcccccc}
  \toprule
  \textbf{Method} & \textbf{S \#0} & \textbf{S \#1} & \textbf{S \#2} & \textbf{S \#3} & \textbf{S \#4} & \textbf{S \#5} \\
  \midrule
  LiteISP~\cite{lite-isp}        & 26.18 & 27.00 & 28.23 & 26.44 & 26.04 & \textbf{\cellcolor{best}{28.63}} \\
  ISPDiffuser~\cite{ispdiffuser} & 24.30 & 23.69 & 26.50 & 24.25 & 23.77 & 25.55 \\
  Ours (w/o 3D LuT)              & \textbf{\cellcolor{best}{27.66}} & \textbf{\cellcolor{best}{27.55}} & \textbf{\cellcolor{best}{29.86}} & \textbf{\cellcolor{best}{27.55}} & \textbf{\cellcolor{best}{27.14}} & 28.51 \\
  Ours (w/o enh., w/o 3D LuT)    & 26.55 & 25.34 & 28.23 & 25.79 & 25.61 & 27.30 \\
  \bottomrule
\end{tabular}}
\end{table}

We retrain the photofinishing module under alternative orders, holding loss, data, and recipe fixed. Ours is best on S24 Style~\#0 and FiveK, second on Zurich, where gamma-first leads by 0.28~dB (Table~\ref{tab:order-ablation}). Reordering costs at most 1.85~dB, well below the 5.3--6.5~dB gap to fine multi-stage baselines on Style~\#0 (Table~\ref{tab:s24}), so the gap comes mainly from our operator designs and training recipe.

\begin{table}[t]
\centering
\caption{Effect of operator order (PSNR\,$\uparrow$). Order is the only variable: all rows share the loss, data, and training recipe, and use denoised pseudo-ground-truth raw without the denoiser, 3D LuT, or enhancement network. Best in \textbf{\colorbox{best}{yellow}}.}
\label{tab:order-ablation}
\resizebox{\linewidth}{!}{
\begin{tabular}{lccc}
  \toprule
  \textbf{Operator order} & \textbf{S24 \#0} & \textbf{FiveK} & \textbf{Zurich} \\
  \midrule
  Gain-GTM-LTM-Chroma-Gamma (ours)                                 & \textbf{\cellcolor{best}{26.55}} & \textbf{\cellcolor{best}{22.47}} & 23.53 \\
  Chroma-GTM-LTM-Gain-Gamma                                        & 25.35 & 21.78 & 23.26 \\
  GTM-LTM-Gamma-Gain-Chroma                                        & 25.58 & 21.92 & 22.54 \\
  Gain-LTM-GTM-Chroma-Gamma                                        & 25.68 & 21.68 & 23.34 \\
  Gain-Chroma-GTM-LTM-Gamma                                        & 24.78 & 21.93 & 23.45 \\
  Gamma-Gain-Chroma-GTM-LTM                                        & 24.70 & 22.30 & \textbf{\cellcolor{best}{23.81}} \\
  \bottomrule
\end{tabular}}
\end{table}

\subsection{Qualitative Results}
\label{sec:qualitative-results}

\begin{figure*}[p]
\centering
\includegraphics[width=\textwidth]{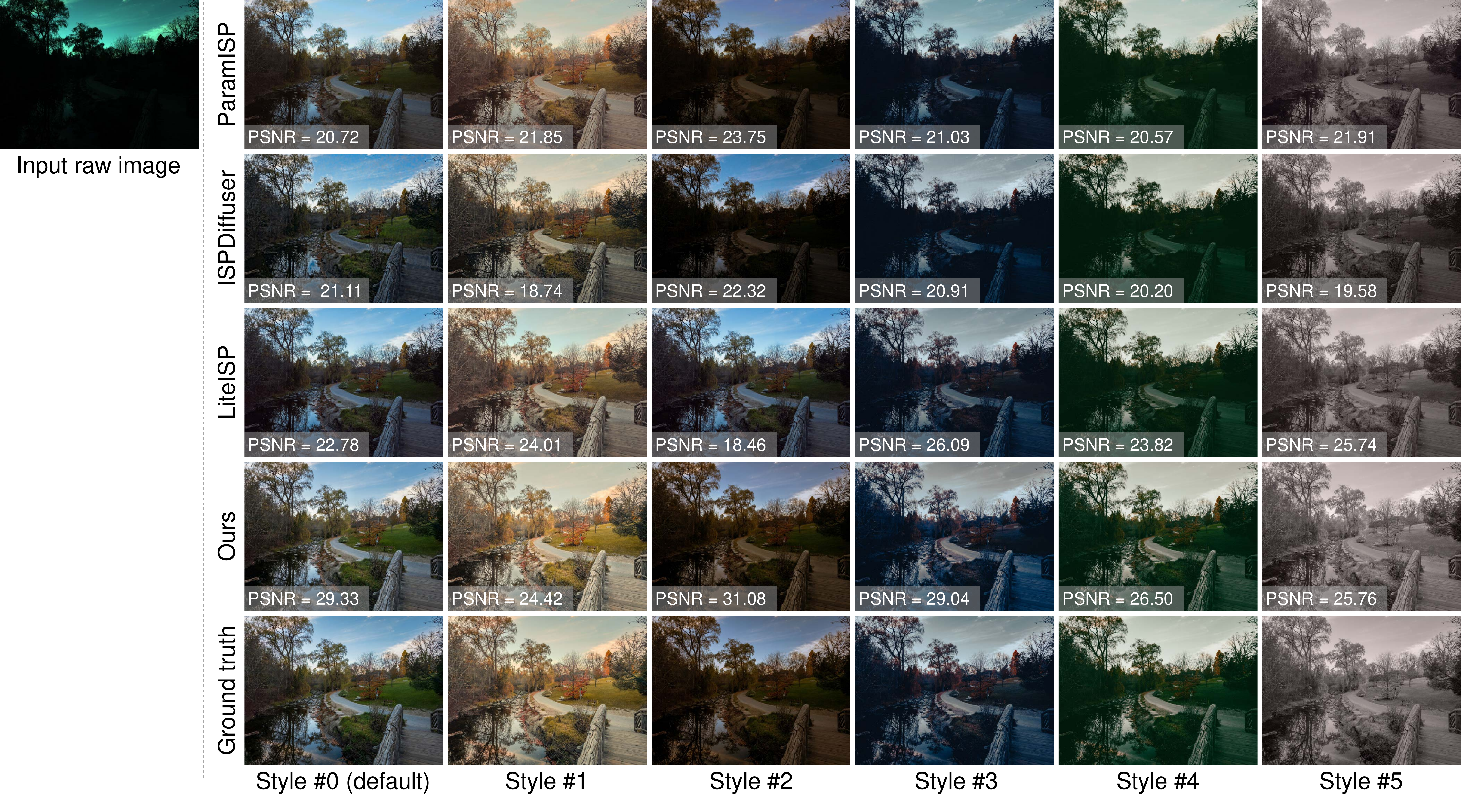}
\caption{
\textcolor{red}{[Best viewed zoomed in]} Qualitative comparison against ISPDiffuser~\cite{ispdiffuser}, LiteISP~\cite{lite-isp}, and ParamISP~\cite{paramisp} on an S24 test image~\cite{s24}, rendered in the default style (Style~\#0) and five artistic styles (Styles~\#1--5). All models are trained per-style on the same S24 train set. Our method consistently achieves higher global/local tone and color fidelity to the ground truth across styles.
}
\label{fig:qualtative-results}
\includegraphics[width=\linewidth]{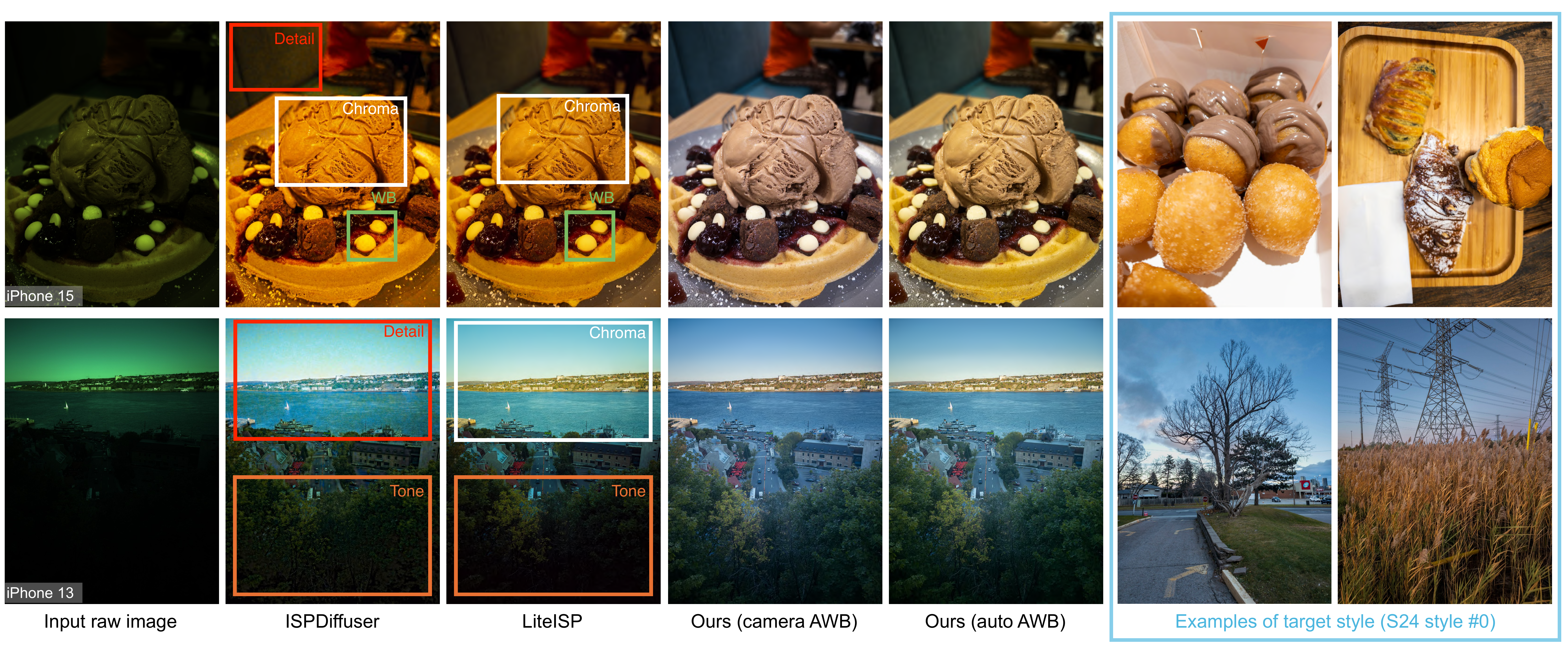}
\caption{
Qualitative comparison on unseen cameras (iPhone~13 and iPhone~15) between our method and LiteISP~\cite{lite-isp} and ISPDiffuser~\cite{ispdiffuser}, all trained on the S24 default style (Style~\#0, see example images in the right column). Our method uses the generic denoiser, and none of the training data included iPhone captures. Artifacted regions are marked with colored boxes. Our renderings track the target style better (e.g., sky color and overall tone in the outdoor scene). ISPDiffuser and LiteISP---being single-stage end-to-end networks without explicit white-balance (WB) control---produce an inaccurate WB on the indoor scene, whereas our renderings are correctly balanced. Because our pipeline includes a standalone color-correction module, the WB can be set or swapped at will: we show results using the camera's AWB and learned AWB models~\cite{afifi2021c5, zhao2025learning} (see Sec.~\myref{sec:awb}{\textcolor{section}{I.3}} for details).
}
\label{fig:cross-camera-comparison-main}
\end{figure*}

Fig.~\ref{fig:qualtative-results} compares our method against recent single-stage black-box~\cite{ispdiffuser, lite-isp} and coarse multi-stage neural ISPs~\cite{paramisp} on an S24 test image, rendered in the default style (\#0) and five artistic styles. Our method consistently delivers higher visual quality across all styles. Additional examples are in Sec.~\myref{sec:supp-additional-results}{\textcolor{section}{K.2}}.

\begin{figure*}[p]
\centering
\includegraphics[width=\linewidth]{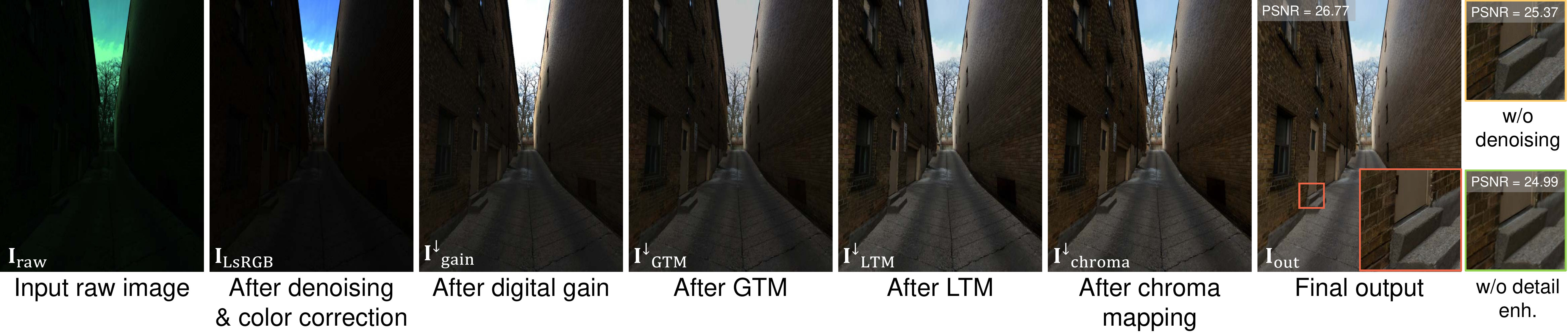}
\caption{Intermediate-stage outputs of our pipeline on an S24 test image.}
\label{fig:intermediate-example-style0}
\includegraphics[width=\textwidth]{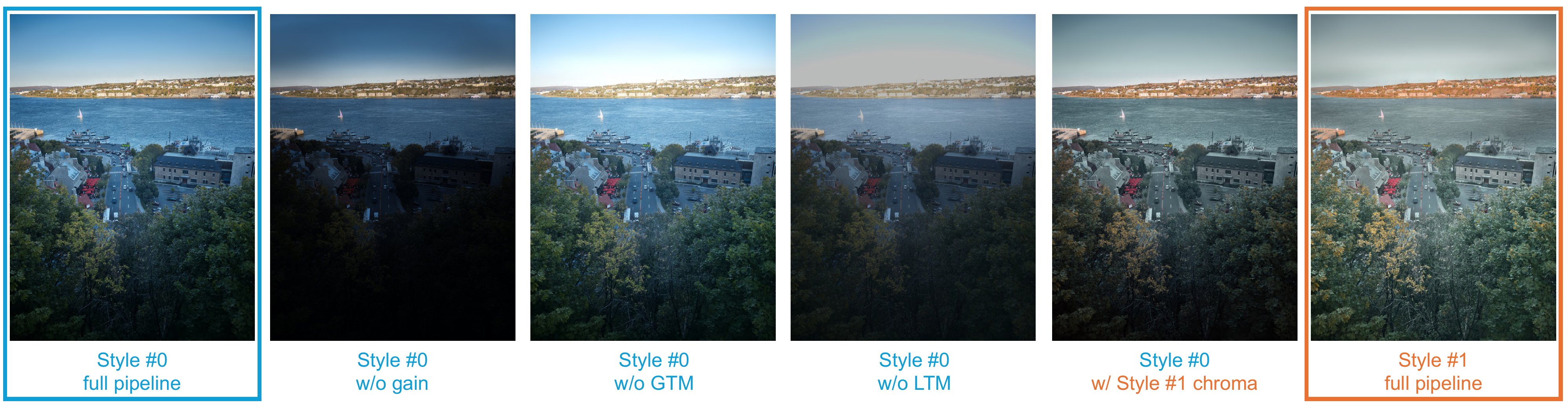}
\caption{Operator disabling and cross-style swapping in the photofinishing module. From the Style~\#0 full pipeline, we drop gain, GTM, LTM, or swap in Style~\#1's chroma. The rightmost column shows the Style~\#1 full pipeline for reference. Each intervention produces a coherent, role-consistent change: dropping LTM (4th~column) leaves the trees very dark, as LTM is responsible for local shadow lifting; swapping in Style~\#1's chroma while keeping Style~\#0's LTM (5th~column) transfers Style~\#1's palette yet retains Style~\#0's darker tree, in contrast with the Style~\#1 full pipeline where the tree is markedly brighter. Input: iPhone~13 (unseen camera). More swapping examples: Fig.~\myref{fig:mix-styles}{\textcolor{section}{37}}.}
\label{fig:mix-styles_new}
\begin{minipage}{\columnwidth}
\centering
\includegraphics[width=0.79\linewidth]{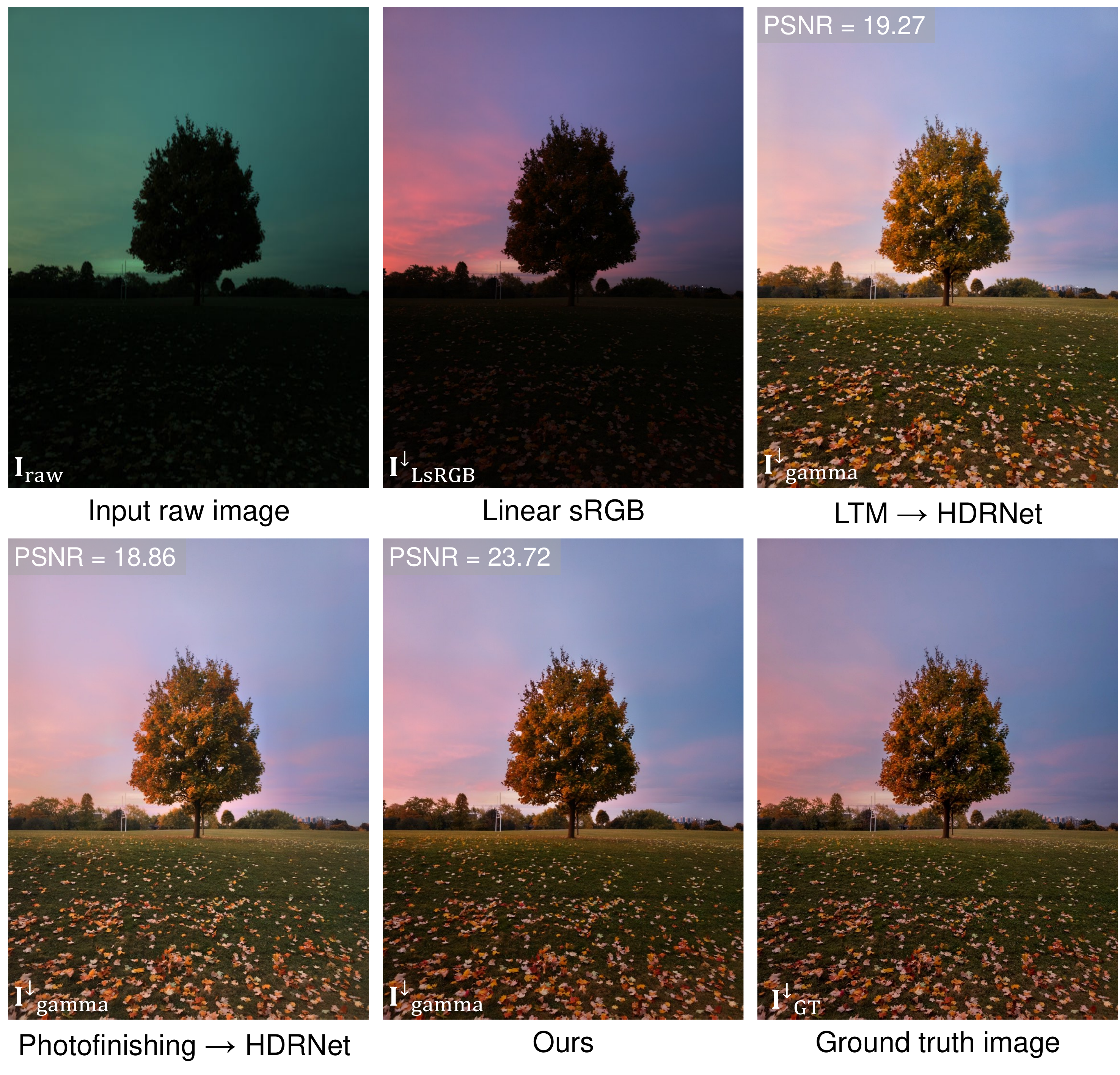}
\caption{Comparison against HDRNet~\cite{hdrnet} on an S24 test image, with two replacement configs jointly trained under the same setup: (1)~HDRNet in place of LTM, and (2)~HDRNet in place of the entire photofinishing module. Both HDRNet variants drift in tone and chroma, while ours tracks the ground truth with markedly higher PSNR. Multi-scale processing and refinement (Sec.~\myref{sec:artifacts}{\textcolor{section}{B.1}}) are applied within our LTM stage.}
\label{fig:ours-vs-hdr-net}
\end{minipage}
\hspace{2mm}
\begin{minipage}{\columnwidth}
\centering
\includegraphics[width=\linewidth]{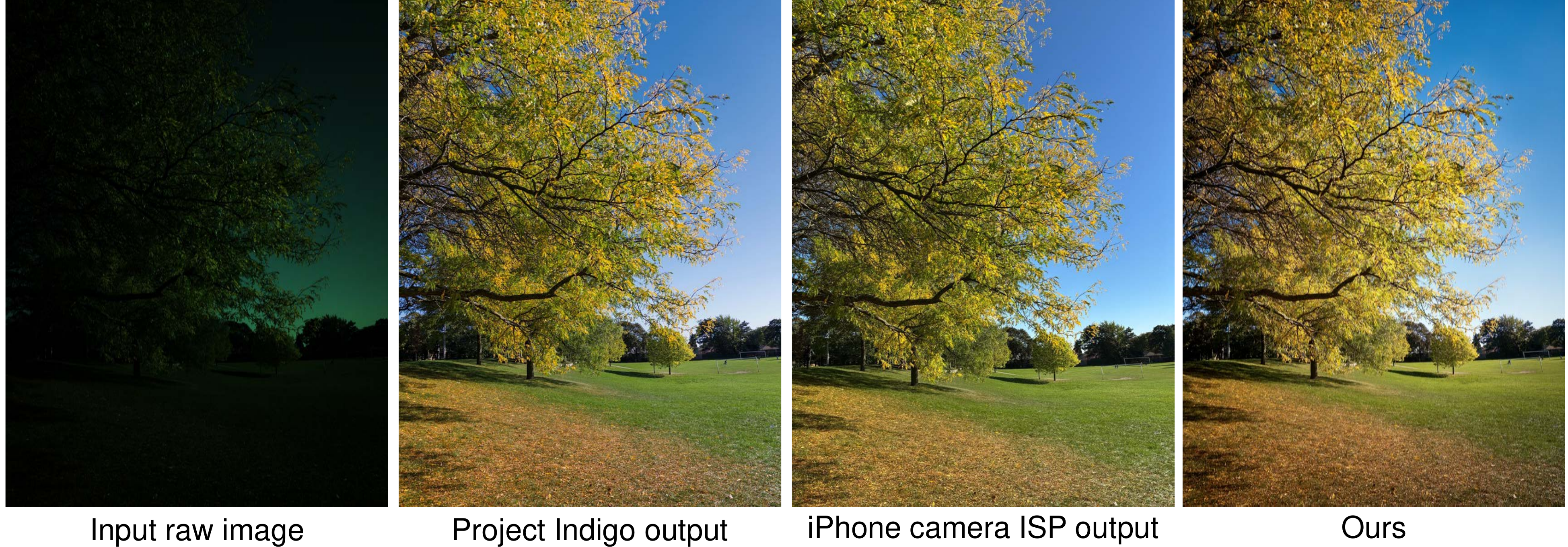}
\caption{
Comparison among Project~Indigo~\cite{adobe_indigo_2025}, the iPhone native camera ISP, and our method (using the generic denoiser and cross-camera auto white balance). The image was captured using the iPhone 13 Pro Max main camera.
}
\label{fig:ours_vs_indigo}
\includegraphics[clip,trim={51cm 70cm 0cm 0cm},width=\linewidth]{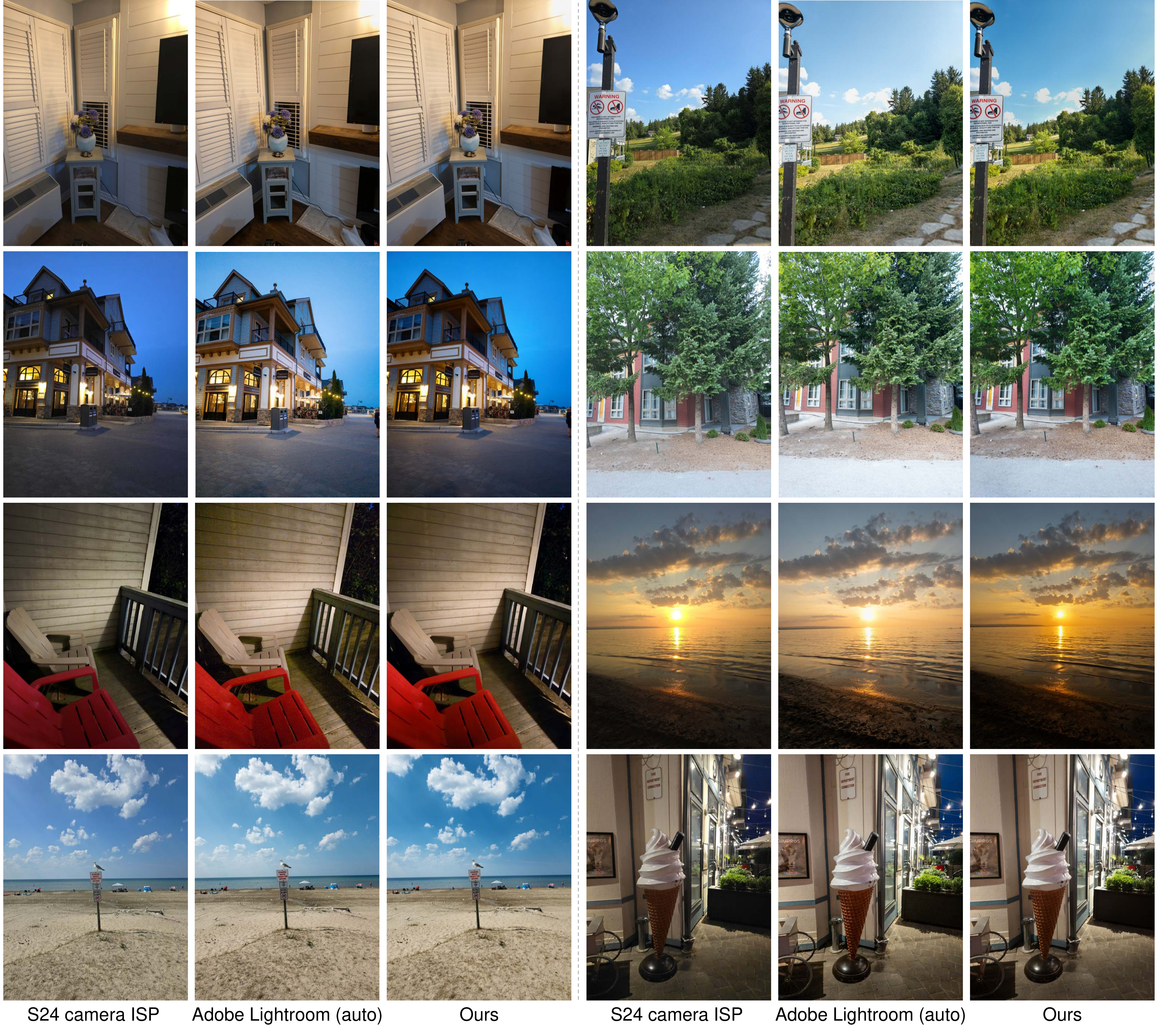}
\includegraphics[width=\linewidth]{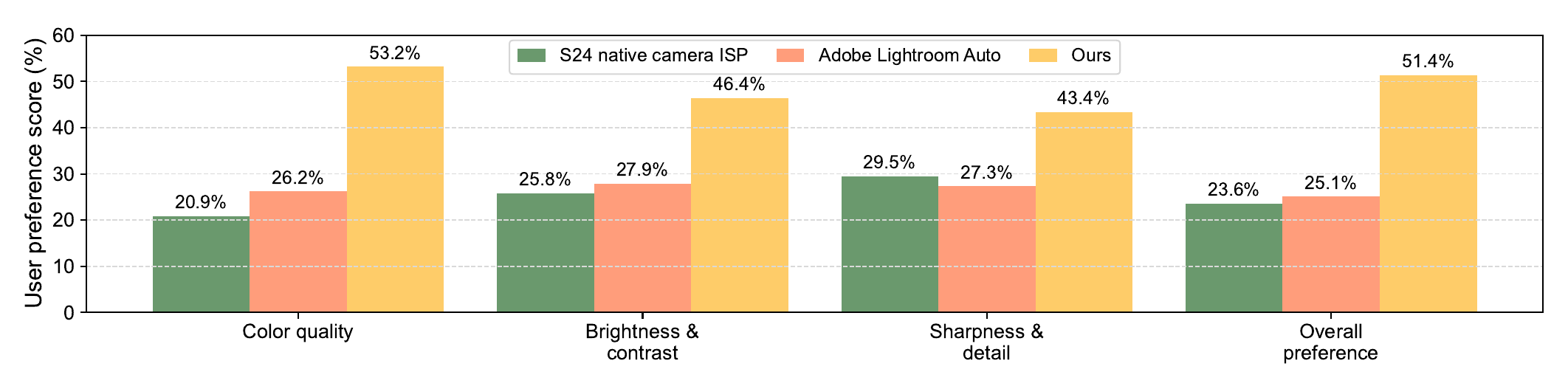}
\caption{User study examples and results. From left to right: Samsung S24 native ISP, Adobe Lightroom auto enhancement, and our method. During the study, the order was randomized for every user and scene.}
\label{fig:user-study-new}
\end{minipage}
\end{figure*}

\paragraph{Cross-Camera Generalization}
Using the generic denoiser and cross-camera AWB, we evaluate on iPhone captures unseen during training. No paired ground truth exists for another camera rendered in an S24 style, so this evaluation is qualitative. The one component that does have ground truth, the generic denoiser, is evaluated quantitatively on SIDD \cite{abdelhamed2018high} (Sec.~\myref{sec:cross-camera}{\textcolor{section}{H}}). Fig.~\ref{fig:cross-camera-comparison-main} compares against LiteISP~\cite{lite-isp} (9~M parameters) and ISPDiffuser~\cite{ispdiffuser} ($\sim$21~M), all trained on the same data. Our lite variant (0.5~M parameters) produces more visually pleasing and controllable results. Additional examples and comparison with commercial ISP systems like the native iPhone ISP and Project Indigo~\cite{adobe_indigo_2025} are in Fig.~\ref{fig:ours_vs_indigo}. More details and examples are shown in Sec.~\myref{sec:cross-camera}{\textcolor{section}{H}} and Sec. ~\myref{sec:awb}{\textcolor{section}{I.3}}.

\paragraph{User Study}
Using our Style~\#0 model, we compared against two commercial ISP systems: Samsung S24 native ISP and Adobe Lightroom (auto enhancement) on 45 scenes captured in S24 Pro mode (DNG, independent of the dataset used for training), spanning indoor, outdoor daylight, sunset, and low-light. Twenty participants performed a three-alternative forced-choice (3AFC) task: for each scene, they viewed the three renderings in randomized order and chose the preferred image under four criteria: `color quality', `brightness \& contrast', `sharpness \& detail', and `overall preference'. Chance level is therefore $33.3\%$. Our method was consistently preferred for all criteria (53.2\%, 46.4\%, 43.4\%, 51.4\%, i.e., $+13.9$--$27.0\%$ over the closest competitor, see Fig.~\ref{fig:user-study-new}), indicating that our pipeline produces renderings more appealing to raters than both commercial baselines on independently captured scenes. Sessions used a calibrated EIZO ColorEdge CG319X reference display under the viewing conditions of ITU-R BT.500-15~\cite{itu2023bt500}, and the study received ethics approval prior to data collection. See Sec.~\myref{sec:user_study}{\textcolor{section}{J}} for details.

\section{Conclusion and Discussion}
\label{sec:conclusion}

We presented a modular neural ISP for raw-to-sRGB rendering with fine-grained control across stages. Our three variants ($\sim$0.5--$3.9$M parameters) render a 12~MP image in about a second on one GPU, reaching state-of-the-art results on nearly all S24 picture styles and competitive results on MIT-Adobe FiveK (Sec.~\myref{sec:supp-additional-results}{\textcolor{section}{K.2}}) with far fewer parameters than the closest competitors. The interpretable modular design also eases debugging and tuning, scales to multiple picture styles, and extends to unseen cameras by swapping the camera-specific modules, as showcased in our photo-editing tool.

Despite these advantages, challenges remain. The fixed order is a role-preservation constraint rather than a performance optimum. We do not reach the best PSNR on FiveK or Zurich, which we attribute to our compact enhancement module. Apart from the generic denoiser, cross-camera rendering lacks paired ground truth and can only be assessed qualitatively. In practice, only a few artist-retouched raw–sRGB pairs may be available, making training difficult. Our modular design may open up promising possibilities for data-efficient adaptation by fine-tuning only a subset of style-specific operators.

\begin{acks}
We thank Luxi Zhao, Raghav Goyal, and Dhruv Verma for their help and useful discussions.
\end{acks}

 \clearpage

\bibliographystyle{ACM-Reference-Format}
\bibliography{main}

\newcommand{\suppteasercaption}{%
Intermediate outputs of our modular neural ISP, including denoising, color correction, digital gain, global tone mapping, local tone mapping, chroma mapping, gamma correction and detail enhancement. Different picture styles and edits are also shown (right). From top to bottom and left to right (for the styles and edits on the right), the first five images correspond to Styles~\#1--5 of the S24 dataset~\cite{s24}, while the remaining ones show results with additional edits and style adjustments. The first example was captured using an iPhone~15 main camera, the second using an iPhone 13 main camera, and the third using a Samsung~S24 main camera. Note that none of the training images were captured with, or include any synthetic data derived from, iPhone devices.
}

\ifarxiv
\twocolumn[{%
\begin{center}
    {\Large\bfseries Supplementary Material}\\[1em]
\end{center}
\vspace{1em}
\centering
\includegraphics[width=0.9\textwidth]{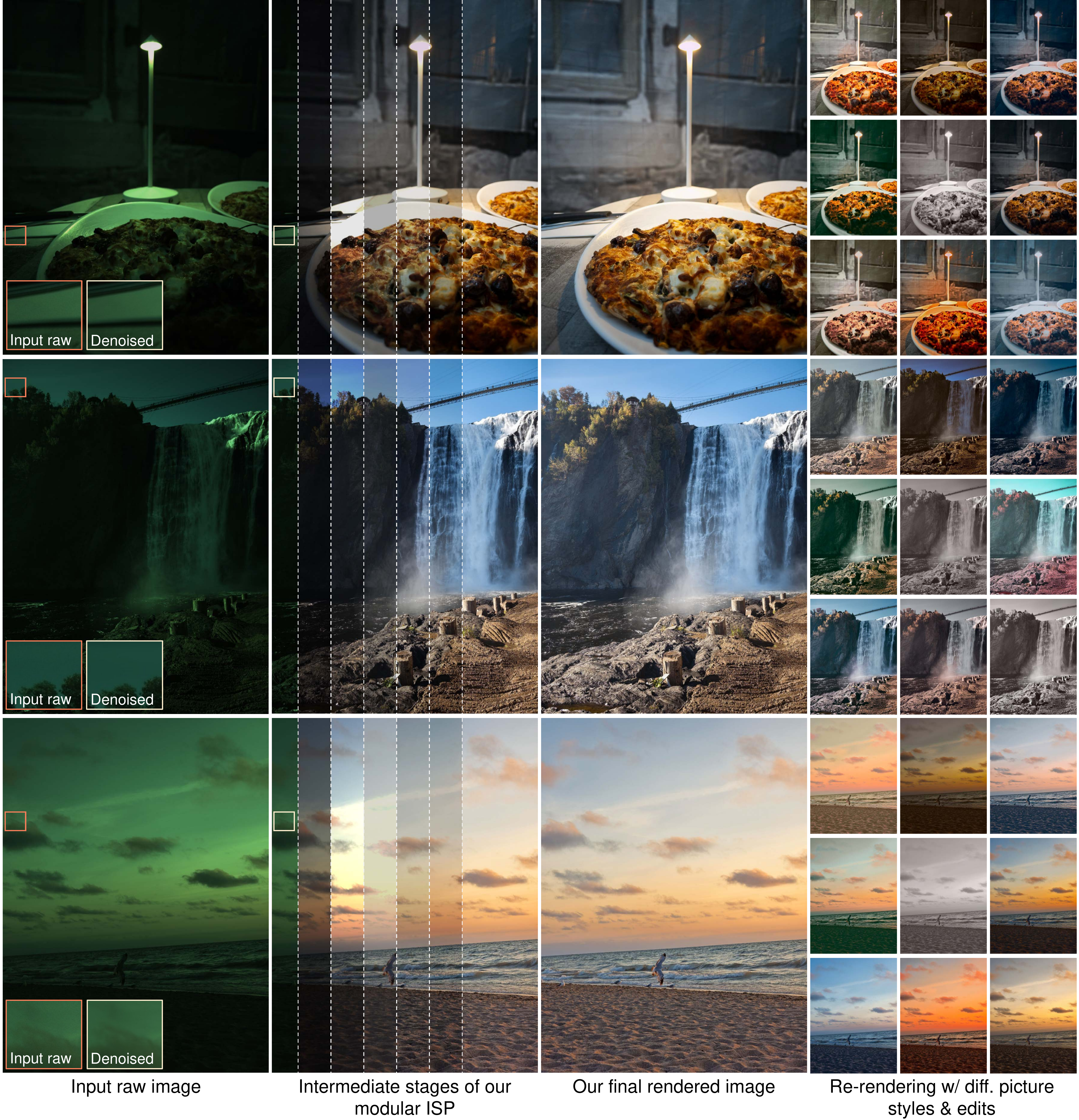}
\vspace{-2mm}
\captionof{figure}{\suppteasercaption}
\label{fig:teaser-supp}
\vspace{1em}
}]
\else
\begin{figure*}[!t]
    \centering
    \includegraphics[width=0.9\textwidth]{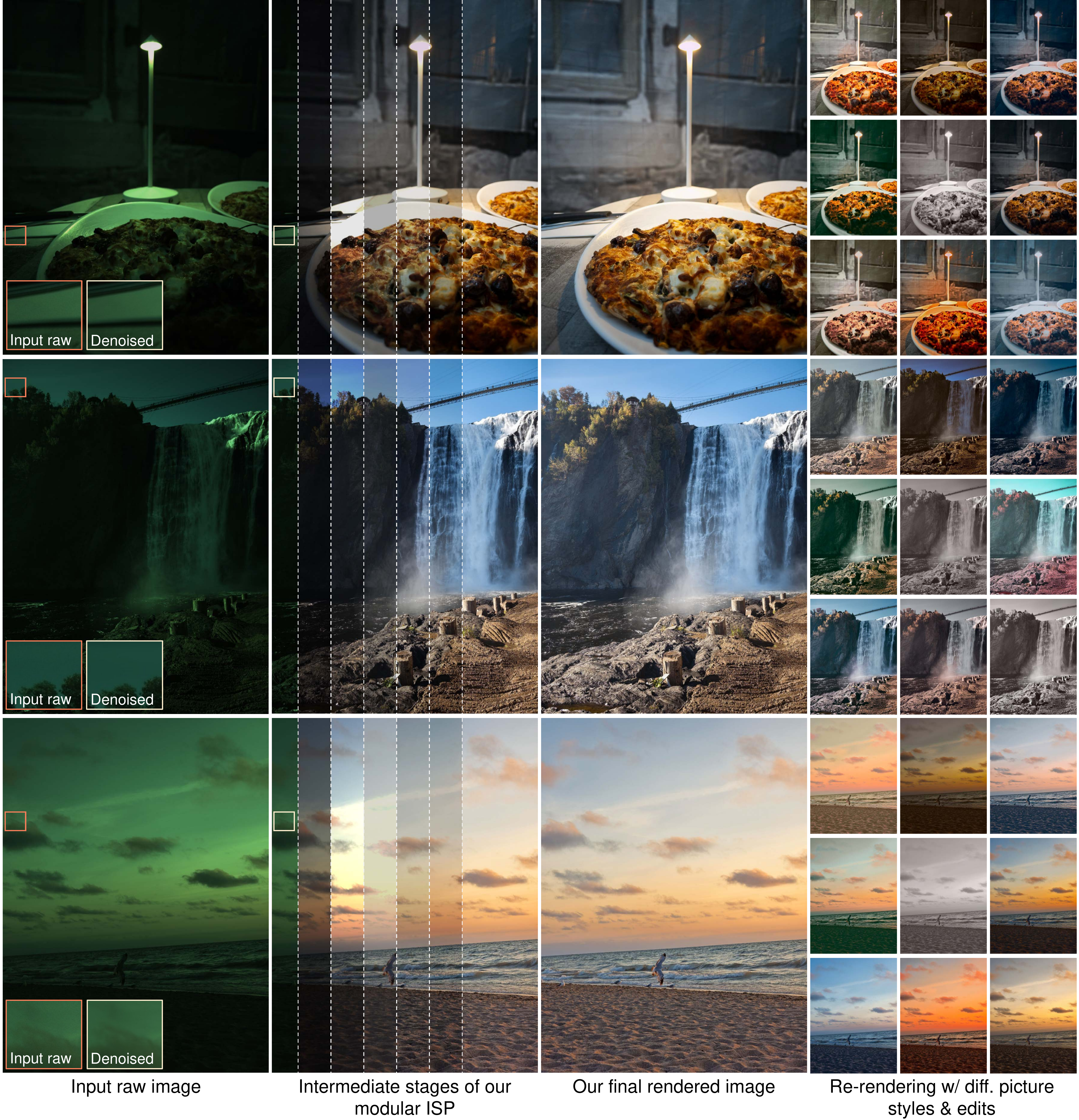}
    \vspace{-2mm}
\caption{\suppteasercaption}
\label{fig:teaser-supp}
\end{figure*}
\fi

In the main paper, we presented our modular neural image signal processing (ISP) framework, which enables accurate raw-to-sRGB rendering through a modular design that provides full control over the rendering process and supports different picture styles. This modularity allows the framework to be flexibly tuned to produce desired capture-time outputs or to function as an interactive photo editor (see Fig.~\ref{fig:teaser-supp}). We provide this supplementary material to further clarify how our method is developed, how it can be configured for capture-time deployment, and how it can alternatively be used as an interactive image-processing tool. Specifically, in Sec.~\ref{sec:accelerated_bilateral_solver}, we describe our GPU-accelerated iterative bilateral solver, which we use to mitigate halo artifacts that may occur in some corner cases. Handling challenging artifacts (with the help of the GPU-accelerated bilateral solver) and training with incomplete datasets are discussed in detail in Sec.~\ref{sec:limitations}.

We then elaborate on the design of the deep networks used in this work in Sec.~\ref{sec:network_design} and provide additional information about the differentiable histogram used in the chroma mapping network in Sec.~\ref{sec:hist}. Afterwards, we describe the guided upsampling regularization used in our implementation in Sec.~\ref{sec:supp-bgu}. Details of the photofinishing loss functions are provided in Sec.~\ref{sec:loss_funcs}, and the training details of our networks are presented in Sec.~\ref{sec:training_details}.

In Sec.~\ref{sec:cross-camera}, we discuss our efforts to improve the generalization of our method across cameras. Sec.~\ref{sec:gui} describes our graphical user interface tool, built on top of our method, which includes additional editing capabilities. In Sec.~\ref{sec:user_study}, we present further details of the user study conducted as part of our evaluation. Lastly, in Sec.~\ref{sec:ablation_studies}, we provide extensive ablation studies of the pipeline components, along with additional results and comparisons. Further evaluation details for both the main paper and this supplementary material are provided in Sec.~\ref{sec:evaluation_details}.

\appendix

\section{GPU-Accelerated Iterative Bilateral Solver}
\label{sec:accelerated_bilateral_solver}

To mitigate residual artifacts that occasionally appear in corner cases of our local tone-mapping (LTM) stage, we add an optional guided refinement step. Edge-aware smoothing is a natural choice for this purpose. The Fast Bilateral Solver (FBS)~\cite{barron2016fast} is particularly effective: it minimizes a quadratic objective with bilateral affinities, thereby enforcing edge adherence while propagating low-frequency information. However, the original solver is not GPU-friendly. FBS constructs a sparse bilateral grid and solves a large symmetric positive definite (SPD) system using preconditioned conjugate gradient (PCG). This involves repeated gather-scatter operations between pixel and grid spaces, dominated by memory access rather than arithmetic, and converges slowly under the Jacobi preconditioner. These factors make direct GPU acceleration of FBS inefficient.

\subsection{Quadratic Objective in Image Space}
We propose a lightweight image-space variant that preserves the FBS energy while avoiding the bilateral grid. Given an initial tensor $\mathbf{M}$ and a guidance image $\mathbf{\mathcal{Z}}$, we solve for a refined output $\mathbf{Y}$ by minimizing the following energy:
\begin{equation}
\min_{\mathbf{Y}} \;\;
\lambda \sum_{p}\lVert \mathbf{Y}_p - \mathbf{M}_p \rVert_2^2
+ \sum_{p}\sum_{q\in\mathcal{N}_k(p)} 
\mathbf{W}_{pq}(\mathbf{\mathcal{Z}})\,\lVert \mathbf{Y}_p - \mathbf{Y}_q \rVert_2^2,
\label{eq:bilateral_objective}
\end{equation}
where $\mathcal{N}_k(p)$ denotes a local $k{\times}k$ neighborhood around pixel $p$, and $\mathbf{W}_{pq}(\mathbf{\mathcal{Z}})$ are bilateral affinities defined based on the guidance image $\mathbf{\mathcal{Z}}$:
\begin{align}
\mathbf{W}_{pq}(\mathcal{Z}) =
\exp\!\Big(
-\frac{\lVert p - q \rVert_2^2}{2\sigma_s^2}
-\frac{(\mathcal{Z}_p - \mathcal{Z}_q)^2}{2\sigma_r^2}
\Big),
\end{align}
with $(\sigma_s,\sigma_r)$ controlling the spatial and range scales.  
The guidance image, $\mathbf{\mathcal{Z}}$, is computed as the luminance of the input RGB image using a weighted combination of the red, green, and blue channels, with respective weights of 0.2989, 0.5870, and 0.1140.

\subsection{Iterative Solver}
Instead of solving Eq.~\ref{eq:bilateral_objective} with PCG, we pre-compute the bilateral weights $\mathbf{W}_{pq}(\mathbf{\mathcal{Z}})$ once (per image) and apply a fixed number of successive over-relaxation (SOR) updates~\cite{SOR}. Initializing $\mathbf{Y}^{(0)}=\mathbf{M}$, each pixel $p$ is updated as:
\begin{align}
\tilde{\mathbf{Y}}^{(t+1)}_p
&=
\frac{\lambda\,\mathbf{M}_p + \sum_{q\in\mathcal{N}_k(p)} \mathbf{W}_{pq}(\mathbf{\mathcal{Z}})\,\mathbf{Y}^{(t)}_q}
{\lambda + \sum_{q\in\mathcal{N}_k(p)} \mathbf{W}_{pq}(\mathbf{\mathcal{Z}})}, \\
\mathbf{Y}^{(t+1)}
&=
\mathbf{Y}^{(t)} + \omega\!\left(\tilde{\mathbf{Y}}^{(t+1)} - \mathbf{Y}^{(t)}\right),
\label{eq:sor_update}
\end{align}
with relaxation $\omega\!\in[1,2)$. In our implementation, $\mathbf{W}_{pq}(\mathbf{\mathcal{Z}})$ are normalized such that $\sum_{q\in\mathcal{N}_k(p)} \mathbf{W}_{pq}(\mathbf{\mathcal{Z}})=1$, making the denominator $\lambda+1$. The same normalized weights are reused across channels and iterations.

\subsection{GPU Implementation and Efficiency}
All steps in Eq.~\ref{eq:sor_update} are implemented using dense tensor primitives (\texttt{unfold}, pointwise operations, and reductions) that are highly optimized on GPUs (see Algorithm~\ref{alg:bilateral_solver}). We use reflective padding to avoid border bias and pre-compute $\mathbf{W}_{pq}(\mathbf{\mathcal{Z}})$ once per image, reusing it across channels and iterations. Unless otherwise stated, our implementation uses the following default values: $k{=}7$, $\sigma_s{=}3.0$\,px, $\sigma_r{=}0.01$, $\lambda{=}10^{-3}$, $n_{\mathrm{iter}}{=}80$, and $\omega{=}1.6$.

\begin{figure}[!t]
\centering
\includegraphics[width=0.65\linewidth]{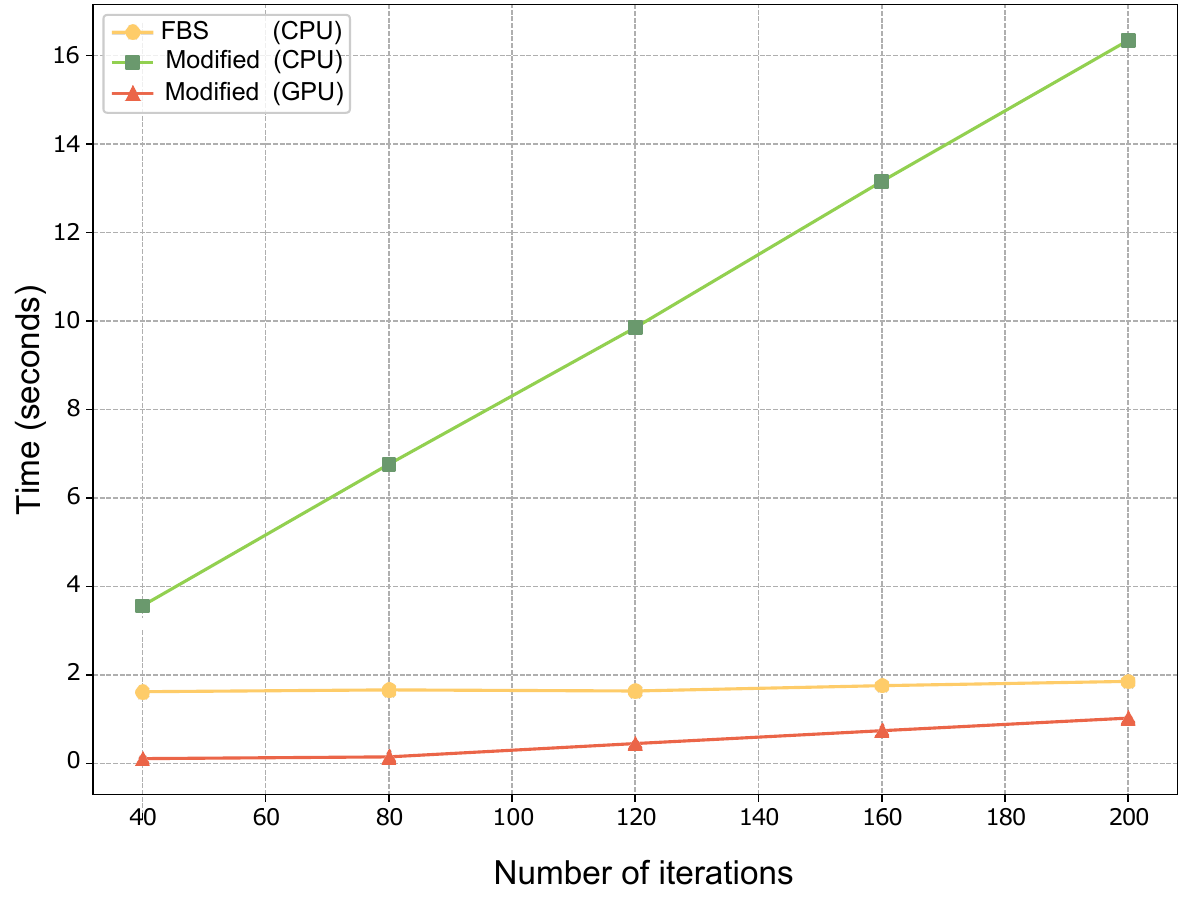}
\vspace{-2mm}
\caption{Runtime of the guided refinement process on a 750$\times$1000 guide 
image for different iteration counts $n_{\mathrm{iter}}$. We compare the Fast 
Bilateral Solver (FBS)~\cite{barron2016fast} on CPU with our modified bilateral 
refinement on both CPU and GPU. Since our pipeline is primarily intended for GPU 
deployment, the modified bilateral refinement provides an efficient and 
practical replacement for FBS. Runtimes were measured on an 
Intel~Core~i7-14700K~CPU and an NVIDIA~GeForce~RTX~4080~SUPER~GPU~(16\,GB\,VRAM). \label{fig:run_time_solver}}
\vspace{-1mm}
\end{figure}

With $n_{\mathrm{iter}}{=}80$, our GPU-based iterative solver achieves an \textbf{$\approx11\times$} wall-clock speedup over the original CPU-based FBS while producing visually comparable refinements. On CPU, however, it is slower than FBS due to the overhead of repeated dense tensor operations (see Fig.~\ref{fig:run_time_solver} for runtime vs.~$n_{\mathrm{iter}}$ on a 750$\times$1000 guidance image). Since our framework assumes GPU availability in most deployment scenarios, the GPU-accelerated solver offers a clear practical advantage.  

\begin{figure}[!t]
\centering
\includegraphics[width=\linewidth]{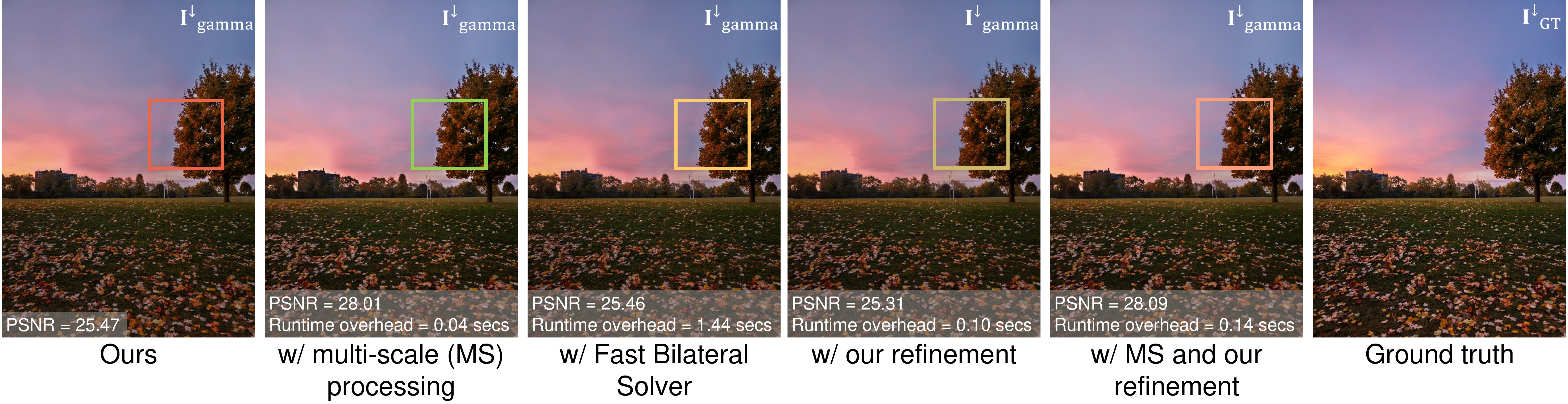}
\vspace{-3mm}
\caption{Comparison of 1) multi-scale processing for generating LTM coefficient maps, 
2) Fast Bilateral Solver (FBS)~\cite{barron2016fast} as a post-processing step, 
3) our modified bilateral refinement in place of FBS, and 
4) our refinement applied after multi-scale processing. 
The input is a linear sRGB image generated from a pseudo ground-truth denoised image in the S24 validation set~\cite{s24}. 
We report PSNR between the photofinishing output (downsampled to one-quarter resolution) and the corresponding ground truth, along with the runtime overhead for each approach, measured on an Intel~Core~i7-14700K~CPU and an NVIDIA~GeForce~RTX~4080~SUPER~GPU.
\label{fig:modified-solver}}
\vspace{-1mm}
\end{figure}

\begin{figure*}[!t]
\centering
\includegraphics[width=\linewidth]{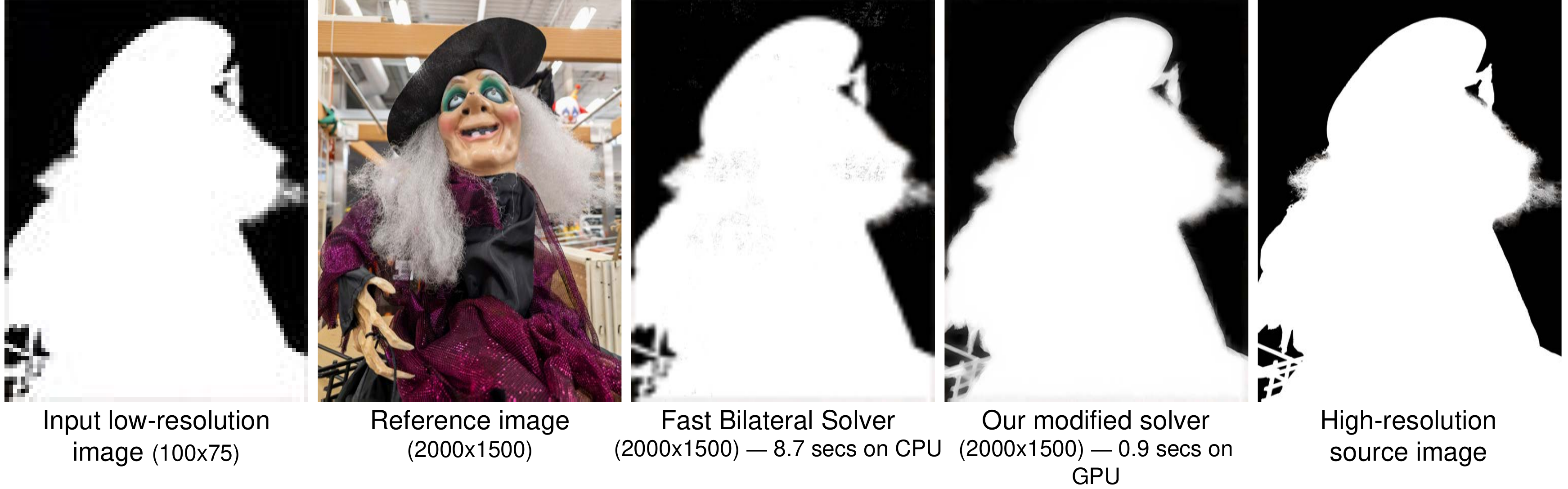}
\vspace{-4mm}
\caption{Comparison of our modified bilateral refinement with the original Fast Bilateral Solver (FBS)~\cite{barron2016fast}. The proposed modified bilateral refinement achieves comparable or improved results while running efficiently on GPU. The reference image is from the S24 test set~\cite{s24}; the high-resolution source was generated in Adobe Photoshop and downsampled to 100$\times$75 to obtain the low-resolution input.  Runtimes were measured on an 
Intel~Core~i7-14700K~CPU and an NVIDIA~GeForce~RTX~4080~SUPER~GPU~(16\,GB\,VRAM). \label{fig:bilateral_general}}
\vspace{-1mm}
\end{figure*}

Our GPU-accelerated solver retains the FBS energy~\cite{barron2016fast} but replaces the bilateral-grid PCG solver with a local SOR scheme. While this sacrifices exact convergence guarantees, it enables efficient and straightforward GPU implementation with sufficient quality in practice. Beyond mitigating artifacts in our LTM module (see Fig.~\ref{fig:modified-solver}), the proposed GPU-accelerated refinement can also serve as a \textit{general} edge-aware post-processing method (see Fig.~\ref{fig:bilateral_general}). Providing a thorough evaluation of our modified solver across different datasets and tasks is beyond the scope of this paper, but we consider this an important direction for future work.

\begin{algorithm}[t]
  \caption{GPU-Accelerated Iterative Bilateral Solver}
  \label{alg:bilateral_solver}
  \begin{algorithmic}[1]
    \REQUIRE Guidance $\mathbf{\mathcal{Z}}$, input tensor $\mathbf{M}$, kernel size $k$, scales $(\sigma_s,\sigma_r)$, smoothness $\lambda$, iterations $n_{\mathrm{iter}}$, relaxation $\omega$
    \STATE Pad $\mathbf{\mathcal{Z}}$ with \texttt{reflect}, extract $k{\times}k$ neighborhoods using \texttt{unfold}
    \STATE Compute range weights $\exp(-(\Delta \mathbf{\mathcal{Z}})^2 / 2\sigma_r^2)$
    \STATE Compute spatial weights $\exp(-\lVert \Delta p\rVert^2 / 2\sigma_s^2)$
    \STATE Form bilateral affinities $\mathbf{W}_{pq}$ and normalize so $\sum_q \mathbf{W}_{pq}=1$
    \STATE Initialize $\mathbf{Y}^{(0)} \gets \mathbf{M}$
    \FOR{$t = 0$ \TO $n_{\mathrm{iter}}-1$}
    \STATE Extract $k{\times}k$ neighborhoods of $\mathbf{Y}^{(t)}$ with \texttt{unfold}
    \STATE Compute smooth term $\sum_q \mathbf{W}_{pq}\,\mathbf{Y}^{(t)}_q$
    \STATE Compute target $(\lambda\,\mathbf{M}_p + \text{smooth})/(\lambda+1)$
    \STATE Update $\mathbf{Y}^{(t+1)} \gets \mathbf{Y}^{(t)} + \omega(\text{target} - \mathbf{Y}^{(t)})$
    \ENDFOR
    \STATE \textbf{return} $\mathbf{Y}^{(n_{\mathrm{iter}})}$
  \end{algorithmic}
\end{algorithm}

\section{Challenges and Potential Solutions}
\label{sec:limitations}

In this section, we discuss key challenges of the proposed framework along with potential solutions. We begin with artifacts in the LTM stage, where in rare cases (particularly under strong backlighting), halo artifacts may appear. To address this, we introduce two optional lightweight steps that can be user-enabled, since applying them indiscriminately may reduce accuracy (as explained in the next subsection). We next consider the case of training on an incomplete dataset that lacks some of the essential information required by our method. For this purpose, we use the Zurich Raw-to-sRGB dataset~\cite{zurich}, which, in addition to the aforementioned challenges, introduces further difficulties due to unaligned raw and ground-truth sRGB training pairs. Specifically, we describe how our framework can still be trained when key data are missing, such as ground-truth denoised raw images or DNG metadata (e.g., illuminant color and color correction matrices).

\subsection{Mitigating Artifacts}
\label{sec:artifacts}

\begin{figure}[t]
\centering
\includegraphics[width=\linewidth]{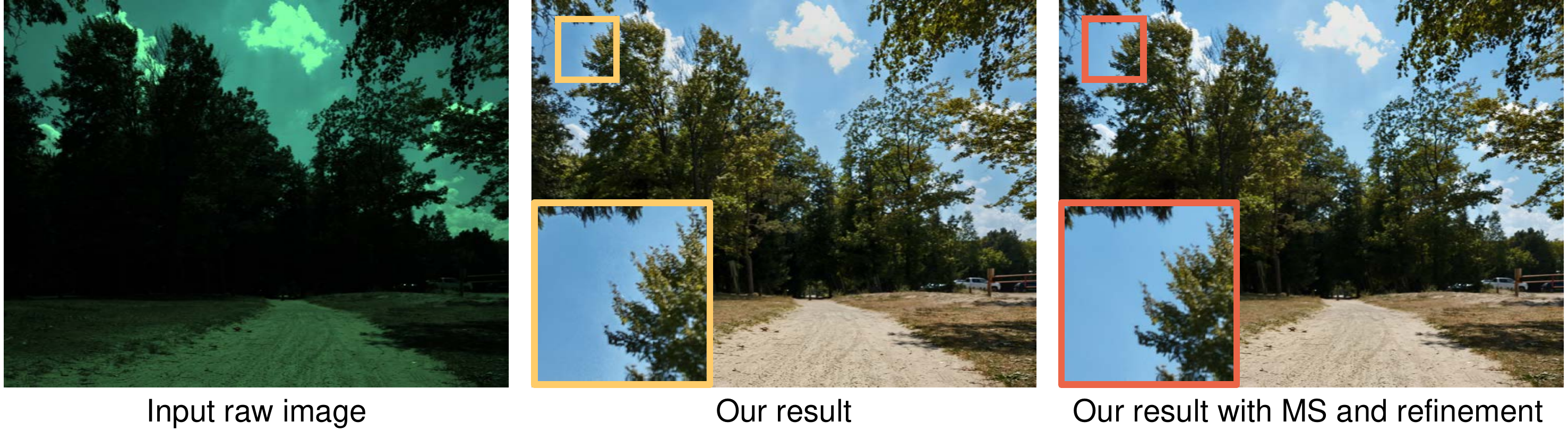}
\vspace{-5mm}
\caption{Our pipeline may produce halo artifacts or color inconsistencies in flat regions near edges. These artifacts can be mitigated via multi-scale (MS) processing and post-refinement of LTM maps. Image captured using the S24 main camera.
\label{fig:halo}}
\vspace{-4mm}
\end{figure}

Despite these advantages, our method faces a few challenges. First, in certain corner cases, halo artifacts may appear near edges in backlit scenes. Because of the modular structure of our framework, we were able to analyze this issue and identify its origin in the LTM process. These artifacts can be mitigated by applying multi-scale (MS) processing and post-processing refinement of the predicted LTM maps. We discuss this in detail in the supplementary material (Sec.~\myref{sec:artifacts}{\textcolor{section}{B.1}}) and show an example in Fig.~\ref{fig:halo}, comparing the original artifact and the mitigated result using MS and refinement.

We optionally apply two lightweight steps that suppress halos while preserving edges: 1) a multi-scale aggregation of LTM coefficient predictions, followed by 2) the edge-aware bilateral refinement discussed in Sec.~\ref{sec:accelerated_bilateral_solver}.

We adopt a multi-scale strategy to improve robustness and suppress halo artifacts in challenging cases (e.g., strong backlighting). Specifically, the input image after digital gain $\mathbf{I}^{\downarrow}_{\texttt{gain}}$ and its global tone-mapped version $\mathbf{I}^{\downarrow}_{\texttt{GTM}}$ are processed at progressively downsampled resolutions using scale factors: $\mathcal{S} = \{1.0, 0.5, 0.25, 0.125, 0.0625\}.$

At each scale $s\in\mathcal{S}$, if $s \neq 1.0$, both $\mathbf{I}^{\downarrow}_{\texttt{gain}}$ and $\mathbf{I}^{\downarrow}_{\texttt{GTM}}$ are bilinearly downsampled; otherwise the original resolution is used. From the input $\mathbf{I}^{\downarrow}_{\texttt{gain}(s)}$ at each scale $s$, we then
generate a guidance map using our multi-scale guidance subnetwork (see Sec.~\ref{sec:supp-ps-ltm-net}), followed by a smoothing step. In particular, the
guidance map is first extended using reflection padding. We then apply average pooling over a 5$\times$5 spatial window, producing a gently
varying guidance signal that stabilizes the slicing coefficients while
preserving the overall luminance structure needed for high-quality upsampling.

The pair $\big(\mathbf{I}^{\downarrow}_{\texttt{gain}(s)},\,
\mathbf{I}^{\downarrow}_{\texttt{GTM}(s)}\big)$ is concatenated and fed into the
grid-prediction subnetwork (see Sec.~\ref{sec:supp-ps-ltm-net}) to produce a coefficient grid. Bilateral slicing using the guidance
map $\mathbf{G}_{\texttt{guide}(s)}$, after the aforementioned smoothing step,
yields scale-specific coefficient maps, which are upsampled to full resolution
when $s \neq 1.0$. The upsampled coefficient maps from all scales are then
averaged to obtain the final output.

\begin{figure}[!t]
\centering
\includegraphics[width=\linewidth]{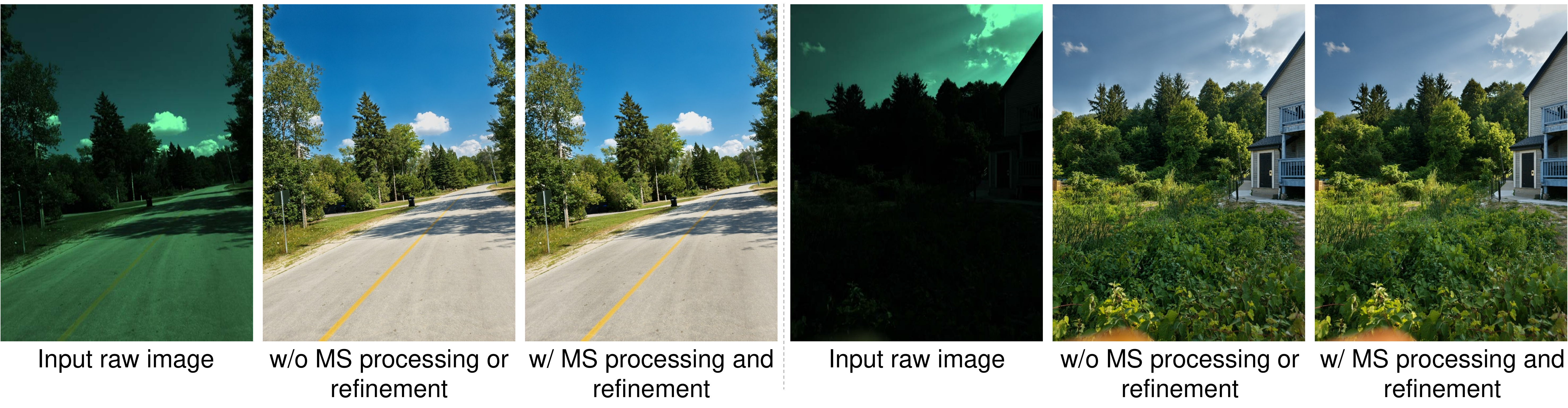}
\vspace{-5mm}
\caption{Halo artifact suppression using the proposed multi-scale (MS) processing and guided refinement of the predicted LTM coefficients. Images were captured with the S24 main camera. \label{fig:halo-supp}}
\vspace{-2mm}
\end{figure}

This strategy integrates information from coarse-to-fine representations, improving stability and reducing artifacts. After multi-scale processing, we apply our GPU-accelerated bilateral refinement (Sec.~\ref{sec:accelerated_bilateral_solver}) to further refine the predicted coefficient maps, using $\mathbf{I}^{\downarrow}_{\texttt{gain}}$ as the guide. These two steps together significantly mitigate halo artifacts in difficult scenes (see Fig.~\ref{fig:halo-supp}).

\begin{figure*}[!t]
\centering
\includegraphics[width=\linewidth]{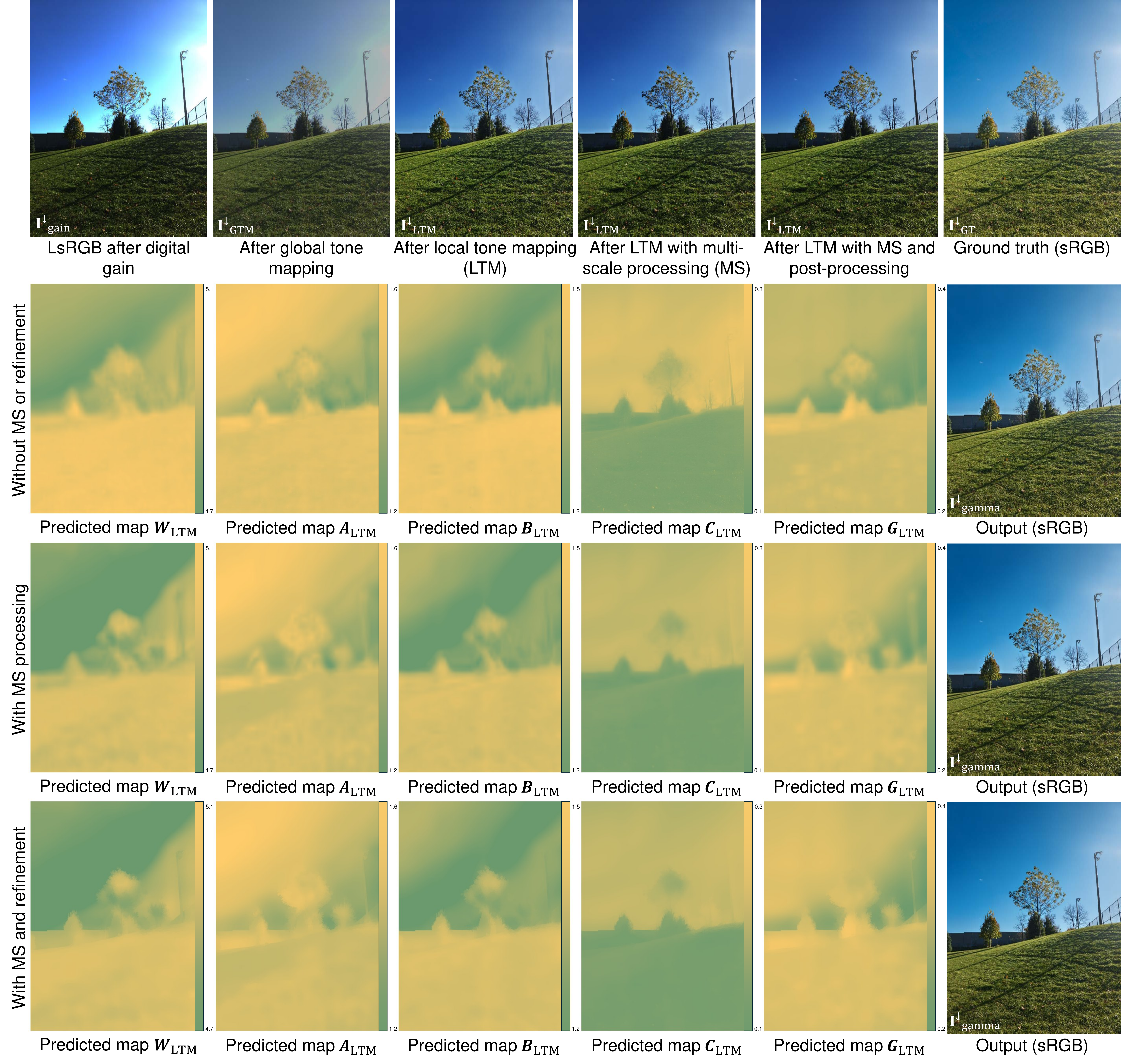}
\vspace{-4mm}
\caption{Comparison of predicted local tone mapping coefficients under different configurations: 1) without multi-scale processing, 2) with multi-scale processing, and 3) with multi-scale processing and guided refinement. For each configuration, we show intermediate outputs after LTM (first row) and the final sRGB outputs. The input images to the LTM module (linear sRGB after digital gain and after global tone mapping) and the ground truth are shown in the top row. The shown example is from the S24 test set~\cite{s24}. \label{fig:bilateral_on_ltm_weights}}
\vspace{-1mm}
\end{figure*}

Applying these steps blindly was found to reduce accuracy, as they also affect the strength of the LTM. To quantify this effect, we report results in Table~\ref{tab:post-process-ltm} using the photofinishing module. We use pseudo ground-truth denoised images at one-quarter resolution as input (after white balancing and color correction) and evaluate against the corresponding sRGB ground-truth images at the same resolution from the S24 test set~\cite{s24}. We compare the accuracy of our trained photofinishing module (Style~\#0, the default style) with and without the multi-scale processing and refinement designed to mitigate the rare halo artifacts observed in challenging scenes.

As shown, applying these steps indiscriminately degrades accuracy, so they remain user-controlled and can be enabled or disabled as needed.

\begin{table}[!t]
\centering
\caption{Results of our method with and without multi-scale processing and refinement of local tone mapping in the photofinishing module. Input images are linear sRGB generated from pseudo ground-truth denoised images in the S24 test set~\cite{s24}, downsampled to one-quarter resolution. Outputs are compared against the ground truth at the same resolution. The best results are highlighted in \textbf{\colorbox{best}{yellow}}.}
\vspace{-1mm}
\scalebox{0.8}{
\label{tab:post-process-ltm}
\begin{tabular}{|l|c|c|}
\hline
\multicolumn{1}{|c|}{} &
  \multicolumn{2}{c|}{\cellcolor{red}\textcolor{white}{\shortstack{\\\textbf{S24 Test Set} \\ \textbf{1/4 (LsRGB/sRGB)}}}} \\ \cline{2-3}
\multicolumn{1}{|c|}{\multirow{-2}{*}{\textbf{Method}}} &
  \textbf{PSNR}\,$\uparrow$ &
  \textbf{SSIM}\,$\uparrow$ \\ \hline
Default (w/o multi-scale and refinement)  & \textbf{\cellcolor{best}{27.49}} & \textbf{\cellcolor{best}{0.939}} \\ \hline
w/ multi-scale and refinement & 26.28 & 0.929 \\ \hline
\end{tabular}
}
\vspace{-2mm}
\end{table}

\subsection{Misaligned and Incomplete Datasets}
\label{sec:misalignment}
One challenge in training our framework is the need for reference denoised images
to supervise the denoising modules, along with camera AWB and CCM data for color correction. While pseudo ground-truths for denoising can be obtained using AI-based third-party denoisers, the absence of DNG files makes it difficult to generate such pseudo ground-truth data or to obtain the necessary color-correction data for training. In Sec.~\myref{sec:misalignment}{\textcolor{section}{B.2}} of the supplementary material, we discuss this issue and present a practical strategy for obtaining the missing data when DNG files are unavailable. Using the Zurich Raw-to-sRGB dataset~\cite{zurich} as a case study, we demonstrate how our workflow can train the proposed method effectively, achieving results comparable to the top-performing methods despite the lack of DNG files and with roughly half or fewer parameters compared to existing alternatives---while maintaining a high degree of modularity not offered by other methods.

Training our method requires access to scene illuminant vectors and color correction matrices (CCMs) for the training images (information typically available in DNG files) along with denoised ``ground-truth'' images. The latter can be generated from DNG files using AI-based blind denoisers such as the one in Adobe Lightroom, as in the S24 dataset~\cite{s24}.
Missing DNG files (or the extracted data they contain) pose a challenge for training our method. In addition, if the ground-truth sRGB images are misaligned with the corresponding raw inputs, this introduces another source of difficulty.

The Zurich Raw-to-sRGB dataset~\cite{zurich} provides an ideal test case for studying these issues, since it lacks both DNG files and the required metadata, and its raw-sRGB pairs are not spatially aligned. We therefore expect a degradation in accuracy, as our design does not explicitly handle misaligned training pairs. Nevertheless, this scenario is valuable for illustrating how our framework can be adapted to operate with missing data and for analyzing the resulting impact on accuracy.

The Zurich dataset consists of raw images captured with a smartphone camera and their corresponding JPEGs from a DSLR camera, but the pairs are not strictly pixel-aligned due to lens distortion, hand-held capture, and other factors. Although such a dataset is useful for evaluating robustness, it does not reflect the typical scenario, as obtaining aligned paired datasets is feasible in practice, as demonstrated by the S24 dataset~\cite{s24} and the MIT-Adobe FiveK dataset~\cite{Adobe5K}.

The remainder of this section describes how we address these missing elements to enable training of our framework and analyzes the effect of misalignment in the Zurich dataset on the resulting performance.

\noindent\\\textbf{Pseudo Ground-Truth Denoised Images.}
To generate pseudo ground-truth denoised images, we apply an inverse 2.2 gamma to the ground-truth sRGB image, producing a ``linearized'' estimate $\mathbf{I}_{\texttt{lin}}$. The use of 2.2 gamma is a practical approximation; in reality, camera ISPs typically apply a sequence of nonlinear operators that cannot be fully inverted by this simple correction~\cite{afifi2019color}.
We then compute a global nonlinear color mapping (CM) function $f_\texttt{CM}:\mathbb{R}^3 \rightarrow \mathbb{R}^3$ that maps $\mathbf{I}_{\texttt{lin}}$ to the input raw domain by solving the following optimization problem:
\begin{align}
\arg\min_{f_{\texttt{CM}}} \;
\big\lVert f_{\texttt{CM}}(\mathbf{I}_{\texttt{lin}}) - \mathbf{I}_{\texttt{raw}} \big\rVert_2^2,
\end{align}
where $f_{\texttt{CM}}$ is computed via linear regression with a polynomial kernel expansion, after saturated pixels are discarded.
The pseudo ground-truth denoised raw is then obtained as:
\begin{align}
\mathbf{I}_{\texttt{pseudo}} = f_\texttt{CM}(\mathbf{I}_{\texttt{lin}}).
\end{align}

\noindent\\\textbf{Illuminant Estimation.}
Since no illuminant metadata is provided in the Zurich dataset, we estimate the camera illuminant as the mean RGB of the demosaiced raw image, equivalent to applying the gray-world assumption~\cite{GW}. While this provides only a rough estimate, the resulting error can be compensated for by the computed CCM described below.

\noindent\\\textbf{Color Correction.}
The Zurich dataset also lacks CCMs, so we solve for a 3$\times$3 matrix $\mathbf{C}$ that maps white-balanced raw colors $\mathbf{r}'$ (with the estimated gray-world illuminant) to the corresponding sRGB values $\mathbf{s}$:
\begin{align}
\min_{\mathbf{C}} \; \lVert \mathbf{r}' \mathbf{C}^\top - \mathbf{s} \rVert_2^2,
\end{align}
subject to each row of $\mathbf{C}$ summing to 1 (to approximately preserve intensity). We solve this constrained least-squares optimization with non-negativity bounds using sequential least squares programming (SLSQP). In practice, the computed CCM compensates for errors in the estimated illuminant to produce colors close to the target images.

After these steps, we obtain the main components required to train our framework: input raw images, pseudo ground-truth denoised images, illuminant vectors, and CCMs to map the denoised raw images into linear sRGB space before training the photofinishing module. This allows us to train the framework as described in the main paper (more details of the training are available in Sec.~\ref{sec:training_details}). Specifically, we train the denoising network and the photofinishing module separately, then generate semi-final images before training the detail-enhancement network. Since the Zurich dataset provides 448$\times$448 semi-aligned patches, we train the detail-enhancement network on patches of the same size, rather than the 512$\times$512 patches used in the main paper experiments. Moreover, we disable the downsampling step before the photofinishing module and the guided upsampling after it when generating paired examples for the enhancement network and during evaluation, since testing is also performed on 448$\times$448 patches.

We report the results of our method in Table~\ref{tab:zurich}.~As shown, our method does not achieve state-of-the-art results on the Zurich dataset, unlike on the S24 dataset~\cite{s24}. Nevertheless, our workflow enables training on incomplete datasets such as Zurich and still yields competitive results, outperforming methods with a comparable number of parameters that lack the modularity of our framework~(e.g., \cite{afifi2021cie, lan}). Compared to the best-performing method~\cite{fourier} in Table~\ref{tab:zurich}, our model requires substantially fewer parameters ($\sim$500~K with the lite denoising network and $\sim$3.9~M with the large variant; see Sec.~\ref{sec:supp-denoising-nets} for configurational details of the denoising network variants), while achieving less than a 1~dB drop in PSNR (compared to $\sim$7.6~M parameters for the reference method~\cite{fourier}).
Moreover, the modular design of our framework provides greater flexibility, making the rendering pipeline easier to control, scale, debug, and customize.

\begin{table}[t]
\caption{Results on the Zurich Raw-to-sRGB dataset~\cite{zurich}. We report PSNR, SSIM~\cite{ssim}, LPIPS~\cite{LPIPS}, and $\Delta$E 2000~\cite{delta-e}, along with the total number of parameters for each method. While the Zurich dataset poses challenges for our framework due to misalignment and missing metadata, our method still achieves competitive results, with about a 1~dB drop in PSNR, while requiring significantly fewer parameters. The best results are highlighted in \textbf{\colorbox{best}{yellow}}.\label{tab:zurich}}
 \centering
\scalebox{0.75}{
\begin{tabular}{|l|c|c|c|c|c|}
\hline
\multicolumn{1}{|c|}{} & \multicolumn{5}{c|}{\cellcolor{red}\textcolor{white}{\textbf{Zurich Raw-to-sRGB Test Set}}} \\ \cline{2-6}
\multicolumn{1}{|c|}{\multirow{-2}{*}{\textbf{Method}}} &
  \textbf{PSNR}\,$\uparrow$ &
  \textbf{SSIM}\,$\uparrow$ &
  \textbf{LPIPS}\,$\downarrow$ &
  $\Delta$\textbf{E 2000}\,$\downarrow$ &
  \textbf{\# params} \\ \hline
PyNet \cite{zurich}                & 21.19 & 0.747 & 0.193 & NA    & 47,548,170 \\ \hline
CIE-XYZ Net \cite{afifi2021cie}    & 19.75 & 0.697 & 0.408 & 9.283 & 1,348,789 \\ \hline
LiteISP \cite{lite-isp}            & 21.55 & 0.749 & 0.187 & NA    & 9,094,000 \\ \hline
FourierISP \cite{fourier}          & \textbf{\cellcolor{best}21.65} & 0.755 & \textbf{\cellcolor{best}0.182} & NA & 7,589,736 \\ \hline
LAN \cite{lan}                     & 19.46 & 0.730 & NA    & NA    & 46,847 \\ \hline
Exposure \cite{hu2018exposure}     & 19.38 & \textbf{\cellcolor{best}0.811} & 0.350 & 11.438 & 6,124,000 \\ \specialrule{1pt}{0pt}{0pt}
Ours (lite,$\texttt{ }$ w/o enhancement)       & 19.57 & 0.717 & 0.401 & 9.448 & 452,447 \\ \hline
Ours (base,\hspace{1.5mm}w/o enhancement)       & 19.57 & 0.718 & 0.405 & 9.415 & 1,139,907 \\ \hline
Ours (large, w/o enhancement)      & 19.73 & 0.721 & 0.397 & 9.257 & 3,841,547 \\ \hline
Ours (lite,$\texttt{ }$ w/$\texttt{ }$ enhancement)        & 20.68 & 0.726 & 0.390 & 8.306 & 503,082 \\ \hline
Ours (base,\hspace{1.5mm}w/$\texttt{ }$ enhancement)        & 20.71 & 0.727 & 0.393 & 8.246 & 1,190,542 \\ \hline
Ours (large, w/$\texttt{ }$ enhancement)       & 20.76 & 0.729 & 0.386 & \textbf{\cellcolor{best}8.221} & 3,892,182 \\ \hline
\end{tabular}}
\end{table}

\section{Design of Networks}
\label{sec:network_design}
This section details the architectures of the networks that form our proposed framework.

\subsection{Raw-Denoising Network}
\label{sec:supp-denoising-nets}

For the raw-denoising network ($\mathcal{D}_{\texttt{raw}}$), we employ three variants of the NAFNet architecture~\cite{nafnet}, which differ only in their base channel width and thus span different points on the accuracy-efficiency trade-off. The lite, base, and large variants differ in their initial channel width (4, 8, and 16 channels, respectively), which is doubled after each encoder stage. This scaling leads to progressively larger bottleneck widths (64, 128, and 256 channels, respectively) and total capacities of approximately 0.25~M, 0.93~M, and 3.6~M parameters. All three networks share the same overall design: a four-stage encoder–decoder with skip connections and a middle stage of four residual blocks. Each network operates in a residual-learning manner, predicting a correction that is added back to the input raw image. The encoder and decoder are composed of [2, 2, 4, 8] and [2, 2, 2, 2] NAF blocks per stage, respectively. Each block consists of lightweight NAF modules with depthwise convolutions, channel attention, and simple gating to realize an activation-free nonlinearity. Layer normalization and learnable scale parameters are applied to stabilize training. The networks are fully convolutional and maintain compatibility with raw images of arbitrary resolution.

\subsection{Attention and Multi-Branch Blocks}
\label{sec:supp-ca-mb}
In our photofinishing module, the networks employ a lightweight attention mechanism and multi-branch convolutional (MBConv) blocks. Specifically, we adopt the coordinate attention (CA) mechanism~\cite{coord-attention} with some modifications. Specifically, we replace Batch Normalization~\cite{ioffe2015batch} with Group Normalization~\cite{wu2018group} to ensure consistent behavior under small batch sizes. We employ the LeakyReLU activation in all CA blocks. To prevent overly narrow bottlenecks in low-channel layers, we impose a lower bound of eight channels on the reduced dimensionality, computed as $\max(8, C_{\texttt{in-CA}} / r_{\texttt{CA}})$, where $C_{\texttt{in-CA}}$ is the input channel count and $r_{\texttt{CA}}$ is the reduction ratio. The reduction ratio is set to~4 by default, except for the luminance-guidance subnetwork within the chroma-mapping network, where we use $r_{\texttt{CA}}{=}2$.

\begin{figure}[!t]
\centering
\includegraphics[width=0.45\linewidth]{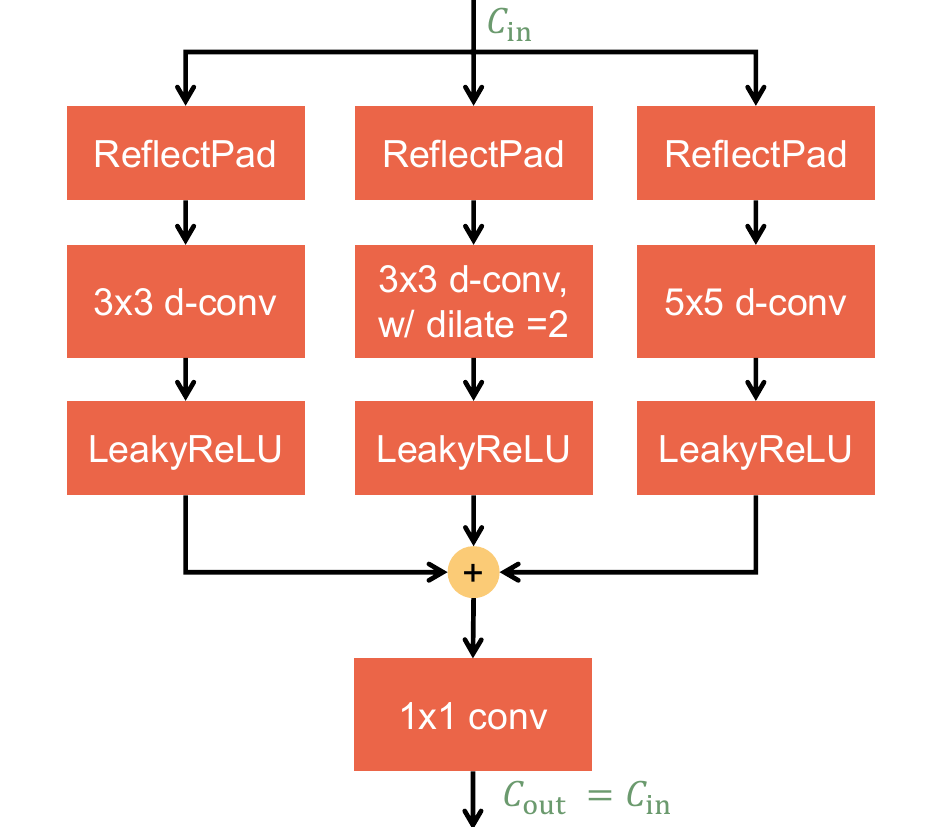}
\vspace{-1mm}
\caption{
Structure of the multi-branch convolutional block (MBConv). The input feature map is processed in three parallel depthwise convolution branches. Each branch applies reflection padding before the convolution and a LeakyReLU activation afterward. The outputs from all branches are summed and then fused through a convolution layer to produce the final aggregated feature map. This block preserves the number of channels (i.e., \textcolor{green}{$C_\texttt{in} = C_\texttt{out}$}).}
\label{fig:mb-block}
\vspace{-2mm}
\end{figure}

The MBConv block is designed to capture features at multiple receptive-field scales while maintaining low computational cost. As shown in Fig.~\ref{fig:mb-block}, the block consists of three depthwise convolution branches that operate in parallel. The first branch uses a 3$\times$3 kernel without dilation to capture fine local details. The second branch also uses a 3$\times$3 kernel but with dilation set to~2, allowing the convolution to gather broader contextual information without increasing parameters. The third branch employs a larger 5$\times$5 kernel to further expand the receptive field and capture smoother structural variations. Each branch begins with reflection padding, followed by a depthwise convolution and a LeakyReLU activation function. The outputs from all branches are summed and then fused by a 1$\times$1 convolution to produce the final aggregated feature map. This design effectively enhances receptive-field diversity without increasing the number of parameters or channels.

We provide ablation studies on the impact of using the MBConv blocks and the CA mechanism on the final photofinishing results in Sec.~\ref{Sec:ablation_studies_ps_desing}.

\subsection{Gain and Gamma-Correction Networks}
\label{sec:supp-ps-gain-gamma-nets}

Both the digital-gain ($\mathcal{D}_{\texttt{gain}}$) and gamma-correction ($\mathcal{D}_{\texttt{gamma}}$) networks share the same lightweight convolutional architecture, designed to predict a single global scalar factor that adjusts exposure or that is used for gamma correction. The two networks differ only in their output ranges.

\begin{figure}[!t]
\centering
\includegraphics[width=0.45\linewidth]{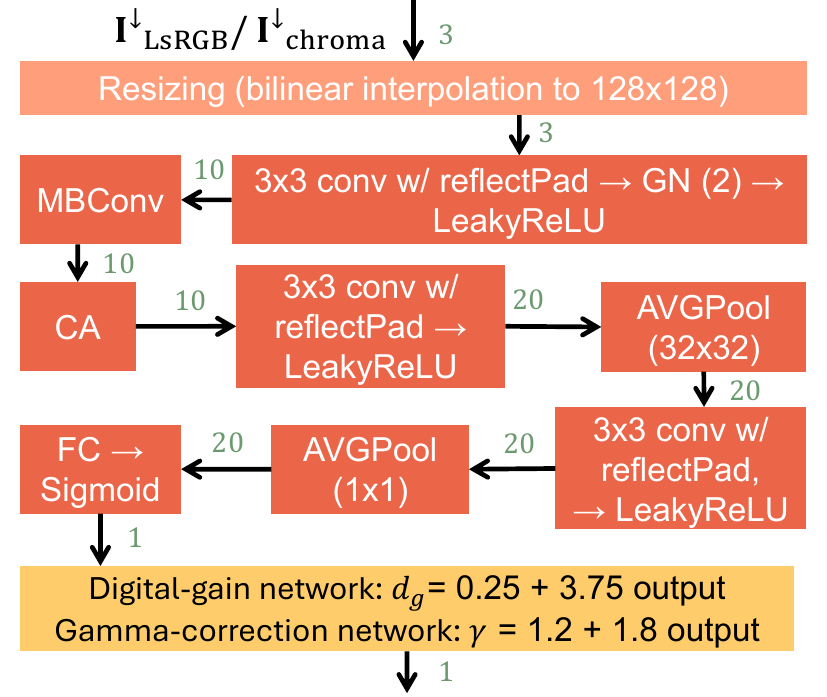}
\vspace{-1mm}
\caption{
Architecture of the network used for both the digital gain and gamma correction (6,587 parameters). The input image is first resized to a fixed resolution 128$\times$128, then processed by a convolutional layers with Group Normalization (2 groups) and LeakyReLU activation, followed by multi-branch convolutional and coordinate-attention blocks. After two stages of adaptive pooling, a fully connected layer with Sigmoid activation predicts a scalar in~$[0,1]$, which is linearly mapped to the target range for either digital gain ($[0.25,\,4.0]$) or gamma ($[1.2,\,3.0]$). The number of channels at each stage is indicated in \textcolor{green}{green}.}
\label{fig:gain-gamma-net}
\vspace{-1mm}
\end{figure}

As illustrated in Fig.~\ref{fig:gain-gamma-net}, the input image is first resized to a fixed resolution of 128$\times$128 using bilinear interpolation. The network then begins with a 3$\times$3 convolution layer with reflection padding, followed by Group Normalization (two groups) and a LeakyReLU activation. The features are processed by an MBConv block to capture multi-scale spatial context, followed by a CA block to encode long-range dependencies along both spatial directions. Another 3$\times$3 convolution and LeakyReLU activation refine the features before global pooling.

Two stages of average pooling are used to progressively compress spatial information: first to one-quarter of the input resolution (i.e., 128$\times$128 $\!\rightarrow\!$ 32$\times$32), and then to a single 1$\times$1 feature vector. The resulting vector is passed through a fully connected layer with a Sigmoid activation to produce a normalized scalar value in the range~$[0,1]$. Reflection padding is applied to all convolutions to prevent border artifacts, and the total number of trainable parameters for each network (digital-gain and gamma-correction) is 6,587.

For the digital-gain stage, the predicted scalar is linearly mapped to the range $[0.25$, $4.0]$ (equivalent to approximately exposure value [EV]~$[-2,+2]$), producing a gain factor~$d_g$ that scales image intensities before tone and color processing. For gamma correction, the normalized output is mapped to the range~$[1.2,\,3.0]$, yielding a gamma factor~$\gamma$. 

\subsection{Global Tone Mapping Network}
\label{sec:supp-ps-gtm-net}

\begin{figure}[!t]
\centering
\includegraphics[width=0.45\linewidth]{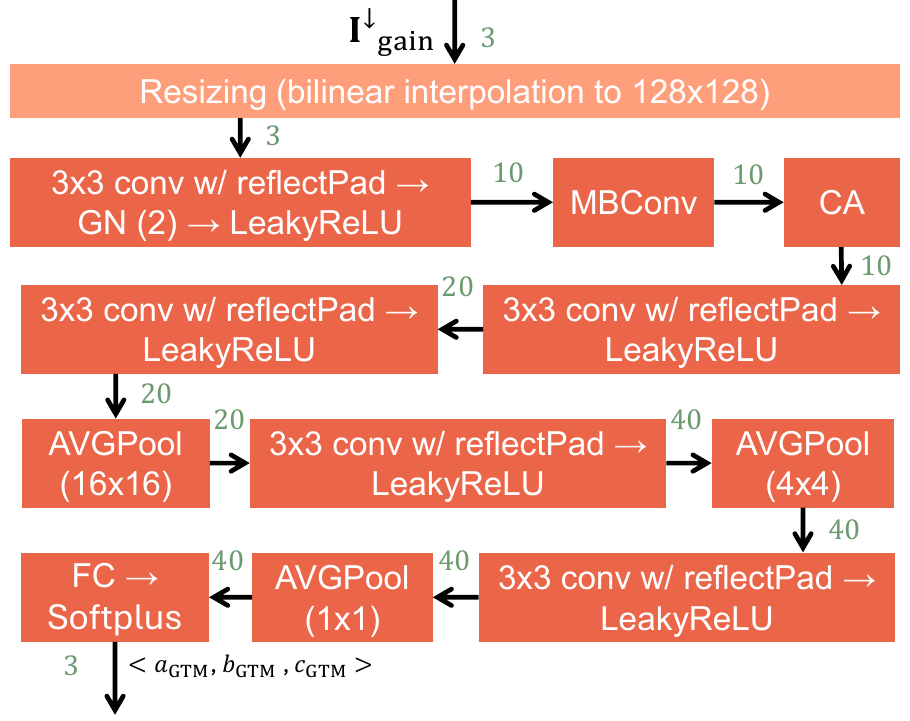}
\vspace{-1mm}
\caption{
Architecture of the global tone-mapping (GTM) network (28,369 parameters). The input linear sRGB image after digital gain ($\mathbf{I}^{\downarrow}_{\texttt{gain}}$) is first resized to 128$\times$128 and processed by convolutional layers with Group Normalization (2 groups) and LeakyReLU activations, followed by multi-branch convolutional and coordinate-attention blocks. Two additional convolutional layers refine the features before a series of adaptive pooling operations (16$\times$16 $\rightarrow$ 4$\times$4 $\rightarrow$ 1$\times$1). A fully connected layer with Softplus activation outputs the three positive parameters $(a_{\texttt{GTM}},\, b_{\texttt{GTM}},\, c_{\texttt{GTM}})$ that define the GTM curve. The number of channels at each stage is shown in \textcolor{green}{green}.}
\label{fig:gtm-net}
\vspace{-3mm}
\end{figure}

The global tone-mapping (GTM) network ($\mathcal{D}_{\texttt{GTM}}$), consisting of 28,369 learnable parameters, predicts three positive parameters that define a parametric tone curve applied to the linear sRGB image after digital gain (\(\mathbf{I}^{\downarrow}_{\texttt{gain}}\)).  
As shown in Fig.~\ref{fig:gtm-net}, the input image is first resized to 128$\times$128 using bilinear interpolation.  
The backbone begins with a 3$\times$3 convolution (10~channels) with reflection padding, followed by Group Normalization (two groups) and a LeakyReLU activation.  
The resulting features are processed by an MBConv block and a CA block to enhance spatial context modeling. Two subsequent 3$\times$3 convolutions (20~channels each, reflection padding) with LeakyReLU activations further refine the features.  
Spatial information is then progressively aggregated through adaptive average pooling to 16$\times$16, followed by a 3$\times$3 convolution (40~channels, reflection padding) and LeakyReLU activation, another adaptive pooling to 4$\times$4, and a final 3$\times$3 convolution (40~channels, reflection padding) with LeakyReLU activation.  
A final global average pooling reduces the feature map to 1$\times$1.  
The flattened feature vector is passed through a fully connected layer that produces three outputs, followed by a Softplus activation to ensure
\((a_{\texttt{GTM}},\, b_{\texttt{GTM}},\, c_{\texttt{GTM}}) > 0\).

\subsection{Local Tone-Mapping Network}
\label{sec:supp-ps-ltm-net}

The LTM network, $\mathcal{D}_{\texttt{LTM}}$ (120,215 learnable parameters), predicts spatially varying coefficients that locally modulate the tone-mapping behavior applied to the linear sRGB image after digital gain ($\mathbf{I}^{\downarrow}_{\texttt{gain}}$) and global tone mapping ($\mathbf{I}^{\downarrow}_{\texttt{GTM}}$).  
Unlike the global tone-mapping network (Sec.~\ref{sec:supp-ps-gtm-net}), which predicts a single set of global parameters $(a_{\texttt{GTM}}, b_{\texttt{GTM}},$ and $c_{\texttt{GTM}})$, the LTM network produces per-pixel coefficient maps ($\mathbf{A}_{\texttt{LTM}}$, $\mathbf{B}_{\texttt{LTM}}$, $\mathbf{C}_{\texttt{LTM}}$, $\mathbf{G}_{\texttt{LTM}}$, and $\mathbf{W}_{\texttt{LTM}})$ that enable locally adaptive tone mapping and exposure adjustment.

\begin{figure}[!t]
\centering
\includegraphics[width=0.8\linewidth]{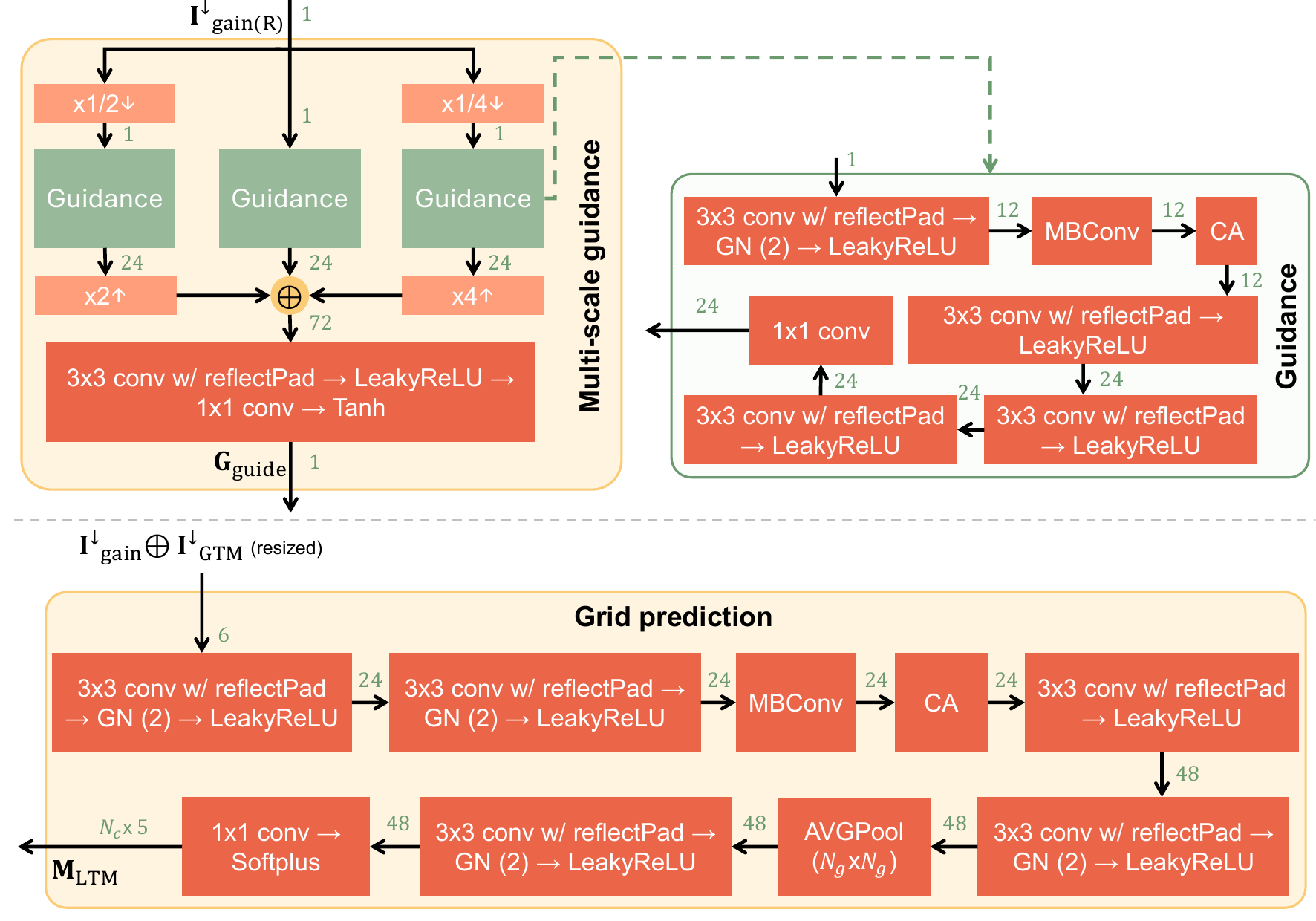}
\caption{
Architecture of the local tone-mapping (LTM) network (120,215 parameters). 
It consists of two main components: 
a multi-scale guidance subnetwork that processes a single channel of the linear sRGB image after digital gain to predict the guidance map $\mathbf{G}_{\texttt{guide}}$, 
and a grid-prediction subnetwork that processes the concatenated ($\oplus$) linear sRGB image after digital gain and the globally tone-mapped image to estimate the coefficient grid $\mathbf{M}_{\texttt{LTM}}$. 
The predicted grid is trilinearly sampled using $\mathbf{G}_{\texttt{guide}}$ to produce five spatial coefficient maps that control local tone-mapping behavior. 
The number of channels at each stage is indicated in \textcolor{green}{green}.}
\label{fig:ltmnet}
\vspace{-2mm}
\end{figure}

The LTM network consists of two main components: 1) a multi-scale guidance subnetwork and 2) a grid-prediction subnetwork, as shown in Fig.~\ref{fig:ltmnet}. 
The multi-scale guidance subnetwork derives a guidance map from one of the color channels of $\mathbf{I}^{\downarrow}_{\texttt{gain}}$. It processes three progressively downsampled versions of the luminance channel ($\times$1, $\times$1/2, and $\times$1/4 scales) using parallel convolutional branches, each composed of a 3$\times$3 convolution (reflection padding), Group Normalization (2 groups), LeakyReLU activation, an MBConv block, and a CA block, followed by four additional convolutional layers.  
The outputs from the three scales are upsampled (bilinear), concatenated, and fused using convolutional layers followed by a Tanh activation to produce the final guidance map $\mathbf{G}_{\texttt{guide}} \in [-1,1]$.
This multi-scale guidance design is intended to improve robustness against scale and contrast variations (see Sec.~\ref{Sec:ablation_studies_ps_desing} for ablation study).

In parallel, the grid-prediction subnetwork estimates a coarse grid of tone-mapping coefficients conditioned on both $\mathbf{I}^{\downarrow}_{\texttt{gain}}$ and $\mathbf{I}^{\downarrow}_{\texttt{GTM}}$.  
The concatenated image pair is downsampled to 384$\times$384 and processed by a series of convolutional layers (with Group Normalization and LeakyReLU activations), MBConv and CA blocks, and average pooling to form a latent grid representation of size $N_g{\times}N_g$ with depth~$N_c$.  
A final 1$\times$1 convolution outputs $N_c{\times}5$ feature channels corresponding to five coefficient volumes, activated by Softplus to ensure non-negativity.

The predicted $N_g{\times}N_g{\times}N_c{\times}5$ coefficient grid, $\mathbf{M}_\texttt{LTM}$, is then sampled via trilinear interpolation using the guidance map $\mathbf{G}_{\texttt{guide}}$ as the depth coordinate, producing the spatial coefficient maps ($\mathbf{A}_{\texttt{LTM}}$,
$\mathbf{B}_{\texttt{LTM}}$,
$\mathbf{C}_{\texttt{LTM}}$,
$\mathbf{G}_{\texttt{LTM}}$, and $\mathbf{W}_{\texttt{LTM}}$). In our implementation, we set $N_c{=}18$ and $N_g{=}64$, corresponding to a $64{\times}64$ spatial grid with 18 depth slices.  
See Sec.~\ref{Sec:ablation_studies_ps_desing} for ablation results on the impact of grid size.

During training, the spatial coefficient maps are regularized using the LTM smoothness loss, $\ell_{\texttt{LTM-s}}$ (see Sec.~\ref{sec:loss_funcs}).  
At inference time, an optional multi-scale processing and bilateral refinement step can be applied to smooth the coefficient maps and mitigate artifacts (see Sec.~\ref{sec:artifacts} for details).

\subsection{Chroma-Mapping Network}
\label{sec:supp-ps-chroma-net}

The chroma-mapping network ($\mathcal{D}_{\texttt{chroma}}$) predicts an image-specific 2D~lookup table (LuT) that transforms the chrominance components (CbCr) of the tone-mapped image produced by the preceding stages. The LuT is learned in an end-to-end fashion and applied via differentiable grid sampling to enable backpropagation through the photofinishing module. The network operates in the YCbCr color space, using the luminance channel of the tone-mapped image~($\mathbf{Y}_\texttt{LTM}$) as an auxiliary guidance signal.

Given the tone-mapped image (processed by both GTM and LTM operators) in YCbCr format, the image is first resized to a fixed spatial resolution of 128$\times$128. A differentiable 2D histogram representation of the CbCr channels, $\hat{\mathbf{H}}^{CbCr}$, is then computed (see Sec.~\ref{sec:hist} for details). This histogram encodes the joint distribution of chroma values across $N_h{\times}N_h$ bins (with a value range of $[-0.5,\,0.5]$) and provides a compact, differentiable summary of the chrominance content. 

\begin{figure}[!t]
\centering
\includegraphics[width=0.63\linewidth]{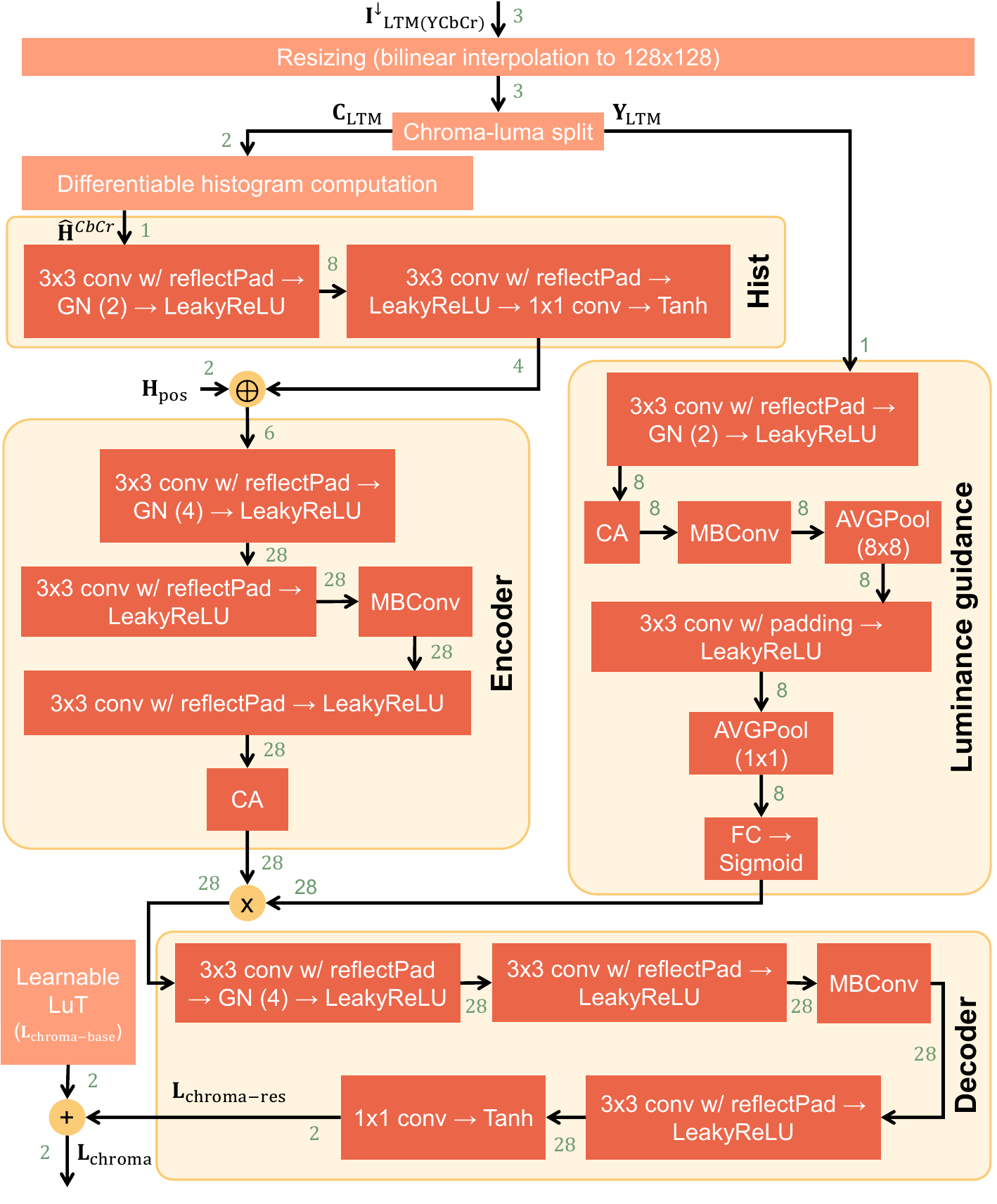}
\caption{
Architecture of the chroma-mapping network (45,466 parameters). The CbCr channels are first converted into a differentiable 2D histogram~($\hat{\mathbf{H}}^{CbCr}$; see Sec.~\ref{sec:hist}), which is processed by a shallow subnetwork (hist). The resulting chroma features are concatenated ($\oplus$) with an identity meshgrid~($\mathbf{H}_\texttt{pos}$) to provide subsequent layers with explicit awareness of the histogram bin positions.
The encoder-decoder backbone (with MBConv and CA blocks) predicts a residual $N_h$$\times$$N_h$$\times$2 LuT that is then added to a learnable base LuT. 
A luminance-guided subnetwork processes the $\mathbf{Y}_{\texttt{LTM}}$ channel to produce an attention vector that modulates the encoder bottleneck features. The number of channels at each stage is indicated in \textcolor{green}{green}.
}
\label{fig:lutnet}
\vspace{-1mm}
\end{figure}

The histogram $\hat{\mathbf{H}}^{CbCr}$ is processed by a shallow convolutional subnetwork (hereafter referred to as the hist subnetwork) to extract chroma features (four channels). These features are concatenated with an identity 2D meshgrid, denoted as $\mathbf{H}_\texttt{pos} \in \mathbb{R}^{N_h \times N_h \times 2}$, which encodes the normalized CbCr coordinates and provides the convolutional layers with explicit knowledge of the histogram bin locations~\cite{afifi2021c5}. The resulting six-channel tensor is then used as input to an encoder–decoder network with luminance-guided attention, as illustrated in Fig.~\ref{fig:lutnet}.

The encoder consists of three convolutional stages. The first encoder stage applies a 3$\times$3 convolution with reflection padding, Group Normalization (four groups), and a LeakyReLU activation. The second encoder stage includes a 3$\times$3 convolution with reflection padding followed by a LeakyReLU activation and an MBConv block, while the third stage applies only a 3$\times$3 convolution with reflection padding followed by a LeakyReLU. All stages preserve the channel dimensionality (28) and spatial resolution. The final encoder output is passed through a CA block to encode long-range dependencies along both chroma dimensions.

In parallel, the luminance channel ($\mathbf{Y}_\texttt{LTM}$) is processed by a lightweight auxiliary subnetwork that produces a 28-D attention vector.  
This luminance-guidance subnetwork begins with a 3$\times$3 convolution (8~channels) followed by Group Normalization (two groups) with LeakyReLU activation, a CA block (reduction ratio~2), and an MBConv block. After adaptive average pooling to 8$\times$8, the features are refined by another 3$\times$3 convolution and LeakyReLU activation, followed by global pooling and a fully connected layer with Sigmoid activation. The resulting normalized vector ($[0,1]^{28}$) modulates the encoder bottleneck features by channel-wise multiplication, allowing luminance-dependent adaptation of the predicted chroma mapping.

The decoder mirrors the encoder structure, consisting of three convolutional stages.  
The final layer uses a 1$\times$1 convolution with two output channels and a Tanh activation to generate the residual chroma LuT ($N_h$$\times$$N_h$$\times$2).  
The predicted LuT ($\mathbf{L}_{\texttt{chroma-res}}$) is added to a learnable, image-independent base LuT ($\mathbf{L}_{\texttt{chroma-base}}$) to produce the final 2D~LuT. The base LuT, $\mathbf{L}_{\texttt{chroma-base}}$, is initialized as an identity mapping in the CbCr space before training. During training, the final constructed LuT is regularized using the chroma LuT smoothness loss, $\ell_{\texttt{LuT-s}}$ (Sec.~\ref{sec:loss_funcs}). The total number of learnable parameters in the chroma-mapping network, including the base LuT, is 45,466.

\subsection{Detail-Enhancement Network}
\label{sec:supp-enhancement-net}

For the detail-enhancement network ($\mathcal{D}_{\texttt{enh}}$), we employ a lightweight variant of the NAFNet architecture~\cite{nafnet}, which predicts a residual correction added back to the input image. The detail-enhancement network follows the same encoder-decoder design with skip connections as our denoising models, but with a significantly reduced depth and width to minimize complexity. Specifically, it is configured with an initial channel width of 8, two encoder stages, two decoder stages, and four blocks in the middle stage. This compact setup results in approximately 50~K trainable parameters while providing sufficient capacity for fine-detail enhancement within our pipeline.

\section{Differentiable Histogram Computation}
\label{sec:hist}

As described in the main paper and in this supplemental material (Sec.~\ref{sec:supp-ps-chroma-net}), our chroma mapping network ($\mathcal{D}_{\texttt{chroma}}$) relies on a differentiable histogram representation of the chrominance channels to enable end-to-end training of the photofinishing module. Given an input image with Cb and Cr channels, we construct a 2D histogram by softly assigning each pixel to its corresponding histogram bins~\cite{afifi2019sensor, afifi2021histogan}.

Let $N_h$ denote the number of bins in the 2D histogram (and, accordingly, in the chroma 2D~LuT constructed by the chroma-mapping network), and let $[v_{\min}, v_{\max}]$ denote the value range. We place $N_h$ uniformly spaced bin centers for both Cb and Cr channels as follows:
\begin{equation}
\begin{aligned}
c_i &= v_{\min} + 
\tfrac{i}{N_h - 1}\,(v_{\max} - v_{\min}), \\[3pt]
&\quad i = 0, \ldots, N_h - 1.
\end{aligned}
\end{equation}

Each pixel $(cb, cr)$ contributes to all bins with a Gaussian weighting:
\begin{equation}
\mathcal{W}_{ij}(cb, cr) = \exp\!\left(-\frac{(cb - c_i)^2 + (cr - c_j)^2}{2\sigma_{\texttt{hist}}^2}\right),
\end{equation}
where $cb$ and $cr$ denote the chrominance values of a pixel, and $c_i, c_j$ are the uniformly spaced bin centers along the Cb and Cr axes, respectively.
The term $\sigma_{\texttt{hist}}$ controls the softness of the bin assignment. This formulation follows the differentiable histogram approach in~\cite{afifi2019sensor, afifi2021histogan}. The histogram is computed by summing over all pixels:
\begin{equation}
\mathbf{H}^{CbCr}_{ij} = \sum_{p} \mathcal{W}_{ij}(cb_p, cr_p),
\end{equation}
then normalized to unit sum and square-rooted:
\begin{equation}
\hat{\mathbf{H}}^{CbCr} = \sqrt{\frac{\mathbf{H}^{CbCr}}{\sum_{i,j} \mathbf{H}^{CbCr}_{ij} + \epsilon}}.
\end{equation}

In our implementation, we set the default values to $N_h{=}24$ bins (see Sec.~\ref{Sec:ablation_studies_ps_desing} for ablation results with $N_h{=}12$), $v_{\min}{=}-0.5$, $v_{\max}{=}0.5$, and $\sigma_{\texttt{hist}}{=}0.075$. 
\section{Guided Upsampling Regularization}
\label{sec:supp-bgu}

In the main paper, we explained that our method employs bilateral guided upsampling (BGU)~\cite{bgu} with a modified regularization that yields better results. Here, we provide additional details of our regularization.  Our photofinishing module predicts a low-resolution gamma-corrected output,
$\mathbf{I}^{\downarrow}_{\texttt{gamma}} \in \mathbb{R}^{\frac{H}{4}\times\frac{W}{4}\times 3}$.
To recover full resolution, we apply BGU using the full-resolution
linear sRGB guide image $\mathbf{I}_{\texttt{LsRGB}}~\in~\mathbb{R}^{H\times W\times 3}$.
The high-resolution photofinishing result (after upsampling) is denoted as 
$\mathbf{I}^{\uparrow}_{\texttt{gamma}}$ $\in$ $\mathbb{R}^{H\times W\times 3}$. The fitting grid is constructed in the bilateral space using the low-resolution guide $\mathbf{I}^{\downarrow}_{\texttt{LsRGB}} \in \mathbb{R}^{\frac{H}{4}\times\frac{W}{4}\times 3}$, while the local affine fitting statistics are accumulated from the corresponding low-resolution input-output pair $(\mathbf{I}^{\downarrow}_{\texttt{LsRGB}}, \mathbf{I}^{\downarrow}_{\texttt{gamma}})$, with guidance derived from the high-resolution image $\mathbf{I}_{\texttt{LsRGB}}$ during upsampling. The grid size is set to $\frac{H}{64}\times\frac{W}{64}\times 16$ to achieve a good balance between speed and accuracy.

Let $\mathbf{S}_m \in \mathbb{R}^{(C+1)\times(C+1)}$ and
$\mathbf{T}_m \in \mathbb{R}^{C\times(C+1)}$ denote the accumulated statistics in grid cell $m$,
with $c_m$ samples. For RGB image, $C = 3$. The affine transform for cell $m$,
$\mathbf{A}_m \in \mathbb{R}^{C\times(C+1)}$, is estimated by solving:

\begin{equation}
\begin{split}
\mathbf{T}_m \mathbf{S}_m^\top + \lambda_m \mathbf{R}_m
&= \mathbf{A}_m \big( \mathbf{S}_m \mathbf{S}_m^\top + \lambda_m \mathbf{I}_{C+1} \big), \\
\lambda_m &= \lambda (c_m + 1),
\end{split}
\label{eq:appendix:upsample:bgu_objective}
\end{equation}
where $\mathbf{I}_{C+1}$ is the $(C{+}1)\times(C{+}1)$ identity matrix and $\lambda$ is a scalar scaling factor. Intuitively, $\mathbf{S}_m$ and $\mathbf{T}_m$ encode the relationship between the
low-resolution input $\mathbf{I}^{\downarrow}_{\texttt{gamma}}$ and the guide
$\mathbf{I}_{\texttt{LsRGB}}$ within grid cell $m$, such that $\mathbf{A}_m$
models a local affine mapping that transfers structure and edges from the guide
to the upsampled output. The regularization matrix $\mathbf{R}_m$ controls the
per-channel stability of this fitting. Thus, the design of $\mathbf{R}_m$
directly affects how well color and fine details are preserved in the final
$\mathbf{I}^{\uparrow}_{\texttt{gamma}}$ result.

\noindent\\\textbf{Original Halide Regularization.}
The Halide implementation of BGU~\cite{bgu} tries to solve:

\begin{equation}
\small{
\mathbf{T}_m^{\text{blur}} {\mathbf{S}_m^{\text{blur}}}^\top + \lambda_m \mathbf{R}^{\text{luma}}_m
= \mathbf{A}_m \big( \mathbf{S}_m^{\text{blur}} {\mathbf{S}_m^{\text{blur}}}^\top + \lambda_m \mathbf{I}_{C+1} \big),}
\end{equation}

\noindent where

\begin{equation}
\mathbf{R}^{\text{luma}}_m =
\begin{bmatrix}
\mathrm{diag}(r^{\text{luma}}_m,\,r^{\text{luma}}_m,\,r^{\text{luma}}_m) & \mathbf{0}
\end{bmatrix},
\end{equation}

\noindent that uses a single luma gain
$r^{\text{luma}}_m \in \mathbb{R}$ applied uniformly to all RGB channels:

\begin{equation}
\begin{split}
r_m^{\text{luma}} &=
\frac{[0.25, 0.5, 0.25] \cdot \mathbf{T}_m^{\text{blur}}[1\!:\!C,\,C\!+\!1] + \lambda^{\text{glob, blur}}}
     {[0.25, 0.5, 0.25] \cdot \mathbf{S}_m^{\text{blur}}[1\!:\!C,\,C\!+\!1] + \lambda^{\text{glob, blur}}}, \\[0.4em]
\lambda^{\text{glob, blur}} &= \lambda \Big( \sum_{n} c_{n}^{\text{blur}} + 1 \Big),
\end{split}
\label{eq:luma_ratio}
\end{equation}

\noindent where $\mathbf{T}_m^{\text{blur}}$ and ${\mathbf{S}_m^{\text{blur}}}$ are the blurred version of $\mathbf{T}_m$ and ${\mathbf{S}_m}$ respectively, and $c_{m}^{\text{blur}}$ is the number of samples in the blurred grid $m$.

Intuitively, this achromatic assumption causes color crosstalk. Although empty cells are handled via spatial grid blurring, this process can oversmooth edges and fine details.

\noindent\\\textbf{Our Gated Regularization.}
Instead of applying grid blurring that may create artifacts, we adopt the original BGU~\cite{bgu} objective (Eq.~\ref{eq:appendix:upsample:bgu_objective}) and use per-channel adaptive gains:
\begin{equation}
\mathbf{R}^{\text{gate}}_m =
\begin{bmatrix}
\mathrm{diag}(\mathbf{r}_m) & \mathbf{0}
\end{bmatrix},
\end{equation}
where $\mathbf{r}_m = [r_{m,1},\,r_{m,2},\,r_{m,3}]$ is defined as:
\begin{equation}
r_{m,c} =
\begin{cases}
\dfrac{\mathbf{T}_m[c,C+1]}{\mathbf{S}_m[c,C+1] + \lambda_m}, & \text{if } c_m > 0, \\[6pt]
r^{\text{glob}}_c, & \text{if } c_m = 0,
\end{cases}
\end{equation}
with global per-channel fallback gains:
\begin{align}
r^{\text{glob}}_c &=
\dfrac{\sum_n \mathbf{T}_n[c,C+1]}{\sum_n \mathbf{S}_n[c,C+1] + \lambda^{\text{glob}}}, \\
\lambda^{\text{glob}} &= \lambda \Big(\sum_n c_n + 1\Big).
\end{align}

This gated regularization applies accurate local per-channel constraints when sufficient samples
exist in a grid cell, while sparsely populated cells are robustly handled using global statistics.
Unlike the original Halide scheme, our regularization eliminates the need for grid blurring, avoids color crosstalk, and preserves sharp details in the upsampled result (see Sec.~\ref{Sec:ablation_studies_upsampling} for ablation studies). 
\section{Loss Function Details}
\label{sec:loss_funcs}

We provide the implementation details of the loss terms used to train our photofinishing module.
Each loss is designed to complement a specific stage of the photofinishing module, ensuring perceptually faithful reproduction of tone, color, and structure.
The overall objective is given in Eq.~\myref{eq:ps-loss-func}{\textcolor{section}{9}} of the main paper.

\noindent\\\textbf{SSIM Loss.}
We adopt a differentiable variant of the structural similarity index~\cite{ssim}, computed independently per color channel.
Given our output and the ground-truth sRGB images $\mathbf{I}^{\downarrow}_{\texttt{gamma}}$ and $\mathbf{I}^{\downarrow}_{\texttt{GT}}$, we compute local statistics over an 11$\times$11 Gaussian window (standard deviation $\sigma{=}1$) using depthwise convolution:
\begin{equation}
\label{eq:ssim-loss}
\begin{aligned}
\mu_1 &= \mathbf{I}^{\downarrow}_{\texttt{gamma}} * \mathbf{w}, \quad
\mu_2 = \mathbf{I}^{\downarrow}_{\texttt{GT}} * \mathbf{w}, \\[3pt]
\sigma_1^2 &= ({\mathbf{I}^{\downarrow}_{\texttt{gamma}}}^2 * \mathbf{w}) - \mu_1^2, \quad
\sigma_2^2 = ({\mathbf{I}^{\downarrow}_{\texttt{GT}}}^2 * \mathbf{w}) - \mu_2^2, \\[3pt]
\sigma_{12} &= (\mathbf{I}^{\downarrow}_{\texttt{gamma}} \mathbf{I}^{\downarrow}_{\texttt{GT}} * \mathbf{w}) - \mu_1 \mu_2.
\end{aligned}
\end{equation}

In Eq.~\ref{eq:ssim-loss}, $\mathbf{w}$ denotes the Gaussian weighting kernel. The SSIM map is then computed as:
\begin{equation}
\texttt{SM}(\mathbf{I}^{\downarrow}_{\texttt{gamma}}, 
           \mathbf{I}^{\downarrow}_{\texttt{GT}}) = \frac{
    (2\mu_1\mu_2 + C_1)(2\sigma_{12} + C_2)
}{
    (\mu_1^2 + \mu_2^2 + C_1)(\sigma_1^2 + \sigma_2^2 + C_2)
},
\end{equation}
where $C_1{=}0.01^2$ and $C_2{=}0.03^2$.
The final loss is defined as $\ell_{\texttt{SSIM}} = 1 - \text{mean}(\texttt{SM})$.

\noindent\\\textbf{Perceptual Loss.}
We compute the perceptual loss using a pretrained VGG-19 network~\cite{simonyan2014very} with frozen weights. We extract features from layers \texttt{relu3\_3} and \texttt{relu4\_2}, following the convention of perceptual supervision in photo enhancement tasks. Both predicted and ground-truth RGB images are resized to $224{\times}224$ and normalized by the ImageNet mean and standard deviation. The loss is given by the following equation:
\begin{equation}
\ell_{\texttt{perc}} =
\sum_{l \in \{\texttt{relu3\_3},\,\texttt{relu4\_2}\}}
\lVert \phi_l(\mathbf{I}^{\downarrow}_{\texttt{gamma}}) - \phi_l(\mathbf{I}^{\downarrow}_{\texttt{GT}}) \rVert_1,
\end{equation}
where $\phi_l(\cdot)$ denotes the activation map of layer $l$.

\noindent\\\textbf{$\Delta E$ Color Loss.}
To enforce perceptual color accuracy, we compute a differentiable approximation of the CIE~1976~$\Delta E^*_{ab}$ (i.e., $\Delta E_{76}$) metric~\cite{robertson1977cie} in the CIE~Lab color space.
Both the predicted and ground-truth images are evaluated before gamma correction.
For the ground-truth, we first remove the sRGB gamma encoding using the gamma scale, $\gamma$, predicted by our gamma-correction network ($\mathcal{D}_{\texttt{gamma}}$), so that both images are evaluated under a consistent, image-specific gamma setting.
The resulting pre-gamma, quasi-linear sRGB images are then converted to the CIE~Lab color space through a differentiable transformation chain:
\[
\text{linear sRGB} \;\rightarrow\; \text{XYZ} \;\rightarrow\; \text{Lab}.
\]
The $\text{XYZ}$ conversion uses the standard transformation matrix:
\begin{equation}
\mathbf{M}_{\texttt{sRGB}\rightarrow\texttt{XYZ}} =
\scalebox{0.9}{$
\begin{bmatrix}
0.4124564 & 0.3575761 & 0.1804375 \\
0.2126729 & 0.7151522 & 0.0721750 \\
0.0193339 & 0.1191920 & 0.9503041
\end{bmatrix}
$}.
\end{equation}
For the $\texttt{XYZ}\!\rightarrow\!\texttt{Lab}$ mapping, 
the standard CIE~Lab definition uses a piecewise cubic-root function as follows:
\begin{equation}
f_{\texttt{cubic}}(t) =
\begin{cases}
t^{1/3}, & t > \delta^3, \\[3pt]
\tfrac{t}{3\delta^2} + \tfrac{4}{29}, & \text{otherwise.}
\end{cases}
\label{eq:cubic_f}
\end{equation}
where $\delta{=}6/29$.
Since this function is non-differentiable at $t{=}\delta^3$, 
we replace it with a smooth blend defined as:
\begin{equation}
f_{\texttt{soft}}(t) = \sigma_{\texttt{s}}\!\big(\alpha_{\texttt{s}}(t - \delta^3)\big)\,t^{1/3} + \big[1 - \sigma_{\texttt{s}}\!\big(\alpha_{\texttt{s}}(t - \delta^3)\big)\big]
    \Big(\tfrac{t}{3\delta^2} + \tfrac{4}{29}\Big),
\end{equation}
where $\sigma_{\texttt{s}}(\cdot)$ denotes the sigmoid function used in our differentiable Lab conversion, and $\alpha_{\texttt{s}}{=}150$ controls the sharpness of the transition.
The resulting $\text{Lab}$ tensors, denoted as $\mathbf{L}^{\texttt{out}}$ and $\mathbf{L}^{\texttt{GT}}$,
represent the full three-channel $(L^*, a^*, b^*)$ values for our output and the ground-truth images, respectively.
The per-pixel $\Delta E_{76}$ distance is then computed as:
\begin{equation}
\ell_{\Delta E} = \frac{1}{N}\sum_{p}
\big\lVert \mathbf{L}_{p}^{\texttt{out}} - \mathbf{L}_{p}^{\texttt{GT}} \big\rVert_2,
\end{equation}
where $N=H\times W$ is the total number of pixels. If the 3D~LuT (pre-chroma mapping network) is included as a learnable component of the photofinishing module, we additionally incorporate an auxiliary color-difference term equal to half the $\Delta E_{76}$ between the de-gammaed ground-truth image and the 3D~LuT output. This auxiliary term encourages the 3D~LuT to approximate the target color rendering prior to subsequent chroma mapping.

\noindent\\\textbf{Chroma Loss.}
This loss enforces chroma consistency by comparing the CbCr channels of the predicted output (after chroma mapping) with those of the de-gammaed ground truth, where the de-gammaing is performed using the image-specific gamma factor $\gamma$ predicted by our gamma-correction network.
The chroma loss, $\ell_{\texttt{CbCr}}$, is defined as:
\begin{equation}
\ell_{\texttt{CbCr}} =
\big\lVert \hat{\mathbf{C}}_{\texttt{LuT}} -
\mathbf{C}_{\texttt{GT}} \big\rVert_1,
\end{equation}
where $\mathbf{C}_{\texttt{GT}}$ denotes the CbCr channels of the de-gammaed ground-truth image (${\mathbf{I}^{\downarrow}_{\texttt{GT}}}^{\gamma}$), and
$\hat{\mathbf{C}}_{\texttt{LuT}}$ represents the mapped CbCr channels obtained
using the predicted 2D LuT $\mathbf{L}_{\texttt{chroma}}$.

\noindent\\\textbf{Tone-Mapping Loss.}
The tone-mapping loss enforces joint consistency between the global and local tone-mapping modules and the ground-truth tone-mapped image.
Let $\mathbf{Y}_{\texttt{GT}}$ denote the ground-truth luminance (Y) channel, de-gammaed using our image-specific gamma factor ($\gamma$). Similarly, let $\mathbf{Y}_{\texttt{GTM}}$ and $\mathbf{Y}_{\texttt{LTM}}$ denote the luminance channels obtained after the global and local tone-mapping stages, respectively. The tone-mapping loss is defined as:
\begin{equation}
\ell_{\texttt{TM}} =
0.6\,\big\lVert \mathbf{Y}_{\texttt{GTM}}^{\downarrow 8} - \mathbf{Y}_{\texttt{GT}}^{\downarrow 8} \big\rVert_1
+ \big\lVert \mathbf{Y}_{\texttt{LTM}} - \mathbf{Y}_{\texttt{GT}} \big\rVert_1,
\end{equation}
where ${\downarrow 8}$ indicates $1/8$ downsampling via bilinear interpolation.
The downsampled term penalizes deviations in the global tone and overall luminance of the tone-mapped output produced by the GTM network, while the second term encourages the LTM network to refine local tonal and luminance variations to better align with the ground truth.

\noindent\\\textbf{Luminance Consistency Loss.}
We regularize the tone-mapping process to preserve the average image brightness:
\begin{equation}
\ell_{\texttt{luma}} =
\big|\,\mathbb{E}[\mathbf{Y}_{\texttt{GTM}}] - \mathbb{E}[\mathbf{Y}_{\texttt{gain}}]\,\big|,
\end{equation}
where $\mathbf{Y}_{\texttt{gain}}$ denotes the Y channel of the linear sRGB image after applying the predicted image-specific digital gain $d_g$ in its YCbCr representation.
This loss prevents exposure drift and encourages the tone-mapping modules to refine contrast while preserving global brightness.

\noindent\\\textbf{Total Variation Regularization Loss.}
The $\ell_{\texttt{LuT-s}}$ and $\ell_{\texttt{LTM-s}}$ terms promote spatial coherence in the chroma LuT and LTM maps. 
We implement these terms as isotropic total variation (TV) losses to encourage smoothness in the learned chroma LuT and local tone-mapping coefficient maps:
\begin{equation}
\ell_{\texttt{TV}}(\mathbf{X}) =
\tfrac{1}{2}\big(\lVert\nabla_h \mathbf{X}\rVert_1 + \lVert\nabla_w \mathbf{X}\rVert_1\big),
\end{equation}
where $\nabla_h$ and $\nabla_w$ denote horizontal and vertical spatial gradients, respectively.

\noindent\\\textbf{Overall Loss Combination.}
The total training objective is defined as a weighted sum of all loss components:
\begin{equation}
\mathcal{L}_{\texttt{total}} =
\sum_j \lambda_j \, \ell_j,
\end{equation}
where each weighting coefficient $\lambda_j$ balances the relative contribution of its corresponding loss term.
Empirically, we found that combining low-level ($\ell_1$, $\ell_{\texttt{SSIM}}$, $\ell_{\texttt{CbCr}}$), perceptual ($\ell_{\Delta E}$, $\ell_{\texttt{perc}}$), and regularization ($\ell_{\texttt{LuT-s}}$, $\ell_{\texttt{LTM-s}}$, $\ell_{\texttt{luma}}$, $\ell_{\texttt{TM}}$) losses yields stable convergence and perceptually consistent results across different picture styles (see Sec.~\ref{Sec:ablation_studies_ps_loss} for ablation studies).

\section{Training Details}
\label{sec:training_details}

As mentioned in the main paper, we train each module separately. In this section, we provide further details of training our framework.

\subsection{Denoising Training}
\label{sec:denoising-training}
We trained the denoising model ($\mathcal{D}_{\texttt{raw}}$) to minimize the $\ell_1$ loss between the pseudo denoising ground truth on 256$\times$256 patches and the model outputs. The network was optimized using AdamW~\cite{loshchilov2017decouple} with 
$\beta_1=0.9$, $\beta_2=0.9$, and no weight decay. The learning rate was 
initialized at $10^{-3}$ and decayed to $10^{-5}$ using cosine 
annealing~\cite{loshchilov2016sgdr}. Training was performed for 100 epochs with 
a batch size of 32, and gradient clipping with a maximum norm of $0.01$ was 
applied to stabilize optimization. For each dataset, including the generic 
denoisers (Sec.~\ref{sec:cross-camera}), we trained three model variants---lite, base, and large. The architectural configurations of these variants are provided in Sec.~\ref{sec:supp-denoising-nets}.

\subsection{Photofinishing Training}
\label{sec:supp-photofinishing-training}

We trained the photofinishing module (which includes: $\mathcal{D}_{\texttt{gain}}$, $\mathcal{D}_{\texttt{GTM}}$, $\mathcal{D}_{\texttt{LTM}}$, $\mathcal{D}_{\texttt{chroma}}$, $\mathcal{D}_{\texttt{gamma}}$, $\mathbf{L}_{\texttt{chroma-base}}$, and optionally the 3D LuT, $\mathbf{L}_{\texttt{RGB}}$) for 600 epochs with a batch size of 8.
The input images were the pseudo ground-truth denoised raw images mapped to linear sRGB space
($\mathbf{I}_{\texttt{LsRGB}}$), and the corresponding ground-truth images were the final sRGB targets
($\mathbf{I}_{\texttt{GT}}$). Both input and ground-truth images were resized to 512$\times$512, with
standard geometric augmentations applied during training.

Optimization was performed using 
Adam~\cite{kingma2014adam} with parameters $\beta_1=0.9$, $\beta_2=0.999$, and 
an $\ell_2$ regularization term of $10^{-7}$. The learning rate was 
initialized at $10^{-4}$ and decayed to $10^{-6}$ using a cosine 
annealing schedule~\cite{loshchilov2016sgdr}. The loss function combined multiple terms as described in the main paper and in 
Sec.~\ref{sec:loss_funcs}. The weights of these losses were empirically set as 
follows: $\lambda_1 = 2.5$ for $\ell_1$, $\lambda_{\texttt{SSIM}} = 0.5$ for 
$\ell_{\texttt{SSIM}}$, $\lambda_{\Delta E} = 0.02$ for $\ell_{\Delta E}$, 
$\lambda_{\texttt{perc}} = 0.01$ for the perceptual loss 
$\ell_{\texttt{perc}}$, $\lambda_{\texttt{CbCr}} = 1.0$ for the CbCr loss 
$\ell_{\texttt{CbCr}}$, $\lambda_{\texttt{LuT-s}} = 0.06$ for the LuT 
smoothness loss $\ell_{\texttt{LuT-s}}$, $\lambda_{\texttt{TM}} = 0.5$ for the 
tone mapping loss $\ell_{\texttt{TM}}$, $\lambda_{\texttt{LTM-s}} = 0.6$ for 
the LTM smoothness loss $\ell_{\texttt{LTM-s}}$, and 
$\lambda_{\texttt{luma}} = 0.2$ for the luminance consistency loss 
$\ell_{\texttt{luma}}$. Note that in the loss function details section (Sec.~\ref{sec:loss_funcs}), the resized photofinishing output
and ground truth are denoted as $\mathbf{I}^{\downarrow}_{\texttt{gamma}}$ (after applying the
predicted gamma correction) and $\mathbf{I}^{\downarrow}_{\texttt{GT}}$, respectively, at a
resolution of 512$\times$512. In all other contexts,
$\mathbf{I}^{\downarrow}_{\texttt{gamma}}$ and $\mathbf{I}^{\downarrow}_{\texttt{GT}}$ refer to the
photofinishing output and ground-truth images at the native $1/4$ raw resolution.

\subsection{Detail-Enhancement Training}
\label{sec:supp-detail-enhancement-training}
After training the denoising models and the photofinishing module, we trained 
our final detail-enhancement network ($\mathcal{D}_{\texttt{enh}}$), which refines the perceptual quality of the 
final output. To construct the training data, we first applied our trained denoisers (lite, base, and large variants, used jointly to enlarge the training 
set as an augmentation strategy). The denoised raw images were converted to linear sRGB using the metadata of each image (illuminant color vector and CCM), downsampled to one quarter of the resolution, processed by the trained photofinishing 
module, and then upsampled using the bilateral guided upsampling method 
\cite{bgu} with our regularization (Sec.~\ref{sec:supp-bgu}). We extracted non-overlapping 512$\times$512 patches from the guided-upsampled results and their corresponding ground-truth images. We used normalized image values in $[0,1]$ (without quantization) to construct the paired training set.

For the Zurich dataset (Sec.~\ref{sec:misalignment}), we did not apply the downsampling and guided-upsampling procedure since the dataset provides only cropped 448$\times$448 patches, and thus we trained directly on these patches. 

Training was performed for 50 epochs with a batch size of 16. The network was optimized using 
Adam~\cite{kingma2014adam} with parameters $\beta_1=0.9$, $\beta_2=0.999$, and $\ell_2$ regularization $10^{-7}$. The learning rate was initialized at 
$10^{-4}$ and decayed to $10^{-6}$ using a cosine annealing  schedule~\cite{loshchilov2016sgdr}. The pixel-wise $\ell_1$ loss between predicted and ground-truth sRGB patches was used as the loss function.

\section{Generalization Across Cameras}
\label{sec:cross-camera}
One of the main advantages of our modular design is the ability to decouple the camera-specific modules of our pipeline from the generic ones that can operate across cameras. Specifically, the raw image denoising module is one of the primary camera-specific components, as noise characteristics differ from camera to camera. Other camera-specific factors include color correction (i.e., auto white balancing and color correction matrices), which can be addressed using recent cross-camera illuminant estimation methods (e.g.,~\cite{afifi2021c5}) or by reading metadata from DNG files---see Sec.~\ref{sec:awb} for details on how we handle cross-camera auto white-balance correction.

To overcome the limitations of camera-dependent denoising, we trained a generic denoiser in addition to our camera-specific models trained for the S24 main camera, as reported in the main paper. To train the generic denoiser, we first modeled the noise characteristics of several cameras and then synthesized noise using these models on the pseudo ground-truth denoised raw images from the S24 dataset~\cite{s24} to generate noisy raw inputs with synthetic noise from different cameras. These synthetic noisy raw images, along with the original noisy raw images from S24, were used to train our generic denoiser models under the same configurations described in Sec.~\ref{sec:denoising-training}. Specifically, we constructed a hybrid S24 dataset comprising 20\% clean pseudo ground-truth denoised images, 40\% real noisy S24 images, and 40\% synthetic noisy images generated using the aforementioned cross-camera noise models.

For each synthetic noisy sample, the ISO value was inherited from its corresponding real image to maintain realistic noise-exposure relationships. We adopted the widely used heteroscedastic Gaussian noise model~\cite{Foi2009ClippedDenoising, Foi2015PracticalRaw-data, Liu2014PracticalImage}:
\begin{equation}
\mathbf{n}_{\texttt{gd}} \sim
\mathcal{N} \left( \mathbf{0}, \mathbf{\sigma}_{\texttt{gd}}^2 \right), \quad
\mathbf{\sigma}_{\texttt{gd}}^2 = \beta_{\texttt{gd},1}\,\mathbf{x}_{\texttt{gd}} + \beta_{\texttt{gd},2},
\end{equation}
where the noise variance $\mathbf{\sigma}_{\texttt{gd}}^2$ increases linearly with the underlying clean pixel intensity $\mathbf{x}_{\texttt{gd}}$. The parameters $\beta_{\texttt{gd},1}$ and $\beta_{\texttt{gd},2}$ represent the shot- and read-noise components, respectively.
To synthesize noise under this model, we collected (\(\mathbf{x}_{\texttt{gd}}, \mathbf{\sigma}_{\texttt{gd}}^2\)) pairs and calibrated $\beta_{\texttt{gd},1}$ and $\beta_{\texttt{gd},2}$ for each ISO and color channel via linear regression. The shot-noise parameter $\beta_{\texttt{gd},1}$ corresponds to the slope, while the read-noise parameter $\beta_{\texttt{gd},2}$ corresponds to the intercept of the linear fit. During calibration, we performed 14-18 captures of an X-Rite ColorChecker chart (24 color patches). Each capture was taken at a different ISO setting, ranging from ISO~50 to ISO~3200. For each ISO, we extracted the 24 color patches and subdivided them into smaller sub-patches to increase the number of samples for fitting. This process yielded multiple small patches of size (35$\times$35) per ISO and color channel, with the total number varying depending on the chart’s coverage within the image.

Next, we computed the mean and variance of each small patch. Using these means as $\mathbf{x}_{\texttt{gd}}$ and their corresponding variances as $\mathbf{\sigma}_{\texttt{gd}}^2$, we fitted a linear regression model to estimate the slope ($\beta_{\texttt{gd},1}$) and intercept ($\beta_{\texttt{gd},2}$).
This produced one pair of $\beta_{\texttt{gd},1}$ and $\beta_{\texttt{gd},2}$ values for each ISO in the calibration data. For ISOs not present, we interpolated the existing parameters (using linear interpolation for $\beta_{\texttt{gd},1}$ and cubic interpolation for $\beta_{\texttt{gd},2}$) to enable noise synthesis for all intermediate ISOs within the calibration range.

To synthesize diverse noise patterns during training, we used noise models calibrated for a total of 12 sensors: S20 FE (main), S20 Ultra (main), Pixel 6 (main), Pixel 9 Pro (main), Pixel 9 Pro (telephoto), S22 Plus (front), S22 Plus (telephoto), S22 Plus (ultra-wide), S22 Plus (main), S22 (ultra-telephoto), S22 (ultra-wide), and S22 Ultra (main). These sensors were selected to represent a diverse range of sensor types (main, telephoto, ultra-wide, and front), enabling the generic denoiser to learn from heterogeneous noise distributions.

Similar to the camera-specific denoiser, we trained three variants of the generic denoiser (lite, base, and large).
Table~\ref{tab:noise-ablation-2} compares the accuracy of camera-specific models---each trained and evaluated on its corresponding dataset (S24 or MIT~Adobe~5K~\cite{Adobe5K})---against the generic denoisers evaluated on the same datasets. The denoised ``ground-truth'' images for Adobe~5K were obtained by running Adobe Lightroom’s AI denoiser on the raw images, following the same procedure used for the S24 dataset.  As expected, camera-specific models performed better on their corresponding cameras, while the generic models were designed to generalize to unseen devices. To further validate this, we evaluated both model types on the Smartphone Image Denoising Dataset (SIDD)~\cite{abdelhamed2018high}, which contains raw images from smartphones not included in the S24, Adobe~5K, or generic model training sets.
Results in Table~\ref{tab:noise-ablation-ssid} show that the generic models outperformed the camera-specific ones, demonstrating stronger cross-camera generalization.

\begin{table}[!t]
\centering
\caption{Comparison of camera-specific and generic denoising models (lite, base, and large) on the S24 \cite{s24} and MIT Adobe 5K \cite{Adobe5K} test sets (noisy/denoised raw). The best results are highlighted in \textbf{\colorbox{best}{yellow}}.}
\scalebox{0.8}{
\label{tab:noise-ablation-2}
\begin{tabular}{|l|c|c|c|c|}
\hline
\multirow{2}{*}{\textbf{Denoising Model}} &
  \multicolumn{2}{c|}{\cellcolor{red}\textcolor{white}{\shortstack{\\\textbf{S24 Test Set} \\ \textbf{(Noisy/Denoised)}}}} &
  \multicolumn{2}{c|}{\cellcolor{orange}\textcolor{white}{\shortstack{\\\textbf{Adobe 5K Test Set} \\ \textbf{(Noisy/Denoised)}}}} \\ \cline{2-5}
 & \textbf{PSNR}\,$\uparrow$ & \textbf{SSIM}\,$\uparrow$ & \textbf{PSNR}\,$\uparrow$ & \textbf{SSIM}\,$\uparrow$ \\ \hline
Camera-specific (lite)  & 53.39 & 0.998 & 55.57 & 0.998 \\ \hline
Camera-specific (base)  & 55.97 & \textbf{\cellcolor{best}{0.999}} & 57.36 & \textbf{\cellcolor{best}{0.999}} \\ \hline
Camera-specific (large) & \textbf{\cellcolor{best}{57.33}} & \textbf{\cellcolor{best}{0.999}} & \textbf{\cellcolor{best}{59.85}} & \textbf{\cellcolor{best}{0.999}} \\ \hline
Generic (lite)          & 51.19 & 0.997 & 51.84 & 0.996 \\ \hline
Generic (base)          & 55.15 & \textbf{\cellcolor{best}{0.999}} & 52.86 & 0.997 \\ \hline
Generic (large)         & 56.46 & \textbf{\cellcolor{best}{0.999}} & 53.07 & 0.997 \\ \hline
\end{tabular}
}
\vspace{-2mm}
\end{table}

\begin{table}[!t]
\centering
\caption{Comparison of denoising models (lite, base, large) trained on S24 \cite{s24}, MIT Adobe 5K \cite{Adobe5K}, and on a wide range of noise profiles for generic denoising, evaluated on the SIDD dataset \cite{abdelhamed2018high}. The best results are highlighted in \textbf{\colorbox{best}{yellow}}.}
\scalebox{0.8}{
\label{tab:noise-ablation-ssid}
\begin{tabular}{|l|c|c|}
\hline
\multicolumn{1}{|c|}{} &
  \multicolumn{2}{c|}{\cellcolor{red}\textcolor{white}{\shortstack{\\\textbf{SIDD Test Set} \\ \textbf{(Raw Noisy/Raw Clean)}}}} \\ \cline{2-3}
\multicolumn{1}{|c|}{\multirow{-2}{*}{\textbf{Denoising Model}}} &
  \textbf{PSNR}\,$\uparrow$ &
  \textbf{SSIM}\,$\uparrow$ \\ \hline
S24 (lite)      & 46.53 & 0.940 \\ \hline
S24 (base)      & 47.28 & 0.952 \\ \hline
S24 (large)     & 47.47 & 0.956 \\ \hline
Adobe (lite)    & 48.05 & 0.966 \\ \hline
Adobe (base)    & 48.28 & 0.967 \\ \hline
Adobe (large)   & 48.12 & 0.971 \\ \hline
Generic (lite)  & 49.09 & 0.971 \\ \hline
Generic (base)  & \textbf{\cellcolor{best}{50.00}} & 0.978 \\ \hline
Generic (large) & 49.55 & \textbf{\cellcolor{best}{0.979}} \\ \hline
\end{tabular}
}
\vspace{-2mm}
\end{table}

\begin{figure*}[!t]
\centering
\includegraphics[width=0.9\linewidth]{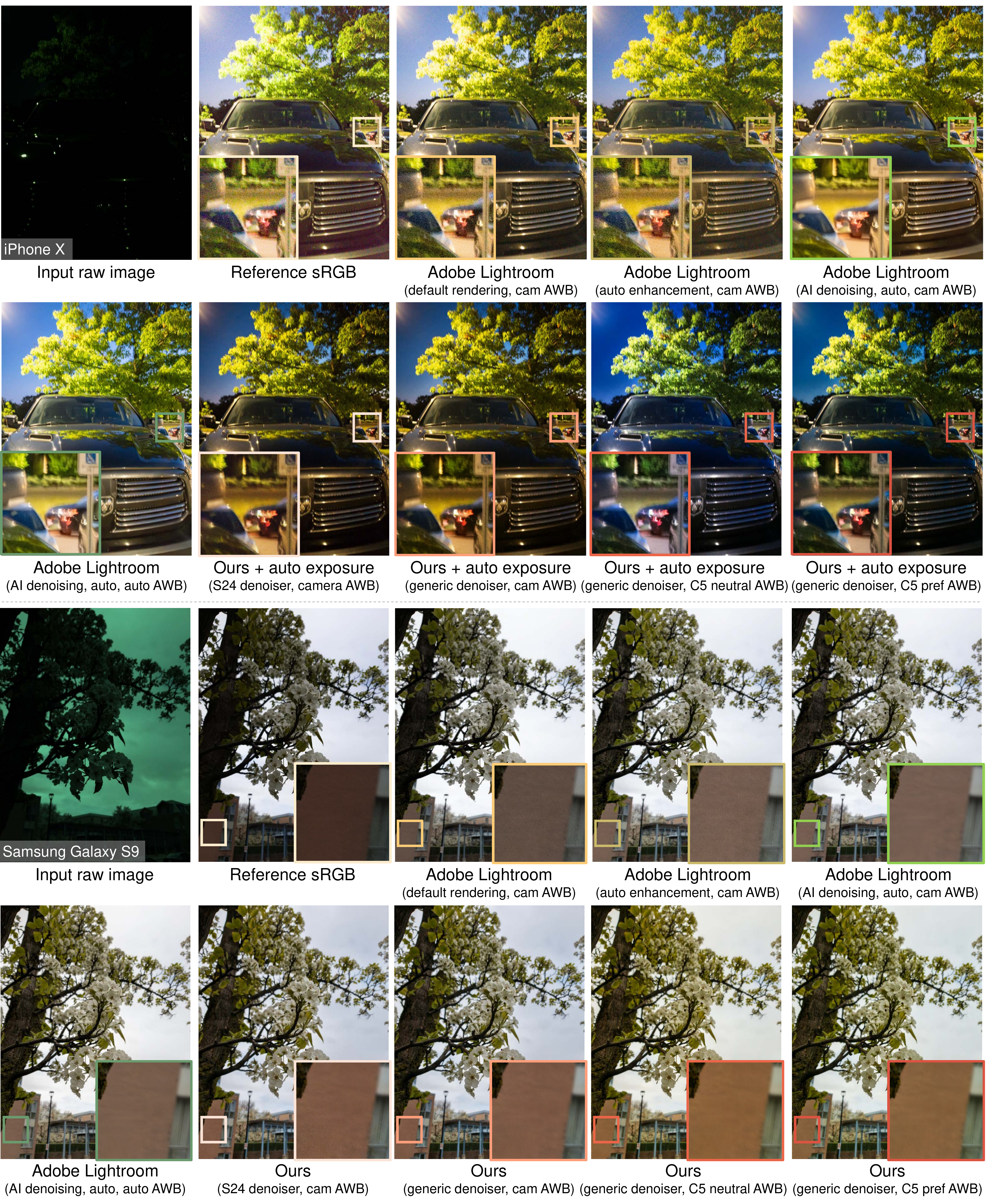}
\vspace{-4mm}
\caption{
Qualitative results on unseen cameras from the Raw-to-Raw dataset~\cite{afifi2021semi}.
The first row shows an iPhone X example and the second a Samsung Galaxy S9 example.
For each case, we include the input raw image, the reference sRGB rendered by the capturing app,
and Adobe Lightroom outputs (default, built-in auto enhancement with/without AI denoising, and results with camera AWB and Lightroom auto AWB).
Our outputs are shown with two denoisers (trained on the S24 dataset~\cite{s24} and a generic one),
using camera AWB, C5 (neutral) AWB~\cite{afifi2021c5}, and user-preference mapping~\cite{zhao2025learning}.
All results are generated with the photofinishing module trained on the S24 dataset.
}
\label{fig:cross_camera}
\vspace{-2mm}
\end{figure*}

Figure~\ref{fig:cross_camera} shows qualitative results of our entire pipeline on unseen cameras (Samsung Galaxy S9 and iPhone X) from the Raw-to-Raw dataset~\cite{afifi2021semi}, using the photofinishing module trained on the S24 dataset. We report results with both the S24-trained denoiser and a generic denoiser, combined with either the camera AWB or the cross-camera AWB model (C5)~\cite{afifi2021c5}, as well as with the post-AWB user-preference mapping~\cite{zhao2025learning} (see Sec.~\ref{sec:awb} for more details). For comparison, we include outputs from Adobe Lightroom, both with and without its built-in auto enhancement and AI-based denoiser features. Our method achieves high-quality cross-camera rendering, competitive with commercial software. For the iPhone X, we observed that raw images are substantially darker than those from most other cameras; thus, we applied the auto-exposure adjustment before the digital gain step (see Sec.~\ref{sec:auto-exposure} for details). This adjustment was facilitated by the modularity of our pipeline, as further discussed in Sec.~\ref{sec:gui}.

\begin{figure*}[!ht]
\centering
\includegraphics[width=0.83\linewidth]{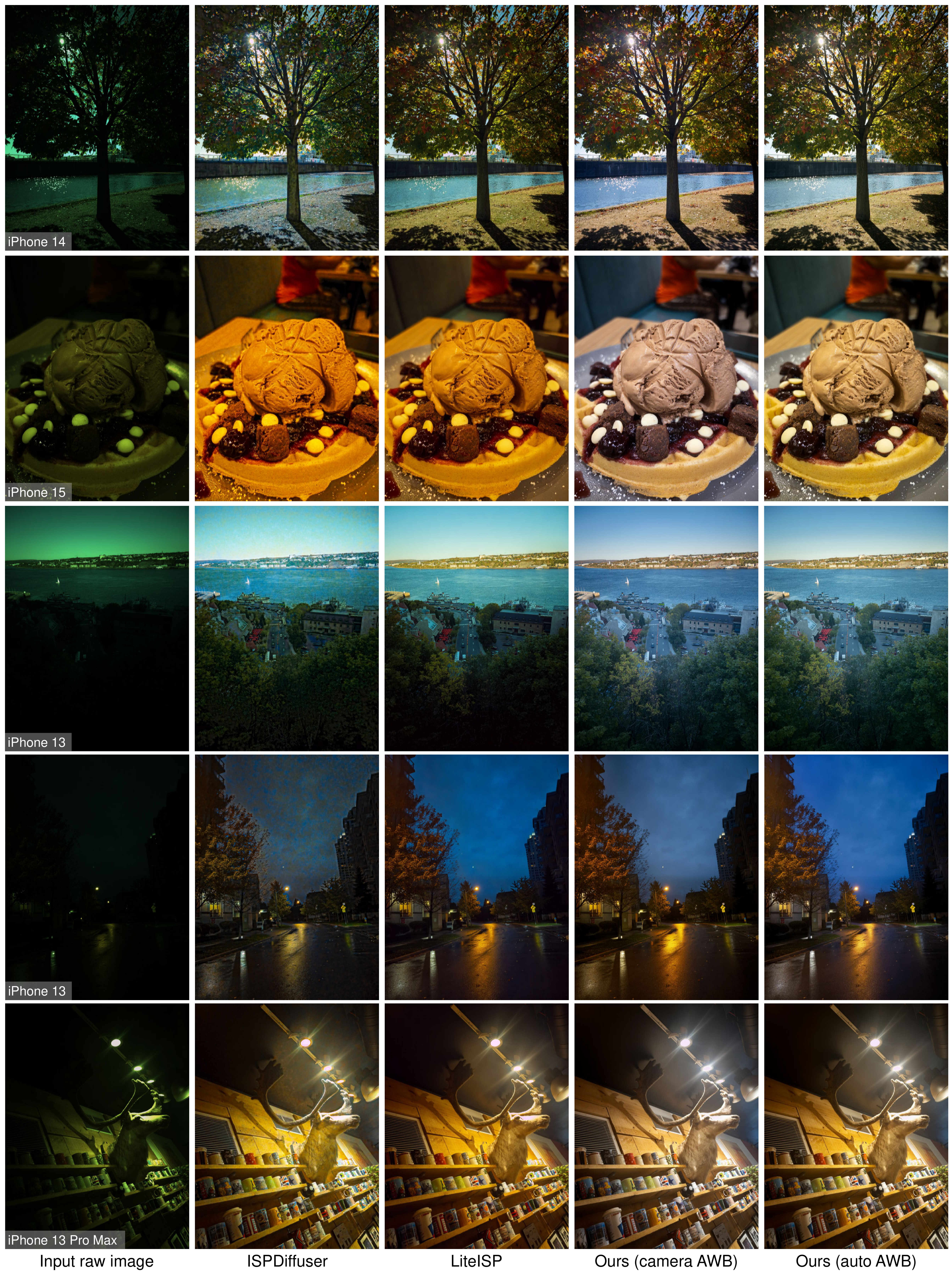}
\vspace{-1mm}
\caption{
Qualitative comparison on unseen cameras (iPhone devices) between our method and the LiteISP~\cite{lite-isp} and ISPDiffuser~\cite{ispdiffuser} models, all trained on the S24 dataset~\cite{s24}. Our method uses the generic denoiser, and none of the training data included iPhone captures. In this example, we show our results under the camera's automatic white balance (AWB) and after applying learned AWB models~\cite{afifi2021c5, zhao2025learning} (see Sec.~\ref{sec:awb} for details).
}
\label{fig:cross-camera-comparison}
\vspace{-2mm}
\end{figure*}

Figure~\ref{fig:cross-camera-comparison}~shows a qualitative comparison between our method (using the generic denoiser along with the photofinishing and detail-enhancement modules trained on the S24 dataset~\cite{s24}) and the LiteISP~\cite{lite-isp} and ISPDiffuser~\cite{ispdiffuser} models, both trained on the same dataset. The comparison is conducted on images captured by various iPhone devices, none of which were included in the training of any model, including ours. Our method produces more consistent and visually pleasing results with greater controllability (e.g., the ability to adjust white balance), while requiring significantly fewer parameters. Specifically, LiteISP contains approximately 9~M parameters and ISPDiffuser around 21~M, whereas our lite version--which was used in this qualitative comparison--requires only about 0.5~M parameters.

Additional results on cameras not seen during training are provided in
Figs.~ \myref{fig:teaser}{\textcolor{section}{1}} and  \myref{fig:ours_vs_indigo}{\textcolor{section}{9}} of the main paper, as well
as in Fig.~\ref{fig:teaser-supp} of this supplemental material. Further
qualitative examples are shown in Figs.~\ref{fig:gui-main-supp},
\ref{fig:s24_vs_adobe5k}, \ref{fig:rerendering}, and
\ref{fig:additional-results} in the following sections.

\section{User Interface for Modular ISP Control}
\label{sec:gui}

\begin{figure*}[!t]
\centering
\includegraphics[width=\linewidth]{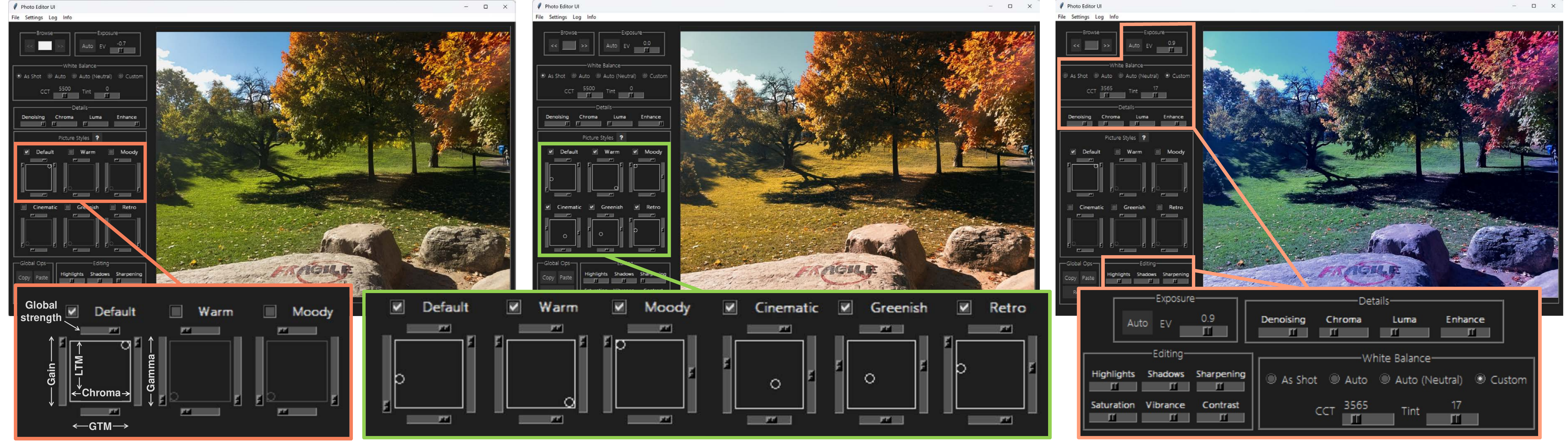}
\vspace{-4mm}
\caption{
User-interactive photo-editing tool built on our modular ISP, providing full control over the rendering process, picture styles, and editing options.
The interface supports selecting or interpolating between styles and adjusting white balance, exposure, color, and overall appearance.}
\label{fig:gui}
\vspace{-2mm}
\end{figure*}

\begin{figure*}[!t]
\centering
\includegraphics[width=\linewidth]{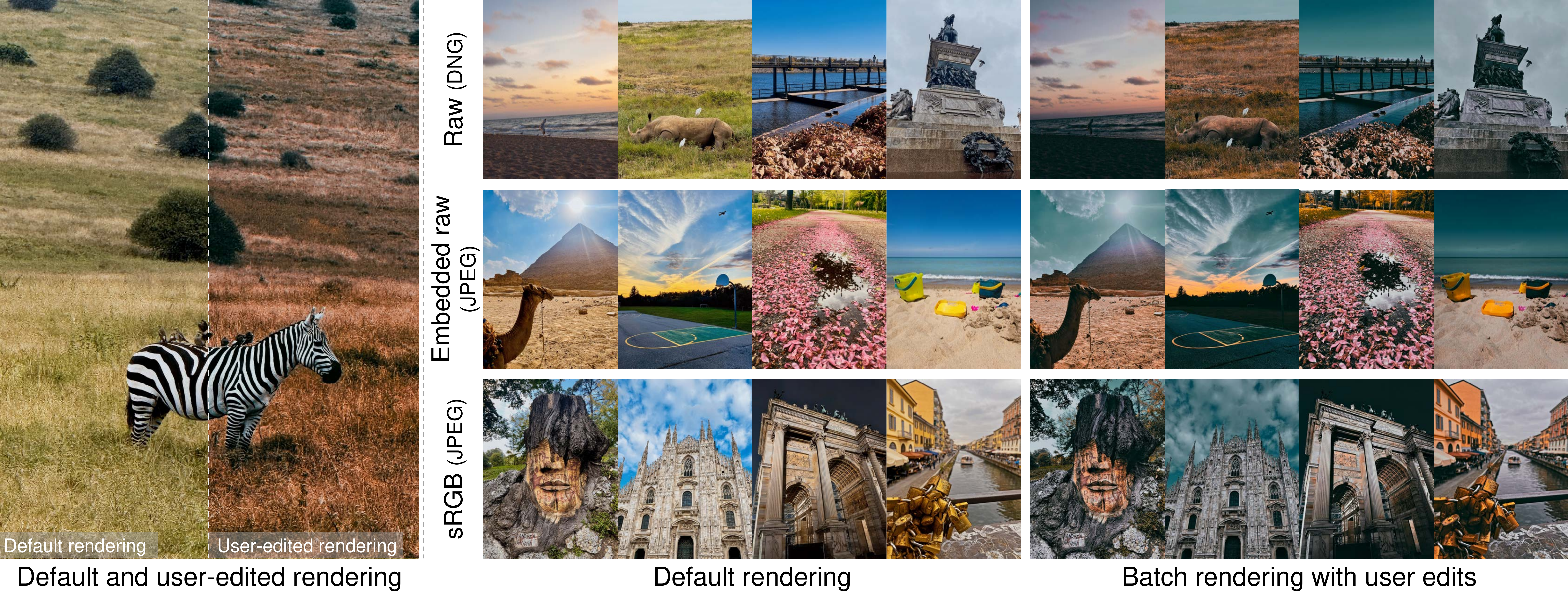}
\vspace{-3mm}
\caption{
Our photo-editing tool offers full control over the modular neural ISP pipeline and includes several interactive editing operators. Our tool supports DNG raw inputs from any camera, as well as rendered sRGB images previously processed by our tool, where compressed raw data is embedded within the JPEG for accurate re-rendering. For sRGB inputs from other cameras, the tool synthesizes raw-like data on the fly for rendering. Users can save custom edits and batch-apply them to multiple images. The shown reference image was captured with a Canon EOS 90D, while other raw and rendered images (with embedded raw) were taken using iPhone 13 and Canon EOS 90D; all sRGB images were captured by the iPhone 13 main camera.}
\label{fig:gui-main-supp}
\vspace{-2mm}
\end{figure*}

\begin{figure*}[!t]
\centering
\includegraphics[width=\linewidth]{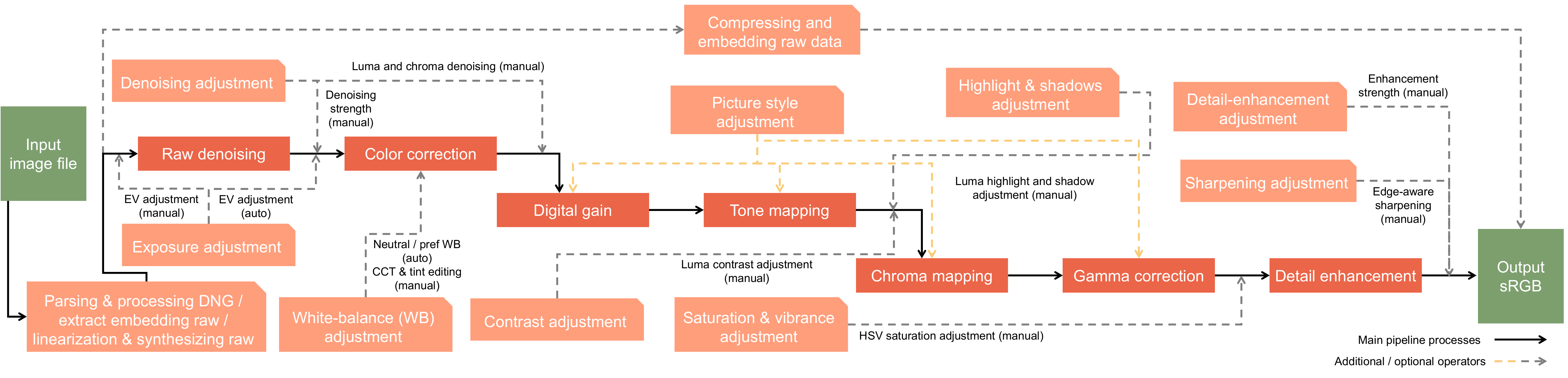}
\vspace{-3mm}
\caption{
Overview of the operators available in our graphical user interface. In addition to the main pipeline processes, the tool provides optional user-controlled operators (e.g., exposure, contrast, highlights, shadows, saturation, and sharpening), allowing fine-grained control over the final rendered image.}
\label{fig:pipeline-gui}
\vspace{-2mm}
\end{figure*}

As described in the main paper, we implemented a graphical user interface (GUI) to enable intuitive interaction with the modular components of our pipeline. Figure~\ref{fig:gui} shows the interface. Our photo-editing tool allows users to control all stages of the pipeline by enabling or disabling modules, adjusting their strengths, selecting from different picture styles, and even blending between styles or intermediate stages. In addition, it provides optional operators (e.g., contrast, highlights, shadows, and others) that can be applied within the pipeline, offering full flexibility to adjust the look and feel of the rendered images. Users can also export their manual edits as new styles and apply them to multiple images in batches. Figure~\ref{fig:gui-main-supp} shows our tool can render DNG raw and rendered sRGB images. Figure~\ref{fig:pipeline-gui} illustrates where these additional operators are integrated within the main pipeline described in the paper. The remainder of this section expands on the implementation details summarized briefly in Sec.~\myref{sec:photo-editing-tool}{\textcolor{section}{2.6}} of the main paper, including camera-agnostic support, raw embedding, external sRGB processing, and memory footprint.

\begin{figure*}[!t]
\centering
\includegraphics[width=0.75\linewidth]{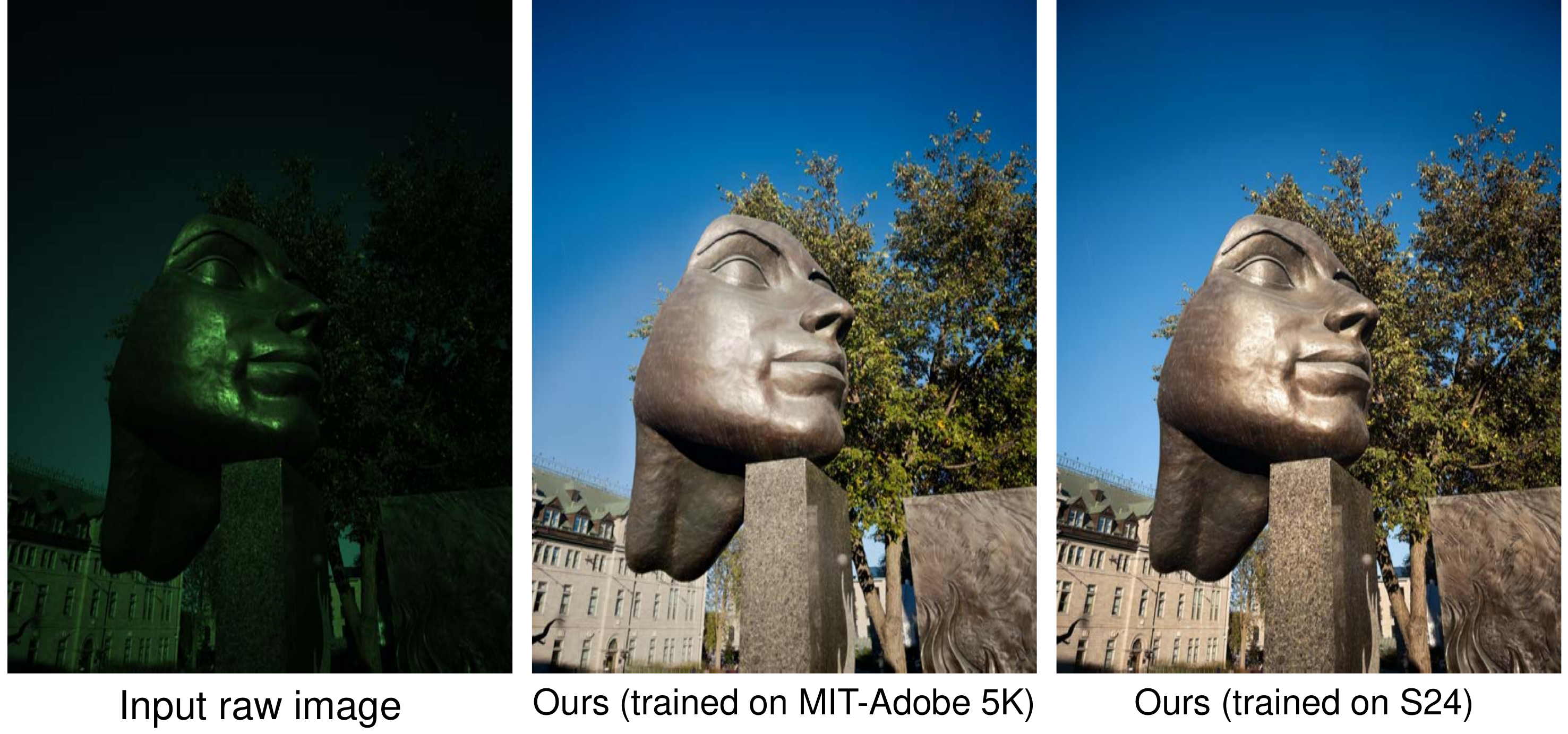}
\vspace{-1mm}
\caption{
Output of our method using photofinishing and detail-enhancement models trained on the S24 dataset~\cite{s24} (default picture style; Style~\#0) and the MIT~Adobe~5K dataset~\cite{Adobe5K} (Expert~C style; see Sec.~\ref{sec:results-adobe-5k} for details). The image was captured with the iPhone~14 main camera. In both cases, our generic base denoising model was used.
}
\label{fig:s24_vs_adobe5k}
\vspace{-2mm}
\end{figure*}

In our implementation, the tool uses the photofinishing models trained on the S24 dataset~\cite{s24} for the default picture style, as we found this model produces more visually plausible results (see Fig.~\ref{fig:s24_vs_adobe5k}). For the LTM, we adopt the proposed multi-scale processing and refinement (Sec.~\ref{sec:artifacts}) as the default configuration, while still allowing the user to disable this option if desired. In addition to the default picture style, we employ models trained on the S24 dataset that support five additional artistic picture styles (Style \#1--5). In our GUI, we refer to them as \texttt{Warm}, \texttt{Moody}, \texttt{Cinematic}, \texttt{Greenish}, and \texttt{Retro}, respectively.
For denoising, we rely on the S24-specific denoising base model for images captured by the S24 main camera, and a generic base denoising model to support images captured by other cameras. For white balancing (Sec.~\ref{sec:awb}), the tool includes two models: a camera-specific model for the S24 main camera and a cross-camera model to support unseen cameras.

Our tool accepts DNG raw images, where we first apply black-level normalization and then perform Menon demosaicing~\cite{menon2007demosaicing} using a parallel, tile-based strategy with partial overlaps between tiles. In practice, we divide the Bayer image into overlapping tiles of 512$\times$512 pixels with a 16-pixel overlap to avoid boundary artifacts. Each tile is demosaiced independently using multiple threads and seamlessly blended into the final full-resolution RGB raw image. This tiled parallel processing improves runtime efficiency (up to a 5$\times$ speed-up on 12-megapixel images compared to the standard full-frame implementation).

The tool further supports post-editable image re-rendering, where compressed raw data is embedded into the final rendered JPEG image (Sec.~\ref{sec:tool-rerendering}). This design allows recovering the raw data for unlimited post-editable re-rendering under different settings, always starting from the raw image and thus avoiding accumulated degradations. Before saving, the raw data is pre-processed by a learned compression model to improve efficiency and keep the JPEG file size compact.

\begin{figure*}[!t]
\centering
\includegraphics[width=\linewidth]{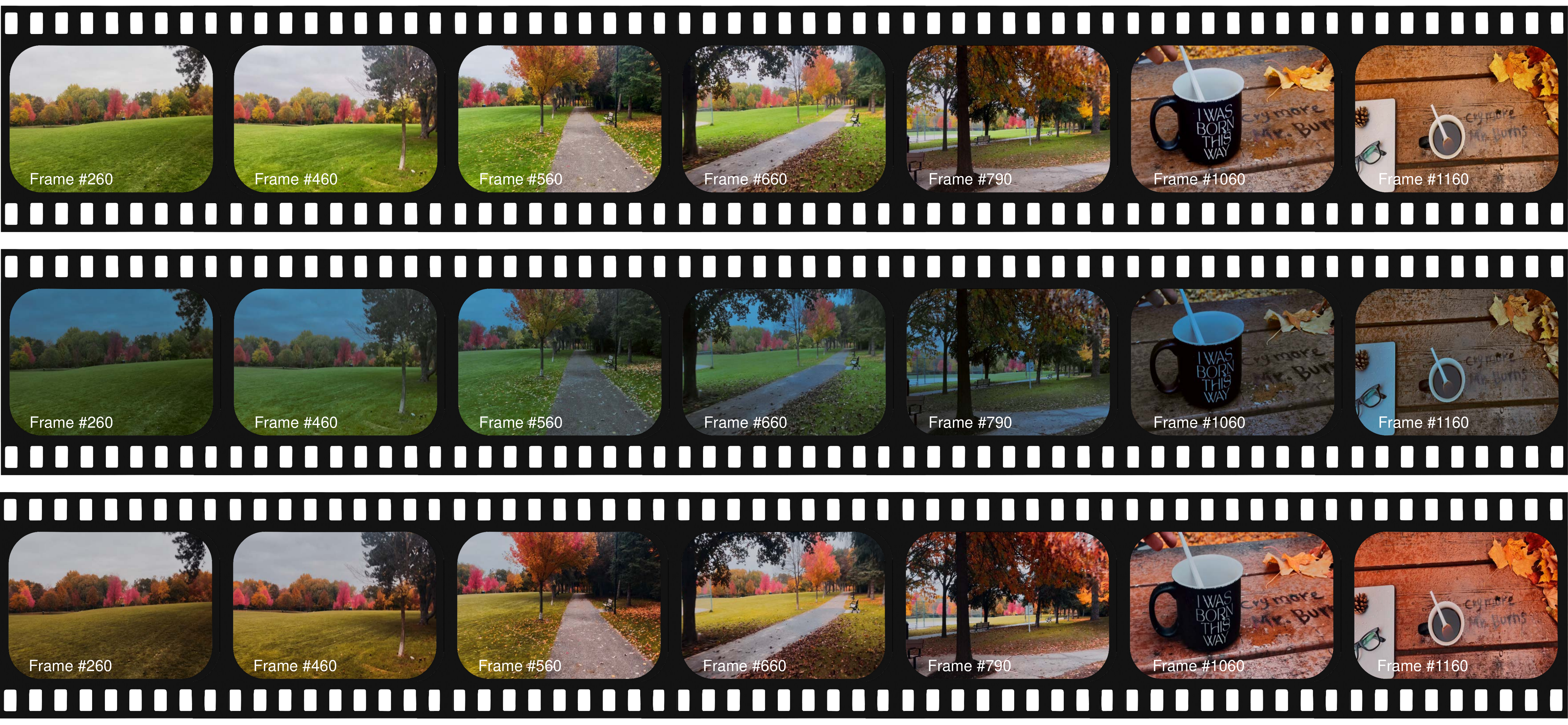}
\vspace{-5mm}
\caption{
Our tool can process sRGB input images (or video frames) and perform in-batch rendering with specific picture styles and edits.
Shown are example frames from sRGB video sequences processed by our method, where each row corresponds to a distinct picture style and edits.
\ifarxiv
See the supplemental \href{https://youtu.be/ByhQjQSjxVM}{video (click to view)}.
\else
\textcolor{highlight}{See the supplemental video.}
\fi
}
\label{fig:video}
\vspace{-2mm}
\end{figure*}

To broaden applicability, the tool can also process sRGB images not originally rendered with our system (and therefore lacking embedded raw data). For this case, we include a learning-based linearization module that maps camera-rendered sRGB images into a camera-agnostic space and synthesizes a raw-like representation, allowing the tool to handle sRGB input images produced outside our pipeline (Sec.~\ref{sec:linearization}). This capability to process sRGB images from third parties, together with the batch-processing option, enables the tool to handle extracted sRGB video frames in a frame-by-frame manner (see Fig.~\ref{fig:video}).
To improve temporal consistency across video frames, we apply a simple temporal smoothing over the current rendering parameters using information from the previous frames within a 9-frame window.
More advanced temporal smoothing approaches are beyond the scope of this paper and are left for future work.

In total, our tool requires loading 3,902,604 parameters (approximately 14.9 MB / 0.015 GB) into GPU memory. This includes the denoising models (camera-specific and generic), photofinishing models for six picture styles (including the default), white balance models (cross-camera and camera-specific), a single detail-enhancement model (we use the one trained for Style \#0 for simplicity, instead of loading style-specific models), the raw pre-processing model (for raw embedding), and the linearization model. As a result, the system remains practical to deploy without heavy GPU memory requirements, especially compared to other methods such as ISPDiffuser~\cite{ispdiffuser}, which alone requires 20,938,890 parameters to render raw to a single picture style (6$\times$ for six styles), or LiteISP~\cite{lite-isp}, which requires 9,094,000 parameters (6$\times$ for six styles). The remainder of this section describes the functionalities of our tool.

\subsection{Denoising Adjustment}
\label{sec:denoising}

The tool provides flexible control over the raw denoising strength.
This control is implemented as a linear interpolation between the image before and after the denoising operation,
modulated by a user-specified scalar parameter in the range $[0,1]$.
A value of $0$ preserves the unprocessed raw image,
a value of $1$ applies the full denoising operation,
and intermediate values yield proportionally mixed results.

In addition to this raw-domain adjustment, we introduce two denoising operators that act in the color-corrected domain—specifically on the linear sRGB image—after the color correction stage. These two operators are applied in the YCbCr representation of the linear sRGB image (where, for simplicity and similar to the photofinishing module, we use the standard RGB-to-YCbCr conversion matrix, which is originally defined assuming the input is in the nonlinear sRGB space).

\noindent\\\textbf{Luma Denoising.}
To reduce luminance noise while preserving structural edges,
we apply an edge-preserving guided filter \cite{gf} to the luma channel, $\mathbf{Y}_{\texttt{LsRGB}}$,
controlled by a radius $r_\texttt{y-d}$ and a regularization term $\epsilon_\texttt{y-d}$.
The output $\mathbf{Y}'_{\texttt{LsRGB}}$ is combined with the original $\mathbf{Y}_{\texttt{LsRGB}}$ through:
\begin{equation}
\mathbf{Y}_{\texttt{y-d}} = (1 - \lambda_{\texttt{y-d}})\,\mathbf{Y}_{\texttt{LsRGB}} + \lambda_{\texttt{y-d}}\,\mathbf{Y}'_{\texttt{LsRGB}},
\end{equation}
where $\lambda_{\texttt{y-d}} \in [0,1]$ determines the luma denoising strength.
Both $r_\texttt{y-d}$ and $\epsilon_\texttt{y-d}$ scale proportionally with $\lambda_\texttt{y-d}$,
allowing a smooth transition from minimal smoothing at low strengths to stronger spatial denoising as $\lambda_\texttt{y-d}$ increases.
This preserves local contrast and fine structure while progressively suppressing luminance grain.
In our implementation, we used $r_\texttt{y-d} = 2 + 10\,\lambda_\texttt{y-d}$ and $\epsilon_\texttt{y-d} = (0.001 + 0.03\,\lambda_\texttt{y-d})^2$,
which provide a balanced trade-off between structure preservation and noise reduction.

\noindent\\\textbf{Chroma Denoising.}
For the chroma components, we apply spatial Gaussian smoothing to suppress color blotches while retaining perceptual color consistency.
The filtered channels $\mathbf{Cb}'_{\texttt{LsRGB}}$ and $\mathbf{Cr}'_{\texttt{LsRGB}}$ are defined as
\begin{equation}
\begin{aligned}
\mathbf{Cb}'_{\texttt{LsRGB}} &= \mathcal{G}(\mathbf{Cb}_{\texttt{LsRGB}};\,\sigma_\texttt{c-d}), \\
\mathbf{Cr}'_{\texttt{LsRGB}} &= \mathcal{G}(\mathbf{Cr}_{\texttt{LsRGB}};\,\sigma_\texttt{c-d}),
\end{aligned}
\end{equation}
where $\mathcal{G}(\cdot;\sigma_\texttt{c-d})$ denotes a separable Gaussian convolution with standard deviation $\sigma_\texttt{c-d}$.
The blur scale $\sigma_\texttt{c-d}$ is adaptively determined based on the user-defined chroma denoising strength $\lambda_\texttt{c-d}$ and the image resolution:
\begin{equation}
\sigma_\texttt{c-d} = \big(3 + 12\,\lambda_\texttt{c-d}\big)\!\left(\frac{H}{3000}\right)^{0.9},
\end{equation}
where $H$ refers to the image height. This adaptive scaling maintains consistent perceptual smoothing across varying image resolutions.
For high chroma denoising strength ($\lambda_\texttt{c-d} > 0.4$), an additional Gaussian pass is applied to further enhance color uniformity in low-texture regions.
The final chroma outputs are blended with the original channels using a nonlinear mixing rule:
\begin{equation}
\begin{aligned}
\mathbf{Cb}_{\texttt{c-d}} &= (1 - \lambda_\texttt{c-d}^{1.5})\,\mathbf{Cb}_{\texttt{LsRGB}}
+ \lambda_\texttt{c-d}^{1.5}\,\mathbf{Cb}'_{\texttt{LsRGB}}, \\
\mathbf{Cr}_{\texttt{c-d}} &= (1 - \lambda_\texttt{c-d}^{1.5})\,\mathbf{Cr}_{\texttt{LsRGB}}
+ \lambda_\texttt{c-d}^{1.5}\,\mathbf{Cr}'_{\texttt{LsRGB}}.
\end{aligned}
\end{equation}

\begin{figure*}[!t]
\centering
\includegraphics[width=0.7\linewidth]{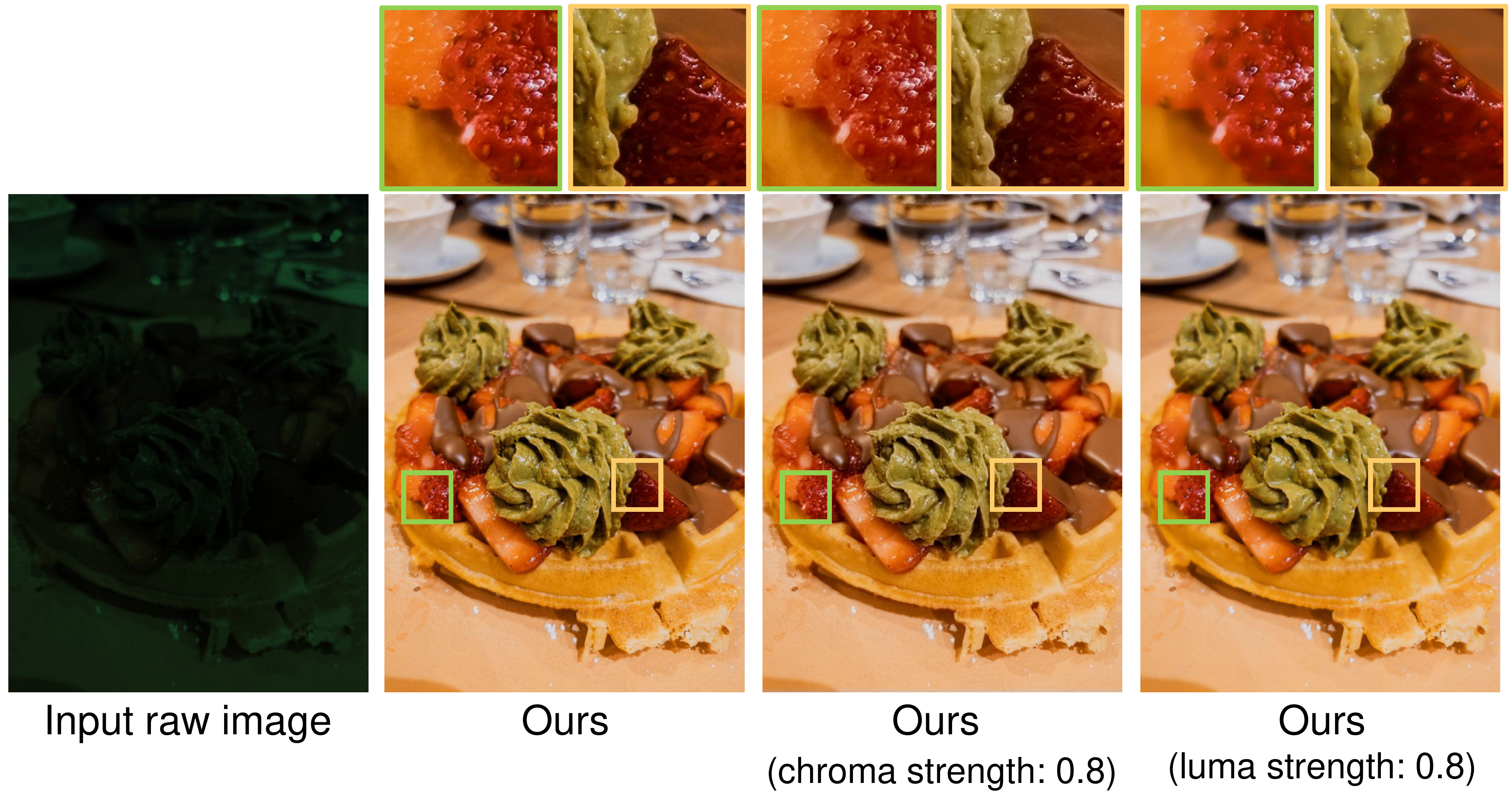}
\vspace{-1mm}
\caption{
Example of luma and chroma denoising.
The shown image was captured using a Samsung~S24 main camera and rendered with our tool.
We show our output without luma/chroma denoising and with both enabled at 80\% strength.
}
\label{fig:chroma_luma_denoising}
\vspace{-2mm}
\end{figure*}

The denoised luma and chroma channels are merged to form the denoised YCbCr version of the linear sRGB image,
which is then transformed back to the linear sRGB space to be processed by the next procedure in our pipeline. Figure~\ref{fig:chroma_luma_denoising} shows an example.

\subsection{Exposure Adjustment}
\label{sec:auto-exposure}
In addition to manual exposure adjustment (defined by an exposure value, EV, applied directly to the raw image and computed as $2^{\text{EV}}$), we provide an optional auto-exposure (AE) adjustment, implemented as a digital brightness compensation applied after conversion to linear sRGB and before the digital gain in the first stage of our photofinishing module.
This step is particularly useful in cross-camera scenarios, where some devices produce raw images that are considerably darker than those encountered during training.
Without such correction, these inputs may be misinterpreted as near-black, leading to insufficient boosting by the photofinishing network.

Our AE approach follows a histogram-based formulation inspired by prior work on exposure correction and tone reproduction~\cite{bernacki2020automatic, tedla2023examining}.
To reduce computation and suppress noise, we first downsample the input to 128$\times$128 pixels using area pooling.
From the downsampled linear sRGB image, we compute a luminance representation $\mathbf{Y}_{E} \in [0,1]^{128 \times 128}$ and construct a 1D histogram with 96 bins.
We then define a Gaussian target histogram over bin centers $b \in [0,1]$ as follows:
\begin{equation}
H^{E}_t(b) \propto \exp\!\left(-\tfrac{1}{2}\left(\tfrac{b - g_{E}}{\sigma_{E}}\right)^2\right),
\end{equation}
centered at a mid-gray value $g_{E}{=}0.08$ with a fixed spread $\sigma_{E}{=}0.05$.
This target distribution encodes the desired tonal balance, with mid-tones dominating while highlights and shadows are preserved.

To determine the exposure correction, we search over a discrete set of EV candidates $\Delta \in [-E, +E]$ with $E{=}1.8$, corresponding to multiplicative scalings of $2^{\Delta}$.
For each candidate, the luminance matrix $\mathbf{Y}_{E}$ is scaled, a histogram is recomputed, and the $\ell_2$ distance to the target distribution is evaluated.
The EV that minimizes this distance is selected:
\begin{equation}
\begin{split}
\Delta^{*}_{E} = \arg\min_{\Delta \in [-E, +E]}
\; \big\| H^{E}(2^{\Delta} \texttt{ }\mathbf{Y}_{E}) - H^{E}_t \big\|_2^2.
\end{split}
\end{equation}
The final output is obtained by scaling the input image by $2^{\Delta^{*}_{E}}$.

\begin{figure*}[!t]
\centering
\includegraphics[width=\linewidth]{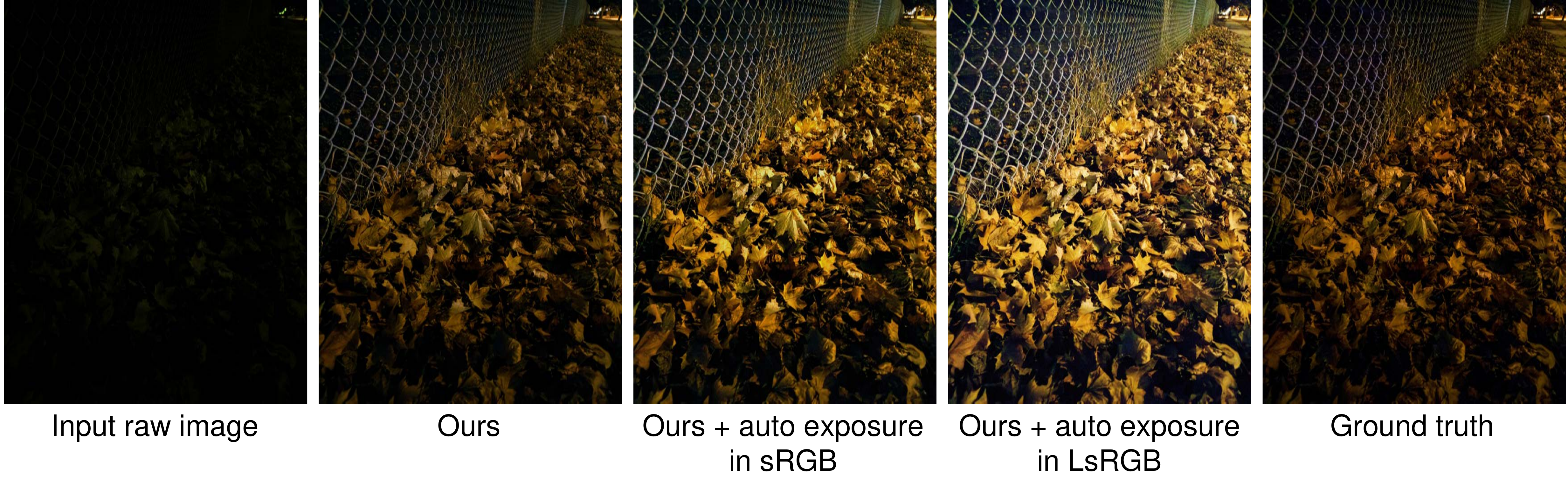}
\vspace{-5mm}
\caption{
We allow applying auto exposure (AE) after conversion to linear sRGB and directly before the photofinishing module.
This figure shows a qualitative example from the S24 validation set~\cite{s24}, including the input raw image, our output without AE, and with AE applied either at the end of the pipeline or in linear sRGB.
The ground-truth sRGB image from the S24 dataset is shown for reference.}
\label{fig:auto-exposure}

\vspace{-2mm}
\end{figure*}

We apply the EV correction in linear sRGB space, since this stage precedes the nonlinear operations in the photofinishing module.
Applying exposure adjustment after nonlinear mapping can limit the ability to adjust exposure effectively~\cite{afifi2021learning}.
Figure~\ref{fig:auto-exposure} illustrates a comparison between applying our AE in linear sRGB versus applying it at the end of the pipeline.

\subsection{White Balance Estimation and Adjustment}
\label{sec:awb}

In all of our main experiments, we relied on the camera's auto white balance (AWB)---the scene illuminant color produced by the on-device AWB estimator embedded in the conventional ISP. Nevertheless, our tool also offers the ability to compute AWB using state-of-the-art illuminant estimation methods. For images captured by the Samsung S24 main camera, we adopted the camera-specific illuminant estimator of \cite{s24}, which we re-trained after removing time and location features since location information is not always available in DNG files. For images captured by other cameras, we employed the cross-camera illuminant estimator C5~\cite{afifi2021c5}. Following \cite{kim2025ccmnet}, we used a modified C5 model that required only the testing image, without the need for additional images from the same camera as in the original C5 method. Specifically, we employed a single-encoder variant that received a 48$\times$48 histogram as input. This model was trained on the NUS dataset~\cite{NUS}, the Cube++ dataset~\cite{cube++}, and the S24 dataset. We followed the training instructions in the original works for both the camera-specific and cross-camera estimators.

Both AWB models (camera-specific and cross-camera) target a neutral white balance, where the goal is for achromatic surfaces to appear gray. However, such neutral estimates do not always match human perceptual preference~\cite{afifi2020deep}, where a bias is often desirable. To account for this, we optionally integrate a post-processing step that applies a nonlinear mapping from the neutral AWB estimate to a perceptual preference-aware AWB, following~\cite{zhao2025learning}.

Our tool also allows manual adjustment of the correlated color temperature (CCT) and tint. For this, we rely on Robertson’s method~\cite{wyszecki2000color} to map between CIE XYZ, CCT, and tint coordinates, combined with an interpolation of the Planckian locus. In this context, the CCT corresponds to the point on the Planckian locus that is closest to the illuminant chromaticity, while the tint represents the signed perpendicular offset of the illuminant from this locus in the CIE 1960 $(u,v)$ chromaticity diagram. By parameterizing AWB in terms of both CCT and tint, we enable intuitive user control: CCT adjusts the overall warmth or coolness of the image, while tint provides fine-grained correction of residual color casts that cannot be addressed by CCT alone.

\begin{figure*}[!t]
\centering
\includegraphics[width=\linewidth]{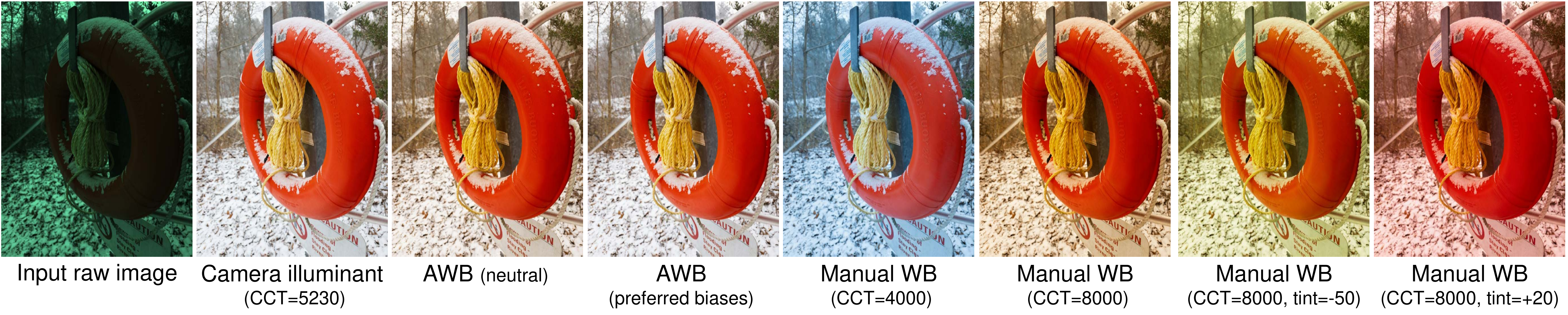}
\vspace{-5mm}
\caption{
Our tool provides different white-balance (WB) options. The image can be rendered using the camera's auto white-balance (AWB) (referred to as ``as shot'' in our tool), or we can re-estimate the illuminant color using a learning-based AWB model. For the latter, we provide two modes: a neutral WB correction that enforces achromatic objects to appear gray, and a preference-aware AWB that applies a bias to better match human perceptual preference. In addition, the user may manually adjust the CCT and tint. Example shown from the test set of the S24 dataset~\cite{s24}.
\label{fig:awb}
}
\end{figure*}

Finally, for the case where the AWB gains are taken from the camera (i.e., the ISP-provided AWB stored in the DNG files), we use the interpolated color correction matrix (CCM) already provided in the DNG files. For all other AWB settings (whether estimated using the camera-specific model or the cross-camera C5 model, in either neutral or preference-biased form, or manually adjusted via CCT and/or tint), we re-compute the CCM that maps the white-balanced raw RGB to linear sRGB. This is achieved by interpolating between the pre-calibrated CCMs available in the DNG metadata, following the DNG specification~\cite{dng}. Figure~\ref{fig:awb} shows an example output from our tool, illustrating different AWB renderings.

\subsection{Contrast Adjustment}
\begin{figure*}[!t]
\centering
\includegraphics[width=\linewidth]{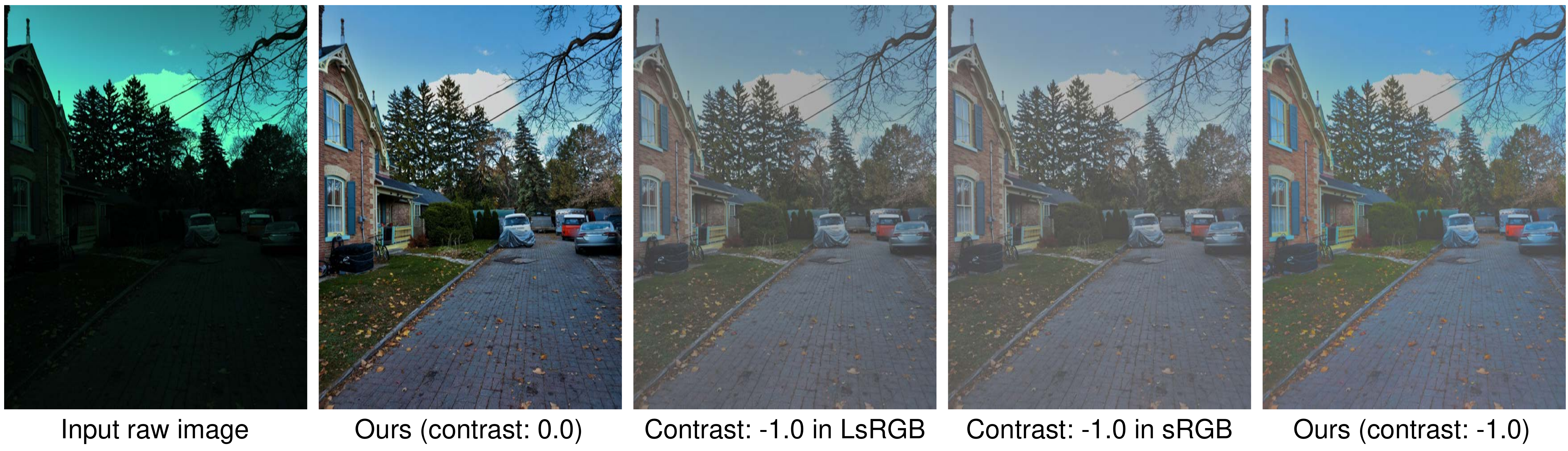}
\vspace{-5mm}
\caption{
We implement a user-controlled contrast adjustment on the luminance channel after local tone mapping.
This figure shows a qualitative example from the S24 validation set~\cite{s24}, including the input raw image,
our output without adjustment (\(\alpha_{\text{contrast}} = 0.0\)),
and with reduced contrast (\(\alpha_{\text{contrast}} = -1.0\)).
For comparison, results are shown when the adjustment is applied after local tone mapping in linear sRGB, after the full photofinishing module, and to the Y channel after local tone mapping
(\(\mathbf{Y}_{\texttt{LTM}}\)) in YCbCr (our choice).
}
\label{fig:contrast}
\end{figure*}

After applying local tone mapping, we optionally include a contrast adjustment step that can be controlled by the user before the chroma mapping stage in our photofinishing module.
This operation modifies the luminance channel \(\mathbf{Y}_{\texttt{LTM}}\) around mid-gray (0.5)
to either increase or decrease the perceived image contrast.
Specifically, given a user-specified adjustment factor \(\alpha_{\text{contrast}} \in [-1,1]\),
we compute:
\begin{equation}
\hat{\mathbf{Y}}_{\texttt{LTM}} = (\mathbf{Y}_{\texttt{LTM}} - 0.5)\,(1 + 0.5 \texttt{ } \alpha_{\text{contrast}}) + 0.5.
\end{equation}
Positive values of \(\alpha_{\text{contrast}}\) increase image contrast
by expanding the difference between dark and bright regions,
while negative values reduce contrast by compressing luminance values toward mid-gray.
This adjustment allows users to fine-tune the global contrast of the image
in addition to the automatic tone mapping provided by the pipeline.
We find that applying contrast adjustment in the luminance channel at this stage
leads to more visually pleasing results compared to applying it directly
to the three channels of the processed linear sRGB either after local tone mapping
or after the full photofinishing module (see Fig.~\ref{fig:contrast}).

\subsection{Highlights and Shadows Adjustment}
We allow the user to optionally adjust highlights and shadows after the local tone-mapping stage.
Both operations are applied to the luminance channel ($\mathbf{Y}_{\texttt{LTM}}$) of the YCbCr local tone-mapped image.
We first construct smooth masks that isolate highlight and shadow regions:
\begin{align}
\mathbf{M}_{\texttt{high}} &= \varsigma(\mathbf{Y}_{\texttt{LTM}};\, \tau_{\texttt{high}},\, \tau_{\texttt{high}}+\epsilon_{\texttt{high}}), \\
\mathbf{M}_{\texttt{shad}} &= 1 - \varsigma(\mathbf{Y}_{\texttt{LTM}};\, \tau_{\texttt{shad}},\, \tau_{\texttt{shad}}+\epsilon_{\texttt{shad}}),
\end{align}
where $\tau_{\texttt{high}}$ and  $\tau_{\texttt{shad}}$ are thresholds for highlight and shadow regions,
$\epsilon_{\texttt{high}}$ and  $\epsilon_{\texttt{shad}}$ control the transition smoothness,
and $\varsigma$ denotes the smoothstep interpolation defined as:
\begin{equation}
\varsigma(x;\, e_0, e_1)
= \left(\frac{x - e_0}{e_1 - e_0}\right)^{\!2},
\end{equation}
with $e_0$ and $e_1$ denoting the lower and upper transition edges.
To prevent over-amplification near saturation, adjustments are limited to the valid luminance range [0.1, 0.9].
In our implementation, we set $\tau_{\texttt{high}}=0.7$, $\tau_{\texttt{shad}}=0.3$,
and $\epsilon_{\texttt{high}}=\epsilon_{\texttt{shad}}=0.1$. The adjusted luminance channels are then computed as follows:
\begin{align}
\mathbf{Y}'_{\texttt{high}} &= \mathbf{Y}_{\texttt{LTM}} + \frac{\alpha_{\texttt{high}}}{20} \texttt{ } \mathbf{Y}_{\texttt{LTM}} \texttt{ } \mathbf{M}_{\texttt{high}}, \\
\mathbf{Y}'_{\texttt{shad}} &= \mathbf{Y}_{\texttt{LTM}} + \frac{\alpha_{\texttt{shad}}}{20} \texttt{ } (1 - \mathbf{Y}_{\texttt{LTM}}) \texttt{ } \mathbf{M}_{\texttt{shad}},
\end{align}
where $\alpha_{\texttt{high}}, \alpha_{\texttt{shad}} \in [-1,1]$ are user-controlled parameters.
Positive $\alpha_{\texttt{high}}$ values boost highlights, while negative values compress highlights to recover detail.
Positive $\alpha_{\texttt{shad}}$ values lift dark regions, whereas negative values deepen them.

\begin{figure*}[!t]
\centering
\includegraphics[width=\linewidth]{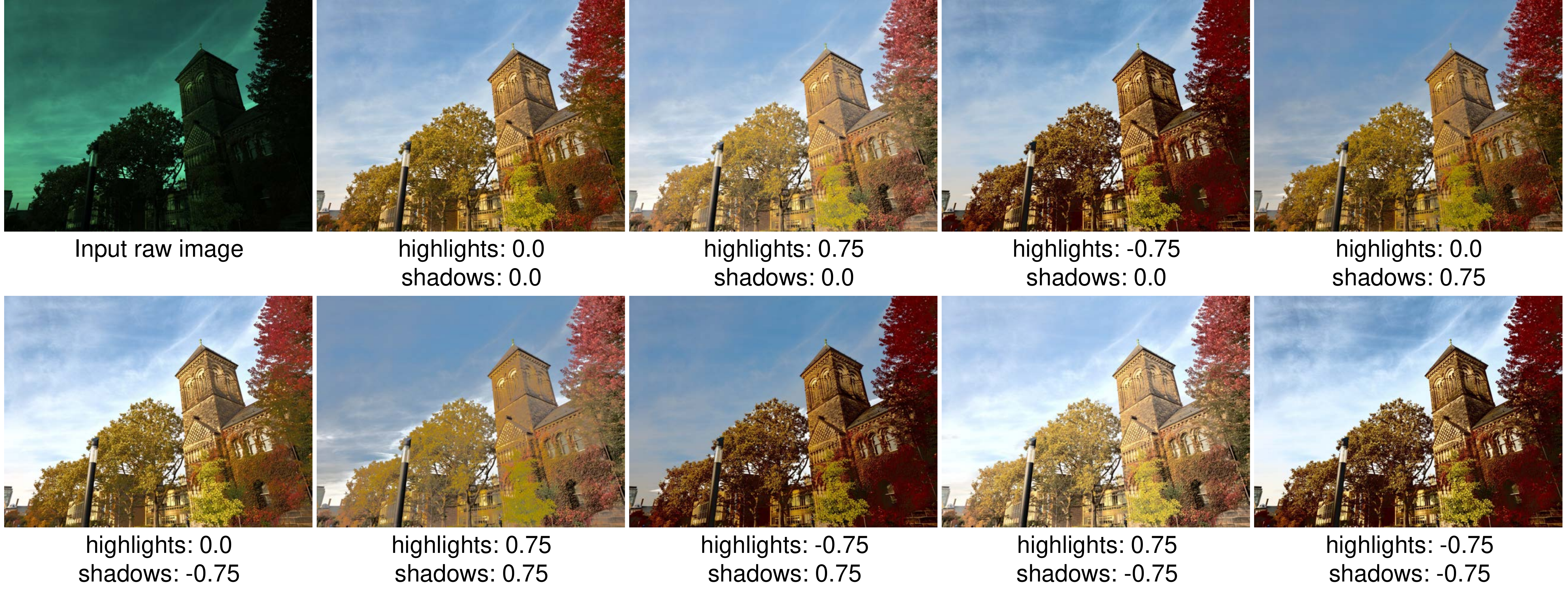}
\vspace{-5mm}
\caption{
We allow the user to optionally adjust highlights and shadows after applying local tone mapping.
Shown here is an example from the S24 validation set~\cite{s24}, including the input raw image and our outputs with different values of the highlight and shadow control parameters,
$\alpha_{\texttt{high}}$ and $\alpha_{\texttt{shad}}$, respectively.
}
\label{fig:highlight-shadow}
\vspace{-1mm}
\end{figure*}

Finally, the adjusted luminance is combined with the original chroma channels to reconstruct the full YCbCr image,
which is then propagated through the remainder of the pipeline.
This provides intuitive, user-controlled tonal adjustments while preserving color fidelity.
See Fig.~\ref{fig:highlight-shadow} for visual examples.

\subsection{Saturation and Vibrance Adjustment}
We optionally allow user-controlled color saturation and vibrance adjustments in the HSV color space after gamma correction in our photofinishing module. Working in the HSV color space is numerically stable because its saturation channel is bounded in $[0,1]$, and it provides an intuitive axis for perceptual editing.

For each pixel, given its HSV representation $(p_h, p_s, p_v)$, the saturation adjustment rescales the saturation channel uniformly across all colors as:
\begin{equation}
p_s' = \min\big(p_s \texttt{ } (1 + \alpha_{\texttt{sat}}), 1\big),
\end{equation}
where $\alpha_{\texttt{sat}}$ is the saturation control parameter. Positive values of $\alpha_{\texttt{sat}}$ increase global saturation, while negative values move the image toward grayscale.

In contrast, vibrance selectively boosts muted colors while preserving already saturated regions. The vibrance transformation is defined as:
\begin{equation}
p_s' = \min\big(p_s \texttt{ } (1 + \alpha_{\texttt{vib}} \texttt{ } (1-p_s)), 1\big),
\end{equation}
where $\alpha_{\texttt{vib}}$ is the vibrance strength. Positive values of $\alpha_{\texttt{vib}}$ increase vibrance by amplifying low-saturation colors more strongly than highly saturated ones, whereas negative values reduce vibrance.

\begin{figure*}[!t]
\centering
\includegraphics[width=\linewidth]{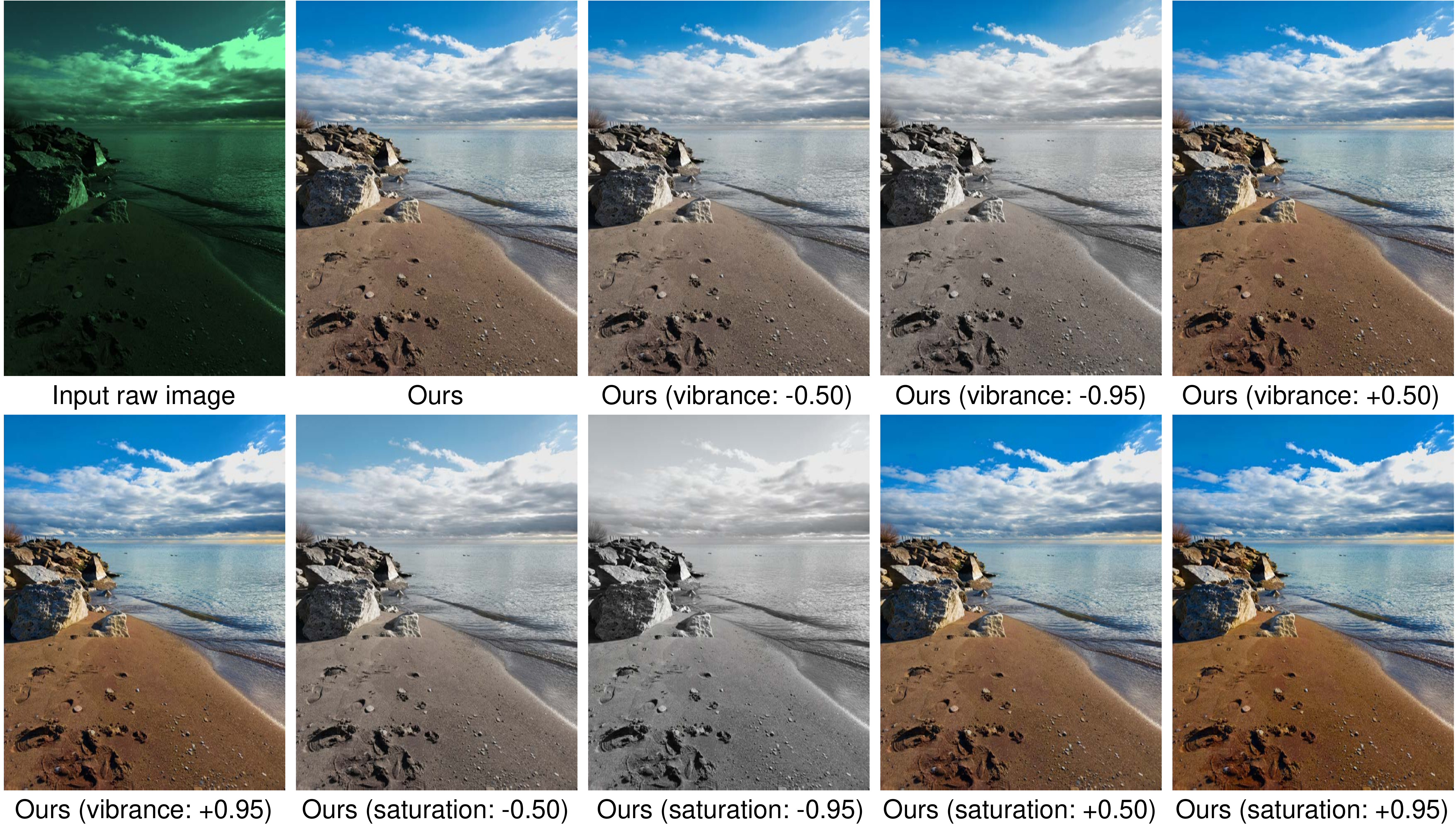}
\vspace{-5mm}
\caption{
We allow the user to optionally adjust the saturation and vibrance in our pipeline. Shown here is an example from the S24 test set~\cite{s24}, including the input raw image, our output without adjustment (\(\alpha_{\texttt{sat}}=\alpha_{\texttt{vib}}=0.0\)), and results with different saturation and vibrance settings (\(\alpha \in \{-0.5, +0.5, -0.95, +0.95\}\)). Saturation uniformly scales chroma across all colors, whereas vibrance selectively enhances muted colors while preserving already saturated regions.
}
\label{fig:sat_vib}
\vspace{-2mm}
\end{figure*}

After applying either transformation, the adjusted HSV values $(p_h, p_s', p_v)$ are converted back to the sRGB space to continue processing in the pipeline. Figure~\ref{fig:sat_vib} illustrates an example from the S24 test set~\cite{s24}, rendered with our photofinishing pipeline using different saturation and vibrance settings. As shown, the saturation adjustment uniformly scales the chroma across all colors, whereas vibrance selectively boosts muted colors while preserving already saturated regions.

\subsection{Detail-Enhancement Control}
\label{sec:enhancement-control}

Similar to the denoising strength control, our tool provides continuous control over the detail-enhancement strength.
This is implemented as a linear interpolation between the image before and after the detail enhancement operation,
modulated by a scalar parameter in the range $[0,1]$.
A value of $0$ preserves the unprocessed image, a value of $1$ applies the full operation,
and intermediate values yield proportional effects.

\subsection{Sharpening Adjustment}
We apply an optional, user-controlled, edge-aware sharpening step at the final stage of our pipeline, operating on the full-resolution sRGB image.
Let $\mathbf{I}_{\texttt{in}}$ denote the input to the sharpening operator (equivalent to $\mathbf{I}_{\texttt{out}}$, the final output of our pipeline).
We first compute a smoothed base layer $\mathbf{B}_{\texttt{sharp}}$ using a Gaussian blur with kernel size $\rho_{\texttt{sharp}}=3$ and standard deviation $\sigma_{\texttt{sharp}}=1.0$.
The high-frequency detail layer is then defined as:
\begin{equation}
\mathbf{D}_{\texttt{sharp}} = \mathbf{I}_{\texttt{in}} - \mathbf{B}_{\texttt{sharp}}.
\end{equation}

\begin{figure*}[!t]
\centering
\includegraphics[width=0.75\linewidth]{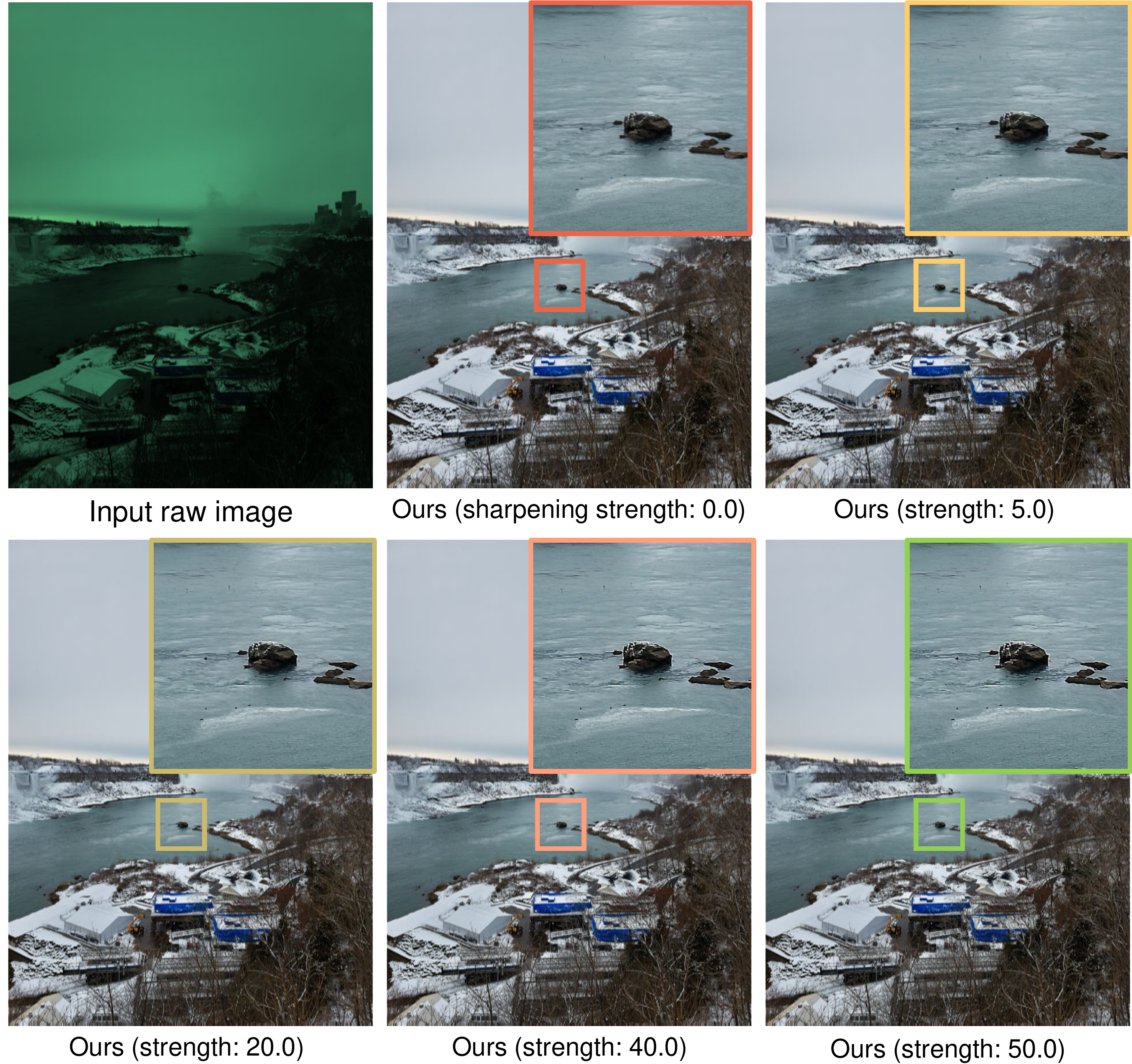}
\vspace{-1mm}
\caption{
We allow the user to apply sharpening as the final step of our pipeline. The example shown is from the validation set of S24~\cite{s24}, including the input raw image and our rendered outputs with different sharpening strengths $\alpha_{\texttt{sharp}}$.
Increasing $\alpha_{\texttt{sharp}}$ enhances edge contrast, with $\alpha_{\texttt{sharp}}=0$ corresponding to no sharpening.
}
\label{fig:sharp}
\vspace{-2mm}
\end{figure*}

To prevent uniform sharpening of noisy flat regions, we construct an edge-aware mask.
Horizontal and vertical gradients are estimated by convolving the input image with simple derivative kernels, producing response maps $\mathbf{G}_x$ and $\mathbf{G}_y$.
Gradients are computed independently per channel using $[-1, 0, 1]$ derivative kernels, and the resulting magnitudes are averaged to form a normalized edge mask.
The edge magnitude map is then computed as:
\begin{equation}
E_{\texttt{sharp}(x,y)} = \sqrt{G_{x(x,y)}^2 + G_{y(x,y)}^2},
\end{equation}
which is normalized to form the edge mask $\mathbf{M}_{\texttt{sharp}} \in [0,1]$.
This mask ensures that sharpening is applied predominantly along edges. The final sharpened image is obtained as:
\begin{equation}
\mathbf{I}'_{\texttt{sharp}} = \mathbf{I}_{\texttt{in}} + \alpha_{\texttt{sharp}} \mathbf{D}_{\texttt{sharp}} \mathbf{M}_{\texttt{sharp}},
\end{equation}
where $\alpha_{\texttt{sharp}}$ controls the sharpening strength.
Higher values of $\alpha_{\texttt{sharp}}$ increase edge contrast, while setting $\alpha_{\texttt{sharp}}=0$ disables sharpening.
Figure~\ref{fig:sharp} shows qualitative results of the sharpening operator applied to an example image.

\subsection{Editing Picture Styles}
As our photofinishing module governs the overall appearance and perceived ``look and feel'' of the final image, and given its modular design, we provide users with fine-grained control over each operator and enable flexible style editing. Specifically, our tool allows: 1) disabling individual operators (e.g., tone mapping, gamma correction) within a selected picture style, 2) interpolating operators' parameters between multiple picture styles, or 3) replacing specific operators of one style with those of another target style. We implement style mixing through simple linear interpolation of the operator parameters predicted by the corresponding photofinishing networks of each selected style. The interpolated parameters are then applied once through our photofinishing module, ensuring consistent and efficient rendering.

\begin{figure*}[!t]
\centering
\includegraphics[width=\linewidth]{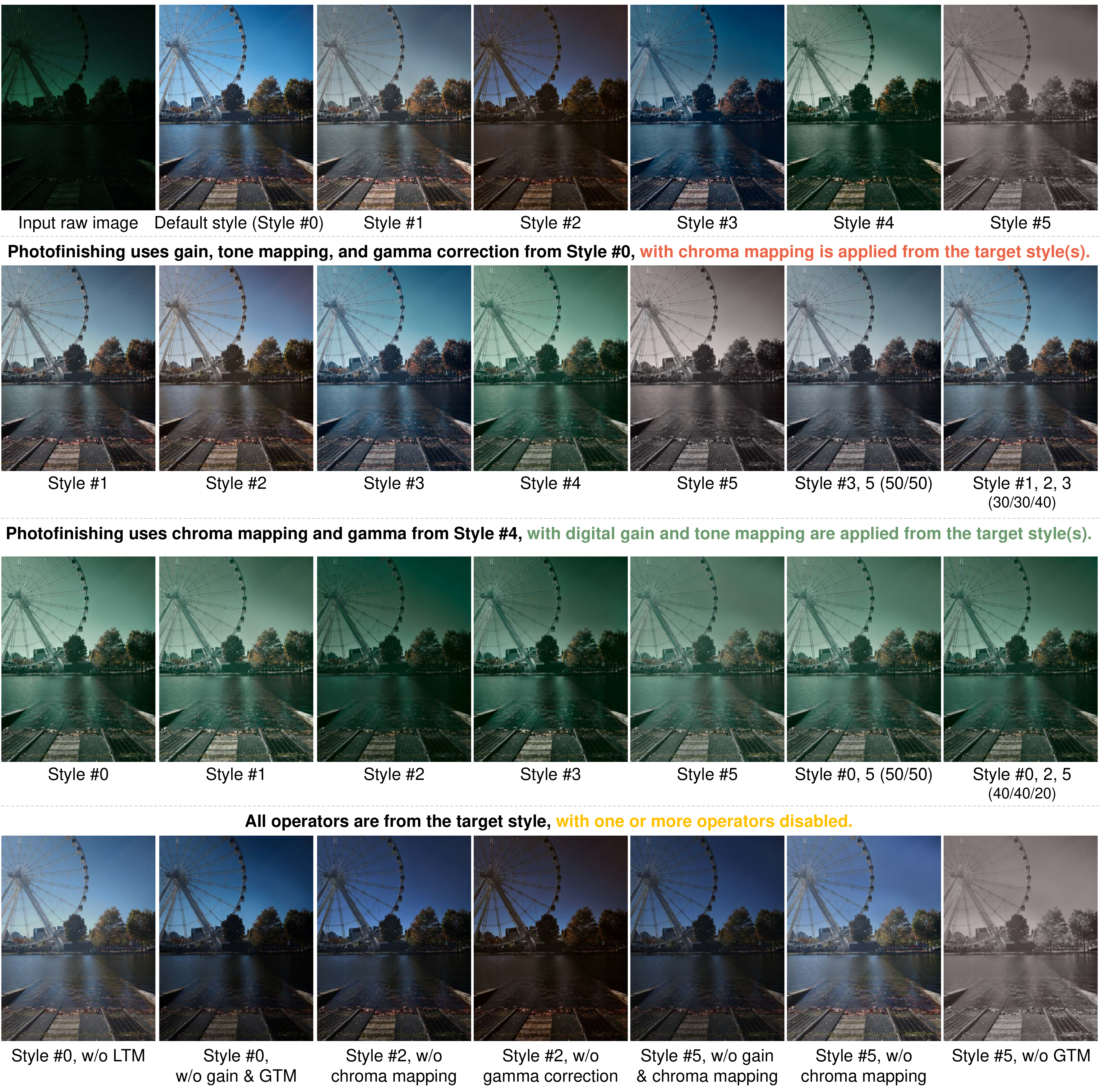}
\vspace{-5mm}
\caption{
Our modular photofinishing design provides users with full control over the picture style of the final image.
In addition to selecting from pre-defined picture styles (each rendered by a pre-trained photofinishing module),
users can also mix operators from different styles or disable specific operators entirely.
The shown example was captured with the iPhone~13 main camera, which is an unseen camera for our pre-trained photofinishing modules.
}
\label{fig:mix-styles}
\vspace{-2mm}
\end{figure*}

Figure~\ref{fig:mix-styles} shows an example image captured with an iPhone~13 (an unseen camera in our training) and processed using our generic denoiser and the photofinishing module (trained on images taken by the S24's main camera) with different picture styles. The figure demonstrates three types of style mixing: 1) operator replacement (e.g., rendering the image with Style~\#0 while substituting chroma mapping from another style), 2) operator interpolation (e.g., performing chroma mapping by interpolating 50\% from Style~\#3 and 50\% from Style~\#5), and 3) operator disabling (e.g., rendering with all operators of Style~\#5 except global tone mapping, which is disabled).

\subsection{Re-Rendering with Embedded Raw}
\label{sec:tool-rerendering}

After rendering the raw image to sRGB, our tool embeds the original raw data within the JPEG container of the rendered output. This was achieved by first compressing the raw image as a JPEG using the Raw-JPEG Adapter \cite{afifi2025jpeg}, and then appending this raw-JPEG file (along with the operator parameters, DNG metadata, and the current editing configuration of the image in the tool) after the end-of-image (EOI) marker of the sRGB JPEG. In this way, the visible JPEG remains fully standards-compliant and decodable by any conventional viewer, while our tool can still access the embedded raw data for re-rendering and further adjustments.

This design offers several advantages. First, it ensures backward compatibility, as the sRGB JPEG remains viewable on all platforms without modification. At the same time, it enables the distribution of a single JPEG file that contains both the standard viewable image (compatible with any JPEG decoder) and the corresponding raw data (accessible by our tool), with a small increase in storage size compared to saving the raw data separately without compression (e.g., in DNG format). Lastly, by embedding the operator parameters, the user can later reset all manual adjustments and reapply a new set exactly as applied during the initial rendering, allowing unlimited post-saving edits without needing to access the original raw DNG file.

For the Raw-JPEG Adapter~\cite{afifi2025jpeg}, we provide two quality settings: raw-JPEG quality~{=}~95, which typically adds about 2-3~MB to the sRGB JPEG file size, and raw-JPEG quality~{=}~75, which adds around 1-2~MB. These values represent a modest storage overhead compared to alternatives such as HEIC on iPhone, which also supports post-capture re-rendering through Apple’s Photos editing tool but often produces larger files (exceeding 10~MB in some scenes or picture styles) and DNG files, which typically range from 12~MB to 35~MB per 12-megapixel image. In contrast, for 12-megapixel images, our approach of appending the JPEG-compressed raw data to the final image requires, on average, only about 5-6~MB in total at raw-JPEG quality~{=} 95 (including approximately 3~MB for the sRGB JPEG) and about 3-4~MB at raw-JPEG quality~{=} 75.

To generate a raw-JPEG compressed file, we pre-process the raw images before JPEG compression using trained models corresponding to each target JPEG quality, as described in~\cite{afifi2025jpeg}. During decoding, and prior to JPEG decompression, we inverted the Raw-JPEG Adapter operators using the stored metadata of the operator parameters embedded in the raw JPEG file.

In our implementation, we employed the Raw-JPEG Adapter model variant without the DCT component, which was recommended for better generalization to unseen cameras during training \cite{afifi2025jpeg}. The Raw-JPEG Adapter model is lightweight, with only 32,076 parameters, and introduces an average overhead of about 0.3 seconds for raw image processing before encoding and 0.1 seconds for restoration.

\begin{figure*}[!t]
\centering
\includegraphics[width=\linewidth]{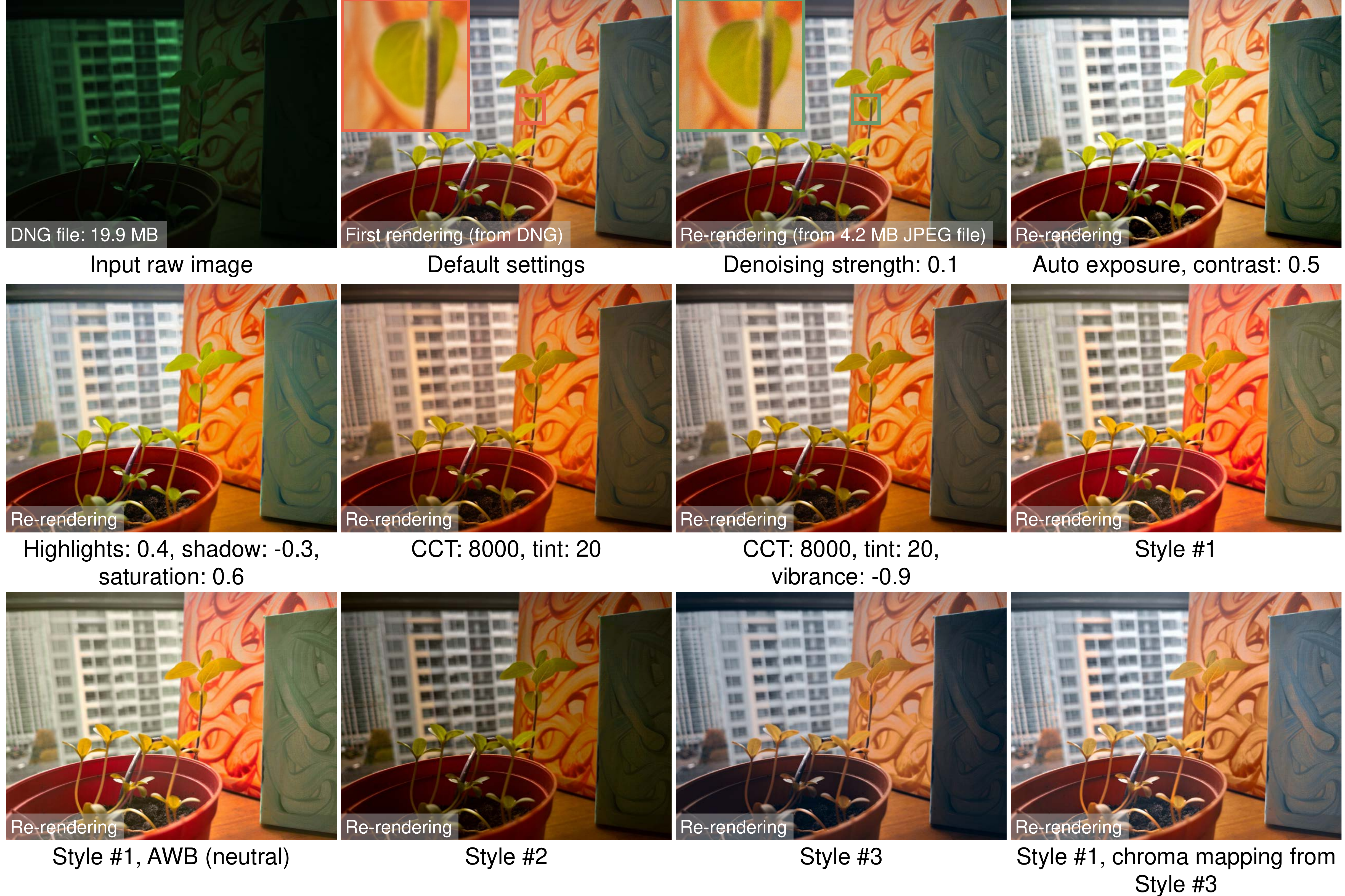}
\vspace{-5mm}
\caption{
Our framework enables post-editable re-rendering of saved JPEG images with the same functionality available to the original raw data. This is achieved by embedding the raw image in a compact form alongside the final sRGB rendering. Shown here is an example image captured with an iPhone 13 and processed by our tool. The first result shows the default rendering, while the remaining results are obtained by re-rendering from the restored raw image using different rendering settings.
}
\vspace{-2mm}
\label{fig:rerendering}
\end{figure*}

With this design, our tool can re-render or edit images with the same level of functionality as when operating directly on the original DNG raw files during the post-render stage, even after multiple re-rendering operations without cumulative degradation in accuracy. This is achieved by preserving the complete raw data within the final JPEG file, with only a slight loss in accuracy due to compression. Figure~\ref{fig:rerendering} demonstrates this capability, showing an example where an image saved using our tool is later re-rendered both with the same settings and with different styles and parameter configurations. A quantitative evaluation is provided in Sec.~\ref{sec:re-rendering-results}, where we compare our design against prior re-rendering alternatives, showing that embedding the raw data within the JPEG enables higher-quality re-rendering results.

\subsection{Processing Input sRGB Images}
\label{sec:linearization}
To broaden the applicability of our framework beyond raw images, we also support standard 8-bit sRGB inputs (e.g., JPEG/PNG) that are not accompanied by embedded raw data. Since our pipeline is designed to operate on raw images, we first linearize the input sRGB image using the method of \cite{afifi2021cie}, where a lightweight network maps the input sRGB image to the CIE XYZ linear space.

We then follow the CIE XYZ-to-raw mapping strategy described in Sec.~4.2.4 of \cite{afifi2021cie}. In this step, we fix the CIE XYZ-to-raw 3$\times$3 matrix to the one corresponding to the S24 main camera under D50 illumination (CCT=5000K). The mapped raw image is then multiplied by the D65 light color in the raw space, yielding a synthetic raw-like representation of the input sRGB image within the S24 main camera’s raw space. Unlike the original design in \cite{afifi2021cie}, which employed two networks to map between sRGB and CIE XYZ in both directions, we only use a single network to map from sRGB to CIE XYZ. To further reduce computational cost and memory usage, we use a lightweight variant of the original architecture. Specifically, we reduced the local sub-network depth in \cite{afifi2021cie} from 16 to 8 convolutional blocks, and the number of channels per convolutional layer from 32 to 24. For the global sub-network in \cite{afifi2021cie}, we use a depth of 4 (instead of 5) with 24 channels (instead of 64), and reduce the final three fully connected layers from [1024, 1024, 1024] to [512, 256, 256], while disabling the dropout layer. After these modifications, the total number of parameters in the linearization model is 693,493 ($\sim$2.7 MB), resulting in only a light additional load on the GPU.

We trained this lightweight linearization model for 300 epochs using the Adam optimizer \cite{kingma2014adam} on 512$\times$512 non-overlapping patches from the sRGB-to-CIE XYZ dataset \cite{afifi2021cie}, with a mini-batch size of 8 and an L2 regularization factor of $10^{-6}$ (in contrast to $10^{-3}$ used in the original work).

\begin{figure*}[!t]
\centering
\includegraphics[width=0.85\linewidth]{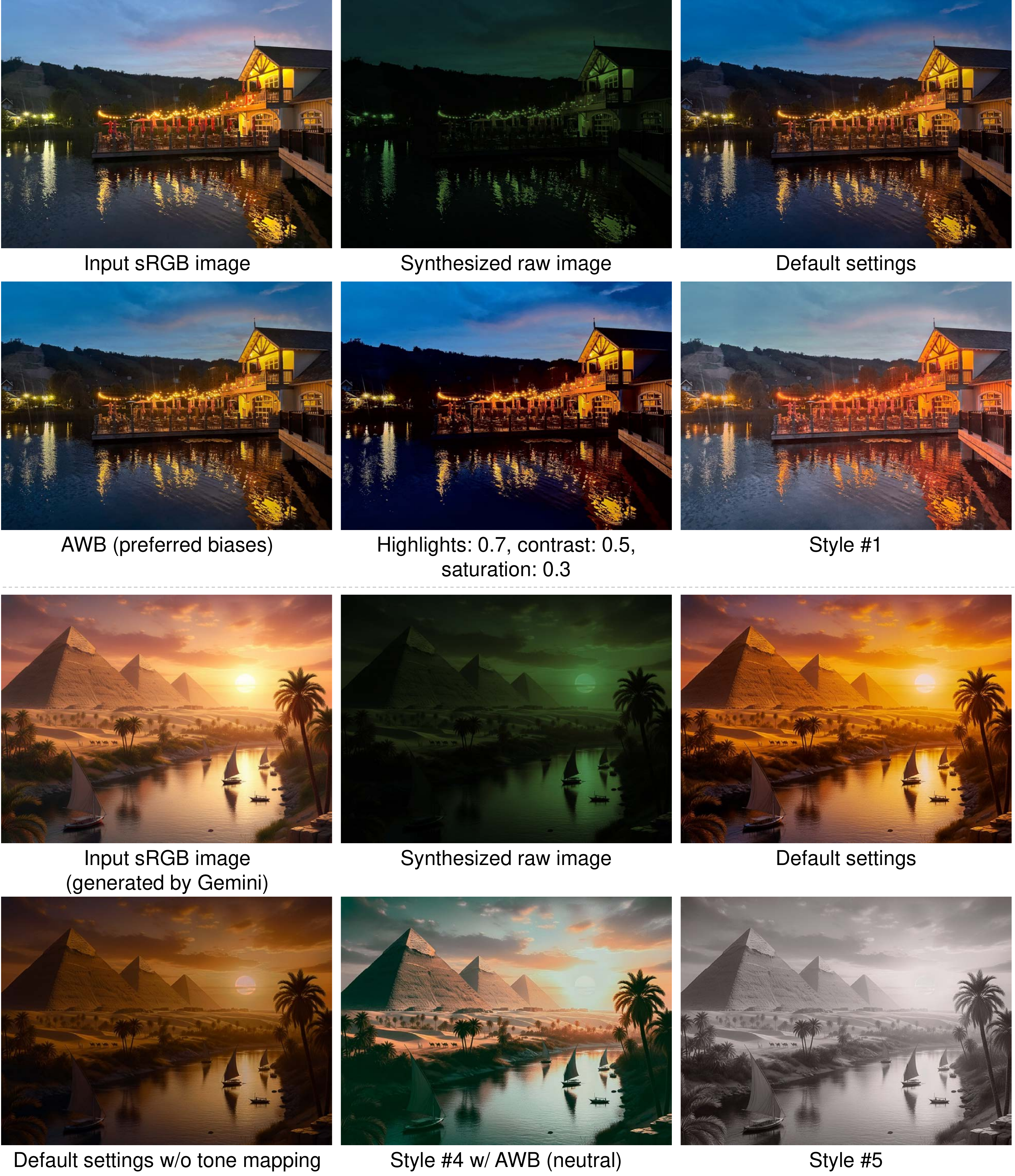}
\vspace{-2mm}
\caption{
To broaden the applicability of our tool to sRGB images saved without embedded raw data (i.e., outside our framework), we linearize the input sRGB images (whether produced by unknown ISPs or generated by AI models) into synthetic raw-like representations, thereby enabling the full functionality of our pipeline. The top example shows an image captured and rendered by the iPhone camera ISP, from which we generate a raw-like image and process it with different settings using our tool. The second example is an image generated by the Gemini 2.5 Flash Image model with the prompt: \texttt{Generate an ancient Egyptian scene with pyramids, the Nile, and palm trees glowing under a golden sunset}. We convert it into a raw-like image and apply our tool with different rendering options.
}

\label{fig:rendering-srgb-images}
\vspace{-2mm}
\end{figure*}

With this synthetic raw conversion process, our tool remains applicable to sRGB inputs rendered by any camera or software ISP, as well as to synthetically generated sRGB images—albeit with reduced accuracy due to residual nonlinearities that cannot be fully removed by linearization. Figure~\ref{fig:rendering-srgb-images} illustrates two examples: the first is an sRGB image captured and rendered by the iPhone’s native camera ISP, and the second is generated by the Gemini 2.5 Flash Image model. Both images are linearized, mapped to the S24 camera raw space, and subsequently rendered with different settings using our tool.

\section{User Study Details}
\label{sec:user_study}

As discussed in the main paper, we conducted a user study to compare our method with the Samsung~S24 native camera ISP and Adobe Lightroom. We captured 45 scenes with the S24 main camera in Pro mode (DNG raw files) and captured the same scenes using the native camera application. The scenes covered diverse conditions, including indoor, outdoor daylight, sunset, and low-light environments. The DNG files were processed using Adobe Lightroom (auto enhancement) and our method, yielding three versions per scene: ours, native ISP, and Lightroom.

\begin{figure*}[!t]
\centering
\includegraphics[width=0.7\linewidth]{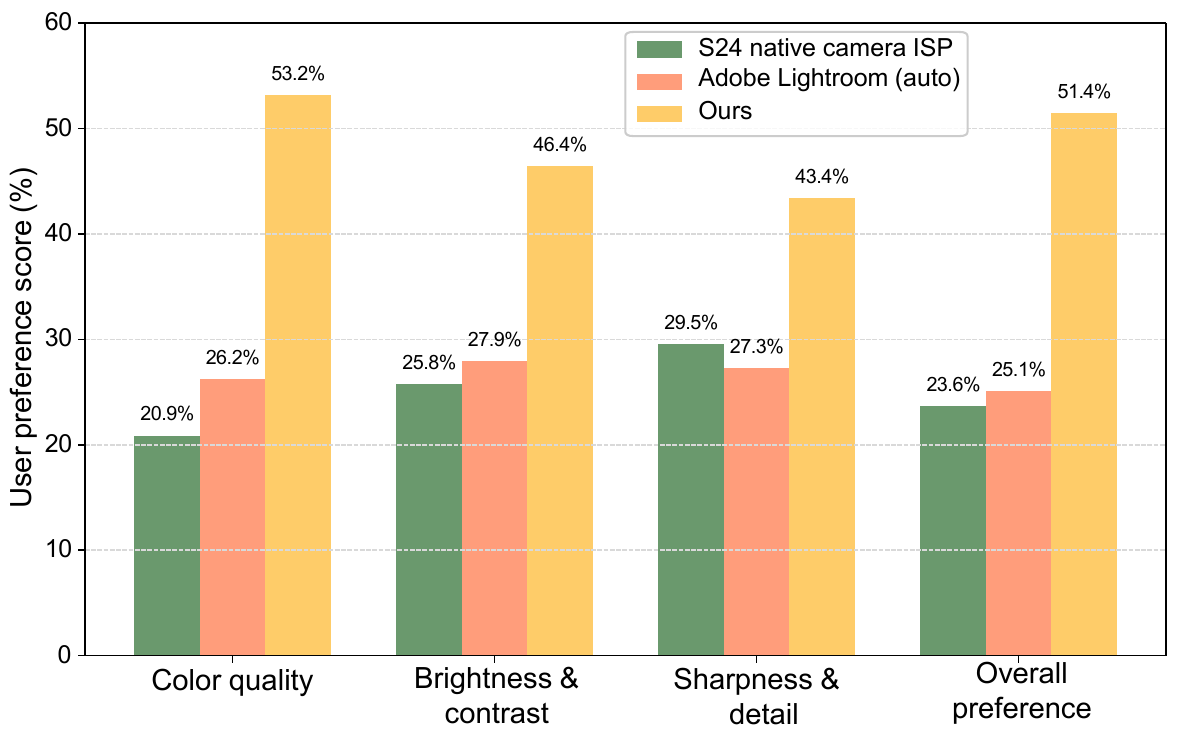}
\vspace{-1mm}
\caption{
User study results. Preference rates (\%) for each criterion comparing our method, the Samsung S24 native ISP, and Adobe Lightroom (auto enhancement).}
\label{fig:user-study}
\vspace{-2mm}
\end{figure*}

\begin{figure*}[!t]
\centering
\includegraphics[width=\linewidth]{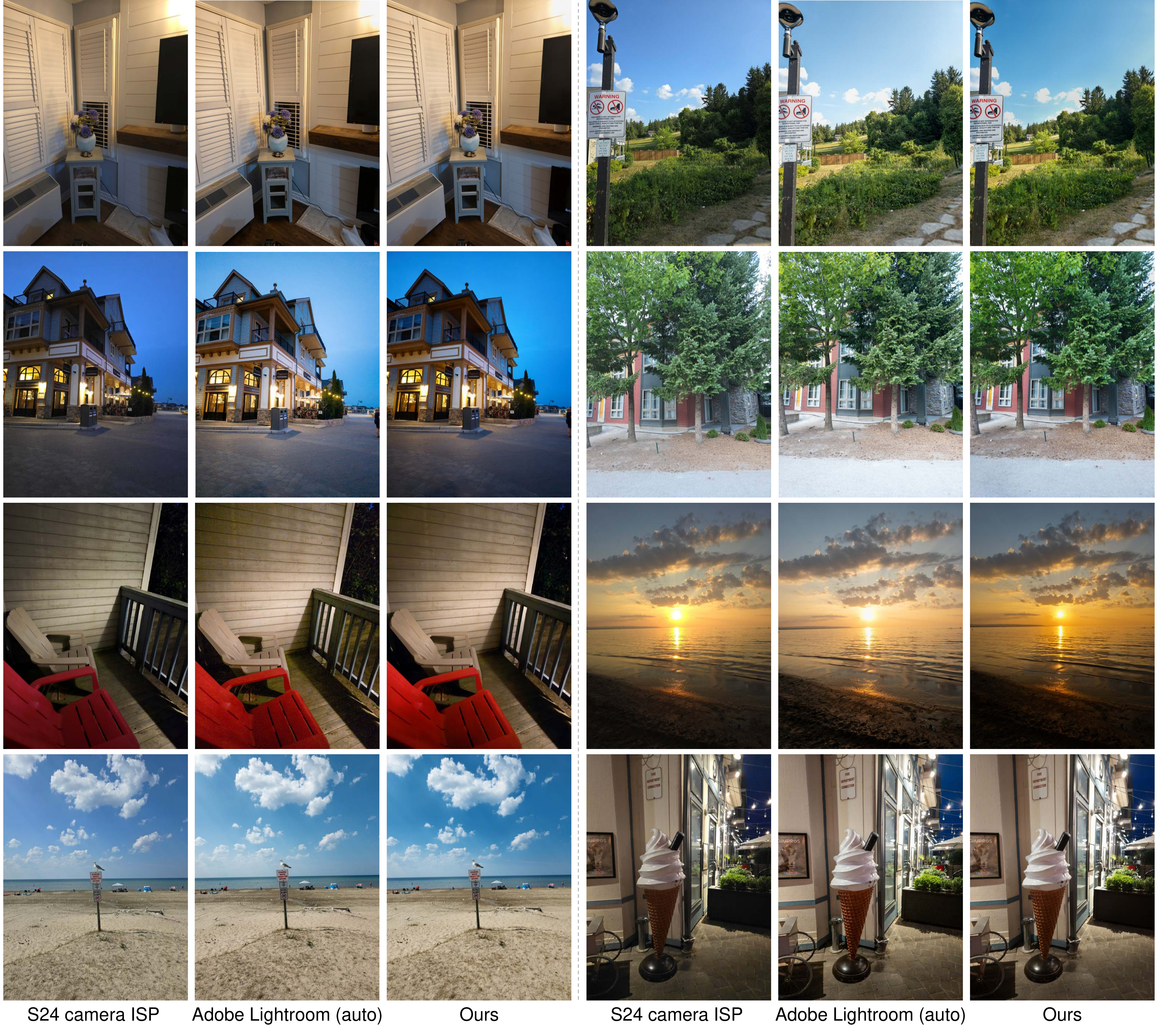}
\vspace{-5mm}
\caption{Example comparisons shown to participants during the user study. For each scene, participants evaluated three versions: our method, the Samsung S24 native ISP, and Adobe Lightroom (auto enhancement). The order was randomized in the GUI for every trial.\label{fig:user-study-examples}}
\vspace{-2mm}
\end{figure*}

Participants viewed the three versions side-by-side in a custom user-study GUI. The display order was randomized for each trial to prevent bias.
All sessions ran on a hardware-calibrated EIZO ColorEdge CG319X (31.1~inch, 4096$\times$2160 IPS, 99\% Adobe~RGB, 98\% DCI-P3), self-calibrated with its built-in sensor to DCI-P3 primaries, a D65 white point, 150~cd/m$^2$, and a gamma of 2.4. Following the viewing-condition recommendations of ITU-R BT.500-15~\cite{itu2023bt500}, participants viewed the display from a distance of 20~inches, approximately 1.4 picture heights. For each scene, participants selected the preferred image under four criteria: `color quality' --- \texttt{How good and natural do the colors look?}, `brightness \& contrast' --- \texttt{Is the image clear, with a good balance between light and dark areas?},\\`sharpness \& detail' --- \texttt{How crisp and detailed does the image look?}, and `overall preference' --- \texttt{Which image do you like more overall?}. \\Twenty participants took part in the study. The study received ethics approval prior to data collection. Each participant evaluated all 45 scenes under the four criteria, yielding 900 votes per criterion (45 scenes $\times$ 20 participants). As shown in Fig.~\ref{fig:user-study}, our method was preferred across all four criteria: 53.2\% in `color quality', 46.4\% in `brightness \& contrast', 43.4\% in `sharpness \& detail', and 51.4\% in `overall preference', with margins of +13.9--27.0\% over the closest competitor. See Fig.~\ref{fig:user-study-examples} for representative examples shown to participants.

\section{Ablation and Additional Results}
\label{sec:ablation_studies}

\subsection{Ablation Studies}
\label{supp-sec:ablation-studies}
We conducted a series of ablation studies to validate the effectiveness of each component in our method. In this subsection, we describe these experiments in detail and present the corresponding results.

\subsubsection{Denoising Ablations}
\label{Sec:ablation_studies_denoising}

As discussed in the main paper, we evaluated three NafNet variants~\cite{nafnet} for raw image denoising: lite (245~K parameters), base (933~K), and large (3.63~M). To compare NafNet with an alternative architecture for raw image denoising, Table~\ref{tab:noise-ablation-1} shows results for the large NafNet model and Restormer~\cite{restormer} (26.1~M), both trained on the S24 raw/denoised pairs~\cite{s24}. The input to each model is a noisy raw image, and the output is a denoised raw image, evaluated against the pseudo ground truth provided in the S24 dataset. The large NafNet variant achieves superior results while requiring substantially fewer parameters.

\begin{table}[!t]
\centering
\caption{Raw image denoising results on the S24 noisy/denoised test set \cite{s24} using NafNet \cite{nafnet} and Restormer \cite{restormer}. The best results are highlighted in \textbf{\colorbox{best}{yellow}}.}
\scalebox{0.8}{
\label{tab:noise-ablation-1}
\begin{tabular}{|l|c|c|}
\hline
\multicolumn{1}{|c|}{} &
  \multicolumn{2}{c|}{\cellcolor{red}\textcolor{white}{\shortstack{\\\textbf{S24 Test Set}\\\textbf{(Noisy/Denoised)}}}} \\ \cline{2-3}

\multicolumn{1}{|c|}{\multirow{-2}{*}{\textbf{Method}}} &
  \textbf{PSNR}\,$\uparrow$ &
  \textbf{SSIM}\,$\uparrow$ \\ \hline
NafNet (large, ours) \cite{nafnet} & \textbf{\cellcolor{best}{57.33}} & \textbf{\cellcolor{best}{0.999}} \\ \hline
Restormer \cite{restormer}         & 55.42 & \textbf{\cellcolor{best}{0.999}} \\ \hline
\end{tabular}
}
\end{table}

Table~\ref{tab:noise-ablation-4} reports results on the S24 test set, where the inputs are noisy raw images and both the outputs and ground truth are sRGB-rendered images. We compare three configurations of our framework (with the photofinishing module enabled in all cases): without denoising, with denoising only (excluding the detail-enhancement network), and with both denoising and detail enhancement, using the base denoising model. The results show that image denoising yields significant improvements.

\begin{table}[!t]
\centering
\caption{Ablation study of raw denoising and detail enhancement on the S24 \cite{s24} test set. Inputs are noisy raw images, and outputs are compared against ground-truth sRGB images. Denoising was performed using the base model ($\sim$933K parameters). The best results are highlighted in \textbf{\colorbox{best}{yellow}}.}
\scalebox{0.85}{
\label{tab:noise-ablation-4}
\begin{tabular}{|l|c|c|}
\hline
\multicolumn{1}{|c|}{} &
  \multicolumn{2}{c|}{\cellcolor{red}\textcolor{white}{\textbf{S24 Test Set}}} \\ \cline{2-3}
\multicolumn{1}{|c|}{\multirow{-2}{*}{\textbf{Framework Variation}}} &
  \textbf{PSNR}\,$\uparrow$ &
  \textbf{SSIM}\,$\uparrow$ \\ \hline
w/o raw denoising                  & 24.48 & 0.731 \\ \hline
w/\hspace{2.5mm}raw denoising                   & 26.48 & 0.883 \\ \hline
w/\hspace{2.5mm}raw denoising + enhancement     & \textbf{\cellcolor{best}{27.52}} & \textbf{\cellcolor{best}{0.922}} \\ \hline
\end{tabular}
}\vspace{-2mm}
\end{table}

\subsubsection{Upsampling Ablations}
\label{Sec:ablation_studies_upsampling}
In both the main paper and this supplementary material (Sec.~\ref{sec:supp-bgu}), we have introduced our regularized variant of BGU~\cite{bgu}. Here, we present ablation results comparing different guided upsampling approaches, including BGU without our regularization. For completeness, we also compare against the regularization used in the original BGU Halide implementation, which we denote as Halide regularization.

A key challenge in this evaluation arises from the ground-truth sRGB images in the~S24 test set.~These ground-truth images include sharpening and detail enhancements that are not present in the pseudo ground-truth denoised raw images. Since the guide image (denoised raw or any of its variants, e.g., linear sRGB) lacks such details, no guided upsampling method can fully recover them. To address this, we conducted experiments under two ground-truth settings:

\begin{enumerate}[leftmargin=*]
    \item Original S24 ground truth: sRGB images with sharpening, detail enhancement, and compression (as provided in the original dataset).
    \item Alternative ground truth: sRGB images rendered directly from the denoised raw DNG files using Adobe Photoshop, with sharpening and detail enhancements disabled and no compression applied.
\end{enumerate}

For both ground-truth settings, we evaluated two guide image versions: 1) the denoised raw image and 2) its linear sRGB conversion. The corresponding source image was generated by downsampling the guide image to one-quarter of its original resolution, while the target image was obtained by downsampling the ground-truth sRGB image to the same resolution.

\begin{figure*}[!t]
\centering
\includegraphics[width=0.6\linewidth]{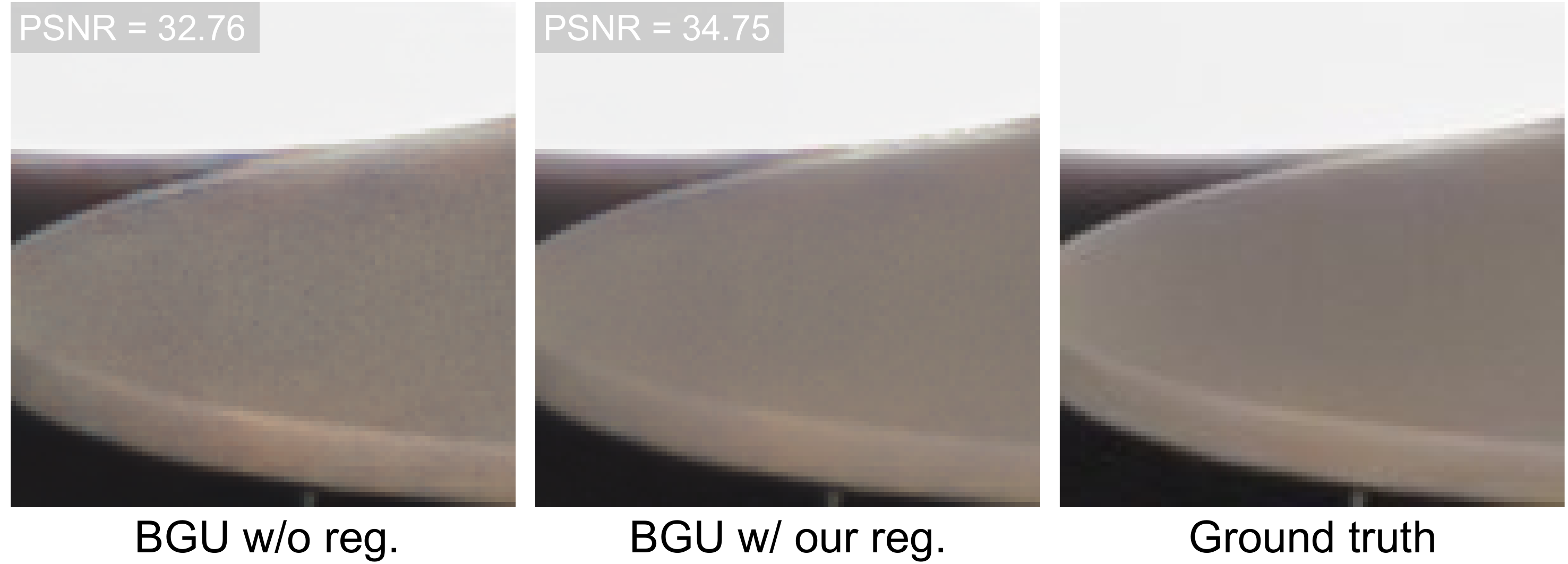}
\vspace{-1mm}
\caption{Visual comparison of guided upsampling on an example from the S24 dataset~\cite{s24}. We compare BGU \cite{bgu} without regularization against our proposed regularization. Our regularization better preserves fine details and produces results closer to the ground truth.}
\label{fig:bgu}
\vspace{-4mm}
\end{figure*}

We utilized ground-truth sRGB images in these experiments, rather than outputs from our photofinishing module, to isolate the evaluation of the guided upsampling methods themselves under the best-case scenario, without indirect effects from the photofinishing stage. Table~\ref{tab:guided-upsampling} reports results comparing BGU \cite{bgu} with our regularization, BGU without regularization, and BGU with the Halide regularization, along with two alternative guided upsampling methods: guided linear upsampling (GLU) \cite{glu} and classical guided image filtering (GF) \cite{gf}. Figure~\ref{fig:bgu} provides a visual comparison of BGU results without and with our proposed regularization. Overall, our regularization produces reconstructions that are both visually and quantitatively closer to the ground truth.

\begin{table*}[!t]
\centering
\caption{Comparison of different guided upsampling methods on the S24 test set  \cite{s24} under two ground-truth settings: original S24 ground truth (with detail enhancements) and alternative ground truth (without enhancements). Results are reported for two guide image versions: denoised raw and linear sRGB (LsRGB). The best results are highlighted in \textbf{\colorbox{best}{yellow}}.}
\scalebox{0.75}{
\label{tab:guided-upsampling}
\begin{tabular}{|l|c|c|c|c|c|c|c|c|}
\hline
\multicolumn{1}{|c|}{} &
  \multicolumn{2}{c|}{\cellcolor{red}\textcolor{white}{\shortstack{\\\textbf{Original S24 GT} \\ \textbf{(Denoised Raw/sRGB)}}}} &
  \multicolumn{2}{c|}{\cellcolor{orange}\textcolor{white}{\shortstack{\textbf{Alt. S24 GT} \\ \textbf{(Denoised Raw/sRGB)}}}} &
  \multicolumn{2}{c|}{\cellcolor{red}\textcolor{white}{\shortstack{\textbf{Original S24 GT} \\ \textbf{(LsRGB/sRGB)}}}} &
  \multicolumn{2}{c|}{\cellcolor{orange}\textcolor{white}{\shortstack{\textbf{Alt. S24 GT} \\ \textbf{(LsRGB/sRGB)}}}} \\ \cline{2-9}
\multicolumn{1}{|c|}{\multirow{-2}{*}{\textbf{Method}}} &
  \textbf{PSNR}\,$\uparrow$ & \textbf{SSIM}\,$\uparrow$ &
  \textbf{PSNR}\,$\uparrow$ & \textbf{SSIM}\,$\uparrow$ &
  \textbf{PSNR}\,$\uparrow$ & \textbf{SSIM}\,$\uparrow$ &
  \textbf{PSNR}\,$\uparrow$ & \textbf{SSIM}\,$\uparrow$ \\ \hline
GLU \cite{glu} & 33.46 & 0.919 & 37.95 & 0.966 & 33.47 & 0.919 & 38.42 & 0.971 \\ \hline
GF \cite{gf}   & 31.58 & 0.907 & 33.74 & 0.943 & 31.97 & 0.912 & 34.84 & 0.959 \\ \hline
BGU (w/o regularization) \cite{bgu} & 33.73 & \textbf{\cellcolor{best}{0.928}} & 40.82 & 0.986 & 33.77 & 0.928 & 41.00 & 0.988 \\ \hline
BGU (w/$\texttt{ }$ Halide regularization) \cite{bgu} & 32.49 & 0.916 & 37.28 & 0.977 & 32.78 & 0.918 & 38.26 & 0.980 \\ \hline
BGU (w/$\texttt{ }$ our regularization) \cite{bgu} & \textbf{\cellcolor{best}{33.85}} & 0.927 & \textbf{\cellcolor{best}{41.69}} & \textbf{\cellcolor{best}{0.987}} & \textbf{\cellcolor{best}{33.92}} & \textbf{\cellcolor{best}{0.928}} & \textbf{\cellcolor{best}{42.01}} & \textbf{\cellcolor{best}{0.989}} \\ \hline
\end{tabular}}
\vspace{-1mm}
\end{table*}

\subsubsection{Photofinishing Loss Ablations}
\label{Sec:ablation_studies_ps_loss}

To analyze the contribution of each loss term in our photofinishing module, we conducted a detailed ablation study on the default picture style (Style~\#0) of the S24 dataset~\cite{s24} (Tables~\ref{tab:ablation_losses-1} and~\ref{tab:ablation_losses-2}). For these experiments, we chose to process images starting from the pseudo ground-truth denoised raw inputs, which were subsequently mapped to the linear sRGB color space. This setup allows us to isolate and evaluate the photofinishing module independently, without the influence of potential degradations introduced by the denoising stage in our full pipeline.

When training with all loss terms enabled, our model achieves the best accuracy in both PSNR and SSIM, as highlighted in Table~\ref{tab:ablation_losses-1}. Removing any single loss term slightly decreases the accuracy. Conversely, progressively adding the losses one by one (Table~\ref{tab:ablation_losses-2}) results in steady improvements, indicating that each term contributes positively to the overall quality.

\begin{table}[t]
\centering
\caption{Ablation study showing the impact of removing each loss term individually on the results of our photofinishing module, evaluated on the S24 test set~\cite{s24}. The input images are linear sRGB frames generated from pseudo ground-truth denoised images in the S24 test set, downsampled to one-quarter resolution. The outputs are compared against the ground truth at the same resolution. The best results are highlighted in \textbf{\colorbox{best}{yellow}}.}
\setlength{\tabcolsep}{4pt}
\renewcommand{\arraystretch}{1.2}
\scalebox{0.72}{
\begin{tabular}{|c|c|c|c|c|c|c|c|c|c|c|c|}
\hline
\rowcolor{red}
\multicolumn{11}{|c|}{\textcolor{white}{\shortstack{\textbf{S24 Test Set (1/4 LsRGB/sRGB)}}}} \\ \hline
$\ell_1$ & $\ell_{\texttt{SSIM}}$ & $\ell_{\texttt{perc}}$ & $\ell_{\Delta E}$ & $\ell_{\texttt{CbCr}}$ & $\ell_{\texttt{LuT-s}}$ & $\ell_{\texttt{luma}}$ & $\ell_{\texttt{TM}}$ & $\ell_{\texttt{LTM-s}}$ & \textbf{PSNR} & \textbf{SSIM} \\ \hline
          & \checkmark & \checkmark & \checkmark & \checkmark & \checkmark & \checkmark & \checkmark & \checkmark &  27.01 & 0.934 \\ \hline
\checkmark &            & \checkmark & \checkmark & \checkmark & \checkmark & \checkmark & \checkmark & \checkmark & 27.01 & 0.931 \\ \hline
\checkmark & \checkmark &            & \checkmark & \checkmark & \checkmark & \checkmark & \checkmark & \checkmark & 26.89 & 0.934 \\ \hline
\checkmark & \checkmark & \checkmark &            & \checkmark & \checkmark & \checkmark & \checkmark & \checkmark & 26.96 & 0.934 \\ \hline
\checkmark & \checkmark & \checkmark & \checkmark &            & \checkmark & \checkmark & \checkmark & \checkmark &  26.99 & 0.934 \\ \hline
\checkmark & \checkmark & \checkmark & \checkmark & \checkmark &            & \checkmark & \checkmark & \checkmark &  27.18 & 0.934 \\ \hline
\checkmark & \checkmark & \checkmark & \checkmark & \checkmark & \checkmark &            & \checkmark & \checkmark &  27.20 & 0.934 \\ \hline
\checkmark & \checkmark & \checkmark & \checkmark & \checkmark & \checkmark & \checkmark &            & \checkmark &  27.13 & 0.934 \\ \hline
\checkmark & \checkmark & \checkmark & \checkmark & \checkmark & \checkmark & \checkmark & \checkmark &            &  27.05 & 0.934 \\ \hline
\checkmark & \checkmark & \checkmark & \checkmark & \checkmark & \checkmark & \checkmark & \checkmark & \checkmark &  \textbf{\cellcolor{best}{27.49}} & \textbf{\cellcolor{best}{0.939}} \\ \hline
\end{tabular}}
\label{tab:ablation_losses-1}
\end{table}

\begin{table}[t]
\centering
\caption{Ablation study showing the cumulative effect of adding each loss term progressively to our photofinishing module, evaluated on the S24 test set~\cite{s24}. Starting from a baseline trained with $\ell_1$ only, additional loss terms are introduced one at a time. Input images are linear sRGB frames generated from pseudo ground-truth denoised images in the S24 test set, downsampled to one-quarter resolution. The outputs are compared against the ground truth at the same resolution. The best results are highlighted in \textbf{\colorbox{best}{yellow}}.}
\setlength{\tabcolsep}{4pt}
\renewcommand{\arraystretch}{1.2}
\scalebox{0.72}{
\begin{tabular}{|c|c|c|c|c|c|c|c|c|c|c|}
\hline
\rowcolor{red}
\multicolumn{11}{|c|}{\textcolor{white}{\shortstack{\textbf{S24 Test Set (1/4 LsRGB/sRGB)}}}} \\ \hline
$\ell_1$ & $\ell_{\texttt{SSIM}}$ & $\ell_{\texttt{perc}}$ & $\ell_{\Delta E}$ & $\ell_{\texttt{CbCr}}$ & $\ell_{\texttt{LuT-s}}$ & $\ell_{\texttt{luma}}$ & $\ell_{\texttt{TM}}$ & $\ell_{\texttt{LTM-s}}$ & \textbf{PSNR} & \textbf{SSIM} \\ \hline
\checkmark &  &  &  &  &  &  &  &  & 26.80 & 0.930 \\ \hline
\checkmark & \checkmark &  &  &  &  &  &  &  & 27.03 & 0.935 \\ \hline
\checkmark & \checkmark & \checkmark &  &  &  &  &  &  & 27.07 & 0.935 \\ \hline
\checkmark & \checkmark & \checkmark & \checkmark &  &  &  &  &  & 27.08 & 0.935 \\ \hline
\checkmark & \checkmark & \checkmark & \checkmark & \checkmark &  &  &  &  & 27.09 & 0.934 \\ \hline
\checkmark & \checkmark & \checkmark & \checkmark & \checkmark & \checkmark &  &  &  & 27.09 & 0.934 \\ \hline
\checkmark & \checkmark & \checkmark & \checkmark & \checkmark & \checkmark & \checkmark &  &  & 27.13 & 0.935 \\ \hline
\checkmark & \checkmark & \checkmark & \checkmark & \checkmark & \checkmark & \checkmark & \checkmark &  & 27.05 & 0.934 \\ \hline
\checkmark & \checkmark & \checkmark & \checkmark & \checkmark & \checkmark & \checkmark & \checkmark & \checkmark & \textbf{\cellcolor{best}{27.49}} & \textbf{\cellcolor{best}{0.939}} \\ \hline
\end{tabular}}
\label{tab:ablation_losses-2}
\vspace{-2mm}
\end{table}

\begin{figure*}[!t]
\centering
\includegraphics[width=\linewidth]{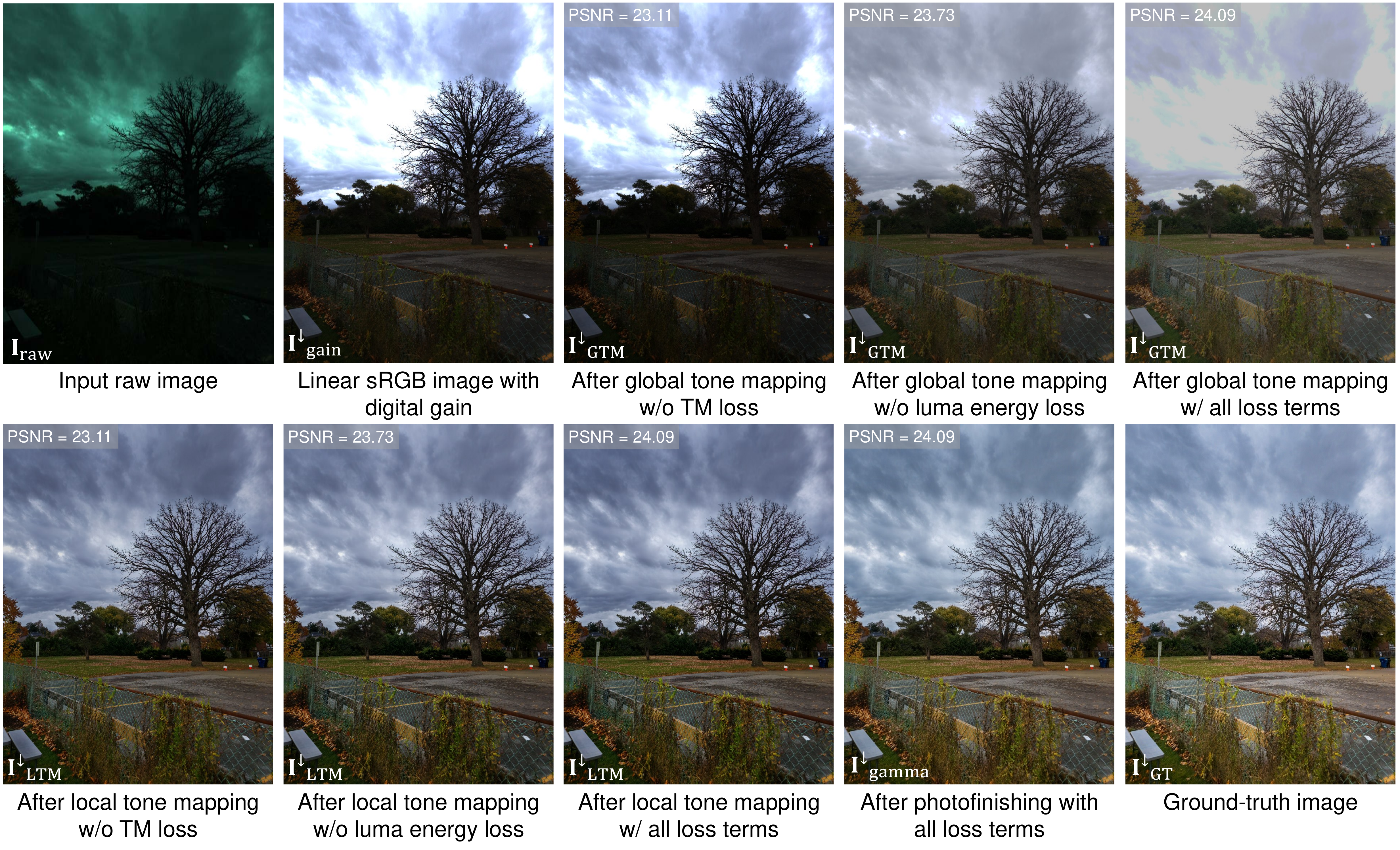}
\vspace{-5mm}
\caption{Visual comparison showing the impact of training with and without the tone mapping loss ($\ell_{\texttt{TM}}$) and the luma energy loss ($\ell_{\texttt{luma}}$). Shown are the input raw image, the color-corrected linear sRGB image after digital gain, the results after global and local tone mapping, and the final photofinishing output compared against the ground truth at the same resolution (i.e., one-quarter of the original raw resolution). The input sample is from the validation split of the S24 dataset~\cite{s24}, where the photofinishing result shown here is obtained by processing the pseudo ground-truth denoised raw image mapped to the linear sRGB space.
\label{fig:ablation-loss-terms-1}}
\vspace{-4mm}
\end{figure*}

\begin{figure*}[!t]
\centering
\includegraphics[width=0.7\linewidth]{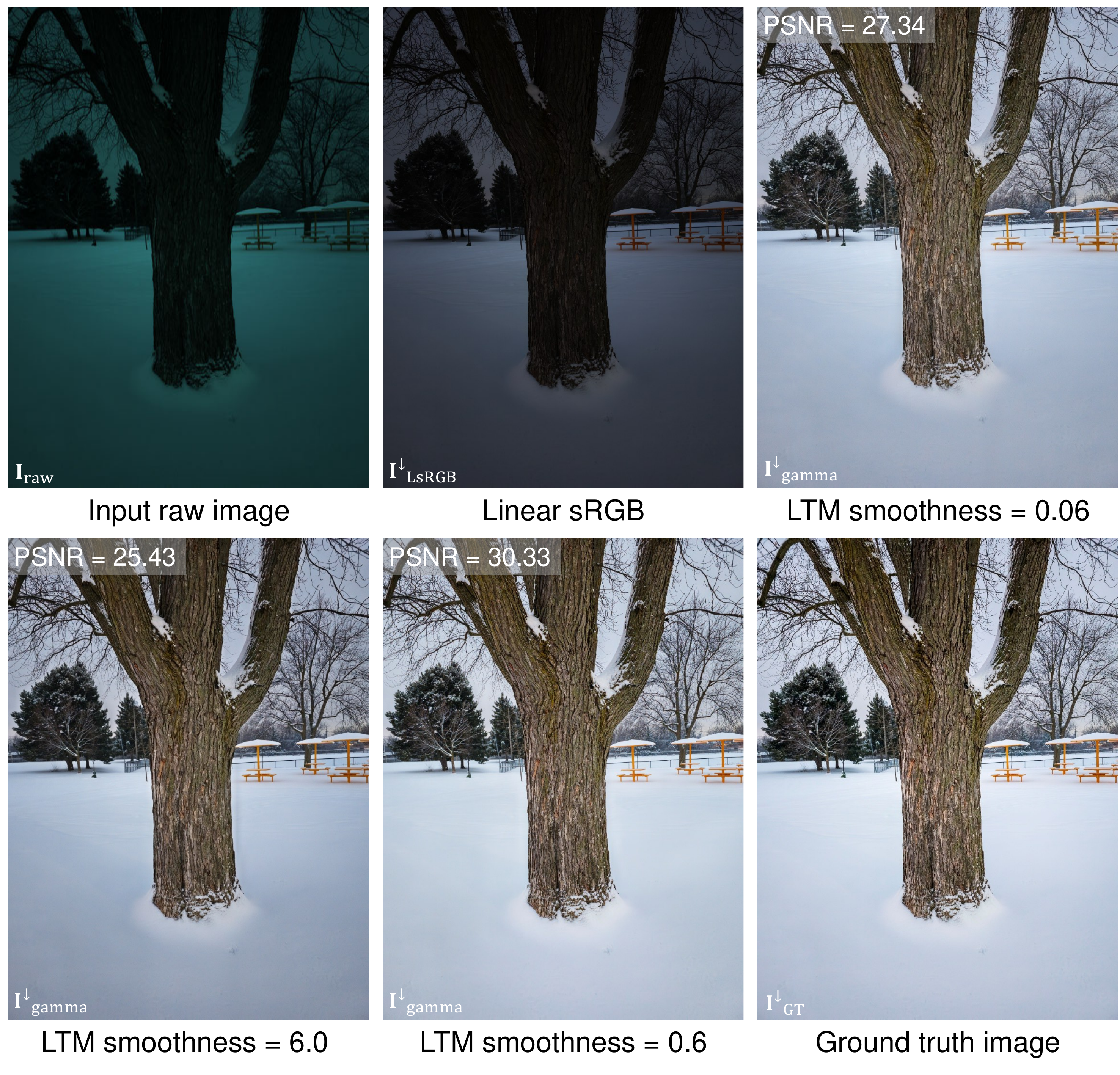}
\vspace{-2mm}
\caption{The impact of different values of $\lambda_{\texttt{LTM-s}}$ on the final output of the photofinishing module. Our chosen value ($\lambda_{\texttt{LTM-s}}{=}0.6$) achieves a good balance between smoothing the local tone-mapping maps and maintaining both qualitative and quantitative quality. The shown image is from the test set of the S24 dataset~\cite{s24}. The displayed result corresponds to the final output of our photofinishing module, obtained by processing the pseudo ground-truth denoised raw image mapped to the linear sRGB space. Both the prediction and ground truth are shown at one-quarter of the original raw resolution.}
\label{fig:ablation-ltm-smoothness}
\vspace{-2mm}
\end{figure*}

Figure~\ref{fig:ablation-loss-terms-1} shows that the tone-mapping loss ($\ell_{\texttt{TM}}$) and luminance-consistency loss ($\ell_{\texttt{luma}}$) lead to more balanced tone reproduction across the image, encouraging the global and local tone-mapping stages of the photofinishing module to function as intended. The global stage refines the overall tone distribution of the gain-adjusted linear sRGB image, while the local stage further adjusts  contrast in spatially varying regions. Both stages maintain a comparable overall luminance level, which encourages the local tone-mapping stage to focus on localized adjustments rather than global shifts. Similarly, Fig.~\ref{fig:ablation-ltm-smoothness} shows that our chosen loss weight for the local tone-mapping smoothness term ($\lambda_{\texttt{LTM-s}}{=}0.6$) achieves a good balance between suppressing local artifacts and preserving fine details (see Table~\ref{tab:ablation-ltm-smoothness} for the quantitative comparison).

\begin{table}[t]
\centering
\caption{Effect of different values of the local tone-mapping smoothness loss weight ($\lambda_{\texttt{LTM-s}}$) on the S24 test set~\cite{s24}. Input images are pseudo ground-truth denoised raw images mapped to the linear sRGB space at one-quarter of the original raw resolution. The best results are highlighted in \textbf{\colorbox{best}{yellow}}.}
\scalebox{0.8}{
\label{tab:ablation-ltm-smoothness}
\begin{tabular}{|l|c|c|}
\hline
\multicolumn{1}{|c|}{} &
  \multicolumn{2}{c|}{\cellcolor{red}\textcolor{white}{\shortstack{\\\textbf{S24 Test Set}\\\textbf{(1/4 LsRGB/sRGB)}}}} \\ \cline{2-3}

\multicolumn{1}{|c|}{\multirow{-2}{*}{\textbf{Loss weight ($\lambda_{\texttt{LTM-s}}$)}}} &
  \textbf{PSNR}\,$\uparrow$ &
  \textbf{SSIM}\,$\uparrow$ \\ \hline
0.06 & 27.13 & 0.935 \\ \hline
0.6 & \textbf{\cellcolor{best}{27.49}} & \textbf{\cellcolor{best}{0.939}} \\ \hline
6.0  & 26.19 & 0.933 \\ \hline
\end{tabular}
}
\vspace{-1mm}
\end{table}

\subsubsection{Photofinishing Design Ablations}
\label{Sec:ablation_studies_ps_desing}

\begin{figure*}[!t]
\centering
\includegraphics[width=0.7\linewidth]{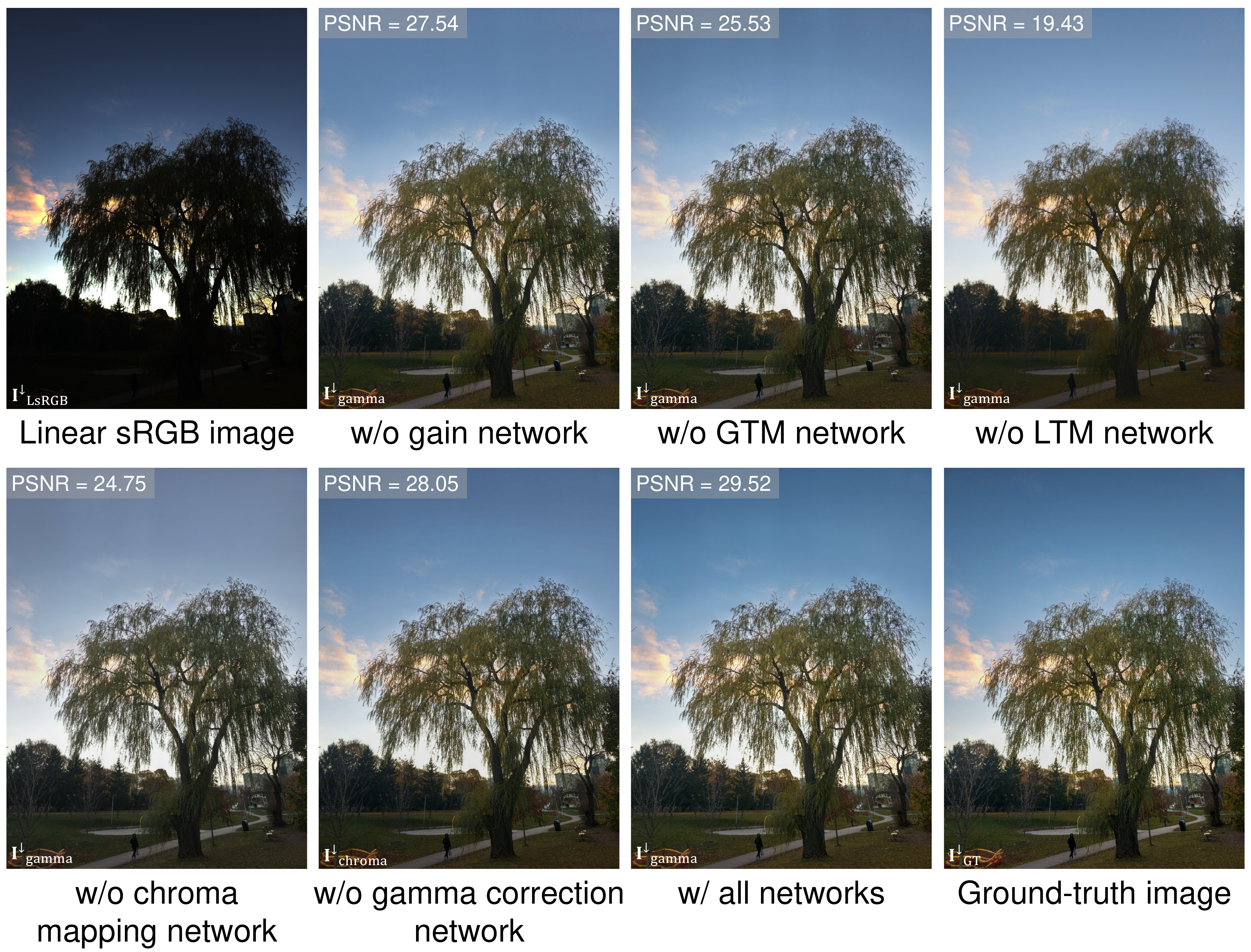}
\vspace{-1mm}
\caption{Output of the photofinishing module trained without one of the processing stages (digital gain, global tone mapping, local tone mapping, chroma mapping, and gamma correction) compared to the full model trained with all stages enabled. The shown image is from the validation set of the S24 dataset~\cite{s24}. The results are obtained by processing the pseudo ground-truth denoised raw image provided in the dataset, mapped to the linear sRGB space at one-quarter of the original raw resolution, and compared against the ground-truth sRGB image at the same resolution.}
\label{fig:ablation-design-1}
\vspace{-1mm}
\end{figure*}

We conducted a set of ablation studies to evaluate our design against alternative configurations. As described earlier, the proposed photofinishing module consists of five networks: digital gain, GTM, LTM, chroma mapping, and gamma correction. Each network predicts the parameters of its corresponding operator. The rest of this subsection details the contribution of each component.

\noindent\\\textbf{Overall Module Composition.} Table~\ref{tab:ablation-photofinishing-design} shows the results of different variants of our photofinishing module, where one network is removed at both training and testing to evaluate the contribution of each stage. As shown, the proposed full design not only provides greater modularity and user control but also achieves the best quantitative results. Figure~\ref{fig:ablation-design-1} shows a qualitative comparison.

\begin{table}[t]
\centering
\caption{Ablation on the design of our photofinishing module. For this ablation, we remove one network of our photofinishing module at both training and inference to evaluate its contribution. Results are shown on the S24 test set~\cite{s24} (default style; Style~\#0), where the input images are pseudo ground-truth denoised raw images mapped to the linear sRGB space and downsampled to one-quarter of the original resolution, and the ground truth is the corresponding sRGB image at the same resolution. The best results are highlighted in \textbf{\colorbox{best}{yellow}}.}
\scalebox{0.8}{
\label{tab:ablation-photofinishing-design}
\begin{tabular}{|c|c|c|c|c|c|c|}
\hline
\multicolumn{7}{|c|}{\cellcolor{red}\textcolor{white}{\textbf{S24 Test Set (1/4 LsRGB/sRGB)}}} \\ \hline
Gain &
GTM &
LTM &
Chroma &
Gamma &
\textbf{PSNR}\,$\uparrow$ &
\textbf{SSIM}\,$\uparrow$ \\ \hline
 & \checkmark & \checkmark & \checkmark &  & 26.68 & 0.931 \\ \hline
\checkmark &  & \checkmark & \checkmark &  & 26.84 & 0.934 \\ \hline
\checkmark & \checkmark &  & \checkmark &  & 24.66 & 0.898 \\ \hline
\checkmark & \checkmark & \checkmark &  &  & 26.45 & 0.928 \\ \hline
\checkmark & \checkmark & \checkmark & \checkmark &  & 26.91 & 0.935 \\ \hline
\checkmark & \checkmark & \checkmark & \checkmark & \checkmark & \textbf{\cellcolor{best}{27.49}} & \textbf{\cellcolor{best}{0.939}} \\ \hline
\end{tabular}
}
\vspace{-2mm}
\end{table}

 \noindent\\\textbf{Local Tone-Mapping and Chroma Design.}
We conducted additional experiments to ablate the design choices of our LTM and chroma mapping components.
The LTM network predicts five spatially varying coefficient maps: $\mathbf{A}_\texttt{LTM}$, $\mathbf{B}_\texttt{LTM}$, $\mathbf{C}_\texttt{LTM}$, $\mathbf{W}_\texttt{LTM}$, and $\mathbf{G}_\texttt{LTM}$.
We progressively enabled these coefficients to evaluate their contribution, and we further examined the effect of the LTM grid size and the design of the chroma LuT used for chroma mapping.

Table~\ref{tab:ablation-ltm-design} reports results of training our photofinishing module with different subsets of the LTM parameters.
We began by predicting a single-channel pixel-wise map representing $\mathbf{A}_\texttt{LTM}$ to modulate the tone-mapping exponent.
Next, we introduced $\mathbf{B}_\texttt{LTM}$ to allow adaptive control over the compression strength in the denominator of the tone-mapping function, followed by adding $\mathbf{C}_\texttt{LTM}$ to model asymmetric behavior between bright and dark regions.
We then incorporated the blending map $\mathbf{W}_\texttt{LTM}$ to locally mix the LTM output with the globally tone-mapped result $\mathbf{I}^{\downarrow}_{\texttt{GTM}}$, and subsequently added the gain map $\mathbf{G}_\texttt{LTM}$, which adjusts local exposure before tone mapping.

\begin{figure*}[!t]
\centering
\includegraphics[width=0.7\linewidth]{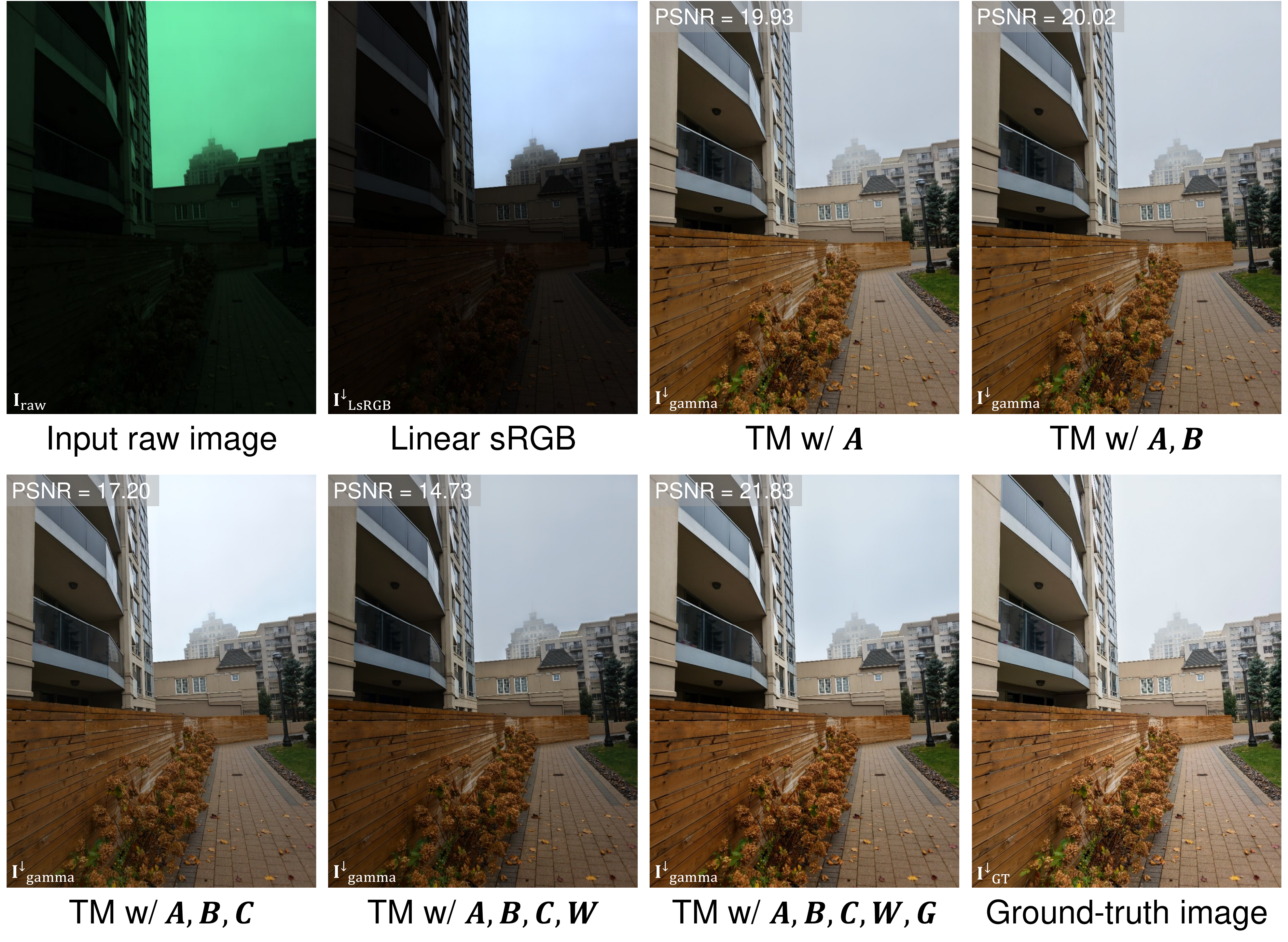}
\vspace{-1mm}
\caption{Qualitative results from our photofinishing module trained with progressively
added LTM parameters ($\mathbf{A}_\texttt{LTM}$, $\mathbf{B}_\texttt{LTM}$,
$\mathbf{C}_\texttt{LTM}$, $\mathbf{W}_\texttt{LTM}$, and $\mathbf{G}_\texttt{LTM}$) during both training and testing.
The input is a pseudo ground-truth denoised raw image mapped to linear sRGB
(1/4 resolution), and the output is compared against the corresponding sRGB
ground truth. Examples are taken from the S24 validation set~\cite{s24}.}
\label{fig:ablation-design-2}
\end{figure*}

This full set of parameters was tested both without and with the multi-scale guidance network to evaluate its contribution. See Fig.~\ref{fig:ablation-design-2} for a qualitative example.

Table~\ref{tab:ablation-lut-ltm} complements this analysis by showing the results of modifying the LTM grid size and the chroma mapping design.~We first reduced the LTM grid size from 64$\times$64$\times$18$\times$5 to 32$\times$32$\times$9$\times$5 to examine the impact of spatial resolution of the LTM parameter grid.
Next, we evaluated predicting the chroma LuT directly instead of learning it as a residual to the image-independent learnable LuT.

Lastly, we evaluated a smaller chroma LuT with $N_h{=}12$ bins (i.e., a LuT size of $12\times12\times2$) instead of $N_h{=}24$ (i.e., $24\times24\times2$) to study the effect of chroma quantization granularity. The results demonstrate that both the residual formulation and higher grid/LuT resolution contribute to improved reconstruction accuracy, with our full configuration achieving the best overall results.

\begin{table}[!t]
\centering
\caption{Ablation on the local tone-mapping (LTM) design. We progressively add each predicted coefficient map  ($\mathbf{A}_\texttt{LTM}$-$\mathbf{G}_\texttt{LTM}$) to evaluate their contribution.  The ``All (w/o multi-scale guide)'' variant includes all five parameters but omits the multi-scale guidance network, using a single guidance network in its place. Results are on the S24 test set~\cite{s24}  using pseudo ground-truth denoised raw inputs mapped to linear sRGB
at one-quarter resolution. The best results are highlighted in  \textbf{\colorbox{best}{yellow}}.}
\scalebox{0.8}{
\label{tab:ablation-ltm-design}
\begin{tabular}{|l|c|c|}
\hline
\multicolumn{1}{|c|}{} &
  \multicolumn{2}{c|}{\cellcolor{red}\textcolor{white}{\shortstack{\\\textbf{S24 Test Set}\\\textbf{(1/4 LsRGB/sRGB)}}}} \\ \cline{2-3}

\multicolumn{1}{|c|}{\multirow{-2}{*}{\textbf{LTM Design}}} &
  \textbf{PSNR}\,$\uparrow$ &
  \textbf{SSIM}\,$\uparrow$ \\ \hline
$\mathbf{A}_\texttt{LTM}$ & 26.49 & 0.927 \\ \hline
$\mathbf{A}_\texttt{LTM}$, $\mathbf{B}_\texttt{LTM}$ & 26.54 & 0.928 \\ \hline
$\mathbf{A}_\texttt{LTM}$, $\mathbf{B}_\texttt{LTM}$, $\mathbf{C}_\texttt{LTM}$ & 26.58 & 0.927 \\ \hline
$\mathbf{A}_\texttt{LTM}$, $\mathbf{B}_\texttt{LTM}$, $\mathbf{C}_\texttt{LTM}$, $\mathbf{W}_\texttt{LTM}$ & 26.86 & 0.931 \\ \hline
All (w/o multi-scale guide) & 26.87 & 0.933 \\ \hline
Ours (all) & \textbf{\cellcolor{best}{27.49}} & \textbf{\cellcolor{best}{0.939}} \\ \hline
\end{tabular}
}
\vspace{-2mm}
\end{table}

\begin{table}[!t]
\centering
\caption{Ablation on the design of the chroma mapping and LTM networks. We evaluate 1) predicting the chroma LuT directly instead of as a residual, 2) using a smaller number of chroma bins ($N_h{=}12$ instead of $N_h{=}24$), and 3) reducing the LTM grid size to $N_g{=}32$ and depth to $N_c{=}9$, instead of $N_g{=}64$ and $N_c{=}18$. Results are shown on the S24 test set~\cite{s24}, where the input images are pseudo ground-truth denoised raw images mapped to linear sRGB and downsampled to one-quarter of the original resolution. The best results are highlighted in \textbf{\colorbox{best}{yellow}}.}
\scalebox{0.8}{
\label{tab:ablation-lut-ltm}
\begin{tabular}{|l|c|c|}
\hline
\multicolumn{1}{|c|}{} &
  \multicolumn{2}{c|}{\cellcolor{red}\textcolor{white}{\shortstack{\\\textbf{S24 Test Set}\\\textbf{(1/4 LsRGB/sRGB)}}}} \\ \cline{2-3}

\multicolumn{1}{|c|}{\multirow{-2}{*}{\textbf{Variant}}} &
  \textbf{PSNR}\,$\uparrow$ &
  \textbf{SSIM}\,$\uparrow$ \\ \hline
Chroma LuT (no residual) & 27.07 & 0.935 \\ \hline
Chroma LuT bins ($N_h$) = 12 & 26.88 & 0.932 \\ \hline
LTM grid size ($N_c$ = 9, $N_g$ = 32) & 26.94 & 0.934 \\ \hline
Ours & \textbf{\cellcolor{best}{27.49}} & \textbf{\cellcolor{best}{0.939}} \\ \hline
\end{tabular}
}
\end{table}

\noindent\\\textbf{Multi-Branch and Coordinate Attention.}
We further evaluated the contribution of the multi-branch convolutional (MBConv) and coordinate attention (CA)~\cite{coord-attention} blocks (Sec.~\ref{sec:supp-ca-mb}) used across the networks of our photofinishing module.~For this ablation, we trained and tested variants with each component removed independently.~Table~\ref{tab:ablation-mb-ca} shows that both the multi-branch structure and the coordinate attention blocks improve the result of our photofinishing module.

\begin{figure*}[!t]
\centering
\includegraphics[width=0.75\linewidth]{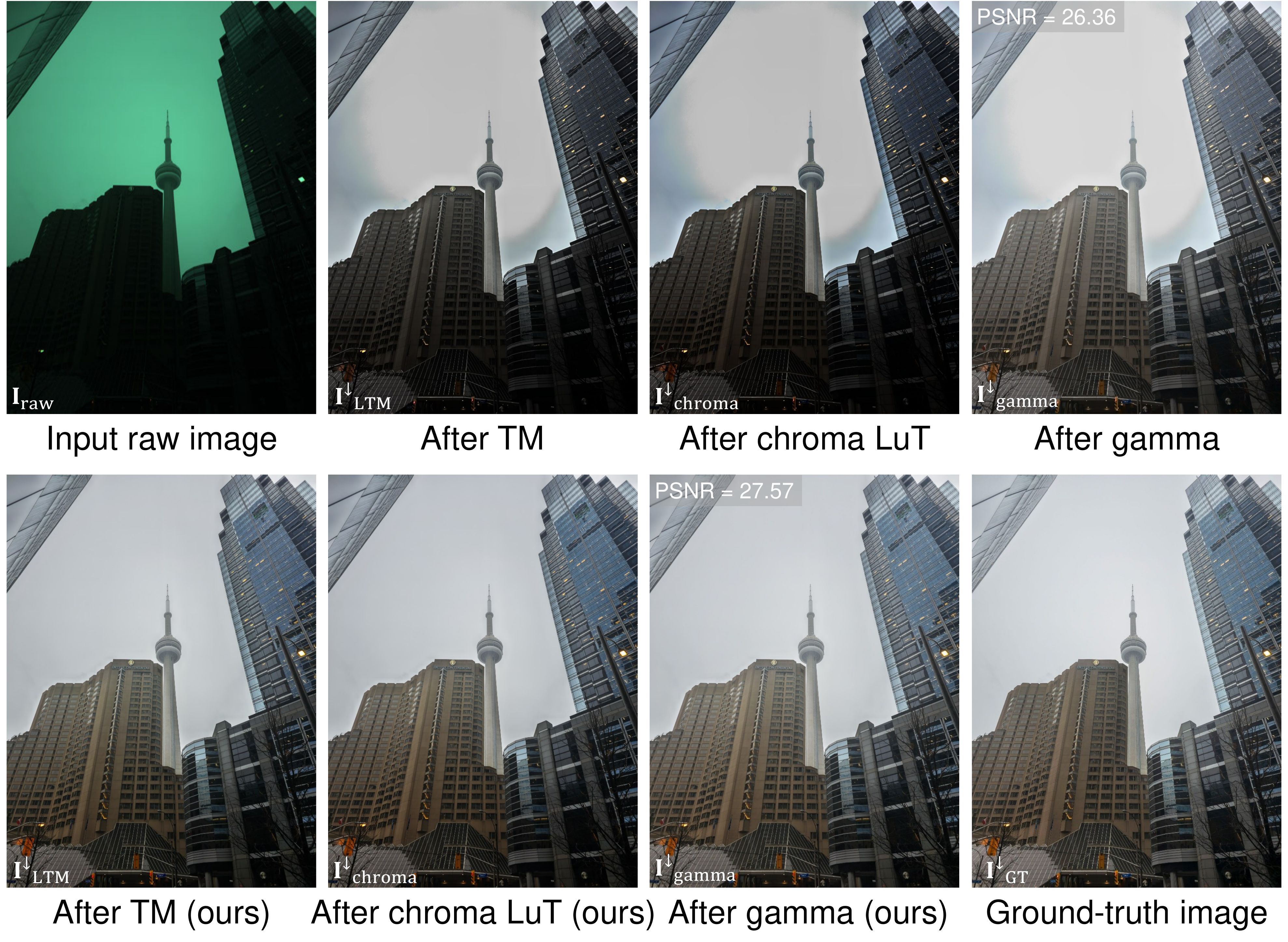}
\vspace{-1mm}
\caption{Qualitative comparison between the luma-based tone-mapping design
and our proposed design.
Inputs to the photofinishing module are pseudo ground-truth denoised raw images
mapped to linear sRGB and downsampled to one-quarter of the original resolution.
Results are shown from the validation set of the S24 dataset~\cite{s24}.}
\label{fig:ablation-old-design}
\vspace{-1mm}
\end{figure*}

\begin{table}[!t]
\centering
\caption{Ablation on the effect of multi-branch (MBConv) and coordinate attention (CA)~\cite{coord-attention} in our photofinishing module.
Results are reported on the S24 test set~\cite{s24} (1/4 linear sRGB/sRGB).
The best results are highlighted in \textbf{\colorbox{best}{yellow}}.}
\scalebox{0.8}{
\label{tab:ablation-mb-ca}
\begin{tabular}{|l|c|c|}
\hline
\multicolumn{1}{|c|}{} &
  \multicolumn{2}{c|}{\cellcolor{red}\textcolor{white}{\shortstack{\\\textbf{S24 Test Set}\\\textbf{(1/4 LsRGB/sRGB)}}}} \\ \cline{2-3}

\multicolumn{1}{|c|}{\multirow{-2}{*}{\textbf{Photofinishing Networks}}} &
  \textbf{PSNR}\,$\uparrow$ &
  \textbf{SSIM}\,$\uparrow$ \\ \hline
w/o MBConv & 26.95 & 0.934 \\ \hline
w/o CA & 26.93 & 0.933 \\ \hline
Ours (w/ MBConv \& CA) & \textbf{\cellcolor{best}{27.49}} & \textbf{\cellcolor{best}{0.939}} \\ \hline
\end{tabular}
}
\end{table}

\noindent\\\textbf{Tone-Mapping Domain.}
In our tone-mapping design, we apply both the global and local tone-mapping operators directly to the linear sRGB representation instead of to the luma (Y) channel of the YCbCr representation at each stage. While operating in the luma domain may seem more intuitive (since it isolates luminance from chroma), we found that this separation leads to performance degradation. Specifically, applying tone mapping (global and local) to the Y channel and then allowing the chroma-mapping network to process the CbCr components resulted in noticeably poorer qualitative and quantitative results.
Table~\ref{tab:ablation-tm-design} compares our proposed design against the luma-based variant on the S24 test set~\cite{s24}.
As shown, applying tone mapping in the linear RGB domain yields superior PSNR and SSIM values.
See Fig.~\ref{fig:ablation-old-design} for a qualitative comparison.

\begin{table}[t]
\centering
\caption{Comparison between our tone-mapping design and a luma-based tone-mapping design.
In the luma-based design, global and local tone-mapping are applied only to the Y
(luminance) channel of the YCbCr color space. In our design, tone-mapping is applied jointly
to all RGB channels in the linear sRGB domain. Results are reported on the S24 test
set~\cite{s24} (1/4 linear sRGB/sRGB). The best results are highlighted in
\textbf{\colorbox{best}{yellow}}.}
\scalebox{0.8}{
\label{tab:ablation-tm-design}
\begin{tabular}{|l|c|c|}
\hline
\multicolumn{1}{|c|}{} &
  \multicolumn{2}{c|}{\cellcolor{red}\textcolor{white}{\shortstack{\\\textbf{S24 Test Set}\\\textbf{(1/4 LsRGB/sRGB)}}}} \\ \cline{2-3}

\multicolumn{1}{|c|}{\multirow{-2}{*}{\textbf{Design}}} &
  \textbf{PSNR}\,$\uparrow$ &
  \textbf{SSIM}\,$\uparrow$ \\ \hline
Luma-based TM design & 25.80 & 0.922 \\ \hline
Ours & \textbf{\cellcolor{best}{27.49}} & \textbf{\cellcolor{best}{0.939}} \\ \hline
\end{tabular}
}
\end{table}

\begin{figure*}[!t]
\centering
\includegraphics[width=0.75\linewidth]{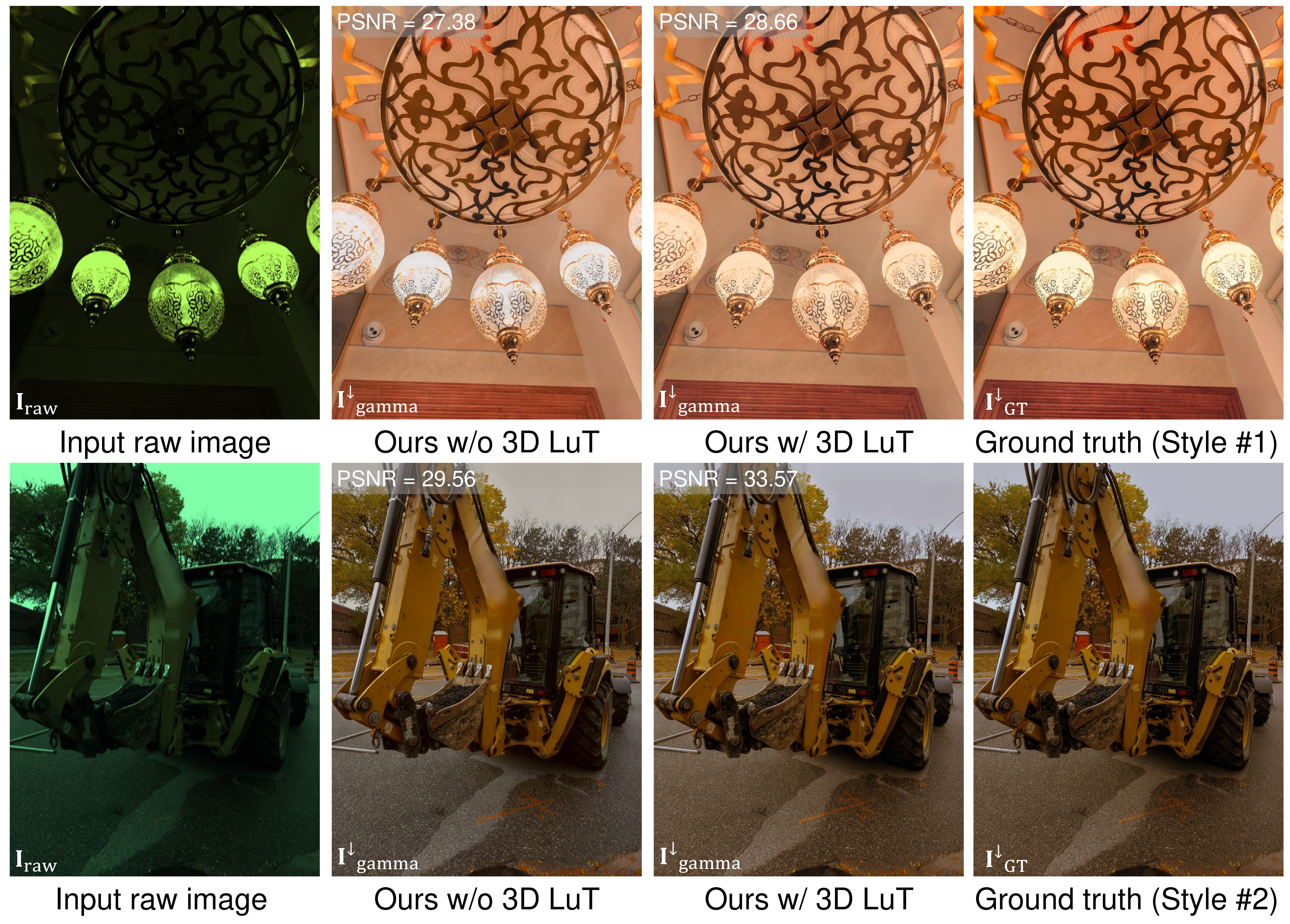}
\vspace{-1mm}
\caption{Impact of incorporating a global 3D LuT when learning artistic picture styles.
Results are shown on the S24 test set~\cite{s24}, where the input to the photofinishing module is the linear sRGB version of the pseudo ground-truth denoised raw image downsampled to one-quarter of the original resolution.}
\label{fig:3d_lut_impact}
\vspace{-2mm}
\end{figure*}

\noindent\\\textbf{3D Lookup Table.}
In the main paper, we described an optional learnable global 3D~LuT designed to improve the rendering of \textit{artistic} picture styles.
Specifically, we learn a global, image-independent 11$\times$11$\times$11 RGB LuT, $\mathbf{L}_\texttt{RGB}$, which is applied before the chroma-mapping operator.
This 3D LuT helps capture more aggressive color transformations that the 2D chroma LuT, $\mathbf{L}_\texttt{chroma}$, constructed by our chroma-mapping network is limited in modeling (see Fig.~\ref{fig:3d_lut_impact} for a qualitative example).

However, we observed that when learning simpler picture styles--such as the default color style used in the S24 dataset~\cite{s24} (Style~\#0)--the 3D LuT provides negligible benefit.
We evaluated our photofinishing module with and without the 3D LuT, where the input is the linear sRGB image derived from the pseudo ground-truth denoised raw data at one-quarter of the original raw resolution. Similar to our other ablations, we perform this analysis in isolation from other components such as guided upsampling or denoising to focus on the photofinishing performance.

Table~\ref{tab:ablation-3dlut-artistic} reports the results on the artistic picture styles of the S24 dataset (Styles~\#1-5).
As shown, incorporating the 3D LuT provides a consistent improvement across styles, with only a minor increase of approximately 4,000 parameters per style. In contrast, Table~\ref{tab:ablation-3dlut} shows that for the default picture style (Style~\#0), the 3D LuT yields no significant improvement and even slightly degrades SSIM.

\begin{table*}[t]
\centering
\caption{Ablation on the effect of the learned 3D~LuT for rendering artistic picture styles using our photofinishing module. Results are shown on the S24 test set~\cite{s24}, where the input consists of pseudo ground-truth denoised raw images mapped to the linear sRGB space and downsampled to one-quarter of the original resolution, and the ground truth is the corresponding sRGB image in the target picture style at the same resolution. The evaluation covers five distinct artistic picture styles (Styles~\#1--5). The best results are highlighted in \textbf{\colorbox{best}{yellow}}.}
\scalebox{0.7}{
\label{tab:ablation-3dlut-artistic}
\begin{tabular}{|l|cccccccccc|c|}
\hline
\multirow{3}{*}{\textbf{Photofinishing Module}} &
  \multicolumn{10}{c|}{\cellcolor{red}\textcolor{white}{\textbf{S24 Test Set (1/4 LsRGB/sRGB)}}} &
  \multirow{3}{*}{\textbf{\# params (per style)}} \\ \cline{2-11}
 &
  \multicolumn{2}{c|}{\cellcolor{orange}\textcolor{white}{\textbf{Style~\#1}}} &
  \multicolumn{2}{c|}{\cellcolor{orange2}\textcolor{white}{\textbf{Style~\#2}}} &
  \multicolumn{2}{c|}{\cellcolor{orange}\textcolor{white}{\textbf{Style~\#3}}} &
  \multicolumn{2}{c|}{\cellcolor{orange2}\textcolor{white}{\textbf{Style~\#4}}} &
  \multicolumn{2}{c|}{\cellcolor{orange}\textcolor{white}{\textbf{Style~\#5}}} &
   \\ \cline{2-11}
 &
  \multicolumn{1}{c|}{\textbf{PSNR}\,$\uparrow$} &
  \multicolumn{1}{c|}{\textbf{SSIM}\,$\uparrow$} &
  \multicolumn{1}{c|}{\textbf{PSNR}\,$\uparrow$} &
  \multicolumn{1}{c|}{\textbf{SSIM}\,$\uparrow$} &
  \multicolumn{1}{c|}{\textbf{PSNR}\,$\uparrow$} &
  \multicolumn{1}{c|}{\textbf{SSIM}\,$\uparrow$} &
  \multicolumn{1}{c|}{\textbf{PSNR}\,$\uparrow$} &
  \multicolumn{1}{c|}{\textbf{SSIM}\,$\uparrow$} &
  \multicolumn{1}{c|}{\textbf{PSNR}\,$\uparrow$} &
  \textbf{SSIM}\,$\uparrow$ &
   \\ \hline
Without 3D~LuT &
  \multicolumn{1}{c|}{26.08} &
  \multicolumn{1}{c|}{0.926} &
  \multicolumn{1}{c|}{28.70} &
  \multicolumn{1}{c|}{0.923} &
  \multicolumn{1}{c|}{26.36} &
  \multicolumn{1}{c|}{0.924} &
  \multicolumn{1}{c|}{26.24} &
  \multicolumn{1}{c|}{0.916} &
  \multicolumn{1}{c|}{28.03} &
  0.949 &
  207,224 \\ \hline
With 3D~LuT &
  \multicolumn{1}{c|}{\textbf{\cellcolor{best}{27.61}}} &
  \multicolumn{1}{c|}{\textbf{\cellcolor{best}{0.935}}} &
  \multicolumn{1}{c|}{\textbf{\cellcolor{best}{30.02}}} &
  \multicolumn{1}{c|}{\textbf{\cellcolor{best}{0.935}}} &
  \multicolumn{1}{c|}{\textbf{\cellcolor{best}{27.78}}} &
  \multicolumn{1}{c|}{\textbf{\cellcolor{best}{0.937}}} &
  \multicolumn{1}{c|}{\textbf{\cellcolor{best}{27.52}}} &
  \multicolumn{1}{c|}{\textbf{\cellcolor{best}{0.932}}} &
  \multicolumn{1}{c|}{\textbf{\cellcolor{best}{29.45}}} &
  \textbf{\cellcolor{best}{0.968}} &
  211,217 \\ \hline
\end{tabular}}

\end{table*}

\begin{table}[t]
\centering
\caption{Ablation on the effect of using the 3D~LuT in our photofinishing module for the default style (Style~\#0) on the S24 dataset~\cite{s24}. The input consists of pseudo ground-truth denoised raw images mapped to the linear sRGB space and downsampled to one-quarter of the original resolution, and the ground truth is the corresponding sRGB image at the same resolution. The 3D~LuT mainly benefits target picture styles with \textit{artistic} appearance, whereas for simpler styles that mainly enhance color and overall tone, the gain is marginal. The best results are highlighted in \textbf{\colorbox{best}{yellow}}.}
\scalebox{0.8}{
\label{tab:ablation-3dlut}
\begin{tabular}{|l|c|c|}
\hline
\multicolumn{1}{|c|}{} &
  \multicolumn{2}{c|}{\cellcolor{red}\textcolor{white}{\textbf{S24 Test Set (1/4 LsRGB/sRGB)}}} \\ \cline{2-3}
\multicolumn{1}{|c|}{\multirow{-2}{*}{\textbf{Photofinishing Module}}} &
  \textbf{PSNR}\,$\uparrow$ &
  \textbf{SSIM}\,$\uparrow$ \\ \hline
Without 3D~LuT & 27.49 & \textbf{\cellcolor{best}{0.939}} \\ \hline
With 3D~LuT & \textbf{\cellcolor{best}{27.53}} & 0.935 \\ \hline
\end{tabular}
}
\end{table}

\subsection{Additional Results}
\label{sec:supp-additional-results}

In this subsection, we provide additional results, including visual outputs of intermediate stages, quantitative results on an additional dataset, detailed evaluations across the S24 dataset styles, and further qualitative examples.

\subsubsection{Visualization of Photofinishing Stages}
\label{sec:supp-intermediate}

\begin{figure*}[!t]
\centering
\includegraphics[width=0.96\linewidth]{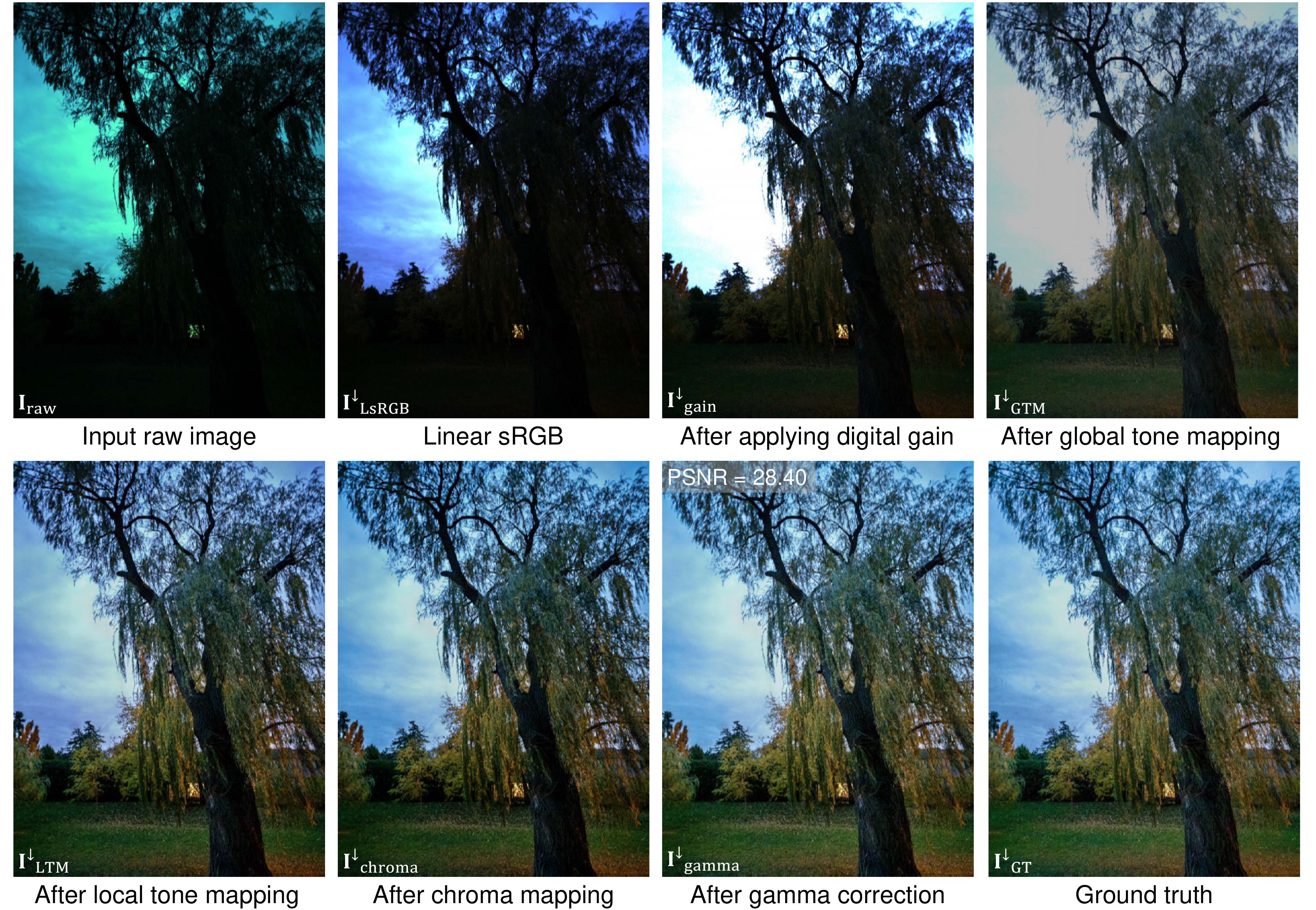}
\vspace{-1mm}
\caption{Intermediate outputs of our photofinishing module for the default picture style (Style~\#0).
The example is from the S24 test set~\cite{s24}.
The input to the photofinishing module is the pseudo ground-truth denoised raw image mapped to the linear sRGB space at one-quarter of the original resolution,
and the ground-truth reference is the corresponding Style~\#0 sRGB image at the same resolution.}
\label{fig:photofinishing-intermediate}
\vspace{-2mm}
\end{figure*}

\begin{figure*}[!t]
\centering
\includegraphics[width=0.81\linewidth]{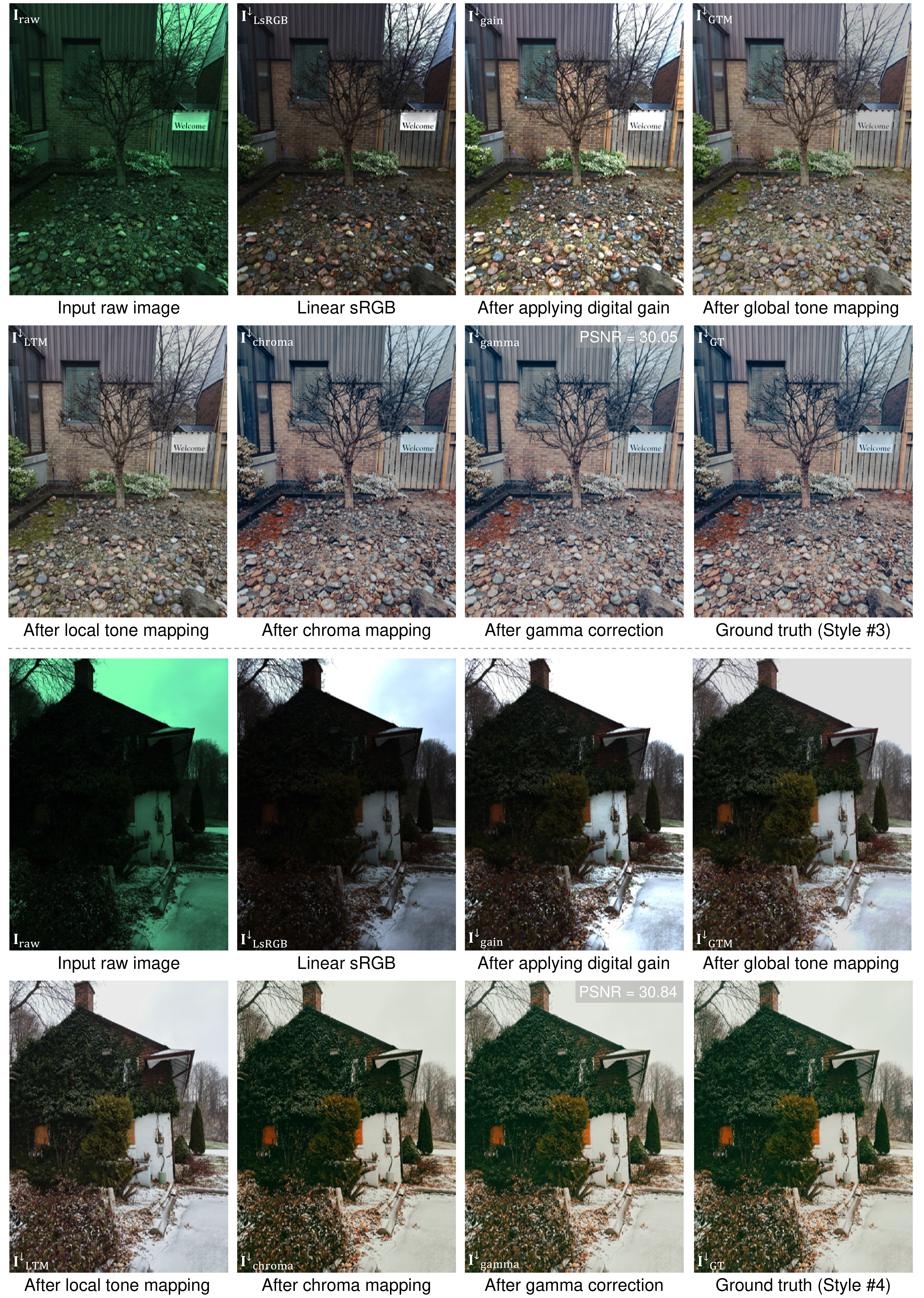}
\caption{Intermediate outputs of our photofinishing module for two artistic picture styles (Style~\#3 and Style~\#4).
Each example is taken from the S24 test set~\cite{s24}.
Inputs are pseudo ground-truth denoised raw images mapped to linear sRGB space at one-quarter of the original resolution,
and the corresponding sRGB ground-truth references are shown at the same resolution.}
\vspace{-3mm}
\label{fig:style-intermediate}
\end{figure*}

\begin{figure*}[!t]
\centering
\includegraphics[width=0.87\linewidth]{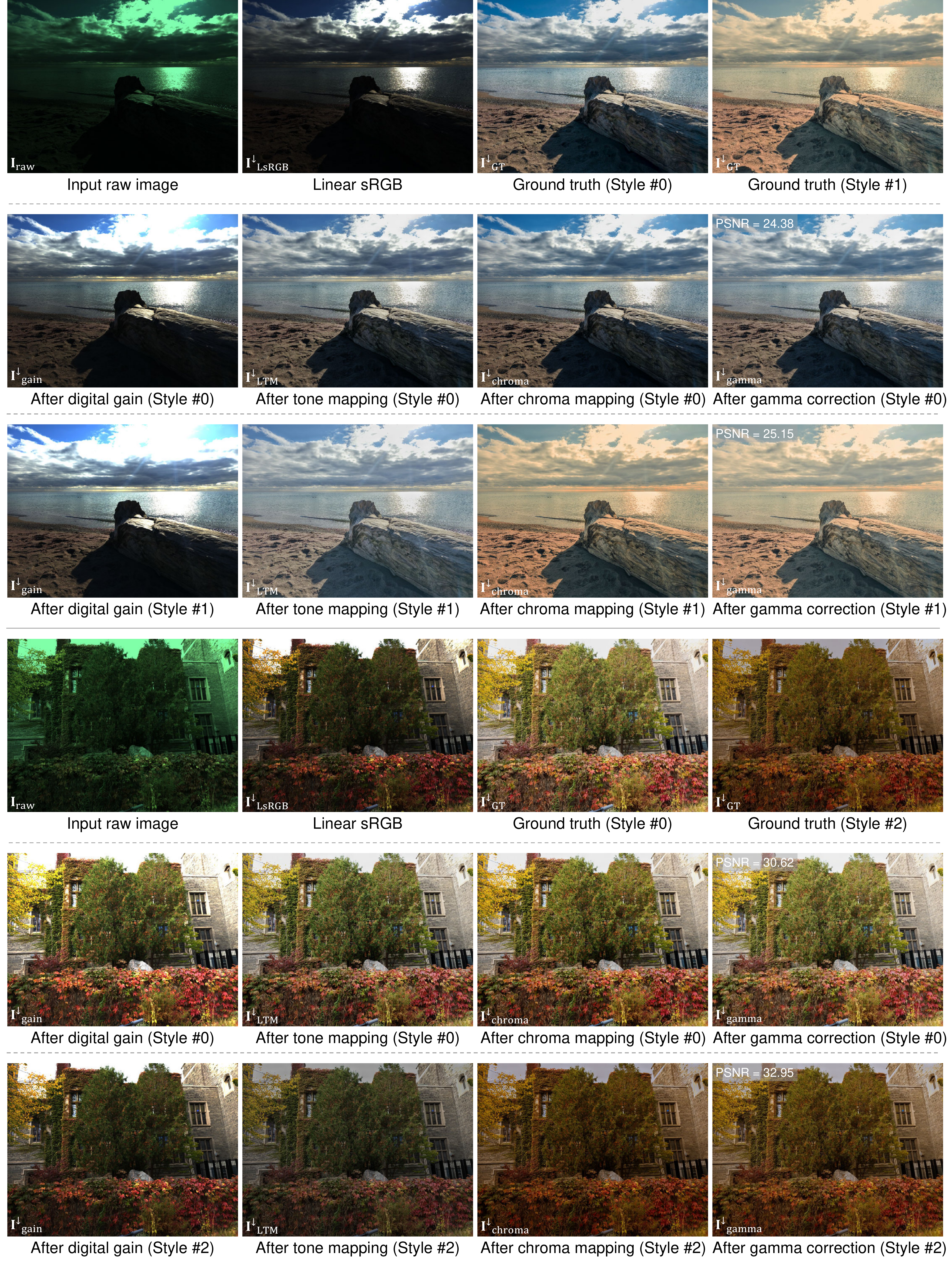}
\vspace{-5mm}
\caption{Comparison of intermediate-stage outputs for the same scene rendered with different picture styles.
Images are taken from the S24 test set~\cite{s24}.
The examples illustrate how the photofinishing stages adapt their tone and chroma adjustments depending on the target picture style.}
\label{fig:styles-intermediate-compare}
\vspace{-2mm}
\end{figure*}

Figure~\myref{fig:intermediate-example-style0}{\textcolor{section}{6}} visualizes intermediate stages of our pipeline (after denoising and color correction, digital gain, tone mapping, chroma mapping, gamma correction, and detail enhancement) using the default picture style (Style~\#0). Another example of the default style is shown in Fig.~\ref{fig:photofinishing-intermediate}. Figure~\ref{fig:style-intermediate} shows two examples rendered with different artistic styles (Styles~\#3 and~\#4), illustrating how the module’s internal representations adapt across styles. Figure~\ref{fig:styles-intermediate-compare} compares the intermediate outputs of the same scene across different picture styles, highlighting how the tone and color transformations learned by each network component vary with style.

\subsubsection{Results on MIT-Adobe 5K Dataset}
\label{sec:results-adobe-5k}

In the main paper, we reported our results on the S24 dataset, which includes multiple picture styles, including artistic styles.
To further evaluate our method against other alternatives, we trained and tested it, along with competing methods, on the MIT-Adobe FiveK dataset \cite{Adobe5K}, using Expert~C as the ground truth. For our method, we generated pseudo ground-truth denoised images following the same procedure used for the S24 dataset, employing the AI-based denoiser available in Adobe Lightroom.
Further details on how we trained and evaluated other methods on this dataset are provided in Sec.~\ref{sec:evaluation_details}. The results are reported in Table~\ref{tab:adobe5k}.~Our method achieves promising results, ranking second in PSNR and first in SSIM, while maintaining a fully modular design with controllable rendering stages.
Moreover, the lite version of our method surpasses several recent methods despite having significantly fewer parameters
(21.27~dB with $\sim$0.5~M parameters vs.\ ISPDiffuser~\cite{ispdiffuser} at 19.44~dB with $\sim$21~M parameters, and ParamISP~\cite{paramisp} at 21.17~dB with $\sim$1.4~M parameters).

\begin{table}[t]
\caption{Results on the MIT-Adobe 5K dataset~\cite{Adobe5K}. We report PSNR, SSIM~\cite{ssim}, LPIPS~\cite{LPIPS}, and $\Delta$E 2000~\cite{delta-e}, along with the total number of parameters for each method. The best results are highlighted in \textbf{\colorbox{best}{yellow}}, and the second best in \textbf{\colorbox{lightgreen}{green}}.}
\centering
\scalebox{0.75}{
\label{tab:adobe5k}
\begin{tabular}{|l|c|c|c|c|c|}
\hline
\multicolumn{1}{|c|}{} &
  \multicolumn{5}{c|}{\cellcolor{red}\textcolor{white}{\textbf{Adobe 5K Test Set}}} \\ \cline{2-6}
\multicolumn{1}{|c|}{\multirow{-2}{*}{\textbf{Method}}} &
  \textbf{PSNR}\,$\uparrow$ &
  \textbf{SSIM}\,$\uparrow$ &
  \textbf{LPIPS}\,$\downarrow$ &
  $\Delta$\textbf{E 2000}\,$\downarrow$ &
  \textbf{\# params} \\ \hline
ISPDiffuser~\cite{ispdiffuser}         & 19.44 & 0.711 & 0.333 & 11.030 & 20,938,890 \\ \hline
PyNet~\cite{zurich}                    & 19.18 & 0.695 & 0.219 & 12.300 & 47,548,170 \\ \hline
CIE-XYZ Net~\cite{afifi2021cie}        & 20.25 & 0.715 & 0.189 & 10.352 & 1,348,789 \\ \hline
Invertible-ISP~\cite{invertible}       & 20.06 & 0.693 & 0.205 & 10.800 & 1,413,760 \\ \hline
LAN~\cite{lan}                         & 16.46 & 0.618 & 0.275 & 14.817 & 46,847 \\ \hline
MicroISP~\cite{microisp}               & 19.84 & 0.708 & 0.217 & 12.188 & 13,560 \\ \hline
ParamISP~\cite{paramisp}               & 21.17 & 0.735 & \textbf{\cellcolor{secondbest}{0.184}} & \textbf{\cellcolor{secondbest}{9.729}} & 1,420,000 \\ \hline
LiteISP~\cite{lite-isp}                & \textbf{\cellcolor{best}{21.45}} & \textbf{\cellcolor{secondbest}{0.740}} & \textbf{\cellcolor{best}{0.169}} & \textbf{\cellcolor{best}{9.687}} & 9,094,000 \\ \hline
FourierISP~\cite{fourier}              & 20.29 & 0.737 & 0.208 & 10.582 & 7,589,736 \\ \hline
Exposure~\cite{hu2018exposure}         & 19.42 & 0.687 & 0.251 & 11.660 & 6,124,000 \\ \specialrule{1pt}{0pt}{0pt}
Ours (lite,$\texttt{ }$ w/o enhancement)  & 21.10 & 0.733 & 0.223 & 11.021 & 452,447 \\ \hline
Ours (base,\hspace{1.5mm}w/o enhancement)  & 21.10 & 0.733 & 0.226 & 11.020 & 1,139,907 \\ \hline
Ours (large, w/o enhancement) & 21.11 & 0.736 & 0.225 & 11.005 & 3,841,547 \\ \hline
Ours (lite,$\texttt{ }$ w/$\texttt{ }$ enhancement)    & 21.27 & 0.739 & 0.219 & 10.765 & 503,082 \\ \hline
Ours (base,\hspace{1.5mm}w/$\texttt{ }$ enhancement)   & 21.28 & \textbf{\cellcolor{secondbest}{0.740}} & 0.221 & 10.762 & 1,190,542 \\ \hline
Ours (large, w/$\texttt{ }$ enhancement)    & \textbf{\cellcolor{secondbest}{21.29}} & \textbf{\cellcolor{best}{0.742}} & 0.220 & 10.743 & 3,892,182 \\ \hline
\end{tabular}}
\end{table}

\subsubsection{Re-Rendering Results with Embedded Raw}
\label{sec:re-rendering-results}
In Sec.~\ref{sec:tool-rerendering}, we described our approach for embedding JPEG-compressed raw data alongside the rendered sRGB image within the saved file. This design enables an unlimited number of post-capture re-rendering operations without cumulative degradation in accuracy or the need to reconstruct raw data from the rendered sRGB image, while introducing only a moderate increase in the final JPEG file size compared to alternatives such as iPhone’s HEIC format for picture styles or saving the original DNG file.

Here, we present a quantitative comparison against alternative methods that enable image re-rendering through raw reconstruction.

Specifically, we compare our embedded-raw approach (Sec.~\ref{sec:tool-rerendering}) with InvISP~\cite{invertible} and ParamISP~\cite{paramisp}, both of which reconstruct raw data from the rendered image to allow re-rendering with different configurations (i.e., using different model weights corresponding to distinct picture styles). Our evaluation scheme is as follows. Each test image in the S24 dataset~\cite{s24} is first rendered into a target style (referred to here as the \textit{source picture style}) using InvISP or ParamISP, followed by raw reconstruction and subsequent re-rendering into the remaining five picture styles using the models trained for those \textit{target picture styles}.

For our method, we directly extract and decode the embedded raw data, then re-render it using our pipeline with the photofinishing and detail-enhancement models of the target picture style. In this experiment, we used the Raw-JPEG Adapter~\cite{afifi2025jpeg} with a raw-JPEG quality setting of 75, which introduces approximately 1-2~MB of overhead for the embedded encoded raw data.
As shown in Tables~\ref{tab:rerender_style0}-\ref{tab:rerender_style5}, our method achieves consistently superior accuracy and avoids the variability observed in other methods, whose re-rendering quality depends on the source picture style of the initially rendered sRGB image.

\begin{table}[t]
\centering
\caption{Re-rendering results for target picture style~\#0 on the S24 dataset~\cite{s24}.
We report PSNR, SSIM~\cite{ssim}, LPIPS~\cite{LPIPS}, and $\Delta$E~2000~\cite{delta-e}.
For Invertible-ISP~\cite{invertible} and ParamISP~\cite{paramisp}, images were first rendered
to sRGB using the source picture style (as indicated in the table) and then reconstructed
to raw before re-rendering to the target style.
For our method, the embedded raw data were used directly for re-rendering, with the base denoising model applied.
The best results are highlighted in \textbf{\colorbox{best}{yellow}}.\label{tab:rerender_style0}}

\scalebox{0.8}{
\begin{tabular}{|l|c|c|c|c|}
\hline
\multicolumn{1}{|c|}{} &
  \multicolumn{4}{c|}{\cellcolor{red}\textcolor{white}{\textbf{S24 Test Set (Target: Style \#0)}}} \\ \cline{2-5}
\multicolumn{1}{|c|}{\multirow{-2}{*}{\textbf{Method}}} &
  \textbf{PSNR}\,$\uparrow$ &
  \textbf{SSIM}\,$\uparrow$ &
  \textbf{LPIPS}\,$\downarrow$ &
  $\Delta$\textbf{E 2000}\,$\downarrow$ \\ \hline
InvISP (source: Style \#1) & 22.69 & 0.817 & 0.152 & 7.319 \\ \hline
InvISP (source: Style \#2) & 22.51 & 0.822 & 0.154 & 7.419 \\ \hline
InvISP (source: Style \#3) & 22.74 & 0.812 & 0.160 & 7.438 \\ \hline
InvISP (source: Style \#4) & 22.50 & 0.809 & 0.160 & 7.550 \\ \hline
InvISP (source: Style \#5) & 22.82 & 0.801 & 0.185 & 7.882 \\ \hline
ParamISP (source: Style \#1) & 21.71 & 0.789 & 0.181 & 8.072 \\ \hline
ParamISP (source: Style \#2) & 22.67 & 0.805 & 0.154 & 7.478 \\ \hline
ParamISP (source: Style \#3) & 21.50 & 0.786 & 0.217 & 9.844 \\ \hline
ParamISP (source: Style \#4) & 22.40 & 0.798 & 0.170 & 7.937 \\ \hline
ParamISP (source: Style \#5) & 16.88 & 0.730 & 0.402 & 16.532 \\ \specialrule{1pt}{0pt}{0pt}
Ours (base denoiser) & \textbf{\cellcolor{best}{26.90}} & \textbf{\cellcolor{best}{0.901}} & \textbf{\cellcolor{best}{0.066}} & \textbf{\cellcolor{best}{4.256}} \\ \hline
\end{tabular}}
\end{table}

\begin{table}[t]
\centering
\caption{Re-rendering results for target picture style~\#1 on the S24 dataset~\cite{s24}.
For Invertible-ISP~\cite{invertible} and ParamISP~\cite{paramisp}, images were first rendered
to sRGB using the source picture style and then reconstructed to raw before re-rendering.
For our method, the embedded raw data were used directly for re-rendering, with the base denoising model applied.
The best results are highlighted in \textbf{\colorbox{best}{yellow}}.\label{tab:rerender_style1}}

\scalebox{0.8}{
\begin{tabular}{|l|c|c|c|c|}
\hline
\multicolumn{1}{|c|}{} &
  \multicolumn{4}{c|}{\cellcolor{red}\textcolor{white}{\textbf{S24 Test Set (Target: Style \#1)}}} \\ \cline{2-5}
\multicolumn{1}{|c|}{\multirow{-2}{*}{\textbf{Method}}} &
  \textbf{PSNR}\,$\uparrow$ &
  \textbf{SSIM}\,$\uparrow$ &
  \textbf{LPIPS}\,$\downarrow$ &
  $\Delta$\textbf{E 2000}\,$\downarrow$ \\ \hline
InvISP (source: Style \#0) & 21.38 & 0.810 & 0.211 & 8.676 \\ \hline
InvISP (source: Style \#2) & 23.27 & 0.833 & 0.164 & 6.509 \\ \hline
InvISP (source: Style \#3) & 23.46 & 0.830 & 0.166 & 6.530 \\ \hline
InvISP (source: Style \#4) & 23.29 & 0.824 & 0.165 & 6.609 \\ \hline
InvISP (source: Style \#5) & 22.85 & 0.828 & 0.181 & 7.881 \\ \hline
ParamISP (source: Style \#0) & 23.11 & 0.814 & 0.169 & 7.065 \\ \hline
ParamISP (source: Style \#2) & 22.80 & 0.814 & 0.166 & 7.135 \\ \hline
ParamISP (source: Style \#3) & 21.84 & 0.796 & 0.215 & 8.687 \\ \hline
ParamISP (source: Style \#4) & 22.74 & 0.810 & 0.180 & 7.517 \\ \hline
ParamISP (source: Style \#5) & 17.36 & 0.754 & 0.374 & 14.479 \\ \specialrule{1pt}{0pt}{0pt}
Ours (base denoiser) & \textbf{\cellcolor{best}{26.68}} & \textbf{\cellcolor{best}{0.897}} & \textbf{\cellcolor{best}{0.088}} & \textbf{\cellcolor{best}{4.417}} \\ \hline
Ours (base denoiser + 3D LuT) & 26.04 & 0.890 & 0.100 & 4.989 \\ \hline
\end{tabular}}
\end{table}

\begin{table}[t]
\centering
\caption{Re-rendering results for target picture style~\#2 on the S24 dataset~\cite{s24}.
The best results are highlighted in \textbf{\colorbox{best}{yellow}}.\label{tab:rerender_style2}}
\scalebox{0.8}{
\begin{tabular}{|l|c|c|c|c|}
\hline
\multicolumn{1}{|c|}{} &
  \multicolumn{4}{c|}{\cellcolor{red}\textcolor{white}{\textbf{S24 Test Set (Target: Style \#2)}}} \\ \cline{2-5}
\multicolumn{1}{|c|}{\multirow{-2}{*}{\textbf{Method}}} &
  \textbf{PSNR}\,$\uparrow$ &
  \textbf{SSIM}\,$\uparrow$ &
  \textbf{LPIPS}\,$\downarrow$ &
  $\Delta$\textbf{E 2000}\,$\downarrow$ \\ \hline
InvISP (source: Style \#0) & 26.17 & 0.843 & 0.124 & 5.301 \\ \hline
InvISP (source: Style \#1) & 26.14 & 0.853 & 0.121 & 5.260 \\ \hline
InvISP (source: Style \#3) & 26.23 & 0.851 & 0.129 & 5.280 \\ \hline
InvISP (source: Style \#4) & 26.19 & 0.847 & 0.125 & 5.283 \\ \hline
InvISP (source: Style \#5) & 25.91 & 0.841 & 0.152 & 5.611 \\ \hline
ParamISP (source: Style \#0) & 25.67 & 0.843 & 0.131 & 5.647 \\ \hline
ParamISP (source: Style \#1) & 24.73 & 0.832 & 0.157 & 6.179 \\ \hline
ParamISP (source: Style \#3) & 24.09 & 0.827 & 0.179 & 7.455 \\ \hline
ParamISP (source: Style \#4) & 24.94 & 0.834 & 0.148 & 6.423 \\ \hline
ParamISP (source: Style \#5) & 19.86 & 0.743 & 0.330 & 11.855 \\ \specialrule{1pt}{0pt}{0pt}
Ours (base denoiser) & \textbf{\cellcolor{best}{29.48}} & \textbf{\cellcolor{best}{0.923}} & \textbf{\cellcolor{best}{0.067}} & \textbf{\cellcolor{best}{3.605}} \\ \hline
Ours (base denoiser + 3D LuT) & 28.74 & 0.915 & 0.082 & 4.216 \\ \hline
\end{tabular}}
\end{table}

\begin{table}[t]
\centering
\caption{Re-rendering results for target picture style~\#3 on the S24 dataset~\cite{s24}.
The best results are highlighted in \textbf{\colorbox{best}{yellow}}.\label{tab:rerender_style3}}
\scalebox{0.8}{
\begin{tabular}{|l|c|c|c|c|}
\hline
\multicolumn{1}{|c|}{} &
  \multicolumn{4}{c|}{\cellcolor{red}\textcolor{white}{\textbf{S24 Test Set (Target: Style \#3)}}} \\ \cline{2-5}
\multicolumn{1}{|c|}{\multirow{-2}{*}{\textbf{Method}}} &
  \textbf{PSNR}\,$\uparrow$ &
  \textbf{SSIM}\,$\uparrow$ &
  \textbf{LPIPS}\,$\downarrow$ &
  $\Delta$\textbf{E 2000}\,$\downarrow$ \\ \hline
InvISP (source: Style \#0) & 23.82 & 0.851 & 0.146 & 7.316 \\ \hline
InvISP (source: Style \#1) & 23.67 & 0.853 & 0.149 & 7.327 \\ \hline
InvISP (source: Style \#2) & 23.56 & 0.858 & 0.147 & 7.394 \\ \hline
InvISP (source: Style \#4) & 23.64 & 0.852 & 0.151 & 7.413 \\ \hline
InvISP (source: Style \#5) & 23.39 & 0.840 & 0.181 & 7.651 \\ \hline
ParamISP (source: Style \#0) & 23.31 & 0.834 & 0.151 & 7.126 \\ \hline
ParamISP (source: Style \#1) & 22.49 & 0.821 & 0.179 & 7.781 \\ \hline
ParamISP (source: Style \#2) & 23.15 & 0.836 & 0.150 & 7.267 \\ \hline
ParamISP (source: Style \#4) & 22.89 & 0.828 & 0.160 & 7.469 \\ \hline
ParamISP (source: Style \#5) & 18.64 & 0.768 & 0.303 & 12.829 \\ \specialrule{1pt}{0pt}{0pt}
Ours (base denoiser) & \textbf{\cellcolor{best}{26.89}} & \textbf{\cellcolor{best}{0.905}} & \textbf{\cellcolor{best}{0.085}} & \textbf{\cellcolor{best}{4.660}} \\ \hline
Ours (base denoiser + 3D LuT) & 26.35 & 0.897 & 0.101 & 5.318 \\ \hline
\end{tabular}}
\end{table}

\begin{table}[t]
\centering
\caption{Re-rendering results for target picture style~\#4 on the S24 dataset~\cite{s24}.
The best results are highlighted in \textbf{\colorbox{best}{yellow}}.\label{tab:rerender_style4}}

\scalebox{0.8}{
\begin{tabular}{|l|c|c|c|c|}
\hline
\multicolumn{1}{|c|}{} &
  \multicolumn{4}{c|}{\cellcolor{red}\textcolor{white}{\textbf{S24 Test Set (Target: Style \#4)}}} \\ \cline{2-5}
\multicolumn{1}{|c|}{\multirow{-2}{*}{\textbf{Method}}} &
  \textbf{PSNR}\,$\uparrow$ &
  \textbf{SSIM}\,$\uparrow$ &
  \textbf{LPIPS}\,$\downarrow$ &
  $\Delta$\textbf{E 2000}\,$\downarrow$ \\ \hline
InvISP (source: Style \#0) & 23.30 & 0.842 & 0.149 & 7.647 \\ \hline
InvISP (source: Style \#1) & 23.24 & 0.843 & 0.148 & 7.686 \\ \hline
InvISP (source: Style \#2) & 23.27 & 0.846 & 0.146 & 7.605 \\ \hline
InvISP (source: Style \#3) & 23.31 & 0.841 & 0.151 & 7.639 \\ \hline
InvISP (source: Style \#5) & 22.85 & 0.828 & 0.181 & 7.881 \\ \hline
ParamISP (source: Style \#0) & 22.90 & 0.824 & 0.141 & 6.585 \\ \hline
ParamISP (source: Style \#1) & 22.01 & 0.806 & 0.171 & 7.398 \\ \hline
ParamISP (source: Style \#2) & 22.88 & 0.825 & 0.140 & 6.622 \\ \hline
ParamISP (source: Style \#3) & 22.09 & 0.811 & 0.167 & 7.441 \\ \hline
ParamISP (source: Style \#5) & 18.64 & 0.744 & 0.323 & 12.482 \\ \specialrule{1pt}{0pt}{0pt}
Ours (base denoiser) & \textbf{\cellcolor{best}{26.44}} & \textbf{\cellcolor{best}{0.897}} & \textbf{\cellcolor{best}{0.080}} & \textbf{\cellcolor{best}{4.428}} \\ \hline
Ours (base denoiser + 3D LuT) & 26.21 & 0.893 & 0.087 & 4.692 \\ \hline
\end{tabular}}
\end{table}

\begin{table}[t]
\centering
\caption{Re-rendering results for target picture style~\#5 on the S24 dataset~\cite{s24}.
The best results are highlighted in \textbf{\colorbox{best}{yellow}}.\label{tab:rerender_style5}}
\scalebox{0.8}{
\begin{tabular}{|l|c|c|c|c|}
\hline
\multicolumn{1}{|c|}{} &
  \multicolumn{4}{c|}{\cellcolor{red}\textcolor{white}{\textbf{S24 Test Set (Target: Style \#5)}}} \\ \cline{2-5}
\multicolumn{1}{|c|}{\multirow{-2}{*}{\textbf{Method}}} &
  \textbf{PSNR}\,$\uparrow$ &
  \textbf{SSIM}\,$\uparrow$ &
  \textbf{LPIPS}\,$\downarrow$ &
  $\Delta$\textbf{E 2000}\,$\downarrow$ \\ \hline
InvISP (source: Style \#0) & 24.89 & 0.870 & 0.168 & 5.844 \\ \hline
InvISP (source: Style \#1) & 24.91 & 0.870 & 0.165 & 5.840 \\ \hline
InvISP (source: Style \#2) & 24.82 & 0.872 & 0.162 & 5.902 \\ \hline
InvISP (source: Style \#3) & 24.83 & 0.863 & 0.175 & 5.941 \\ \hline
InvISP (source: Style \#4) & 24.78 & 0.863 & 0.174 & 5.960 \\ \hline
ParamISP (source: Style \#0) & 24.47 & 0.841 & 0.160 & 4.963 \\ \hline
ParamISP (source: Style \#1) & 24.20 & 0.834 & 0.180 & 5.122 \\ \hline
ParamISP (source: Style \#2) & 24.50 & 0.847 & 0.153 & 5.000 \\ \hline
ParamISP (source: Style \#3) & 23.95 & 0.833 & 0.170 & 5.088 \\ \hline
ParamISP (source: Style \#4) & 24.18 & 0.840 & 0.159 & 5.098 \\ \specialrule{1pt}{0pt}{0pt}
Ours (base denoiser) & 27.75 & 0.916 & 0.102 & 3.537 \\ \hline
Ours (base denoiser + 3D LuT) & \textbf{\cellcolor{best}{28.27}} & \textbf{\cellcolor{best}{0.921}} & \textbf{\cellcolor{best}{0.094}} & \textbf{\cellcolor{best}{3.332}} \\ \hline
\end{tabular}}
\end{table}

\subsubsection{Detailed Results on S24 Styles~\#1--5}
\label{sec:detailed-results-styles}

In the main paper, we reported the PSNR scores of our method compared to alternative approaches across different picture styles available in the S24 dataset.
Tables~\ref{tab:detailed-style-result-s-1}--\ref{tab:detailed-style-result-s-5} provide a more comprehensive evaluation, including SSIM~\cite{ssim}, LPIPS~\cite{LPIPS}, and $\Delta$E~2000~\cite{delta-e}.
The detailed results for Styles~\#1--5.

\begin{table}[t]
\centering
\caption{Detailed results for Style~\#1 on the S24 test set~\cite{s24}.
Our method is evaluated with different denoising model capacities (lite, base, and large), with and without the enhancement network and optional 3D~LuT.
The best results are highlighted in \textbf{\colorbox{best}{yellow}} and the second best in \textbf{\colorbox{lightgreen}{green}}.
\label{tab:detailed-style-result-s-1}}
\vspace{-2mm}
\scalebox{0.75}{
\begin{tabular}{|l|c|c|c|c|}
\hline
\multicolumn{1}{|c|}{} &
  \multicolumn{4}{c|}{\cellcolor{red}\textcolor{white}{\textbf{S24 Test Set (Style \#1)}}} \\ \cline{2-5}
\multicolumn{1}{|c|}{\multirow{-2}{*}{\textbf{Method}}} &
  \textbf{PSNR}\,$\uparrow$ &
  \textbf{SSIM}\,$\uparrow$ &
  \textbf{LPIPS}\,$\downarrow$ &
  $\Delta$\textbf{E 2000}\,$\downarrow$ \\ \hline
Exposure~\cite{hu2018exposure}       & 19.77 & 0.782 & 0.209 & 10.709 \\ \hline
ReconfigISP ~\cite{yu2021reconfigisp}& 19.56 & 0.775 & 0.208 & 9.848\\ \hline
Neural Photo-Finishing~\cite{neural_photo_finishing} & 20.43 & 0.792 & 0.222 & 10.372 \\ \hline
ISPDiffuser~\cite{ispdiffuser}       & 25.60 & 0.910 & 0.117 & 5.100 \\ \hline
PyNet~\cite{zurich}                  & 24.36 & 0.875 & 0.097 & 5.759 \\ \hline
CIE-XYZ Net~\cite{afifi2021cie}      & 22.40 & 0.859 & 0.176 & 8.291 \\ \hline
PP (cmKAN) \cite{cmkan}  & 20.93 &  0.875 & 0.139  & 8.436 \\ \hline
Deep-FlexISP \cite{flexisp}  & 24.86  & 0.891  &  0.108 &  6.150 \\ \hline
Invertible-ISP~\cite{invertible}     & 23.48 & 0.832 & 0.157 & 6.642 \\ \hline
LAN~\cite{lan}                       & 22.98 & 0.812 & 0.126 & 6.669 \\ \hline
MicroISP~\cite{microisp}             & 20.30 & 0.747 & 0.183 & 11.298 \\ \hline
ParamISP~\cite{paramisp}             & 24.97 & 0.856 & 0.123 & 5.724 \\ \hline
LiteISP~\cite{lite-isp}              & 26.66 & 0.915 & \textbf{\cellcolor{best}{0.067}} & 4.702 \\ \hline
FourierISP~\cite{fourier}            & 25.19 & \textbf{\cellcolor{best}{0.925}} & 0.099 & 5.432 \\ \specialrule{1pt}{0pt}{0pt}

Ours (lite,$\texttt{ }$ w/o enhancement)   & 25.16 & 0.866 & 0.124 & 5.609 \\ \hline
Ours (base,\hspace{1.5mm}w/o enhancement)   & 25.28 & 0.873 & 0.116 & 5.483 \\ \hline
Ours (large, w/o enhancement)  & 25.31 & 0.874 & 0.114 & 5.463 \\ \hline
Ours (lite,$\texttt{ }$ w/o enhancement, w/ 3D LuT)    & 26.39 & 0.874 & 0.086 & 4.486 \\ \hline
Ours (base,\hspace{1.5mm}w/o enhancement, w/ 3D LuT)    & 26.52 & 0.880 & 0.078 & \textbf{\cellcolor{secondbest}{4.348}} \\ \hline
Ours (large, w/o enhancement, w/ 3D LuT)   & 26.56 & 0.882 & \textbf{\cellcolor{secondbest}{0.076}} & \textbf{\cellcolor{best}{4.329}} \\ \hline
Ours (lite,$\texttt{ }$ w/$\texttt{ }$ enhancement, w/ 3D LuT)    & 26.56 & 0.906 & 0.092 & 4.807 \\ \hline
Ours (base,\hspace{1.5mm}w/$\texttt{ }$ enhancement, w/ 3D LuT)    & \textbf{\cellcolor{secondbest}{26.71}} & 0.914 & 0.085 & 4.692 \\ \hline
Ours (large, w/$\texttt{ }$ enhancement, w/ 3D LuT)   & \textbf{\cellcolor{best}{26.75}} & \textbf{\cellcolor{secondbest}{0.916}} & 0.083 & 4.668 \\ \hline
\end{tabular}}
\end{table}

\begin{table}[t]
\centering
\caption{Detailed results for Style~\#2 on the S24 test set~\cite{s24}.
The best results are highlighted in \textbf{\colorbox{best}{yellow}} and the second best in \textbf{\colorbox{lightgreen}{green}}.
\label{tab:detailed-style-result-s-2}}
\vspace{-2mm}
\scalebox{0.75}{
\begin{tabular}{|l|c|c|c|c|}
\hline
\multicolumn{1}{|c|}{} &
  \multicolumn{4}{c|}{\cellcolor{red}\textcolor{white}{\textbf{S24 Test Set (Style \#2)}}} \\ \cline{2-5}
\multicolumn{1}{|c|}{\multirow{-2}{*}{\textbf{Method}}} &
  \textbf{PSNR}\,$\uparrow$ &
  \textbf{SSIM}\,$\uparrow$ &
  \textbf{LPIPS}\,$\downarrow$ &
  $\Delta$\textbf{E 2000}\,$\downarrow$ \\ \hline
Exposure~\cite{hu2018exposure}       & 22.01 & 0.795 & 0.207 & 8.873 \\ \hline
ReconfigISP ~\cite{yu2021reconfigisp}& 23.46 & 0.817 & 0.140 & 7.153 \\ \hline
Neural Photo-Finishing~\cite{neural_photo_finishing} & 24.54 & 0.814 & 0.140 & 6.766 \\ \hline
ISPDiffuser~\cite{ispdiffuser}       & 27.30 & 0.912 & 0.116 & 4.600 \\ \hline
PyNet~\cite{zurich}                  & 25.94 & 0.890 & 0.095 & 5.432 \\ \hline
CIE-XYZ Net~\cite{afifi2021cie}      & 24.05 & 0.872 & 0.120 & 6.634 \\ \hline
PP (cmKAN) \cite{cmkan}  & 23.04  &  0.875 &  0.131 & 7.613 \\ \hline
Deep-FlexISP \cite{flexisp}  &  27.47 &  0.911 &  0.089 &  4.554 \\ \hline
Invertible-ISP~\cite{invertible}     & 26.35 & 0.861 & 0.115 & 5.229 \\ \hline
LAN~\cite{lan}                       & 23.74 & 0.782 & 0.112 & 6.436 \\ \hline
MicroISP~\cite{microisp}             & 23.66 & 0.801 & 0.155 & 8.885 \\ \hline
ParamISP~\cite{paramisp}             & 27.11 & 0.869 & 0.101 & 4.947 \\ \hline
LiteISP~\cite{lite-isp}              & 28.33 & 0.922 & \textbf{\cellcolor{best}{0.064}} & 4.284 \\ \hline
FourierISP~\cite{fourier}            & 28.03 & \textbf{\cellcolor{best}{0.928}} & 0.082 & 4.488 \\ \specialrule{1pt}{0pt}{0pt}

Ours (lite,$\texttt{ }$ w/o enhancement)   & 28.09 & 0.890 & 0.094 & 4.589 \\ \hline
Ours (base,\hspace{1.5mm}w/o enhancement)   & 28.19 & 0.893 & 0.090 & 4.516 \\ \hline
Ours (large, w/o enhancement)  & 28.22 & 0.893 & 0.089 & 4.503 \\ \hline
Ours (lite,$\texttt{ }$ w/o enhancement, w/ 3D LuT)    & 29.21 & 0.898 & 0.068 & 3.641 \\ \hline
Ours (base,\hspace{1.5mm}w/o enhancement, w/ 3D LuT)    & \textbf{\cellcolor{secondbest}{29.29}} & 0.901 & \textbf{\cellcolor{secondbest}{0.065}} & \textbf{\cellcolor{secondbest}{3.594}} \\ \hline
Ours (large, w/o enhancement, w/ 3D LuT)   & \textbf{\cellcolor{best}{29.31}} & 0.901 & 0.064 & \textbf{\cellcolor{best}{3.588}} \\ \hline
Ours (lite,$\texttt{ }$ w/$\texttt{ }$ enhancement, w/ 3D LuT)    & 28.92 & 0.920 & 0.078 & 4.153 \\ \hline
Ours (base,\hspace{1.5mm}w/$\texttt{ }$ enhancement, w/ 3D LuT)    & 28.99 & 0.923 & 0.075 & 4.135 \\ \hline
Ours (large, w/$\texttt{ }$ enhancement, w/ 3D LuT)   & 29.01 & \textbf{\cellcolor{secondbest}{0.924}} & 0.075 & 4.130 \\ \hline
\end{tabular}}
\end{table}

\begin{table}[t]
\centering
\caption{Detailed results for Style~\#3 on the S24 test set~\cite{s24}.
The best results are highlighted in \textbf{\colorbox{best}{yellow}} and the second best in \textbf{\colorbox{lightgreen}{green}}.
\label{tab:detailed-style-result-s-3}}
\vspace{-2mm}
\scalebox{0.75}{
\begin{tabular}{|l|c|c|c|c|}
\hline
\multicolumn{1}{|c|}{} &
  \multicolumn{4}{c|}{\cellcolor{red}\textcolor{white}{\textbf{S24 Test Set (Style \#3)}}} \\ \cline{2-5}
\multicolumn{1}{|c|}{\multirow{-2}{*}{\textbf{Method}}} &
  \textbf{PSNR}\,$\uparrow$ &
  \textbf{SSIM}\,$\uparrow$ &
  \textbf{LPIPS}\,$\downarrow$ &
  $\Delta$\textbf{E 2000}\,$\downarrow$ \\ \hline
Exposure~\cite{hu2018exposure}       & 19.69 & 0.803 & 0.234 & 12.029 \\ \hline
ReconfigISP ~\cite{yu2021reconfigisp}& 19.69 & 0.782 & 0.230 & 11.884\\ \hline
Neural Photo-Finishing~\cite{neural_photo_finishing} & 21.18 & 0.770 & 0.221 & 11.624 \\ \hline
ISPDiffuser~\cite{ispdiffuser}       & 25.02 & 0.899 & 0.125 & 5.600 \\ \hline
PyNet~\cite{zurich}                  & 24.70 & 0.882 & 0.096 & 5.999 \\ \hline
CIE-XYZ Net~\cite{afifi2021cie}      & 22.00 & 0.854 & 0.162 & 8.906 \\ \hline
PP (cmKAN) \cite{cmkan}  &  21.85 &  0.876 & 0.130  & 8.077 \\ \hline
Deep-FlexISP \cite{flexisp}  &  25.23 & 0.901  &  0.106 &  5.807 \\ \hline
Invertible-ISP~\cite{invertible}     & 23.84 & 0.852 & 0.144 & 7.347 \\ \hline
LAN~\cite{lan}                       & 23.47 & 0.830 & 0.118 & 6.902 \\ \hline
MicroISP~\cite{microisp}             & 21.45 & 0.792 & 0.169 & 10.751 \\ \hline
ParamISP~\cite{paramisp}             & 24.77 & 0.872 & 0.121 & 6.417 \\ \hline
LiteISP~\cite{lite-isp}              & 26.31 & 0.913 & \textbf{\cellcolor{best}{0.073}} & 5.220 \\ \hline
FourierISP~\cite{fourier}            & 25.38 & \textbf{\cellcolor{best}{0.919}} & 0.100 & 5.703 \\ \specialrule{1pt}{0pt}{0pt}

Ours (lite,$\texttt{ }$ w/o enhancement)   & 25.66 & 0.877 & 0.113 & 5.930 \\ \hline
Ours (base,\hspace{1.5mm}w/o enhancement)   & 25.73 & 0.881 & 0.110 & 5.894 \\ \hline
Ours (large, w/o enhancement)  & 25.75 & 0.882 & 0.110 & 5.883 \\ \hline
Ours (lite,$\texttt{ }$ w/o enhancement, w/ 3D LuT)    & 26.69 & 0.884 & 0.085 & 4.727 \\ \hline
Ours (base,\hspace{1.5mm}w/o enhancement, w/ 3D LuT)    & \textbf{\cellcolor{secondbest}{26.79}} & 0.889 & 0.081 & \textbf{\cellcolor{secondbest}{4.675}} \\ \hline
Ours (large, w/o enhancement, w/ 3D LuT)   & \textbf{\cellcolor{best}{26.83}} & 0.890 & \textbf{\cellcolor{secondbest}{0.080}} & \textbf{\cellcolor{best}{4.665}} \\ \hline
Ours (lite,$\texttt{ }$ w/$\texttt{ }$ enhancement, w/ 3D LuT)    & 26.78 & 0.913 & 0.090 & 5.209 \\ \hline
Ours (base,\hspace{1.5mm}w/$\texttt{ }$ enhancement, w/ 3D LuT)    & 26.78 & 0.913 & 0.090 & 5.213 \\ \hline
Ours (large, w/$\texttt{ }$ enhancement, w/ 3D LuT)   & 26.83 & \textbf{\cellcolor{secondbest}{0.914}} & 0.089 & 5.196 \\ \hline
\end{tabular}}
\end{table}

\begin{table}[t]
\centering
\caption{Detailed results for Style~\#4 on the S24 test set~\cite{s24}.
The best results are highlighted in \textbf{\colorbox{best}{yellow}} and the second best in \textbf{\colorbox{lightgreen}{green}}.
\label{tab:detailed-style-result-s-4}}
\vspace{-2mm}
\scalebox{0.75}{
\begin{tabular}{|l|c|c|c|c|}
\hline
\multicolumn{1}{|c|}{} &
  \multicolumn{4}{c|}{\cellcolor{red}\textcolor{white}{\textbf{S24 Test Set (Style \#4)}}} \\ \cline{2-5}
\multicolumn{1}{|c|}{\multirow{-2}{*}{\textbf{Method}}} &
  \textbf{PSNR}\,$\uparrow$ &
  \textbf{SSIM}\,$\uparrow$ &
  \textbf{LPIPS}\,$\downarrow$ &
  $\Delta$\textbf{E 2000}\,$\downarrow$ \\ \hline
Exposure~\cite{hu2018exposure}       & 17.34 & 0.731 & 0.297 & 15.120 \\ \hline
ReconfigISP ~\cite{yu2021reconfigisp}& 19.04 & 0.759 & 0.248 & 11.099 \\ \hline
Neural Photo-Finishing~\cite{neural_photo_finishing} & 20.59 & 0.745 & 0.220 & 10.334 \\ \hline
ISPDiffuser~\cite{ispdiffuser}       & 25.93 & 0.904 & 0.083 & 5.100 \\ \hline
PyNet~\cite{zurich}                  & 24.34 & 0.876 & 0.094 & 5.745 \\ \hline
CIE-XYZ Net~\cite{afifi2021cie}      & 22.26 & 0.846 & 0.147 & 7.624 \\ \hline
PP (cmKAN) \cite{cmkan}  & 20.91  & 0.852  & 0.140  &  8.394 \\ \hline
Deep-FlexISP \cite{flexisp}  & 23.96  &  0.878 &  0.108 & 5.965 \\ \hline
Invertible-ISP~\cite{invertible}     & 23.33 & 0.842 & 0.145 & 7.694 \\ \hline
LAN~\cite{lan}                       & 22.80 & 0.790 & 0.116 & 6.736 \\ \hline
MicroISP~\cite{microisp}             & 20.34 & 0.750 & 0.175 & 11.499 \\ \hline
ParamISP~\cite{paramisp}             & 24.18 & 0.853 & 0.118 & 6.138 \\ \hline
LiteISP~\cite{lite-isp}              & 25.04 & 0.894 & 0.082 & 5.517 \\ \hline
FourierISP~\cite{fourier}            & 24.74 & 0.906 & 0.100 & 5.591 \\ \specialrule{1pt}{0pt}{0pt}

Ours (lite,$\texttt{ }$ w/o enhancement)   & 25.47 & 0.868 & 0.107 & 5.486 \\ \hline
Ours (base,\hspace{1.5mm}w/o enhancement)   & 25.55 & 0.872 & 0.104 & 5.437 \\ \hline
Ours (large, w/o enhancement)  & 25.58 & 0.873 & 0.104 & 5.424 \\ \hline
Ours (lite,$\texttt{ }$ w/o enhancement, w/ 3D LuT)    & 26.35 & 0.880 & 0.079 & 4.414 \\ \hline
Ours (base,\hspace{1.5mm}w/o enhancement, w/ 3D LuT)    & \textbf{\cellcolor{secondbest}{26.47}} & 0.884 & \textbf{\cellcolor{secondbest}{0.076}} & \textbf{\cellcolor{secondbest}{4.365}} \\ \hline
Ours (large, w/o enhancement, w/ 3D LuT)   & 26.51 & 0.885 & 0.075 & \textbf{\cellcolor{best}{4.355}} \\ \hline
Ours (lite,$\texttt{ }$ w/$\texttt{ }$ enhancement, w/ 3D LuT)    & 26.66 & 0.905 & 0.081 & 4.601 \\ \hline
Ours (base,\hspace{1.5mm}w/$\texttt{ }$ enhancement, w/ 3D LuT)    & 26.79 & \textbf{\cellcolor{secondbest}{0.910}} & 0.077 & 4.564 \\ \hline
Ours (large, w/$\texttt{ }$ enhancement, w/ 3D LuT)   & \textbf{\cellcolor{best}{26.84}} & \textbf{\cellcolor{best}{0.911}} & \textbf{\cellcolor{best}{0.075}} & 4.553 \\ \hline
\end{tabular}}
\end{table}

\begin{table}[t]
\centering
\caption{Detailed results for Style~\#5 on the S24 test set~\cite{s24}.
The best results are highlighted in \textbf{\colorbox{best}{yellow}} and the second best in \textbf{\colorbox{lightgreen}{green}}.
\label{tab:detailed-style-result-s-5}}
\vspace{-2mm}
\scalebox{0.75}{
\begin{tabular}{|l|c|c|c|c|}
\hline
\multicolumn{1}{|c|}{} &
  \multicolumn{4}{c|}{\cellcolor{red}\textcolor{white}{\textbf{S24 Test Set (Style \#5)}}} \\ \cline{2-5}
\multicolumn{1}{|c|}{\multirow{-2}{*}{\textbf{Method}}} &
  \textbf{PSNR}\,$\uparrow$ &
  \textbf{SSIM}\,$\uparrow$ &
  \textbf{LPIPS}\,$\downarrow$ &
  $\Delta$\textbf{E 2000}\,$\downarrow$ \\ \hline
Exposure~\cite{hu2018exposure}       & 22.52 & 0.858 & 0.155 & 6.570 \\ \hline
ReconfigISP ~\cite{yu2021reconfigisp}& 22.60 & 0.838 & 0.172 & 7.021\\ \hline
Neural Photo-Finishing~\cite{neural_photo_finishing} & 22.14 & 0.842 & 0.249 & 9.619 \\ \hline
ISPDiffuser~\cite{ispdiffuser}       & 26.83 & 0.929 & 0.116 & 3.800 \\ \hline
PyNet~\cite{zurich}                  & 26.32 & 0.920 & 0.089 & 3.982 \\ \hline
CIE-XYZ Net~\cite{afifi2021cie}      & 24.67 & 0.897 & 0.133 & 5.270 \\ \hline
PP (cmKAN) \cite{cmkan}  &  21.30 &  0.880 & 0.147  & 7.338 \\ \hline
Deep-FlexISP \cite{flexisp}  &  24.65 &  0.912 &  0.124 &  8.448\\ \hline
Invertible-ISP~\cite{invertible}     & 24.90 & 0.875 & 0.163 & 5.927 \\ \hline
LAN~\cite{lan}                       & 25.38 & 0.878 & 0.108 & 4.662 \\ \hline
MicroISP~\cite{microisp}             & 22.68 & 0.823 & 0.221 & 10.500 \\ \hline
ParamISP~\cite{paramisp}             & 25.43 & 0.867 & 0.134 & 4.499 \\ \hline
LiteISP~\cite{lite-isp}              & 28.07 & 0.935 & \textbf{\cellcolor{best}{0.071}} & 3.434 \\ \hline
FourierISP~\cite{fourier}            & 27.41 & \textbf{\cellcolor{best}{0.947}} & 0.089 & 3.561 \\ \specialrule{1pt}{0pt}{0pt}

Ours (lite,$\texttt{ }$ w/o enhancement)   & 27.08 & 0.898 & 0.112 & 3.868 \\ \hline
Ours (base,\hspace{1.5mm}w/o enhancement)   & 27.19 & 0.906 & 0.108 & 3.831 \\ \hline
Ours (large, w/o enhancement)  & 27.23 & 0.908 & 0.107 & 3.820 \\ \hline
Ours (lite,$\texttt{ }$ w/o enhancement, w/ 3D LuT)    & 27.93 & 0.911 & 0.096 & 3.398 \\ \hline
Ours (base,\hspace{1.5mm}w/o enhancement, w/ 3D LuT)    & 28.13 & 0.921 & 0.088 & 3.336 \\ \hline
Ours (large, w/o enhancement, w/ 3D LuT)   & 28.19 & 0.924 & 0.086 & 3.323 \\ \hline
Ours (lite,$\texttt{ }$ w/$\texttt{ }$ enhancement, w/ 3D LuT)    & 28.73 & 0.930 & 0.090 & 3.234 \\ \hline
Ours (base,\hspace{1.5mm}w/$\texttt{ }$ enhancement, w/ 3D LuT)    & \textbf{\cellcolor{secondbest}{28.95}} & 0.938 & 0.084 & \textbf{\cellcolor{secondbest}{3.177}} \\ \hline
Ours (large, w/$\texttt{ }$ enhancement, w/ 3D LuT)   & \textbf{\cellcolor{best}{29.03}} & \textbf{\cellcolor{secondbest}{0.941}} & \textbf{\cellcolor{secondbest}{0.082}} & \textbf{\cellcolor{best}{3.160}} \\ \hline
\end{tabular}}
\end{table}

\subsubsection{Additional Qualitative Results}
\label{sec:add-qualitative-results}

\begin{figure*}[t]
\centering
\includegraphics[width=\linewidth]{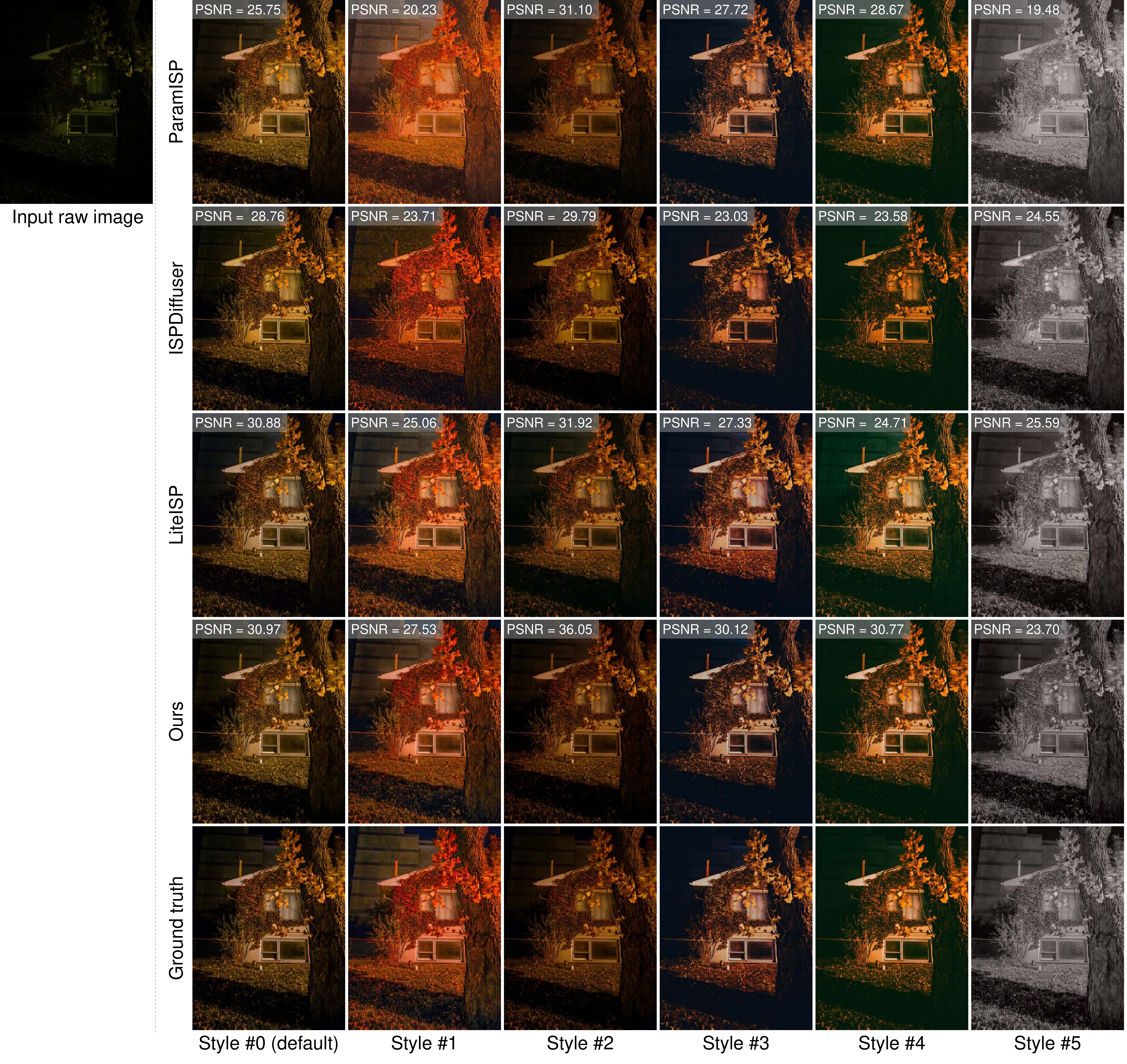}
\vspace{-5mm}
\caption{Additional qualitative comparison between our method and recent neural ISP methods (ISPDiffuser~\cite{ispdiffuser}, LiteISP~\cite{lite-isp}, and ParamISP~\cite{paramisp}). Results are shown for the default style of the S24 dataset~\cite{s24} (Style~\#0) and the remaining artistic picture styles (Styles~\#1--5). The shown images are from the S24 test set.}
\label{fig:qualtative-results-supp}
\vspace{-3mm}
\end{figure*}

\begin{figure*}[!t]
\centering
\includegraphics[width=0.93\linewidth]{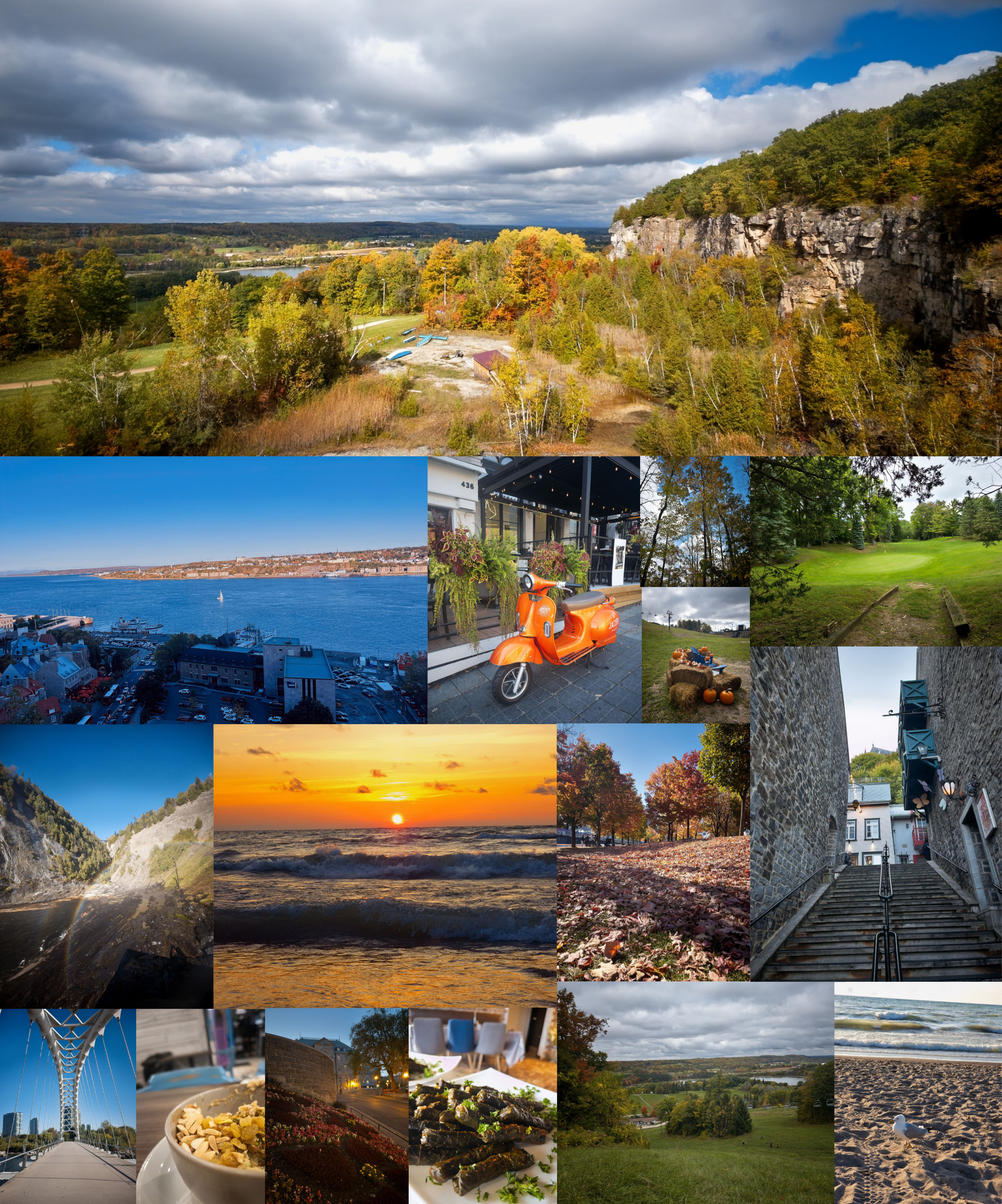}
\vspace{-4mm}
\caption{Additional results produced by our method. The shown images were rendered from raw captures taken with different iPhone devices (13, 13~Pro~Max, 14, and 15) and across various camera modules (main, ultra-wide, and telephoto). For these results, we used the photofinishing and detail-enhancement models trained on the S24 dataset~\cite{s24}, along with the generic denoising and cross-camera AWB models. None of the models were trained on data from any iPhone device.}
\label{fig:additional-results}
\vspace{-4mm}
\end{figure*}

We present additional qualitative results obtained using our method. Figure~\ref{fig:qualtative-results-supp} shows a qualitative comparison between our method and recent state-of-the-art neural ISP methods on the S24 test set \cite{s24}, along with the PSNR results of each method for every style. As shown, our method consistently produces high-quality results while using fewer parameters overall, since only the photofinishing and detail-enhancement modules employ style-specific weights, owing to the modularity of our design. Lastly, Figure~\ref{fig:additional-results} presents additional results on images captured by different iPhone devices. None of our training data includes iPhone images, further demonstrating the cross-device generalization ability of our method.

\section{Evaluation Details}
\label{sec:evaluation_details}

Except for the reported results on the Zurich Raw-to-sRGB dataset~\cite{zurich} (Sec.~\ref{sec:misalignment}), we trained and evaluated all other methods ourselves, following their original papers and released codebases. We generally followed the training procedures and hyperparameters described in the original works, with minor modifications where necessary (as elaborated in the following). For the S24 dataset~\cite{s24}, only demosaiced linear RGB raw images are
available. For methods requiring single-channel mosaiced inputs, we simulated the mosaicing process by applying a fixed Bayer pattern to the RGB images. For three-channel input methods (such as ours), the raw input images from the S24~\cite{s24}, MIT-Adobe FiveK~\cite{Adobe5K}, and Zurich Raw-to-sRGB~\cite{zurich} datasets were generated by applying black-level normalization (using black and saturation levels specified in the DNG files or dataset metadata), followed by demosaicing with the Menon algorithm~\cite{menon2007demosaicing} when RGB demosaiced raw images were not provided.

For the S24 and Zurich Raw-to-sRGB datasets, training was performed using each dataset’s respective training split. For the MIT-Adobe FiveK dataset, we followed the train/validation/test split provided in~\cite{hu2018exposure}; specifically, the split consists of 493 validation images, 989 test images, and the remaining images are used for training.

Since the MIT-Adobe FiveK dataset~\cite{Adobe5K} (Sec.~\ref{sec:results-adobe-5k}) contains images captured with multiple DSLR cameras, training and evaluating single-stage black-box neural ISP methods can often be unstable. To ensure fairness and stability, we trained and tested other methods
on linear sRGB inputs, obtained by converting the raw images using the
illuminant and CCM provided in the DNG metadata.  For methods that require single-channel mosaiced inputs, we used mosaiced data from the MIT-Adobe FiveK dataset only after applying white balancing and color correction.

For Deep-FlexISP~\cite{flexisp}, which employs a three-stage framework (denoising, white balance, and Bayer-to-sRGB mapping), we first trained the denoising stage on the S24 dataset using paired noisy raw and denoised ground-truth images, and then fine-tuned it for each picture style in the S24 dataset, following the training scheme described in the original Deep-FlexISP paper~\cite{flexisp}.

For ReconfigISP~\cite{yu2021reconfigisp}, the codebase depends on external assets that are not included (specifically, pretrained proxy checkpoints and ISP-kernel libraries for differentiable fixed operators). Therefore, we were unable to run the original architecture-search pipeline. Instead, we retrained a fixed architecture on the S24 dataset following the optimal architecture reported for a similar dataset in the paper~\cite{yu2021reconfigisp}. We used denoised and demosaiced RGB inputs and re-implemented the sRGB-domain operators:
\textit{GTM-Manual} $\rightarrow$ \textit{Gamma} $\rightarrow$ \textit{WB-Quadratic}.
These operators are fixed and contain only 34 trainable parameters in total ($3 + 1 + 30$). The model optimizes these parameters for each target style.

For Neural Photo-Finishing (NPF)~\cite{neural_photo_finishing}, the original method is built around Adobe Lightroom. Since Adobe Lightroom does not provide public access to intermediate rendering outputs, the released NPF codebase instead uses Darktable as its rendering backend. However, the repository does not include implementations of the Differentiable Program Operations or the raw style transfer pipeline, so we re-implemented these missing components. Our reimplementation uses the following pipeline:
temperature $\rightarrow$ exposure $\rightarrow$ \textit{colorin} (metadata-driven color space transform) $\rightarrow$ \textit{colorbalancergb} $\rightarrow$ \textit{filmicrgb} (tone mapping) $\rightarrow$ \textit{colorout} (color space transform). We implemented \textit{colorbalancergb} using a Neural Pointwise network and \textit{filmicrgb} using a Neural Area-wise network, while the remaining modules were implemented as Differentiable Program Operations. The rendering pipeline contains 7,849,062 parameters and was trained once, after which it is shared across all target styles. The style encoder contains 483,153 parameters and was retrained for each target style. We replace the original unsupervised style transfer loss with L1 since we are in a supervised setting. During reimplementation, we found that NPF is tightly coupled to a specific version of the third-party rendering engine, which introduces two key caveats: 1) intermediate outputs can be difficult to access or interpret, and 2) the internal module ordering can be non-transparent. Both pieces of information are critical for constructing the proxy pipeline, but may be difficult to obtain.

Among all methods, ParamISP~\cite{paramisp} and ISPDiffuser~\cite{ispdiffuser} required special handling during training compared to the default configurations described in their original papers.~For ParamISP~\cite{paramisp}, originally
designed for two-stage training (multi-camera pretraining followed by
camera-specific finetuning), we found that omitting the pretraining stage
yielded better results on the S24 dataset. For ISPDiffuser~\cite{ispdiffuser},
we observed that training on image patches both reduced training time and
improved accuracy. In both cases, we report the best-performing configurations
we identified.

For the Zurich dataset (Sec.~\ref{sec:misalignment}), results are taken from
prior publications, except for CIE XYZ Net~\cite{afifi2021cie} and Exposure~\cite{hu2018exposure}, which we
trained and tested ourselves. LAN~\cite{lan} results are reported from
Rawformer~\cite{perevozchikov2024rawformer}, while the remaining numbers are
taken from~\cite{zurich, fourier}.  We report quantitative results for all methods using PSNR, SSIM~\cite{ssim},
LPIPS~\cite{LPIPS}, and $\Delta E_{2000}$~\cite{delta-e}. For LPIPS, both
outputs and ground-truth images are resized to 1024$\times$1024 prior to computation.

\end{document}